\documentclass[lettersize]{IEEEtran}
\usepackage{amsmath,amsfonts}
\usepackage{algorithmic}
\usepackage{array}
\usepackage{subcaption} % Use subcaption instead of subfig
\usepackage{textcomp}
\usepackage{stfloats}
\usepackage{url}
\usepackage{verbatim}
\usepackage{graphicx}
\def\BibTeX{{\rm B\kern-.05em{\sc i\kern-.025em b}\kern-.08em
    T\kern-.1667em\lower.7ex\hbox{E}\kern-.125emX}}
\usepackage{balance}

\newcommand{\E} {{\mathbb E}}
\newcommand{\R}{\mbox{$\mathbb{R}$}}

\newcommand{\B}{\mathbf}
\newcommand{\Bs}{\boldsymbol}

\newcommand{\Av}[1] {\underset {#1} {\operatorname{Ave}}\,}
\newcommand{\KL}{\mathbb{KL}}

\DeclareMathOperator*{\Exp}{\mathbb{E}}

\DeclareMathOperator*{\argmin}{arg\,min}

\newcounter{row}
\makeatletter
\@addtoreset{subfigure}{row}
\makeatother

\usepackage[unicode=true]{hyperref}
\hypersetup{
    colorlinks=true,
    citecolor=blue,
    urlcolor=blue,
    linkcolor=blue,
} % Load hyperref first
\usepackage{cleveref} % Then load cleveref

\crefname{section}{Section}{Sections}
\Crefname{section}{Section}{Sections}

\crefname{figure}{Fig.}{Figs.}
\Crefname{figure}{Fig.}{Figs.}

\crefname{subfigure}{Fig.}{Figs.} % Customize cleveref for multiple figures
\Crefname{subfigure}{Fig.}{Figs.}

\crefname{equation}{}{}
\Crefname{equation}{}{}

\crefname{subequations}{}{}
\Crefname{subequations}{}{}

\usepackage{bookmark}
\usepackage{microtype}
\begin{document}
\title{Multi-scale clustering and source separation of InSight mission seismic data}
  \author{
    \IEEEauthorblockN{Ali Siahkoohi,
    Rudy Morel,
    Randall Balestriero,
    Erwan Allys,
    Grégory Sainton,
    Taichi Kawamura, \\
    and Maarten V. de Hoop}
\thanks{
Ali Siahkoohi is with the Department of Computer Science, University of
Central Florida, Orlando, FL 32816, USA (e-mail: \href{mailto:alisk@ucf.edu}{alisk@ucf.edu}), and was formerly affiliated with Rice University.

Rudy Morel is jointly affiliated with the Département d'Informatique,
École Normale Supérieure, 45 rue d’Ulm, 75005, Paris, France and the Center for Computational Mathematics Flatiron Institute 162 5th Avenue New York, NY 10010, USA (e-mail:
\href{mailto:rmorel@flatironinstitute.org}{rmorel@flatironinstitute.org}).

Randall Balestriero is with the Department of Computer Science, Brown
University, Providence, RI 02912, USA (e-mail:
\href{mailto:randall_balestriero@brown.edu}{randall\_balestriero@brown.edu}).

Erwan Allys is with the Laboratoire de Physique de l’Ecole normale supérieure, ENS, Université PSL, CNRS, Sorbonne Université, Université Paris-Cité, 75005 Paris, France (e-mail: \href{mailto:erwan.allys@phys.ens.fr}{erwan.allys@phys.ens.fr}).

Grégory Sainton is with the LUX, Observatoire de Paris, Université PSL, Sorbonne Université, CNRS, 75014 Paris, France (e-mail: \href{mailto:gregory.sainton@obspm.fr}{gregory.sainton@obspm.fr}).

Taichi Kawamura is with the Université Paris Cité, Institut de Physique du Globe de Paris, CNRS, 75005 Paris, France (e-mail: \href{mailto:kawamura@ipgp.fr}{kawamura@ipgp.fr}).

Maarten V. de Hoop is with the Department of Computational Applied Mathematics and Operations Research, Rice University, Houston, TX 77005, USA (e-mail: \href{mailto:mdehoop@rice.edu}{mdehoop@rice.edu}).
}
}

% \markboth{Submitted to the IEEE Transactions on Neural Networks and Learning Systems}{}
\maketitle

% \IEEEpeerreviewmaketitle

\begin{abstract}
Unsupervised source separation involves unraveling an unknown set of
source signals recorded through a mixing operator, with limited prior
knowledge about the sources, and only access to a dataset of signal
mixtures. This problem is inherently ill-posed and is further challenged
by the variety of timescales exhibited by sources in time series data
from planetary space missions. As such, a systematic multi-scale
unsupervised approach is needed to identify and separate sources at
different timescales. Existing methods typically rely on a preselected
window size that determines their operating timescale, limiting their
capacity to handle multi-scale sources. To address this issue, we
propose an unsupervised multi-scale clustering and source separation
framework by leveraging wavelet scattering spectra that provide a
low-dimensional representation of stochastic processes, capable of
distinguishing between different non-Gaussian stochastic processes.
Nested within this representation space, we develop a factorial
variational autoencoder that is trained to probabilistically cluster
sources at different timescales. To perform source separation, we use
samples from clusters at multiple timescales obtained via the factorial
variational autoencoder as prior information and formulate an
optimization problem in the wavelet scattering spectra representation
space. When applied to the entire seismic dataset recorded during the
NASA InSight mission on Mars, containing sources varying greatly in
timescale, our approach disentangles such different sources, e.g.,
minute-long transient one-sided pulses (known as ``glitches'') and
structured ambient noises resulting from atmospheric activities that
typically last for tens of minutes, and provides an opportunity to
conduct further investigations into the isolated sources.
\end{abstract}


\section{Introduction}
\IEEEPARstart{S}{ource} separation involves the often ill-posed problem of retrieving an \textit{a priori} unknown number of source signals from an observed signal, often in the form of a time series. In order to tackle the ambiguity in retrieving the source signals, source separation methods incorporate available prior information about the sources---traditionally, by making assumptions about the regularity of the sources \cite{Cardoso_1989, JuttenEtAl_1991, BinghamEtAl_2000, NandiEtAl_1996, Cardoso_1998, JuttenEtAl_2004, ZhenEtAl_2017, HwangEtAl_2018, LiuEtAl_2018, GuzikEtAl_2024}, e.g., sparsity or low-rank structure in some transform domain and certain distributional assumptions. While these traditional methods have been extensively studied and are well understood, their underlying regularity assumptions, if not realistic, can introduce bias into the outcome of the source separation process \cite{Cardoso_1998,ParraEtAl_2003}. For example, the eponymous classical independent component analysis (ICA) \cite{Lee_1998} source separation method and its variations, e.g., nonlinear ICA \cite{Hyv_arinenEtAl_1999}, generally impose assumptions such as non-Gaussianity and stationarity of sources, which makes their application extremely challenging in noisy environments and when dealing with data that we have very little prior knowledge, e.g., when dealing with data obtained from extraterrestrial missions. In fact, most real-world applications of ICA fall back to separating noise from signal \cite{Salimi_khorshidiEtAl_2014,Hoyer_1999} as opposed to separating the sources within the signal that is embedded in noise.

In contrast to the conventional methods, data-driven source separation approaches are able to learn the prior information on sources (either implicitly or explicitly) from data. For example, supervised source separation methods \cite{KameokaEtAl_2019, WangEtAl_2018, LuoEtAl_2023, ZhangEtAl_2024, OzerEtAl_2024, LutatiEtAl_2024, CohenEtAl_2024, PonsEtAl_2024} aim to separate sources by utilizing labeled training data, comprising pairs of sources and their mixtures. However, in domains with limited expert knowledge, obtaining labeled training data is often challenging, making supervised learning methods unsuitable. Furthermore, the use of synthetic data to artificially generate labels also proves challenging when the signal to be generated is not thoroughly understood. On the other hand, while unsupervised source separation methods can be applied in domains with no access to labeled data \cite{JangEtAl_2003, FevotteEtAl_2009, HersheyEtAl_2016, LinEtAl_2018, DrudeEtAl_2019, KeEtAl_2020, DoEtAl_2020, WisdomEtAl_2020, NeriEtAl_2021, LiuEtAl_2022, DentonEtAl_2022}, they are not well-suited to handle source separation problems where sources exhibit vastly varying timescales. This is partially due to their reliance on preselected window sizes that limits their usage to separating sources that span within the window's timescale. This limitation can be a critical obstacle in analyzing complex phenomena that involve sources with vastly different timescales.

\begin{figure*}[t]
    \centering
    \includegraphics[width=1.0\textwidth]{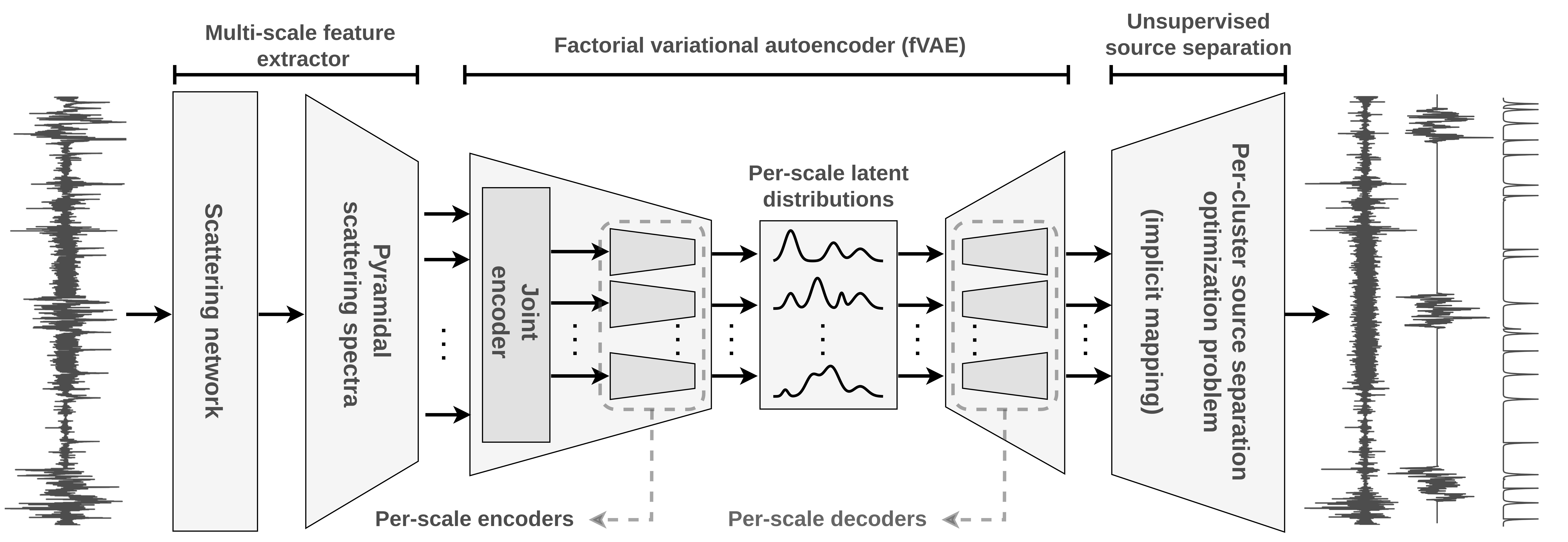}
    \caption{A schematic diagram depicting the proposed unsupervised
    multi-scale clustering and source separation framework, consisting
    of three stages. In the initial stage, indicated on the left in
    \cref{fig:schematic}, we convert the data into the pyramidal
    scattering spectra representation space. The objective of this stage
    is to represent the data across significantly diverse timescales,
    ranging from less than a minute to an hour. The second stage
    involves training a generative model, namely a factorial variational
    encoder, which learns a joint probabilistic description of the
    multi-scale representations (indicated in the middle of
    \cref{fig:schematic} with arrows going from one block to another
    indicating representations associated with different timescales).
    The goal of this stage is to identify prominent sources in the
    dataset across several timescales, which will, in turn, be used as
    ``prior information'' in source separation. In the final stage,
    indicated on the right in \cref{fig:schematic}, we separate the
    prominent source signals identified by the fVAE in the previous
    stage via solving an optimization problem in the scattering spectra
    space with the goal of separating a specific source of interest in
    the time domain. \cref{app:algorithm-overview} of the Appendix
    provides a more detailed walkthrough of the proposed framework. The
    code for partially reproducing the results will be made available
    after the review process.}
    \label{fig:schematic}
\end{figure*}

To address this problem, we present an unsupervised multi-scale clustering and source separation framework that is capable of detecting and separating prominent sources in a dataset of signal mixtures. Our framework consists of the following building blocks (see \cref{fig:schematic} for a schematic diagram),

\begin{itemize}
    \item \textbf{Pyramidal scattering spectra.} In light of the
    multi-scale nature of the sources, we develop our framework nested
    within the wavelet scattering spectra representation space
    \cite{MorelEtAl_2022}. This multi-scale representation of time
    series is based on the second moment of scattering network features
    \cite{BrunaEtAl_2013} and is capable of capturing the non-Gaussian
    and multi-scale characteristics of stationary stochastic processes
    in the data. To take into account the existence of multiple sources
    with varying timescales, we propose the pyramidal scattering spectra
    (first contribution), which involves averaging the wavelet
    scattering spectra representation at different timescales, resulting
    in a description of the data that captures non-Gaussian
    characteristics within different timescales;
    \item \textbf{Factorial variational autoencoder.} Within the
    pyramidal scattering spectra representation space, we introduce a
    factorial variant of Gaussian-mixture variational autoencoders
    (VAEs) \cite{KingmaEtAl_2014a, NEtAl_2017, JangEtAl_2017} (second
    contribution), hereon referred to as fVAE, that simultaneously
    learns to probabilistically cluster sources at different
    timescales---the different factors---in the latent space and
    independently sample scattering spectra representations associated
    with each cluster. The former provides clusters of prominent sources
    in the dataset at different timescales, and the latter models the
    wavelet scattering spectra distribution of these sources, providing
    prior information necessary for source separation;
    \item \textbf{Unsupervised source separation.} We perform source
    separation by complementing the method proposed by
    \cite{SiahkoohiEtAl_2023} by the learned prior information regarding
    each source (third contribution). Our method for source separation
    involves solving an optimization problem over the unknown sources in
    the wavelet scattering spectra representations space. This is
    achieved by minimizing carefully selected and normalized loss
    functions that incorporate prior knowledge about the source of
    interest, ensure data-fidelity, and promote statistical independence
    between the recovered sources.
\end{itemize}

To demonstrate the applicability of our approach, we apply our approach
to the entire seismic dataset recorded by a seismometer on Mars during
NASA's Interior Exploration using Seismic Investigations, Geodesy and
Heat Transport (InSight) mission \cite{GiardiniEtAl_2020,
GolombekEtAl_2020, Knapmeyer_endrunEtAl_2020}. The InSight lander's
seismometer---known as the SEIS instrument---detected marsquakes
\cite{HorlestonEtAl_2022, CeylanEtAl_2022, PanningEtAl_2023,
Service_2023} and transient atmospheric signals, such as wind and
temperature changes, that provide information about the Martian
atmosphere \cite{BanfieldEtAl_2020} and enable studying the interior
structure and composition of the Red Planet \cite{BegheinEtAl_2022}. The
signal recorded by the InSight seismometer is heavily influenced by
atmospheric activity and surface temperature \cite{Lognonn_eEtAl_2020,
LorenzEtAl_2021}. These non-seismic signals largely vary in timescale
from minutes to hours and it would be crucial to capture this in
multiple timescale. In addition to providing examples on the seismic
dataset recorded during the InSight mission, we demonstrate the results
of our approach in a controlled setting when applied to a stylized
example involving multi-scale clustering and source separation of time
series consisting of sources with three different timescale (cf.
\cref{app:stylized} of the Appendix).

Through analysis of this complex dataset via our approach, we observe
that clusters identified at the finest timescale reveal distinct
features such as glitches with and without precursors, with the former
showing increased occurrence around Martian sunset. As the timescale
broadens, clusters exhibit longer-duration events like oscillatory
signals and bursts of high-frequency energy, possibly associated with
wind dynamics. Moreover, coarser timescales unveil phenomena like wind
gusts and atmospheric interactions, evident from characteristic
waveforms and their occurrence patterns aligned with Martian day-night
cycles. By leveraging clusters of glitches and wind-burst noises, we
applied our source separation method and successfully separated these
sources from the given time series. The results, illustrated through
waveform visualizations, indicate successful separation with minimal
alterations to the original waveform, validating the effectiveness of
the approach.  Additionally, as we show in
\cref{app:marsquake_separation} of the Appendix, our method is also
capable of separating background noise and glitches from a marsquake,
which indicates the applicability of our method in cleaning sources that
are not prominent enough to have their own clusters. Finally, the
exploration of latent space clusters via the fVAE architecture provides
insights into the learned representation of the dataset. Through
low-dimensional visualizations, we examined clusters across different
timescales. We observed that clusters in finer timescales exhibit
clearer boundaries, suggesting more structured data. Moreover, the
distribution of quality ``A'' broadband events and pressure drops across
different timescales highlights the multi-scale representation of data
captured by the fVAE architecture. Broadband events concentrate more in
coarser timescales, while pressure drops are compactly situated in finer
timescales, indicating the efficacy of the learned representation in
capturing diverse temporal phenomena. Overall, these analyses offer
valuable insights into the separation of prominent sources and the
structure of latent space clusters, showcasing the capabilities of the
proposed methods in analyzing complex datasets.

In the following sections, we introduce the utilization of wavelet
scattering spectra as a representation rich in domain knowledge for
analyzing multi-scale time series. Furthermore, we explain how we extend
this representation to handle non-stationary processes. The core of our
approach lies in the description of factorial Gaussian-mixture VAEs,
which enable multi-scale probabilistic clustering, which will be used as
prior information for source separation. In the final stage of our
framework, we detail our source separation approach, which involves
solving an optimization problem using loss functions defined in the
wavelet scattering spectra space. Lastly, we present the results of
applying our approach to the seismic dataset obtained from the InSight
mission.

\section{Pyramidal scattering spectra}
\label{sec:scattering}

\noindent Time-series data $\B{x}$ recorded during space missions are
typically non-stationary non-Gaussian noise. Such time-series are
multi-scale in two respects. First, the signal $\B{x}$ is a mixture of
sources occurring at different timescales, which contributes to its
non-stationarity. Second, each individual source has variations on a
wide range of scales, which can be observed by looking at a scalogram.

This section constructs a representation adapted to this doubly
multi-scale nature of the data by building pyramidal wavelet scattering
spectra. Such representation can be described as a multiple timescales
average over diagonal correlation features on a two-layer convolutional
neural network with predefined wavelet filters.

\subsection{Wavelet scattering networks}

\noindent A scattering network \cite{BrunaEtAl_2013} is a cascade of
wavelet operators $\B{W}$ followed by nonlinear activation functions
(akin to a typical convolutional neural network). A wavelet transform
operator $\B{W}$ is a convolutional operator with predefined wavelet
filters that extracts variations at separate scales. These filters
include a low-pass filter $\varphi_J(t)$ and $J$ complex-valued
band-pass filters $\psi_j(t)=2^{-j}\psi(2^{-j}t),\ 1\leq j\leq J$, which
are obtained by the dilation of a mother wavelet $\psi(t)$ that have
zero mean and a fast decay away from $t=0$. The wavelet coefficient
$\B{W}\B{x}(t,j)=x\star\psi_j(t)$ extracts variations of the input
signal $\B{x}(t)$ around time $t$ at scale $2^j$. In order to
characterize time-evolution of the wavelet coefficients e.g., envelope
modulation, we apply a modulus $|\cdot|$ and cascade a second wavelet
operator. The output of a two-layer scattering network $S$ is $S(\B{x})
:= \big( \B{W}\B{x} \,,\, \B{W}|\B{W}\B{x}| \big)^{\top}$, it extracts
variations of signal $\B{x}$ and its multi-scale envelopes
$|\B{W}\B{x}|$ at different times and different scales. Even though
these networks have been successfully employed in tasks such as
intermittency analysis \cite{BrunaEtAl_2015}, clustering
\cite{SeydouxEtAl_2020}, event detection and segmentation
\cite{Rodr_iguezEtAl_2021} (with learnable wavelets), they are not
sufficient to build an accurate description of a multi-scale process, as
they fail to capture crucial dependencies across different scales
\cite{MorelEtAl_2022}.

The architecture we use for wavelet scattering
spectra computation throughout this paper is a two-layer scattering
network with $J=8$ different octaves with one wavelet per octave.

\subsection{Capturing non-Gaussian properties through scale dependencies}

\noindent Sources $\B{x}$ encountered in time-series studied in this
paper are very often non-Gaussian processes. For sake of simplicity, let
us assume $\B{x}$ is a stationary source. If $\B{x}$ were Gaussian then
the different scale channels of a scattering network would be
independent, however it is not the case in practice and these
dependencies were shown to be crucial to characterize the non-Gaussian
stochastic structure of $\B{x}$ \cite{MorelEtAl_2022}. Such dependencies
can be captured by considering the correlation matrix
$\E[S(\B{x})S(\B{x})^\top]$:
\begin{equation}
\label{eq:scat-cov-full}
\E
\begin{bmatrix}
\B{W}\B{x} \left(\B{W}\B{x}\right)^\top & \B{W}\B{x} \left(\B{W}|\B{W}\B{x}|\right)^\top\\
\B{W}|\B{W}\B{x}| \left(\B{W}\B{x}\right)^\top & \B{W}|\B{W}\B{x}| \left(\B{W}|\B{W}\B{x}|\right)^\top
\end{bmatrix}.
\end{equation}

This matrix contains three types of coefficients. Correlation coefficients $\E[\B{W}\B{x} \left(\B{W}\B{x}\right)^\top]$ come down to the wavelet power spectrum, which characterizes in particular the \textit{roughness} of the signal. Correlation coefficients $\E[\B{W}\B{x} \left(\B{W}|\B{W}\B{x}|\right)^\top]$ capture signed interaction between wavelet coefficients. In particular, they detect sign-asymmetry and time-asymmetry in $\B{x}$ \cite{MorelEtAl_2022}. Finally, coefficients $\E\{\B{W}|\B{W}\B{x}| \left(\B{W}|\B{W}\B{x}|\right)^\top\}$ capture correlations between signal envelopes $|\B{W}\B{x}|$ at different scales. These correlations account for intermittency and envelope time-asymmetry \cite{MorelEtAl_2022}.

Owing to the compression properties of wavelet operators \cite{Wornell_1993} for the type of signals considered in this paper, these matrices are quasi-diagonal. We denote $\E[\operatorname{diag}\left(S(\B{x})S(\B{x})^\top\right)]$ an appropriate diagonal approximation of the full sparse matrix in \cref{eq:scat-cov-full}. The expectation $\E$ is replaced by a time average denoted by $\Av{}$ (average pooling) whose size should be chosen as the typical duration of event $\B{x}$. The wavelet scattering spectra representation is
\begin{equation}
\label{eq:scattering-cov}
\Psi(\B{x}) := \Av{} \big( S(\B{x})
\,,\,
\operatorname{diag}\left(S(\B{x})S(\B{x})^\top\right)
\big).
\end{equation}

This representation extracts average and correlation features on a two-layer convolutional neural network with predefined wavelet filters. They are analogous to the features extracted in \cite{GatysEtAl_2015} for generation. However, we do not train any weights in our representation. Owing to the compression properties of wavelet operators we obtain a low-dimensional representation. For a signal $\B{x}\in\R^L$ of length $L$ the scattering spectra contain approximately $\log^3_2(L)/6$ coefficients \cite{MorelEtAl_2022}. As a consequence, our representation, composed of order 1 and order 2 moments, can be estimated with low-variance, which can be crucial for clustering and disentangling sources.
One can also consider scattering cross-spectra between two signals, $\B{x}$ and $\B{y}$, defined by  $\Psi(\B{x},\B{y})=\operatorname{Ave} \operatorname{diag} \big(S(\B{x})S(\B{y})^\top\big)$. They capture nonlinear, non-Gaussian dependencies between signals $\B{x}$ and $\B{y}$.

\subsection{Pyramidal averaging}

\noindent Non-stationarity in the data $\B{x}$ is in part explained by the presence of sources at different timescales. Our representation $\Psi(\B{x})$ averages scattering spectra features on a certain window. If a source has a time-duration that is much smaller than the average window size, it will be averaged out, and the representation will contain little information about such a source. To take into account the variety of source timescales, we consider different averaging sizes in a causal manner. We replace $\Av{}{}$ in \cref{eq:scattering-cov} by a multi-scale average pooling operator $\Av{} = \big(\Av{t\in w_1},\ldots,\Av{t\in w_K} \big)$. The windows $w_1\subset\ldots\subset w_K$ have a pyramidal structure, they are of increasing size, and all ending at the same time, where $\Av{t\in w_1}$ considers recent past while $\Av{t\in w_K}$ considers distant past. To cover a large range of timescales, we choose $w_{k+1}$ to be four times longer than $w_k$. This defines a \emph{pyramidal scattering spectra} representation $\Psi(x) = (\Psi_1(x),\ldots,\Psi_K(x))$ whose factor $k$ is
\begin{equation}
\label{eq:pyramidal-scattering-cov}
\Psi_k (\B{x}) := \Av{t\in w_k}
\big(
S(\B{x}),
\operatorname{diag}\left(S(\B{x})S(\B{x})^\top\right)
\big).
\end{equation}

The full representation $\Psi(\B{x})$, consisting of all factors decomposes the variation in the stochastic structure of process $\B{x}$ over time through the multi-scale pooling operator $\Av{}$.

\section{Factorial Gaussian-mixture variational autoencoder (fVAE)}
\label{sec:fvae}

\noindent To perform source separation on our multi-scale representation, we require a generative model that can simultaneously cluster and sample. As Gaussian-mixture VAEs \cite{KingmaEtAl_2014a, NEtAl_2017, JangEtAl_2017} are capable of learning highly structured, low-dimensional latent representations of data, they are a promising candidate for achieving our goals. The major open question remains on the structure of the mapping between the input space time-series and latent space cluster variables. As our aim is to cluster---or separate---sources co-occurring with different timescales, we propose a factorial variant to Gaussian-mixture VAEs to (i) jointly encode the wavelet scattering spectra representations of different timescales; (ii) learn a low-dimensional Gaussian mixture latent variable for each timescale, enabling clustering; and (iii) independently decode the latent representations of each timescale to be used as prior information in source separation. In the next few subsections we describe our proposed generative model.

\subsection{Generative model}

\noindent Denote $\B{u} := (\B{u}_0, \ldots, \B{u}_{s-1})$ as the pyramidal wavelet scattering spectra representation of an input signal for $s$ scales. Our goal is to approximate the target joint distribution $p(\B{u})$ using variational inference (VI) \cite{JordanEtAl_1999, BleiEtAl_2017}, leveraging samples from this distribution as training data. We achieve this by defining the following generative model,
\begin{subequations}\label{eq:generative}
\begin{align}
p_{\theta}(\B{u}, \B{y}, \B{z}) & = p_{\theta}(\B{u} | \B{z}) p_{\theta}(\B{z} | \B{y}) p_{\theta}(\B{y}) \\
& = \prod_{i=0}^{s-1} p_{\theta}(\B{u}_i | \B{z}_i)\,p_{\theta}(\B{z}_i | y_i)\,p_{\theta}(y_i).
\end{align}
\end{subequations}

In this expression, $\B{z}_i$ and $y_i$ for $i = 0, \ldots, s-1$ represent the Gaussian mixture and categorical latent variables for the $i^{\text{th}}$ timescale, respectively. Furthermore, $\B{z} := (\B{z}_0, \ldots, \B{z}_{s-1})$ and $\B{y} := (y_0, \ldots, y_{s-1})$ represent the collection of latent variables for all timescales. We choose the following parametric distributions for these random variables for all timescales $i = 0, \ldots, s-1$,
\begin{subequations}
\begin{align}
p_{\theta}(y_i) & = \text{Cat} \left( c_i^{-1} \B{1}_{c_i}\right), \\
p_\theta(\B{z}_i | y_i) & =\mathcal{N}\left(\B{z}_i | \Bs{\mu}_{z,i}(y ; \Bs{\theta}), \operatorname{diag}(\Bs{\sigma}_{z, i}^2(y_i ; \Bs{\theta}))\right), \\
p_\theta(\B{u}_i | \B{z}_i) & =\mathcal{N}\left(\B{u}_i | \Bs{\mu}_{u,i}(\B{z}_i ; \Bs{\theta}), \operatorname{diag}(\Bs{\sigma}_{u,i}^2(\B{z}_i ; \Bs{\theta}))\right),
\end{align}
\label{eq:generative-latents}
\end{subequations}

\noindent where  $c_i$ represents the number of components in the Gaussian mixture latent distribution. $p_\theta(\B{z}_i | y_i)$ is modeled as a Gaussian distribution with $\Bs{\mu}_{z,i}(y ; \Bs{\theta})$ and $\Bs{\sigma}_{z, i}(y_i ; \Bs{\theta})$ simply being learnable vectors for each $y_i \in \{0, \ldots, c_i-1 \}$, which amounts to a Gaussian mixture model for $\B{z}_i$. Conditioned on $\B{z}_i$, $\B{u}_i$ is also modeled as a Gaussian distribution with mean $\Bs{\mu}_{u,i}(\B{z}_i ; \Bs{\theta})$ and diagonal covariance $\Bs{\sigma}_{u,i}(\B{z}_i; \Bs{\theta})$ parameterized using deep nets. The generative model setup outlined above translates to having independent decoders---i.e., mappings from latent variables to scattering spectra representations---for each timescale. This approach enables the independent synthesis of scattering spectra representations for each timescale for the downstream source separation step. To train this generative model tractably using VI, we define an inference model in the next section that approximates the latent posterior distribution.

\subsection{Inference model}

\noindent Evaluating the likelihood of the parametric distribution $p_{\theta}(\B{u})$ requires marginalizing out the Gaussian mixture and the categorical latent variables from the joint distribution in \cref{eq:generative}. Unfortunately, this process is computationally infeasible due to the high-dimensionality of these distributions. To overcome this obstacle, we utilize amortized VI to approximate the latent posterior distribution $q(\B{y}, \B{z} | \B{u})$. This approximation uses the Evidence Lower Bound (ELBO) \cite{KingmaEtAl_2014a, BleiEtAl_2017} to approximate the model likelihood conditioned on the latent posterior on the wavelet scattering spectra representations for all timescales. To enable multi-scale clustering of the input data, we employ the following factorization of the posterior distribution,
\begin{subequations}
\begin{align}
q_{\phi}(\B{z}, \B{y} | \B{u}) & = q_{\phi}(\B{z} | \B{y}, \B{u}) q_{\phi}(\B{y} | \B{u}) \\
& = \prod_{i=1}^s q_{\phi}(\B{z}_i | y_i, \B{u}) q_{\phi}(y_i | \B{u}).
\end{align}
\label{eq:inference}
\end{subequations}

In the above expression, the pyramidal scattering spectra representation is used to infer the per-scale cluster, which in turn determines the per-scale, per-cluster Gaussian latent distribution (associated component in the Gaussian mixture model). We use the following parameterizations to learn an amortized latent posterior model for each timescale $i = 0, \ldots, s-1$,
\begin{subequations}
\begin{align}
q_{\phi}(y_i | \B{u})  & = \text{Cat} \left( \Bs{\pi}_i (\B{u}; \Bs{\phi}) \right), \\
q_{\phi}(\B{z}_i | y_i, \B{u}) & = \mathcal{N}\big(\B{z}_i | \Bs{\mu}_{z,i}(\B{u}, y_i ; \Bs{\phi}),  \\
& \quad \quad \quad \ \ \operatorname{diag}(\Bs{\sigma}_{z, i}^2(\B{u}, y_i ; \Bs{\phi}))\big),
\end{align}
\label{eq:inference-latents}
\end{subequations}

\noindent where $\Bs{\pi}_i (\B{u}; \Bs{\phi})$, parameterized by a neural network, represents the cluster membership probabilities for pyramidal scattering spectra input $\B{u}$ at the $i^{\text{th}}$ timescale. Since the inferred latent variable encodes information regarding the cluster membership of $\B{u}$, we explicitly input both $\B{u}$ and $y_i$ to the neural network parameterizations of the mean $\Bs{\mu}_{z,i}(\B{u}, y_i ; \Bs{\phi})$ and diagonal covariance $\Bs{\sigma}_{z, i}^2(\B{u}, y_i ; \Bs{\phi})$. With the generative and inference models defined, we derive the objective function for training the fVAE in the next section.

\subsection{Training objective function}

\noindent Training the fVAE involves minimizing the reverse Kullback-Leibler (KL) divergence between the parameterized and true joint distribution,
\begin{subequations}
\begin{align}
& \argmin_{\Bs{\theta}}\,\KL\,\big( p(\B{u}) \,||\,  p_{\theta}(\B{u}) \big) \\
& = \argmin_{\Bs{\theta}}\,\Exp_{\B{u} \sim  p(\B{u})}
\big[ \underbrace{\log p(\B{u})}_{\text{const. w.r.t. }\Bs{\theta}}-\log p_{\theta}(\B{u})  \big] \\[4pt]
& = \argmin_{\Bs{\theta}}\,\Exp_{\B{u} \sim  p(\B{u})}
\big[ -\log p_{\theta}(\B{u})  \big].
\end{align}
\label{eq:reverse-kl}
\end{subequations}

Evaluating the likelihood $p_{\theta}(\B{u})$ is intractable due to the
required marginalization over $\B{y}$ and $\B{z}$ (cf.
\cref{eq:generative}). To overcome this intractability, we maximize a
lower bound to the likelihood (or minimize the negative lower bound),
commonly referred to as the ELBO. This lower bound to the likelihood is
obtained by computing the expectation of joint distribution in
\cref{eq:generative} with respect to the latent posterior distribution,
leading to the following training optimization problem:
\begin{subequations}
\begin{align}
& \min_{\Bs{\theta}}\, \Exp_{\B{u} \sim  p(\B{u}) }\,\big[-\log p_{\theta}(\B{u}) \big] \\
\leq & \min_{\Bs{\theta}, \Bs{\phi}}  \sum_{i=0}^{s-1} \Exp_{\B{u} \sim  p(\B{u}) }\, \Big[ \Exp_{
\substack{y_i \sim  q_{\phi}(y_i | \B{u})\\[-1pt]
\B{z}_i \sim  q_{\phi}(\B{z}_i | y_i, \B{u})}} \\
&\quad \Big[ -\log  \frac{p_{\theta}(\B{u}_i | \B{z}_i)\,p_{\theta}(\B{z}_i | y_i)\,p_{\theta}(y_i)}{q_{\phi}(\B{z}_i | y_i, \B{u})\,q_{\phi}(y_i | \B{u})} \Big] \Big]
\label{eq:reverse-kl-2-i-new} \\
= &  \min_{\Bs{\theta}, \Bs{\phi}}\, \sum_{i=1}^s  \Big[\Exp_{
\substack{\B{u} \sim  p(\B{u})\\[-1pt]
\B{z}_i \sim  q_{\phi}(\B{z}_i | y_i, \B{u})}} \, \underbrace{\big[ -\log p_\theta(\B{u}_i | \B{z}_i) \big] }_{\substack{
    \text{Per-scale reconstruction loss}}} \\[7pt]
&   + \Exp_{\B{u} \sim  p(\B{u}) }\, \underbrace{\KL\,\big( q_{\phi}(y_i | \B{u}) \,||\, p_{\theta}(y_i) \big)}_{\substack{
    \text{Categorical prior on } y_i}}  \\[4pt]
& +  \Exp_{
\substack{\B{u} \sim  p(\B{u})\\[-1pt]
y_i \sim  q_{\phi}(y_i | \B{u}) }}\, \underbrace{\big[ \KL\,\big( q_{\phi}(\B{z}_i | y_i, \B{u}) \,||\,  p_\theta(\B{z}_i | y_i)  \big) \big] }_{\substack{
    \text{Gaussian mixture prior on } \B{z}_i}} \Big].
\end{align}
\label{eq:reverse-kl-2}
\end{subequations}

The expectations in the optimization problem above can be approximated
using Monte Carlo integration over samples from their respective
distributions. For a detailed derivation, please refer to
\cref{app:loss_derivation} of the Appendix.

\section{Unsupervised source separation}\label{sec:src-sep}

\noindent In this section, we aim to perform the source separation in the time domain, while the previous section focused on the scattering spectra space.
We introduce an unsupervised source separation algorithm based on an optimization problem in the time domain, with a loss function defined in the scattering spectra space.
The pretrained fVAE provides a model for the distribution of sources in the scattering spectra space. We use this as prior information in our optimization problem. Combining these techniques allows for identifying and separating unknown multi-scale sources within a given time window, which aligns with the objectives of unsupervised source separation. This approach was initially introduced by \cite{Regaldo_saint_blancardEtAl_2021, DelouisEtAl_2022} and recently adapted to single timescale unsupervised source separation by \cite{SiahkoohiEtAl_2023}.

Denote $\B{x}$ as a given time window, which is the sum of unknown independent sources $\B{s}_i$, where $i=1, \ldots, M$, possibly occurring at different timescales, with measurement noise $\Bs{\nu}(t)$, so that $\B{x}(t) = \sum_{i=1}^M \B{s}_i(t) + \Bs{\nu}(t)$, where $\B{n}(t) = \Bs{\nu}(t) + \sum_{i=2}^{M} \B{s}_i(t)$. Our approach to source separation involves detecting the prominent sources in $\B{x}(t)$ and separating them one-by-one. We assume that the source $\B{s}_1(t)$ is associated to one of the clusters identified by our fVAE model and we wish to separate $\B{s}_1$ from the mixture. To address this ill-posed problem, we incorporate prior knowledge in the form of realizations $\{\B{s}_1^i\}_{i=1}^N$, which the fVAE identifies as samples from the same cluster as the unknown source. Using these samples, we define three loss terms to ensure that the reconstructed source $\B{\widetilde{s}}_1$ has statistics consistent with the collected samples $\B{s}_1^i$, as well as that $\B{x}-\B{\widetilde{s}}_1$ has the same statistics as $\B{n}$. It also promotes statistical independence between $\B{s}_1$ and $\B{x}-\B{\widetilde{s}}_1$. These loss terms are:
\begin{subequations}
\begin{align}
\mathcal{L}_\text{prior}(\B{s}_1) & = \sum_{i=1}^N
\frac{\Big\|\Psi_k\big(\B{s}_1\big) - \Psi_k\big(\B{s}^i_1\big)\Big\|_2^2}{\sigma^2\big(\Psi_k\big(\B{s}^i_1\big)\big)}, \label{eq:loss-prior}\\
\mathcal{L}_\text{cross}(\B{s}_1) & =  \sum_{i=1}^N \frac{\Big\| \Psi_k\left(\B{s}^i_1, \B{x}-\B{s}_1 \right)\Big\|^2_2}{\sigma^2\big(\Psi_k\big(\B{s}^i_1,\B{x}\big)\big)}, \label{eq:loss-cross}\\
\mathcal{L}_\text{data}(\B{s}_1) & =  \sum_{i=1}^N \frac{\Big\| \Psi_k\left(\B{x} - \B{s}_1 + \B{s}^i_1 \right) - \Psi_k\left(\B{x}\right)\Big\|^2_2}{\sigma^2\big(\Psi_k\big(\B{x}+\B{s}^i_1\big)\big)}. \label{eq:loss-data}
\end{align}
\end{subequations}

The loss in \cref{eq:loss-prior} ensures consistency between the statistics of the recovered signal $\B{s}_1$ with the statistics of the observed signals $\B{s}^i_1$.
The loss in \cref{eq:loss-cross} ensures that the reconstruction $\B{x}-\B{s}_1$ of $\B{n}$ has implicitly the correct statistics by ensuring $\B{n}+\B{s}^i_1$ has consistent statistics with $\B{x}$.
Finally, the loss in \cref{eq:loss-data} promotes statistical independence between the recovered source $\B{s}_1$ and the recovered noise $\B{n}$. To facilitate the optimization and to avoid having to choose weighting parameters, each loss term is normalized with respect to the standard deviation of each coefficient in $\Psi_k$.
In this manner, $\sigma^2\left( \Psi_k(\B{s}_1^i)\right)$ represents the vector of variances for each coefficient in $\Psi_k$ computed across different realizations $\B{s}_1^i$. The same applies to $\sigma^2\left( \Psi_k(\B{x}+\B{s}_1^i)\right)$ and $\sigma^2\left( \Psi_k(\B{s}_1^i,\B{x})\right)$.
We then sum the normalized loss terms to define the reconstruction $\B{\widetilde{s}}_1$ as the solution to the optimization problem:
\begin{equation}
\begin{aligned}
    \label{eq:ss_optimization}
\B{\widetilde{s}}_1 := \argmin_{\B{s}_1}\, \Big[ \mathcal{L}_{\text{data}} (\B{s}_1) + \mathcal{L}_{\text{prior}} (\B{s}_1) + \mathcal{L}_{\text{cross}} (\B{s}_1) \Big].
\end{aligned}
\end{equation}

The algorithm relies solely on constraints in the scattering spectra
space, except for the initialization $\B{\widetilde{s}}_1 = \B{x}$,
which contains valuable information about the mixture of sources in the
time domain. We describe the optimal value that the objective function
in \cref{eq:ss_optimization} converges to in \cref{app:optimal_value} of
the Appendix, which can serve as an approximate quantitative indicator
of the quality of the source separation.

\section{Results}

\noindent The purpose of our approach unsupervised multi-scale clustering and source separation framework is to address challenges associated with source separation in domains where limited expert knowledge on the sources is available and source also exhibit vastly different timescales.

We demonstrate the effectiveness of our approach on seismic data recorded by the InSight mission. The data consists of seismic records from the SEIS instrument, which has been operational on Mars since 2018. The dataset contains a variety of signals, including marsquakes, glitches, and wind imprints, each with different timescales. We aim to cluster and separate these sources using our proposed framework. We present the results of our approach in the following subsections. Additional results regarding this experiment are provided in \cref{app:all_results,app:fvae_reconstruction,app:marsquake_separation} of the Appendix. Finally, we also showcase our approach in a controlled setting through a stylized example that mimics some of the sources observed in the Mars dataset (cf. \cref{app:stylized} of the Appendix).

\subsection{Seismic records from the InSight mission}

\noindent Martian seismic ambient signals consist of signals from various sources, each with different timescales. For instance, at shorter timescales, we expect to observe transient one-sided pulses known as glitches (tens of seconds in duration) \cite{ScholzEtAl_2020}. These glitches likely stem from thermal cracks within the instrument's subsystems or atmospheric phenomena like dust devils---local low-pressure structures moving along the ambient wind (also lasting tens of seconds) \cite{BanfieldEtAl_2020}. Conversely, at longer timescales, we anticipate observing different phenomena such as regional winds whose direction and speed vary over time. These atmospheric phenomena strongly correlate with temperature, exhibiting a notable dependency on the local time at the station \cite{BanfieldEtAl_2020}.

Moreover, the InSight SEIS instrument---a three-axis instrument (separated by $120^{\circ}$) called U, V, W---has recorded several major marsquakes, which hold significant importance for the insights they offer into the Martian subsurface \cite{St_ahlerEtAl_2022}. However, due to their rarity in comparison to the aforementioned signal types (only 39 events over the course of four years \cite{Service_2023}), clustering approaches are not naturally inclined to assign a cluster to marsquake recordings. Nevertheless, as we shall demonstrate, these signals tend to concentrate in the fVAE's latent space associated with the correct timescale. This concentration serves as an indication of the rich structure of our multi-scale representation of this dataset.

\subsection{Architecture}\label{sec:architecture}

\noindent The architecture of the fVAE primarily relies on fully connected layers. The input consists of concatenated pyramidal wavelet scattering spectra, which are then passed through the joint encoder. This module is composed of a series of fully connected residual blocks, totaling four blocks. Each block includes a fully connected layer that reduces the dimensionality of the concatenated features to a hidden dimension, here set to 1024. This is followed by a Batchnorm layer and LeakyReLU nonlinearity. The output of the residual block is then brought back to the input dimensionality using another fully connected layer, and the result is added to the input (skip connection) to form the output of the residual block. Thanks to the pyramidal scattering spectra representation, the joint encoder with skip connection preserves maximal information from each scale and learns to extract useful information for generative modeling and clustering of representations.

Moving on to the per-scale encoders, the output of the joint encoder is split among different scales, and each scale is fed to a per-scale encoder. These encoders consist of compositions of four fully connected layers, Batchnorm, and LeakyReLU activations. The first layer changes the dimensionality of each scale's representation to the hidden dimension (1024), while the last layer reduces this hidden dimension to the latent dimension of 32. We parameterize the latent distribution of each scale as a Gaussian mixture model with nine components, where the mean and diagonal covariances are unknown vectors. On the decoding side, the per-scale decoders mirror the per-scale encoders and reconstruct the input multi-scale representation from the latent space.

\subsection{Training details of the factorial variational autoencoder}\label{sec:training_details}

\noindent We utilized the entire data recorded during the InSight
mission \cite{Service_2019} to train the fVAE. During training, we
randomly reserved $10\%$ of the data as validation data for tuning a set
of hyperparameters. Following the approach of \cite{BarkaouiEtAl_2021},
we employed nine clusters at each timescale (see \cref{app:all_results}
of the Appendix for a discussion on correspondence with results in
\cite{BarkaouiEtAl_2021}). Key hyperparameters—such as the number of
clusters, dimensionality of the latent space, learning rate, batch size,
and network architecture—were selected based on a combination of prior
knowledge, grid search, and empirical validation. Specifically, the fVAE
was trained using the architecture outlined above, employing a hidden
dimension of 1024 and a latent dimension of 32. The Adam optimization
algorithm \cite{KingmaEtAl_2014b} was employed with a learning rate of
$10^{-3}$, chosen via grid search for stability. A batch size of $16384$
was used to maximize parallelism within the limits of GPU memory.
Network architecture choices were guided by the complexity of the input
features and validated through the accuracy in reconstructing the input
pyramidal scattering spectra (cf.
\cref{fig:app_decoder_scale-1,fig:app_decoder_scale-2,fig:app_decoder_scale-3,fig:app_decoder_scale-4}
in \cref{app:fvae_reconstruction} of the Appendix), while avoiding
overfitting. To address non-differentiability concerns associated with
the categorical distribution learning, we utilized the Gumbel-Softmax
distribution, enabling a differentiable approximate sampling mechanism
for categorical variables \cite{JangEtAl_2017}. The initial temperature
parameter for the Gumbel-Softmax distribution was set to $1.0$, and we
exponentially decayed the temperature to a minimum value of $0.5$.
Training takes approximately 51 hours on a Tesla V100 GPU, typically
converging within this time frame as indicated by the plateauing of the
ELBO and stabilization of cluster assignments. We repeated the training
with multiple different random seeds and observed relatively good
stability; specifically, the main clusters remained consistent across
runs, though minor histogram variations were present, possibly due to
data points lying near the boundaries of different mixture components.

\subsection{Identification of clusters across different timescales}
\label{sec:clusters}

\noindent To obtain the aforementioned representation, we consider four different timescales to cover the range of timescales that the sources within the dataset of interest might exhibit. We utilize nine-component Gaussian mixture latent variables motivated by a previous study conducted by \cite{BarkaouiEtAl_2021}, where the authors provided a single-scale clustering of this dataset using Gaussian mixture models. Since our approach will provide nine clusters per timescale, we expect some of the clusters to be redundant, i.e., share a lot of similarities to one another. Based on existing studies on seismic data from the InSight mission \cite{ScholzEtAl_2020, BanfieldEtAl_2020, BarkaouiEtAl_2021}, we selected the finest timescale to be $51.2$ seconds, which is equivalent to a window size of $4^5$ samples with $20$ samples per second. This choice allows us to capture the diversity present in the one-sided pulses. To determine the subsequent timescales, we multiplied the previous timescale by a factors of four. While we do not expect to cluster broadband marsquakes \cite{CeylanEtAl_2022} due to their infrequent occurrence compared to other sources, we set the largest window size to be $54.6$ minutes, i.e., a window size of $4^8$ samples. This window size covers marsquakes and other sources related to atmospheric-surface interactions \cite{BanfieldEtAl_2020}.

\begin{figure*}[!t]
  \centering

  \begin{subfigure}[t]{0.495\textwidth}
      \centering
      \includegraphics[width=0.64\textwidth]{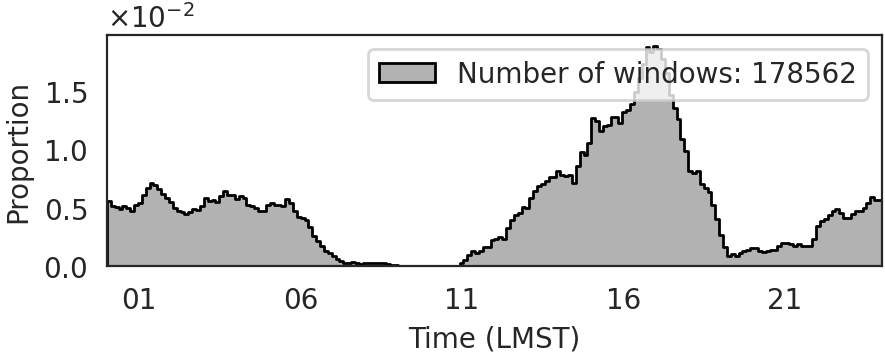}
  \end{subfigure}
  \begin{subfigure}[t]{0.495\textwidth}
    \centering
      \includegraphics[width=0.64\textwidth]{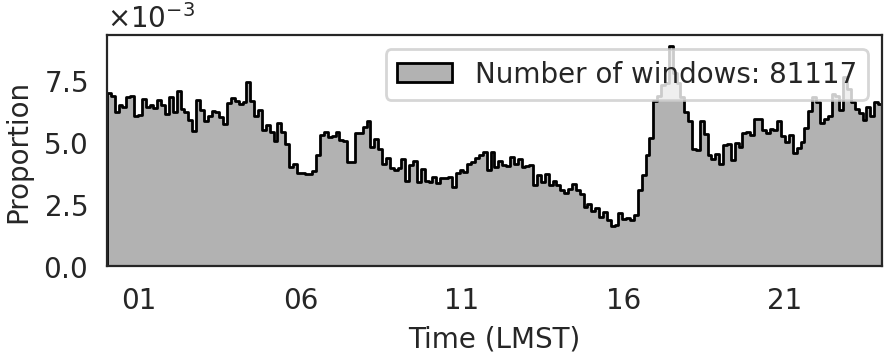}
  \end{subfigure}

  \begin{subfigure}[t]{0.495\textwidth}
      \includegraphics[width=\textwidth]{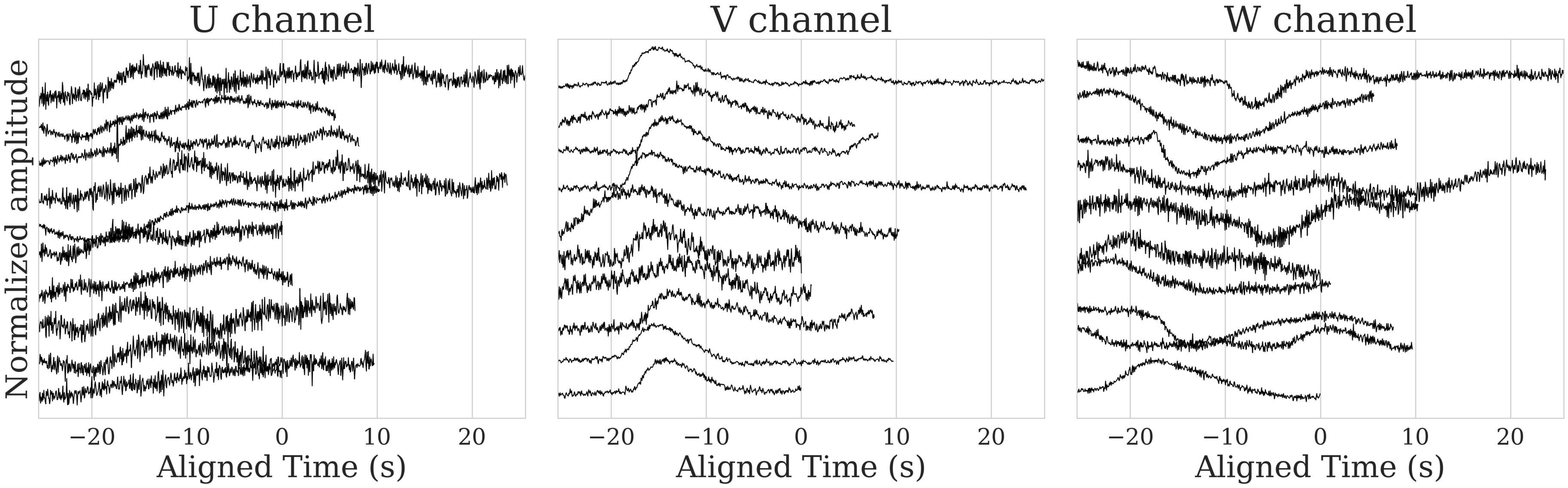}
      \caption{Glitches without precursor (51.2-second timescale)}
      \label{fig:cluster-6_scale-1}
  \end{subfigure}\hspace{0.1em}
  \begin{subfigure}[t]{0.495\textwidth}
      \includegraphics[width=\textwidth]{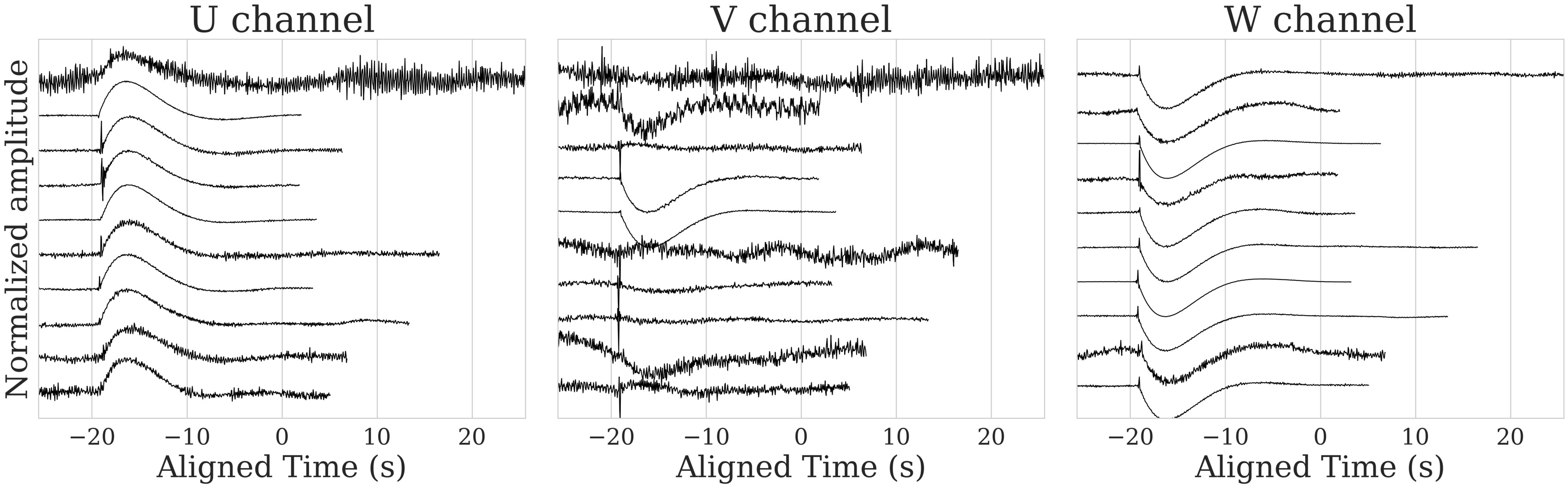}
      \caption{Glitches with precursor (51.2-second timescale)}
      \label{fig:cluster-7_scale-1}
  \end{subfigure}

  \begin{subfigure}[t]{0.495\textwidth}
      \centering
      \includegraphics[width=0.64\textwidth]{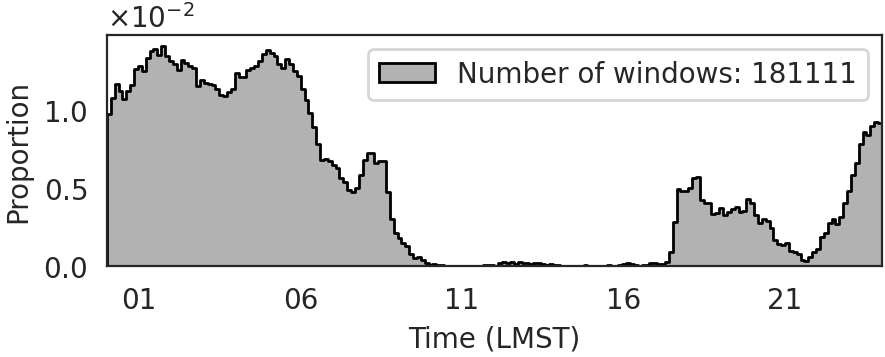}
  \end{subfigure}
  \begin{subfigure}[t]{0.495\textwidth}
      \centering
      \includegraphics[width=0.64\textwidth]{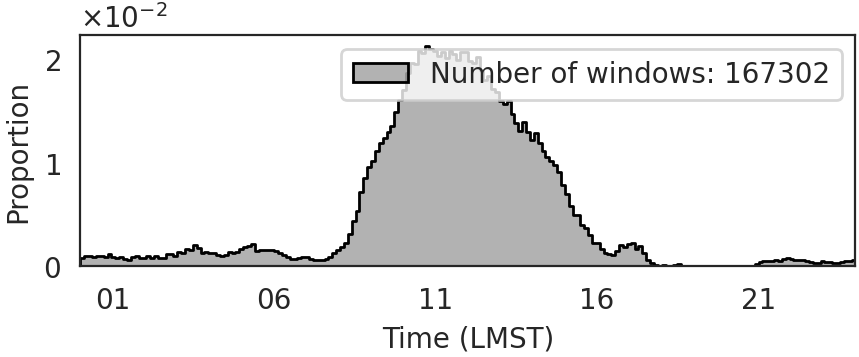}
  \end{subfigure}

  \begin{subfigure}[t]{0.495\textwidth}
      \includegraphics[width=\textwidth]{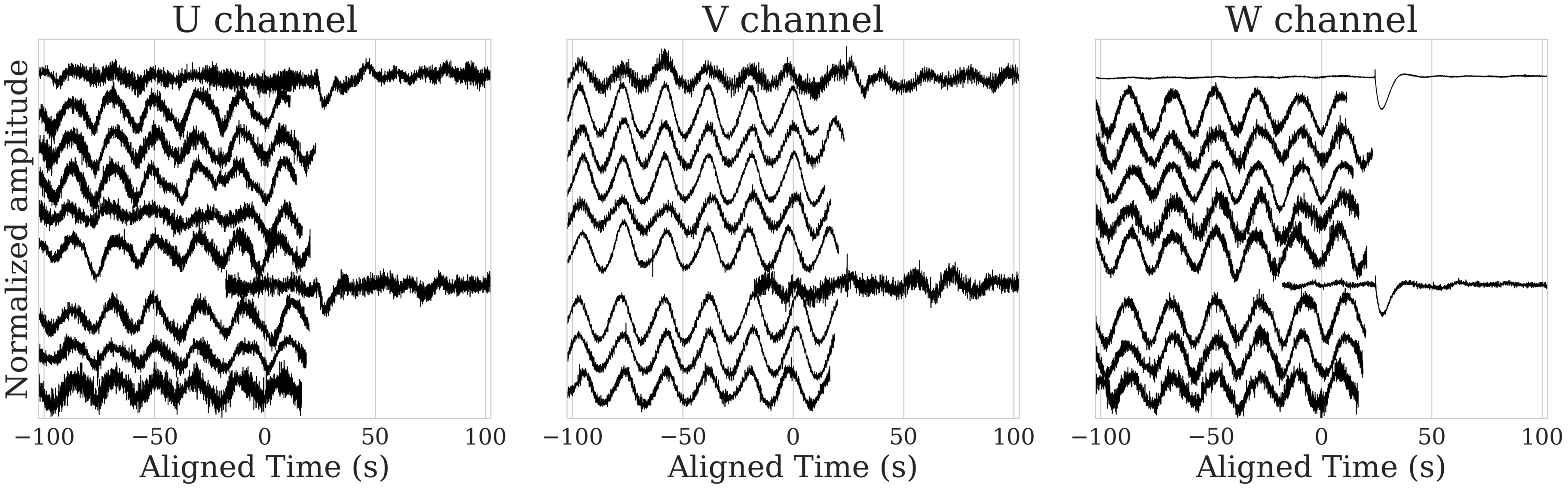}
      \caption{Oscillatory signals (3.4-minute timescale)}
      \label{fig:cluster-6_scale-2}
  \end{subfigure}\hspace{0.1em}
  \begin{subfigure}[t]{0.495\textwidth}
      \includegraphics[width=\textwidth]{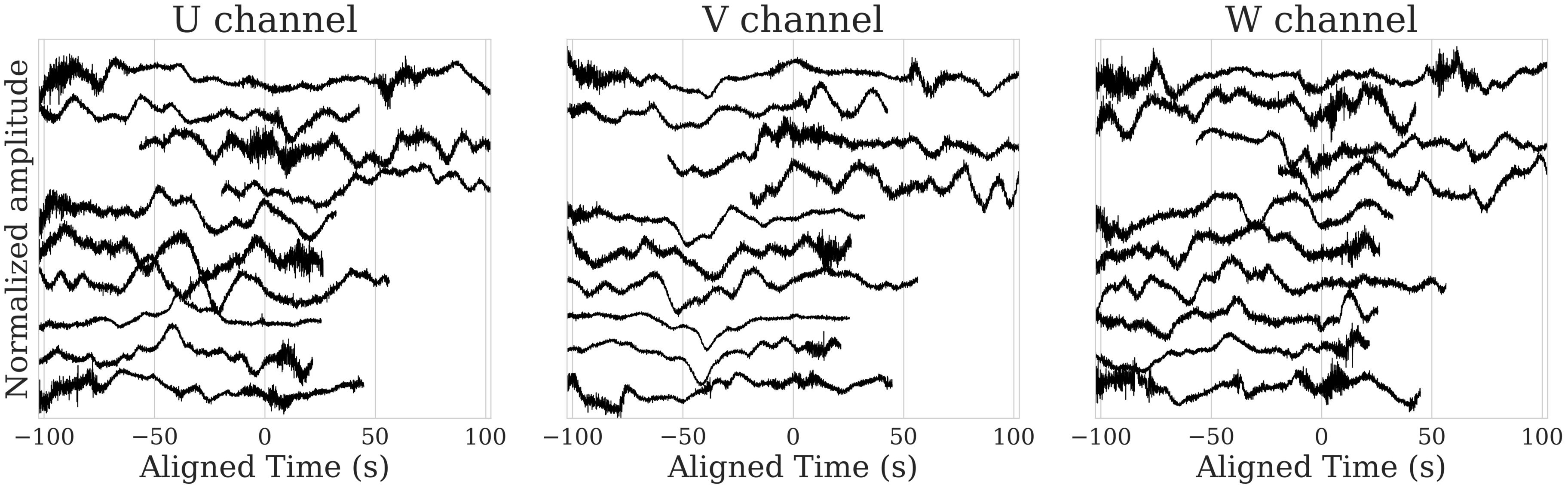}
      \caption{High-frequency wind bursts (3.4-minute timescale)}
      \label{fig:cluster-8_scale-2}
  \end{subfigure}

  \caption{The visualization of the cluster occurrence time histogram, obtained by aggregating data across the entire mission, and ten aligned waveforms of two clusters belonging to the finer timescales. The horizontal axis on the time histograms represents local mean solar time (LMST). As expected, the finest time scale is capable of distinguishing between two types of glitches: with (\cref{fig:cluster-6_scale-1}) and without (\cref{fig:cluster-7_scale-1}) a precursor. According to their occurrence time histogram (horizontal axis on the time histograms is one Martian day), both of these glitch clusters have a tendency to appear more frequently around the Martian sunset. However, the glitch cluster without a precursor, seems to be more localized in time. In the $3.4$-minute timescale, clusters are characterized by longer timescales. The cluster in \cref{fig:cluster-6_scale-2} exhibits an oscillatory signal with an approximate 25-second period, predominantly occurring during Martian night and the cluster in \cref{fig:cluster-8_scale-2} shows bursts of high-frequency, rapidly dissipating energy.}
  \label{fig:clusters_short}
\end{figure*}

\begin{figure*}[!t]
  \centering

  \begin{subfigure}[t]{0.495\textwidth}
      \centering
      \includegraphics[width=0.64\textwidth]{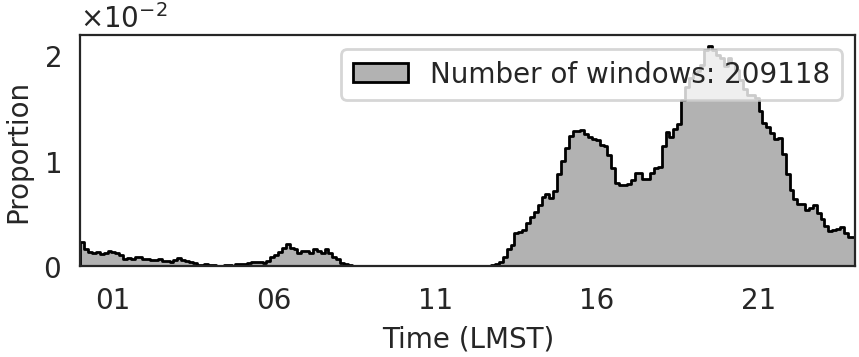}
  \end{subfigure}
  \begin{subfigure}[t]{0.495\textwidth}
      \centering
      \includegraphics[width=0.64\textwidth]{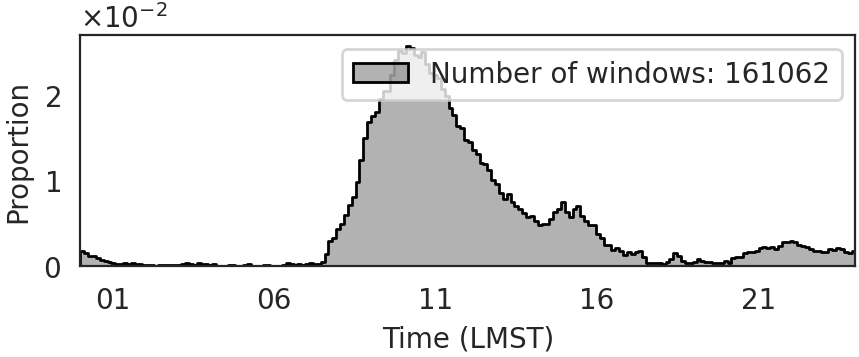}
  \end{subfigure}

  \begin{subfigure}[t]{0.495\textwidth}
      \includegraphics[width=\textwidth]{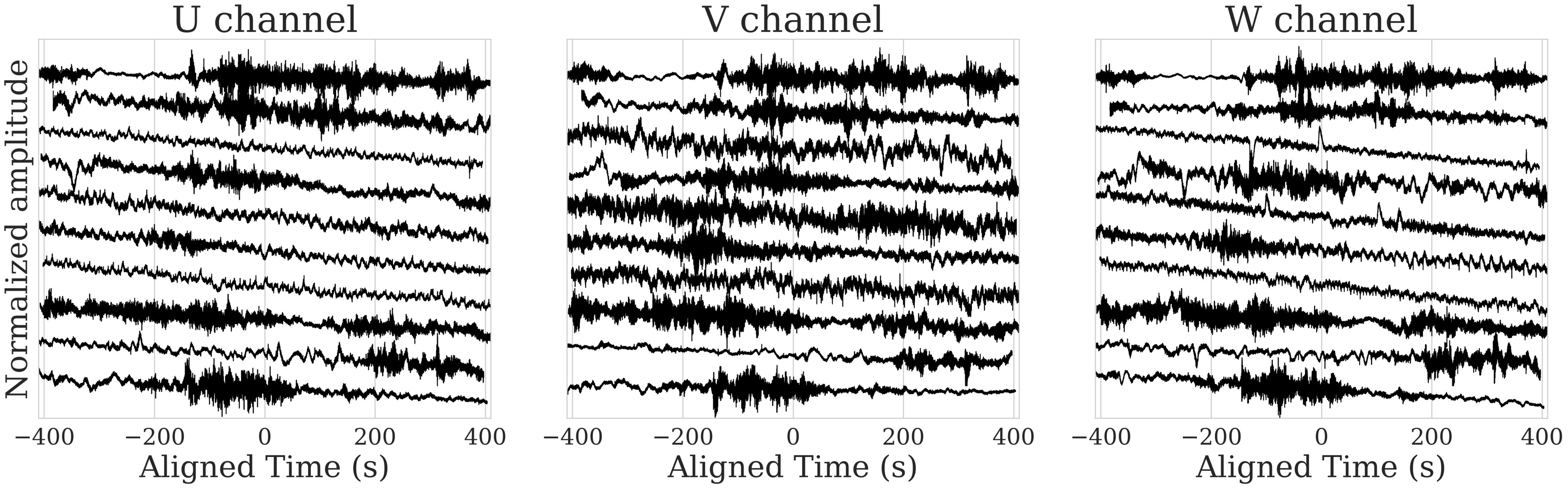}
      \caption{Wind gusts with ringing oscillations (13.6-minute timescale)}
      \label{fig:cluster-2_scale-3}
  \end{subfigure}\hspace{0.1em}
  \begin{subfigure}[t]{0.495\textwidth}
      \includegraphics[width=\textwidth]{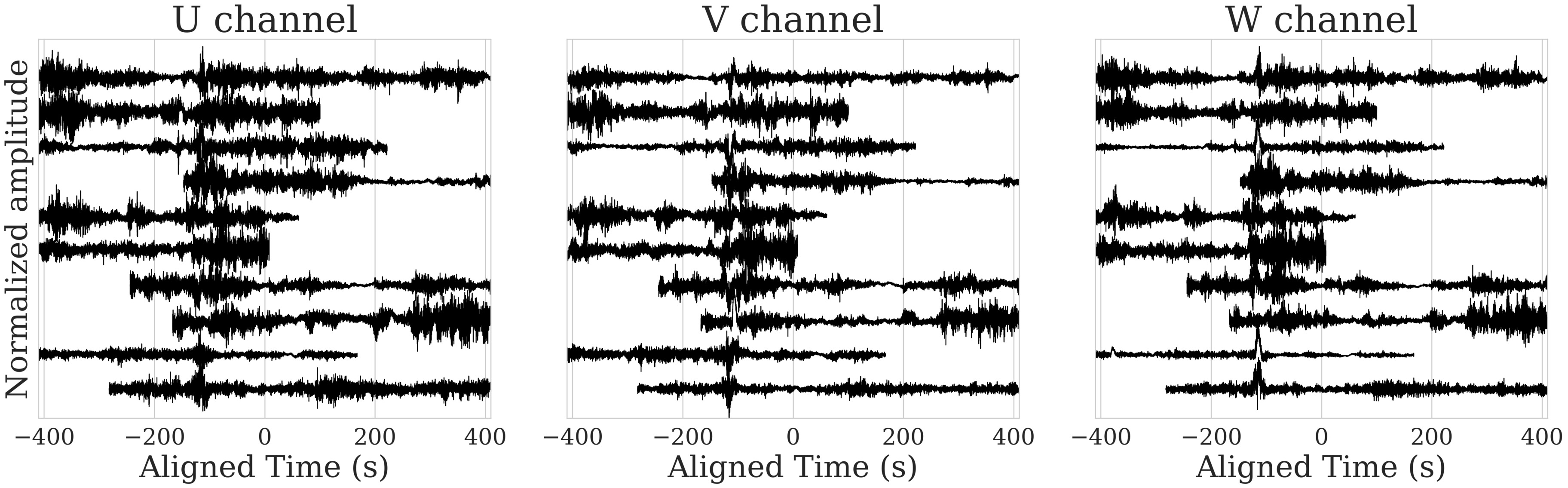}
      \caption{Dust devil signatures (13.6-minute timescale)}
      \label{fig:cluster-6_scale-3}
  \end{subfigure}

  \begin{subfigure}[t]{0.495\textwidth}
      \centering
      \includegraphics[width=0.64\textwidth]{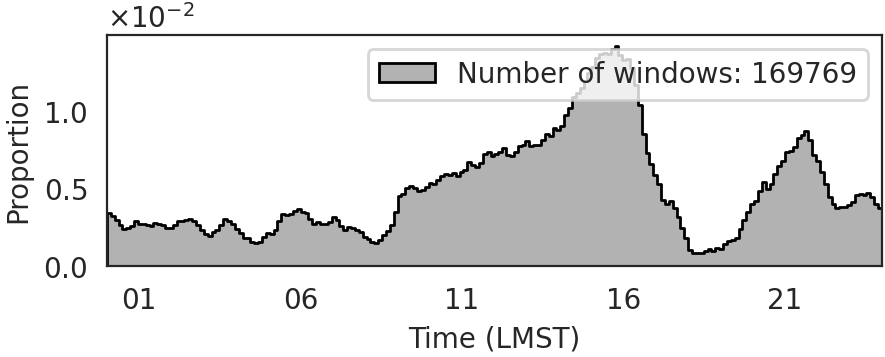}
  \end{subfigure}
  \begin{subfigure}[t]{0.495\textwidth}
      \centering
      \includegraphics[width=0.64\textwidth]{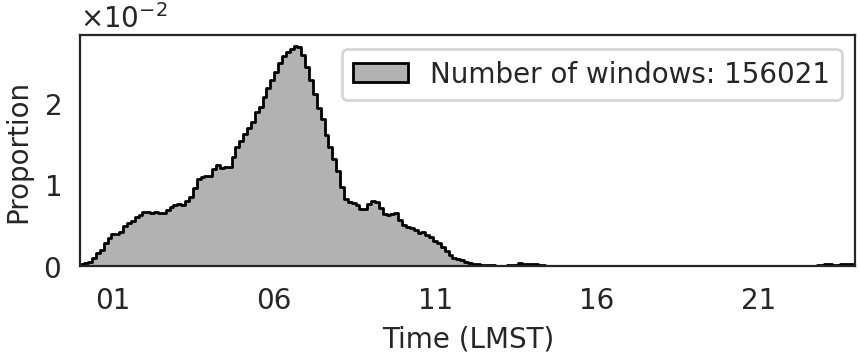}
  \end{subfigure}

  \begin{subfigure}[t]{0.495\textwidth}
      \includegraphics[width=\textwidth]{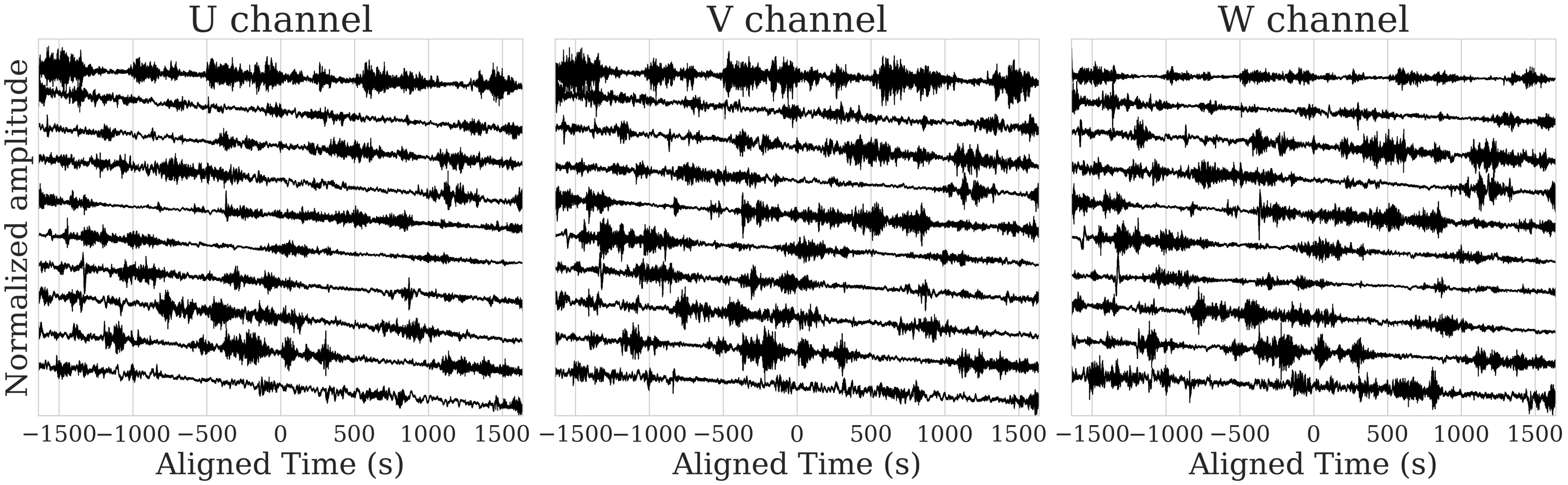}
      \caption{Sustained wind events (54.6-minute timescale)}
      \label{fig:cluster-0_scale-4}
  \end{subfigure}\hspace{0.1em}
  \begin{subfigure}[t]{0.495\textwidth}
      \includegraphics[width=\textwidth]{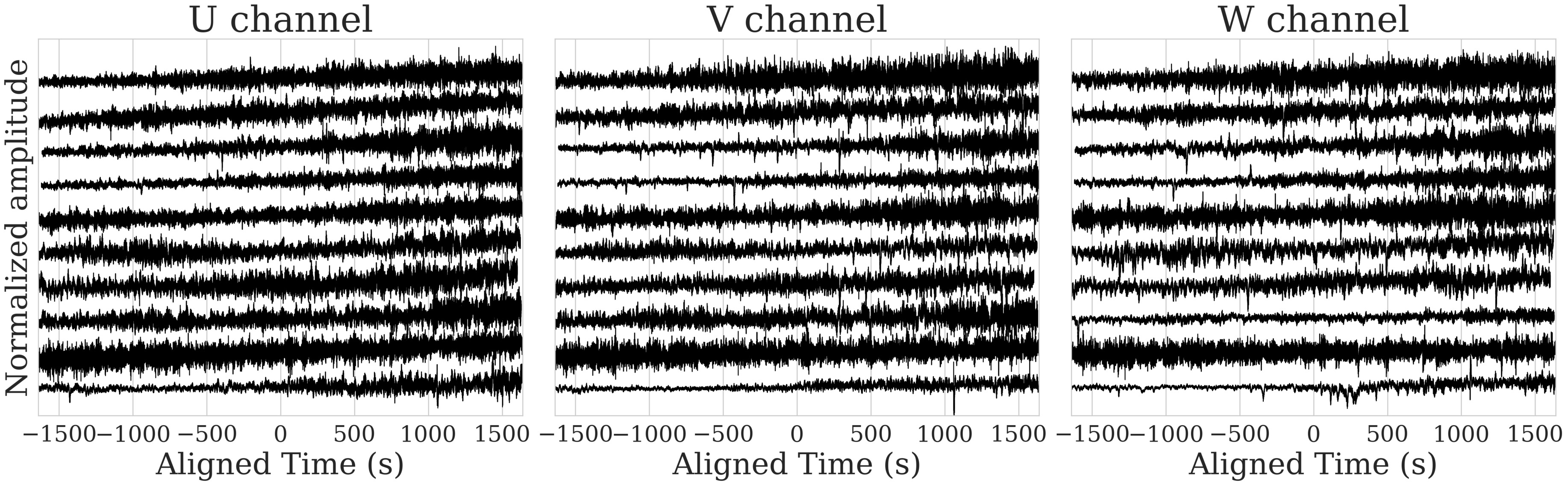}
      \caption{Sunrise atmospheric interactions (54.6-minute timescale)}
      \label{fig:cluster-1_scale-4}
  \end{subfigure}

  % \vspace{0.5em}
  \caption{The visualization of the cluster occurrence time histogram, obtained by aggregating data across the entire mission, and ten aligned waveforms of two clusters belonging to the coarser timescales. The horizontal axis on the time histograms represents local mean solar time (LMST). When focusing on coarser timescales, we start to see events that were not captured in the fine timescale clusters. Notably, \cref{fig:cluster-2_scale-3,fig:cluster-0_scale-4} show some characteristic waveforms with a sharp onset and a following ringing oscillations. Such waveforms are observed when a strong wind gust is blowing. This is also consistent with the occurrence time histogram of these clusters, which are localized in time \cite{BanfieldEtAl_2020}. \Cref{fig:cluster-1_scale-4} shows a cluster of waveform dominated by sunrise-related surface-atmospheric interactions. This can be inferred based on the cluster's occurrence time histogram as well as the gradual increase in waveform amplitude, which is related to the higher ambient seismic noise during the Martian daytime.}
  \label{fig:clusters_long}
\end{figure*}

While further investigation will be necessary to fully uncover the nature of the different clusters detected by our study, the results we obtained already demonstrate that the multi-scale approach successfully distinguishes the various phenomena observed in the Martian data. We provide information on all identified clusters within the four timescales considered in \cref{app:all_results} of the Appendix and present the notable clusters in this section. Specifically, \cref{fig:clusters_short,fig:clusters_long} illustrate two noteworthy identified clusters for the finer ($51.2$ seconds and $3.4$ minutes) and coarser ($13.6$ minutes and $54.6$ minutes) timescales, respectively. To visualize clusters across the four timescales, we depict both the cluster occurrence time histogram along with ten aligned waveforms from that cluster. The cluster occurrence time histogram, i.e., the distribution of windows belonging to that cluster throughout the entire mission, will allow us to identify what times of the day a cluster may exhibit more occurrence, which in turn might reveal clusters that contain atmosphere-surface interaction signals. The aligned waveforms, on the other hand, are useful in identifying the main feature of each cluster. We adopt the approach outlined in \cite{BarkaouiEtAl_2021} to generate the aligned waveforms. This procedure entails calculating, for each cluster within each timescale, the Pearson correlation coefficient between the most probable three component waveform assigned to that cluster (identified via the fVAE) and the remaining three components waveforms in that cluster, adjusted for lag to maximize correlation. Finally, we present the most probable waveform with each component plotted separately, along with nine other aligned waveforms (based on the lag producing the maximum correlation) for the waveforms exhibiting the highest correlation.

\paragraph{51.2-second timescale} At the finest timescale, we identified two clusters where the primary features in both of the clusters are glitches (see \cref{fig:cluster-6_scale-1,fig:cluster-7_scale-1}). The glitches in \cref{fig:cluster-7_scale-1} contain a precursor, i.e., a high amplitude spike right before the glitch, and while not being restricted to a time interval during one Martian day, they tend to be observed more frequently close to the sunset. This observation, albeit for glitches in general and not just glitches with precursors, has been independently made in prior work \cite{BarkaouiEtAl_2021}, which further confirms the nature of this cluster. Note that while the glitches in \cref{fig:cluster-7_scale-1} uniformly contain one-sided pulses with increasing amplitude in the U components and decreasing amplitudes in the V and W components, this is due to visualizing aligned waveforms and we do indeed detect other combinations in this cluster. Cluster 6 of this timescale \cref{fig:cluster-6_scale-1}, however, contains glitches without a precursor, has been detected in almost twice as many windows as cluster 7 (glitches with precursor, \cref{fig:cluster-7_scale-1}), and has a significant peak around the sunset.

\paragraph{3.4-minute timescale} In the this timescale, the main feature of the clusters is often longer in timescale.
For example, we obtain a distinct cluster that contains an oscillatory signal with an approximate $25$-second period (see \cref{fig:cluster-6_scale-2}). This cluster appears to mostly occur during Martian night, however, the oscillatory signal might also be present during the day but is drowned in high amounts of ambient daytime noise. No clear root cause for this oscillatory signal has yet been identified, but the analyses carried out show that there is no correlation with the pressure drops, but one has been noted with the wind. For the moment, any instrumental origin seems to have been ruled out. Note that at least two of the aligned waveforms in \cref{fig:cluster-6_scale-2} also contain glitches, but a closer look reveals the existence of the oscillatory signal. In addition, we identify a cluster that contains one or more bursts of high-frequency, quickly dissipating energy (see \cref{fig:cluster-8_scale-2}). It is worth noting that due to the high occurrence rate of glitches, the clusters identified in coarser times scales often also include glitches. However, except for the multi-glitch clusters, we argue glitches in coarser timescales are not the main characterizing feature of the clusters according to the aligned waveform plots.

\paragraph{13.6-minute timescale} While in this timescale we still obtain clusters dominated by glitches, we also see clusters associated with surface-atmosphere interactions to emerge. The noteworthy clusters are presented in \cref{fig:cluster-2_scale-3,fig:cluster-6_scale-3}. The cluster in \cref{fig:cluster-2_scale-3} clearly contains waveforms with a sharp onset and a following ringing oscillations. This feature is observed when a strong wind gust blows. Note that the length of this feature is longer than the high-frequency events with quickly dissipating energy in \cref{fig:cluster-8_scale-2}. This might suggest that the winds captured in this cluster seem to be sustained longer. Interestingly, the cluster occurrence time histogram of this cluster in  \cref{fig:cluster-8_scale-2} suggests that these winds occur mostly before and after sunset. Another cluster that we observe in this timescale in shown in \cref{fig:cluster-6_scale-3}. The waveforms associated with this cluster seem to have a distinct ``noisy'' pulse in the middle, with high-frequency, high-amplitude waveforms before and after it. These might suggest a cluster of dust devils. The occurrence time histogram of this cluster also consistent with the time interval that dust devils are more frequently observed (09:00 to 15:00 LMST), which is confirmed by the InSight mission pressure sensors \cite{ChatainEtAl_2021} as well as through the Pathfinder lander \cite{FerriEtAl_2003}.

\paragraph{54.6-minute timescale} When focusing on the largest timescale, we again observe larger scale phenomena that were not captured in fine-scale clusters (recall \cref{fig:clusters_short}). Notably, a wind cluster, as depicted in \cref{fig:cluster-0_scale-4}, exhibits similar characteristic waveforms to the wind cluster in the $13.6$-minute timescale \cref{fig:cluster-2_scale-3}, such as a sharp onset followed by ringing oscillations. However, upon examining the time histograms in \cref{fig:cluster-2_scale-3,fig:cluster-0_scale-4}, it becomes evident that the cluster associated with \cref{fig:cluster-2_scale-3} indicates a more significant dip after sunset compared to the other. Knowing that the wind speeds consistently decrease to very low levels approximately 2-4 hours after sunset \cite{BanfieldEtAl_2020}, the significant dip in the time histogram of  \cref{fig:cluster-0_scale-4} can be due to the fact that there are fewer $54.6$-minute long windows during this period that do not coincide with the low-wind interval compared to the number of $13.6$-minute long windows. Finally, the cluster in \cref{fig:cluster-1_scale-4} contains atmospheric signals related to the sunrise based on the cluster's occurrence time histogram and the gradual increase in waveform amplitude, which is expected as the ambient noise recorded via the SEIS instrument during the day is consistently higher.

\begin{figure*}[!t]
  \centering

  \begin{subfigure}[t]{0.35\textwidth}
      \includegraphics[width=\textwidth]{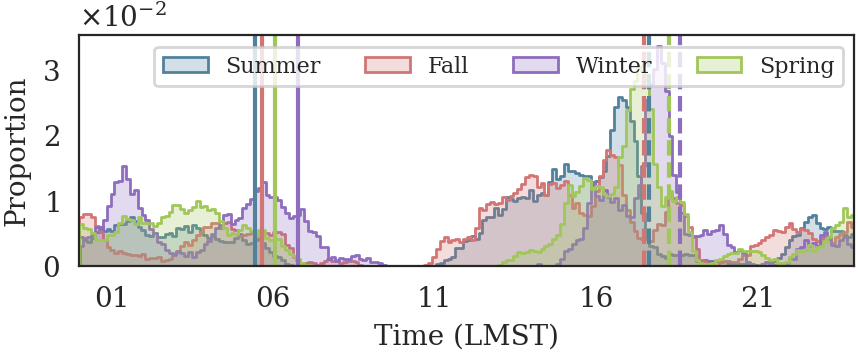}
      \caption{Glitches without precursor (51.2s timescale)}
      \label{fig:cluster-6_scale-1_season}
  \end{subfigure}\hspace{3em}
  \begin{subfigure}[t]{0.35\textwidth}
      \includegraphics[width=\textwidth]{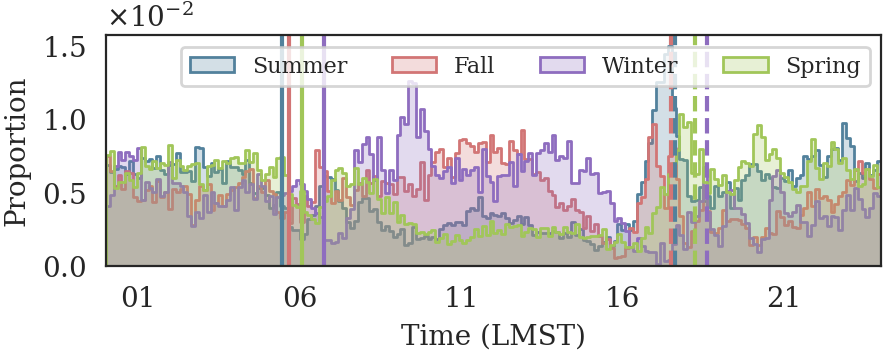}
      \caption{Glitches with precursor (51.2s timescale)}
      \label{fig:cluster-7_scale-1_season}
  \end{subfigure}

  \begin{subfigure}[t]{0.35\textwidth}
      \includegraphics[width=\textwidth]{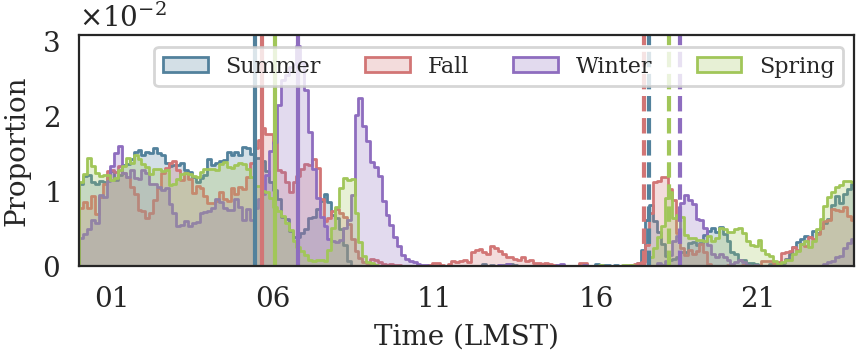}
      \caption{Oscillatory signals (3.4m timescale)}
      \label{fig:cluster-6_scale-2_season}
  \end{subfigure}\hspace{3em}
  \begin{subfigure}[t]{0.35\textwidth}
      \includegraphics[width=\textwidth]{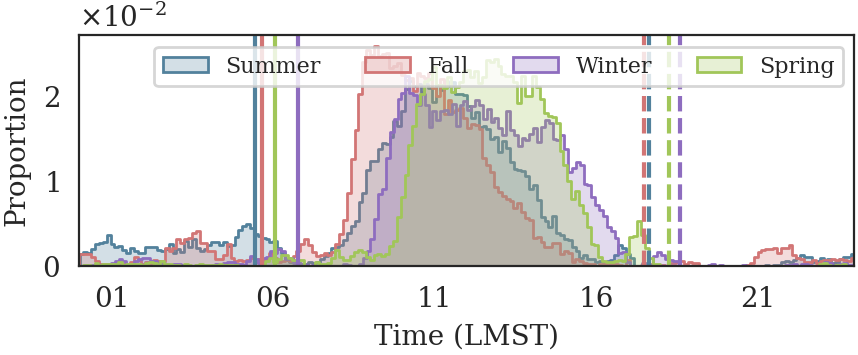}
      \caption{High-frequency wind bursts (3.4m timescale)}
      \label{fig:cluster-8_scale-2_season}
  \end{subfigure}

  \begin{subfigure}[t]{0.35\textwidth}
      \includegraphics[width=\textwidth]{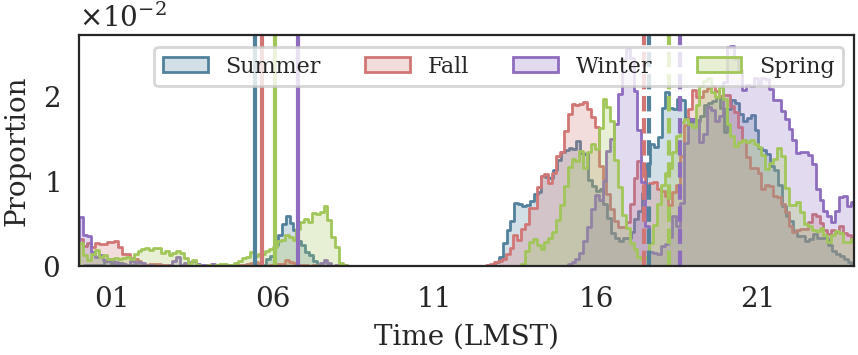}
      \caption{Wind gusts with ringing (13.6m timescale)}
      \label{fig:cluster-2_scale-3_season}
  \end{subfigure}\hspace{3em}
  \begin{subfigure}[t]{0.35\textwidth}
      \includegraphics[width=\textwidth]{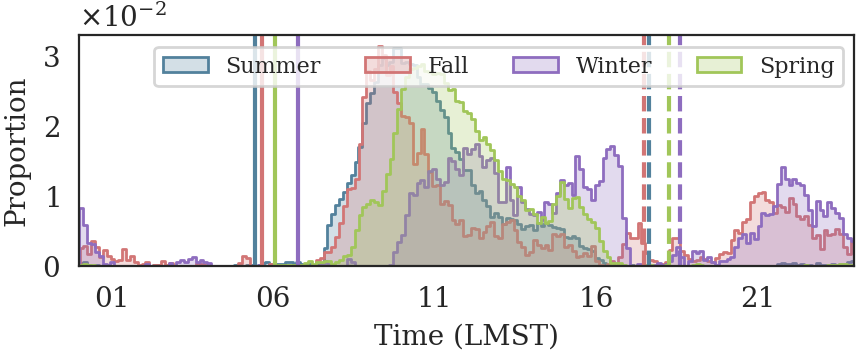}
      \caption{Dust devil signatures (13.6m timescale)}
      \label{fig:cluster-6_scale-3_season}
  \end{subfigure}

  \begin{subfigure}[t]{0.35\textwidth}
      \includegraphics[width=\textwidth]{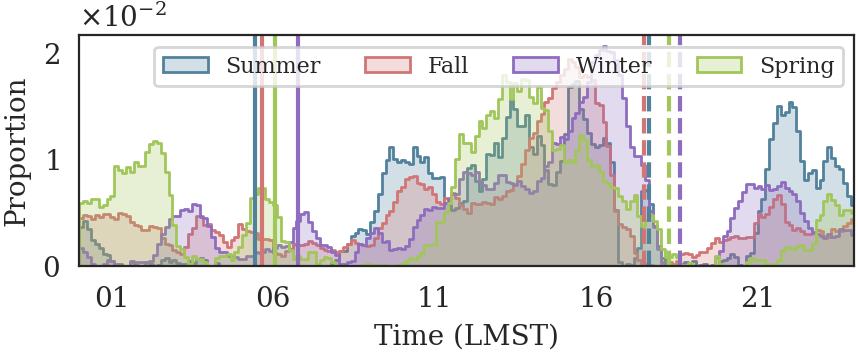}
      \caption{Sustained wind events (54.6m timescale)}
      \label{fig:cluster-0_scale-4_season}
  \end{subfigure}\hspace{3em}
  \begin{subfigure}[t]{0.35\textwidth}
      \includegraphics[width=\textwidth]{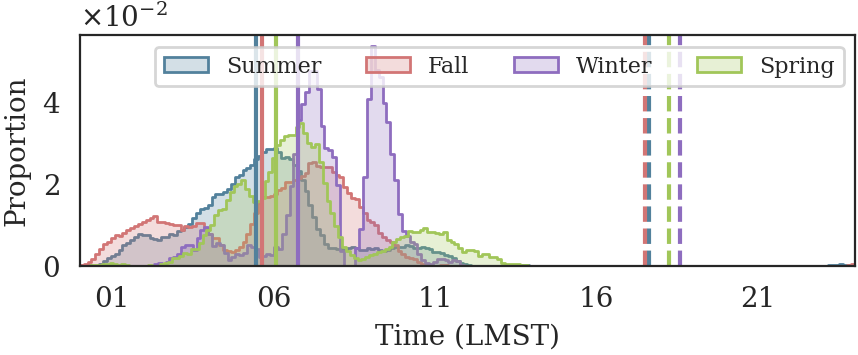}
      \caption{Sunrise-related interactions (54.6m timescale)}
      \label{fig:cluster-1_scale-4_season}
  \end{subfigure}

  % \vspace{0.5em}
  \caption{An overlay of occurrence time histograms of selected clusters within four timescales where time histograms are obtained by aggregating data from Martian local spring (green), summer (blue), fall (red), and winter (purple). The horizontal axis on the time histograms represents local mean solar time (LMST). Each time histogram is associated with the clusters presented in \cref{fig:clusters_short,fig:clusters_long} where each row represents a distinct timescale, with timescale increasing from top to bottom. The vertical and dashed lines correspond to average sunrise and sunset times over the associated season, respectively. Clusters with Martian daytime occurrences exhibited a tendency to shift in their peaks. Additionally, some clusters exhibit slight changes in their distribution, mainly involving a reduction or increase of their peaks as seasons change.}
  \label{fig:season_all_clusters}
\end{figure*}

\subsection{Seasonal impacts on surface-atmosphere interactions}

\noindent To gain deeper insights into atmospheric-surface interactions, we investigate the impact of seasonal changes on the cluster occurrence time histograms. Using the clustered windows from all timescales, we computed occurrence time histograms for each cluster using aggregated data from all four seasons on Mars. We superimposed the per-season occurrence time histograms for all the clusters presented in \cref{fig:clusters_short,fig:clusters_long}, and overlay the per-season average sunrise (solid lines) and sunset (dashed lines) times to facilitate the interpretation of the histograms. \Cref{fig:season_all_clusters} summarizes the results.

\begin{figure*}[!t]
  \centering
  \begin{subfigure}[t]{0.325\textwidth}
      \includegraphics[width=\textwidth]{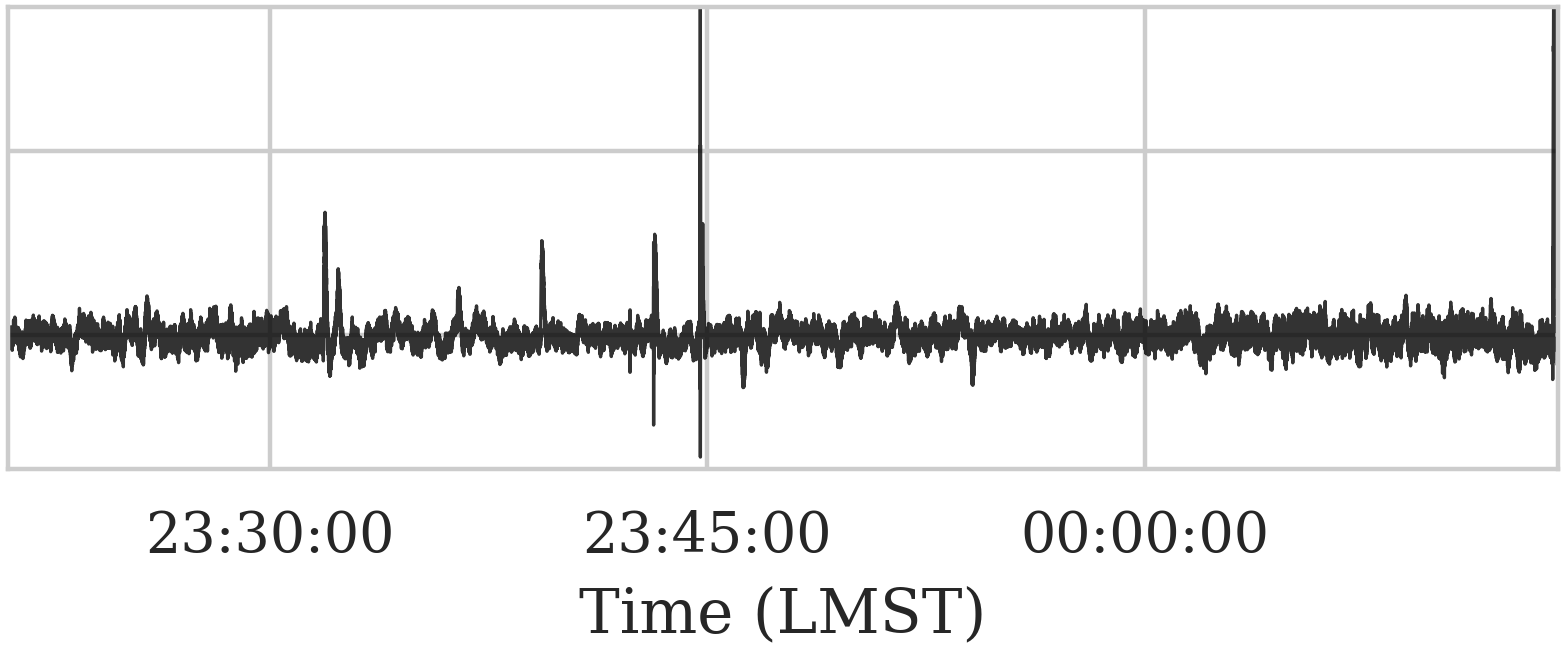}
      \caption{Raw mixed signal with glitches}
      \label{fig:source_separation_glitch_real}
  \end{subfigure}\hspace{0em}
  \begin{subfigure}[t]{0.325\textwidth}
      \includegraphics[width=\textwidth]{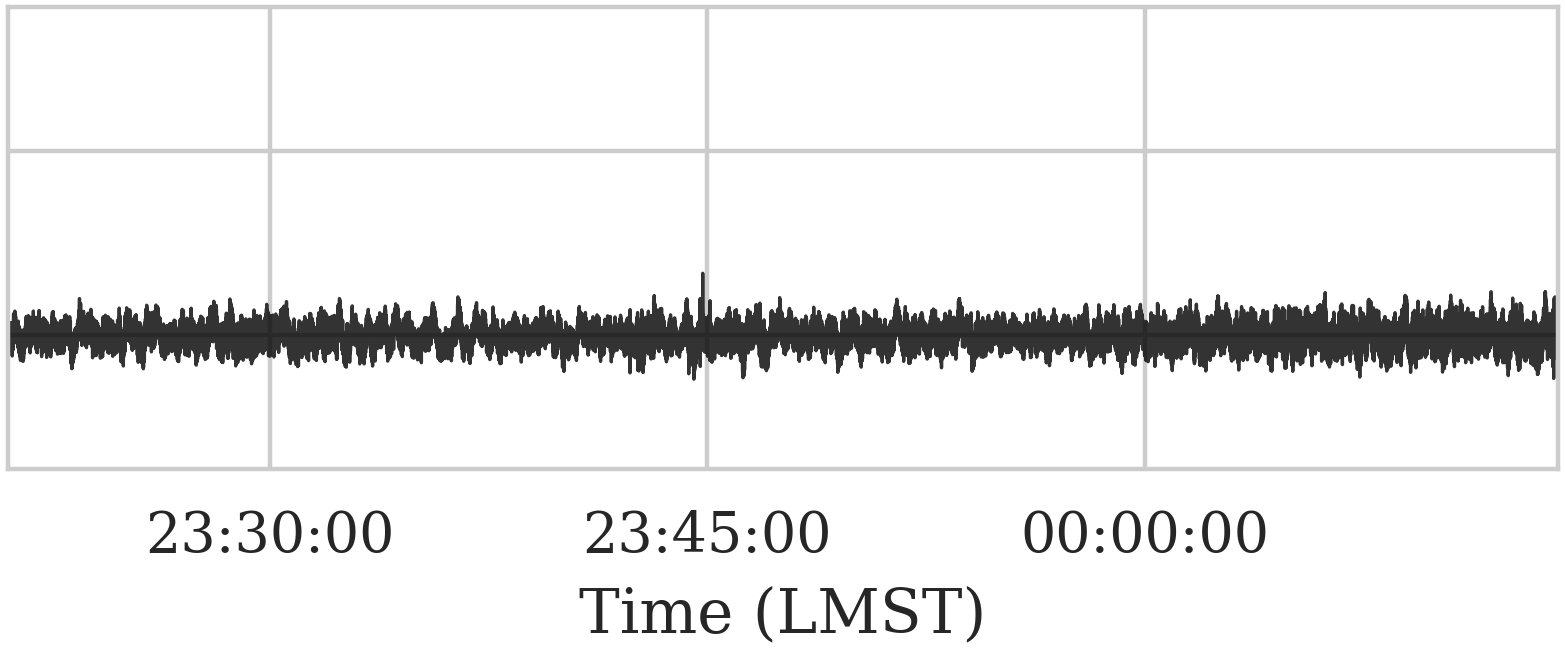}
      \caption{Background signal after glitch separation}
      \label{fig:source_separation_glitch_background}
  \end{subfigure}\hspace{0em}
  \begin{subfigure}[t]{0.325\textwidth}
      \includegraphics[width=\textwidth]{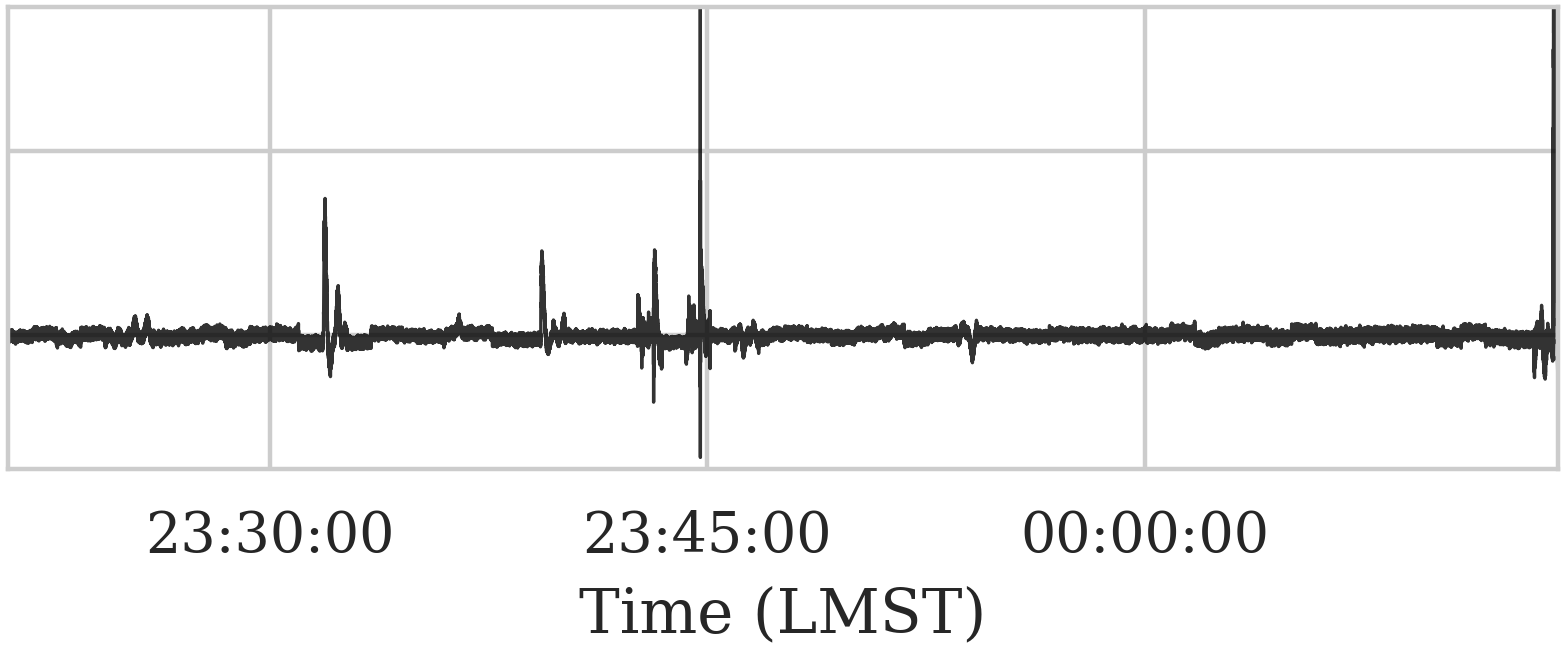}
      \caption{Extracted glitch signals}
      \label{fig:source_separation_glitch_source}
  \end{subfigure}

  \begin{subfigure}[t]{0.325\textwidth}
      \includegraphics[width=\textwidth]{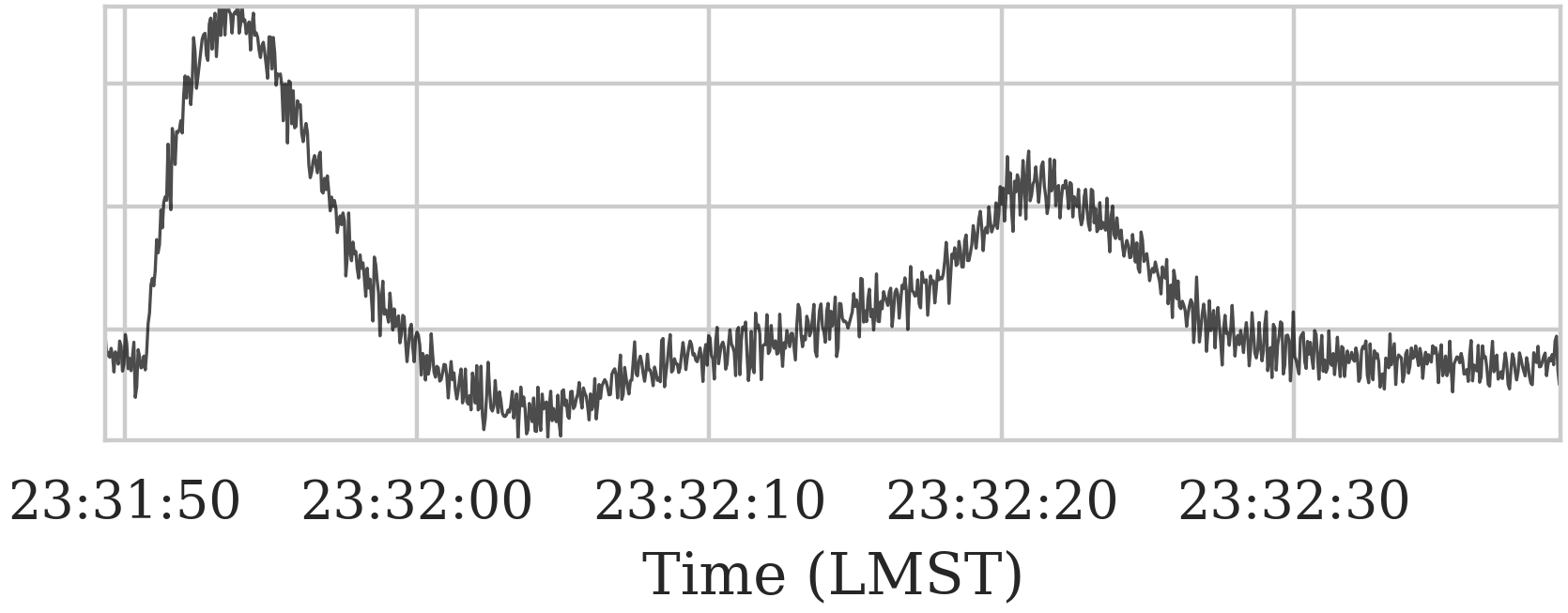}
      \caption{Multiple glitches (raw)}
      \label{fig:source_separation_glitch_real_zoom_13}
  \end{subfigure}\hspace{0em}
  \begin{subfigure}[t]{0.325\textwidth}
      \includegraphics[width=\textwidth]{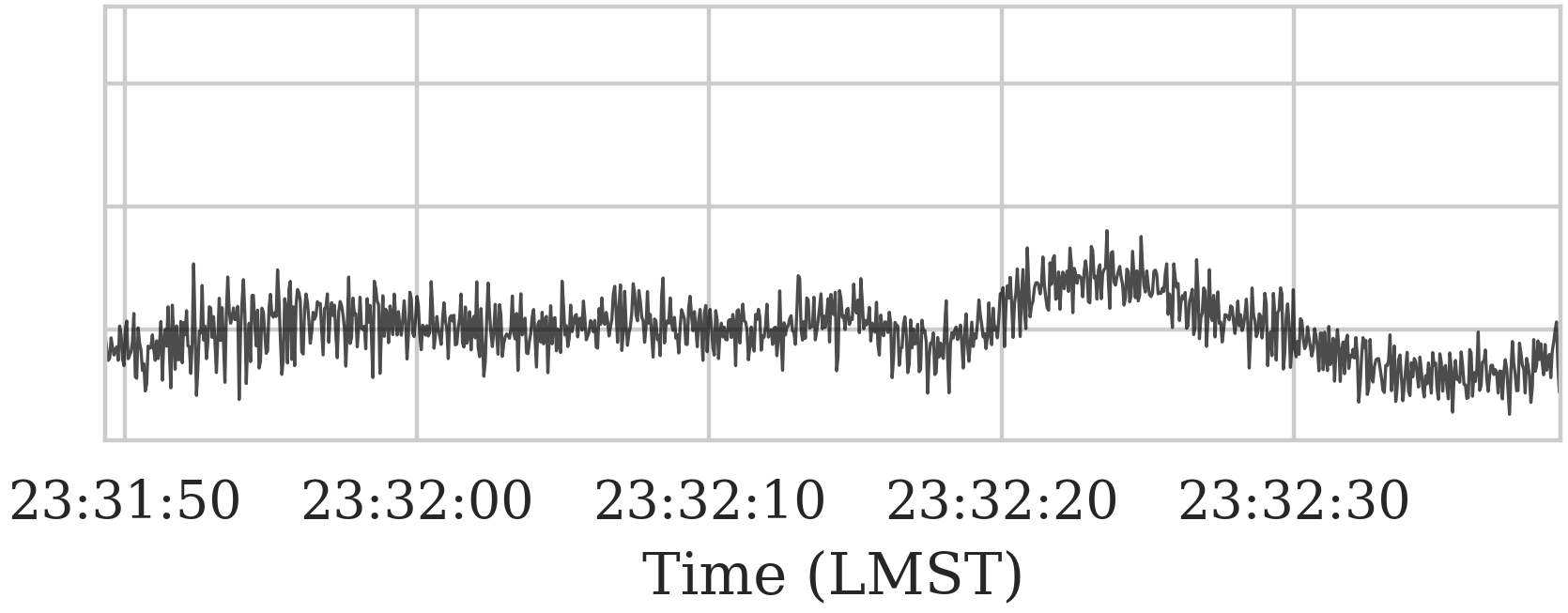}
      \caption{Background after separation}
      \label{fig:source_separation_glitch_background_zoom_13}
  \end{subfigure}\hspace{0em}
  \begin{subfigure}[t]{0.325\textwidth}
      \includegraphics[width=\textwidth]{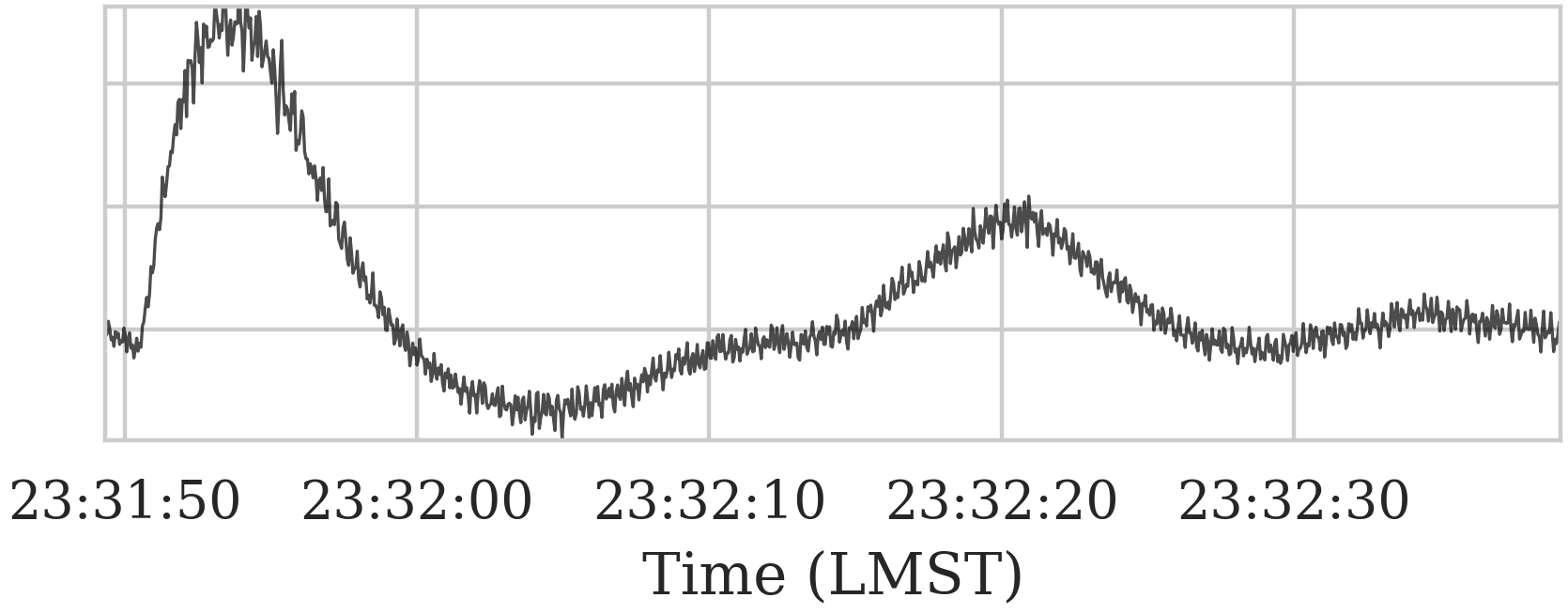}
      \caption{Separated multiple glitches}
      \label{fig:source_separation_glitch_source_zoom_13}
  \end{subfigure}

  \begin{subfigure}[t]{0.325\textwidth}
      \includegraphics[width=\textwidth]{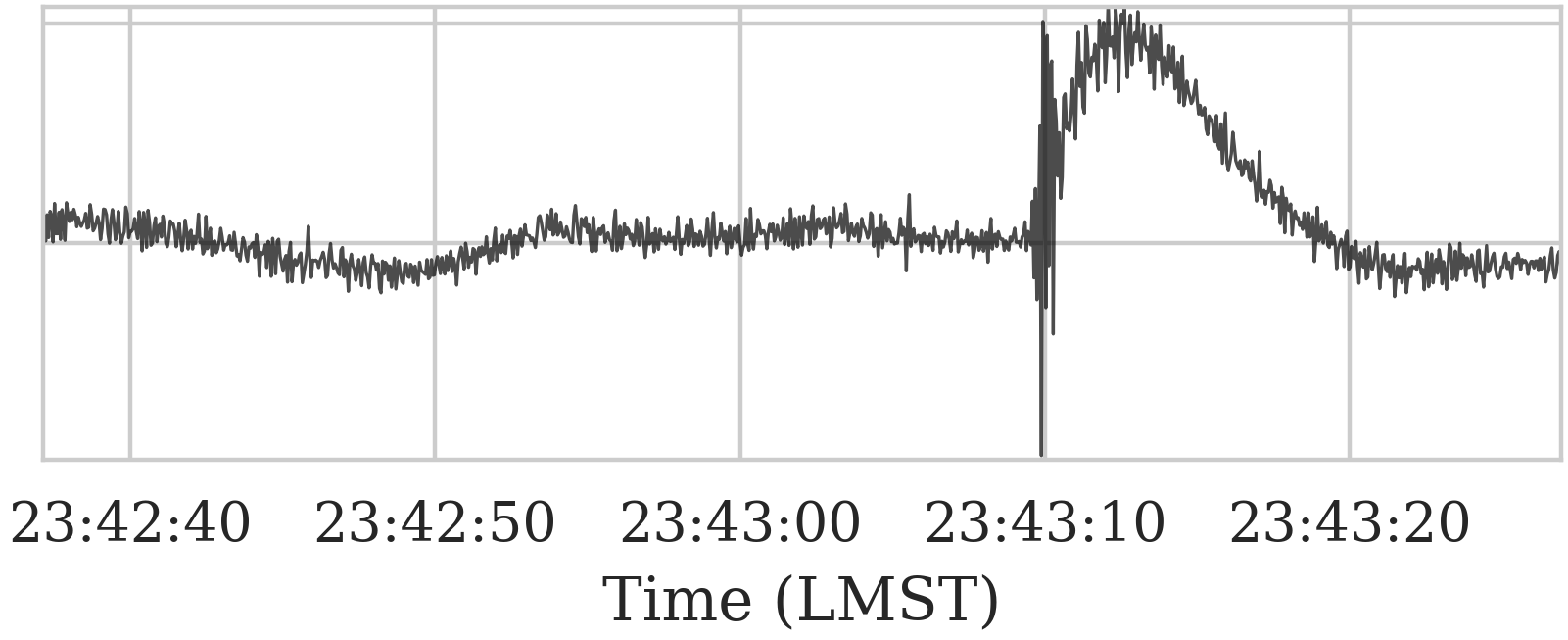}
      \caption{Glitch with precursor (raw)}
      \label{fig:source_separation_glitch_real_zoom_26}
  \end{subfigure}\hspace{0em}
  \begin{subfigure}[t]{0.325\textwidth}
      \includegraphics[width=\textwidth]{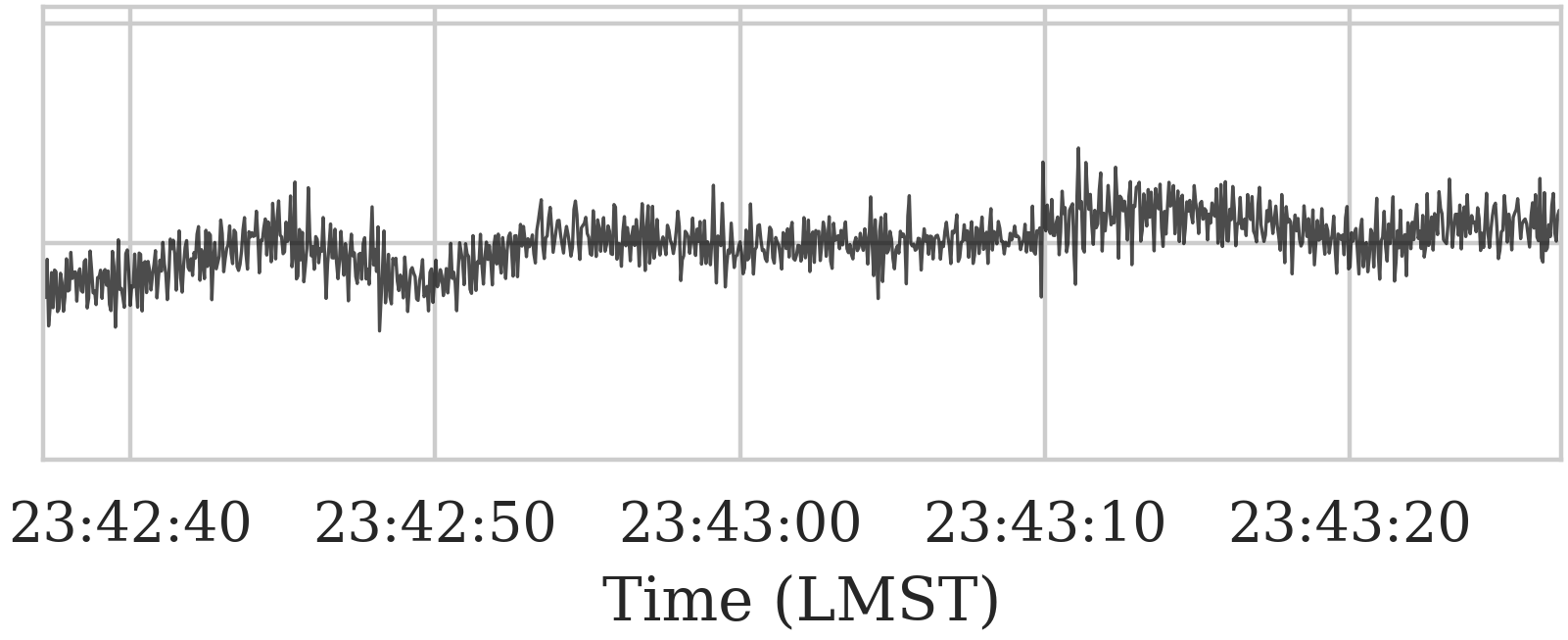}
      \caption{Background after separation}
      \label{fig:source_separation_glitch_background_zoom_26}
  \end{subfigure}\hspace{0em}
  \begin{subfigure}[t]{0.325\textwidth}
      \includegraphics[width=\textwidth]{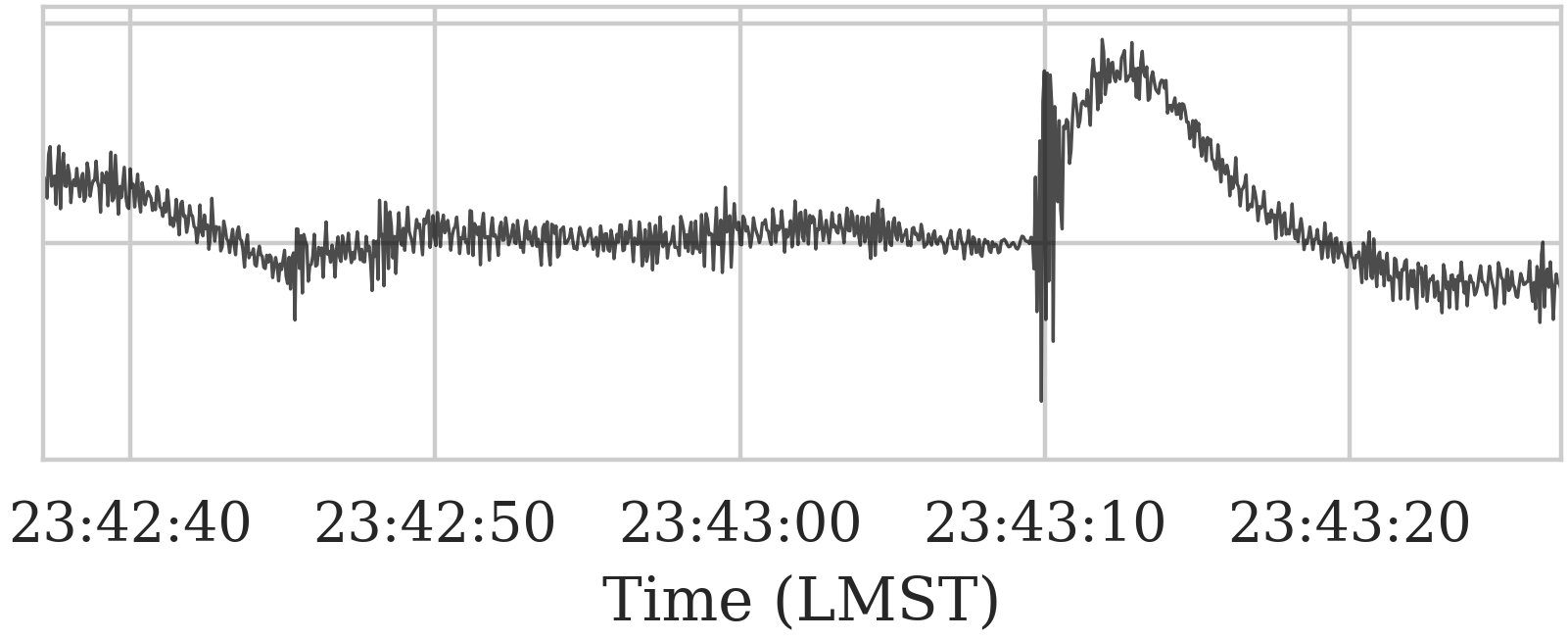}
    \caption{Separated glitch with precursor}
      \label{fig:source_separation_glitch_source_zoom_26}
  \end{subfigure}

  \begin{subfigure}[t]{0.325\textwidth}
      \includegraphics[width=\textwidth]{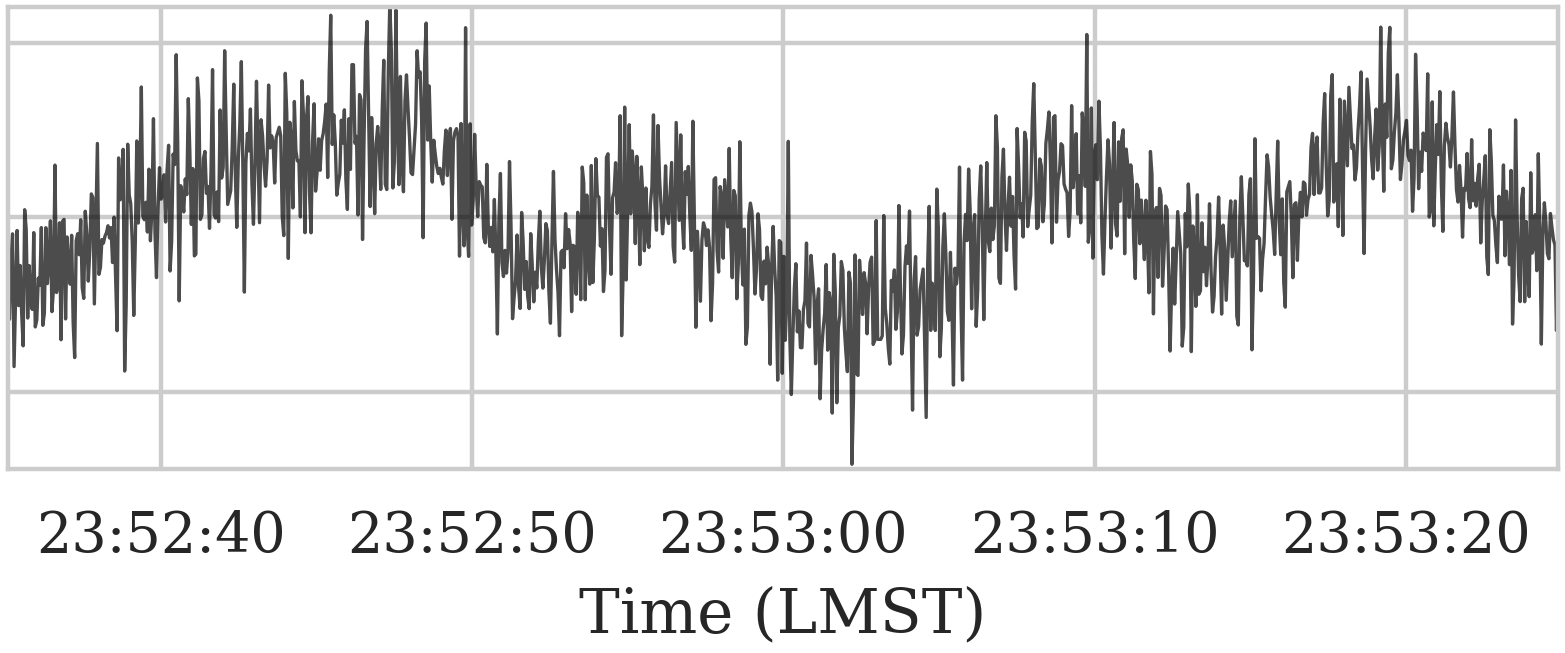}
      \caption{Clean background region (raw)}
      \label{fig:source_separation_glitch_real_zoom_38}
  \end{subfigure}\hspace{0em}
  \begin{subfigure}[t]{0.325\textwidth}
      \includegraphics[width=\textwidth]{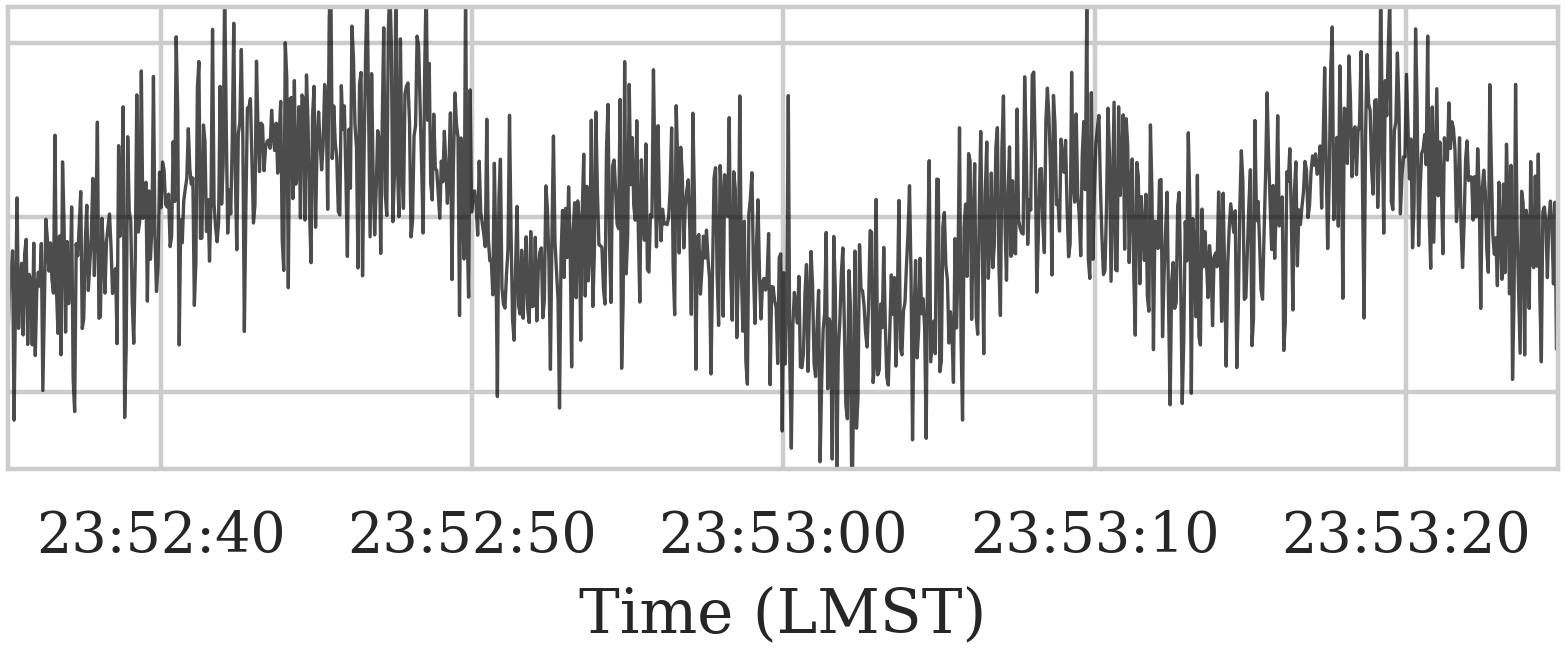}
      \caption{Preserved background signal}
      \label{fig:source_separation_glitch_background_zoom_38}
  \end{subfigure}\hspace{0em}
  \begin{subfigure}[t]{0.325\textwidth}
      \includegraphics[width=\textwidth]{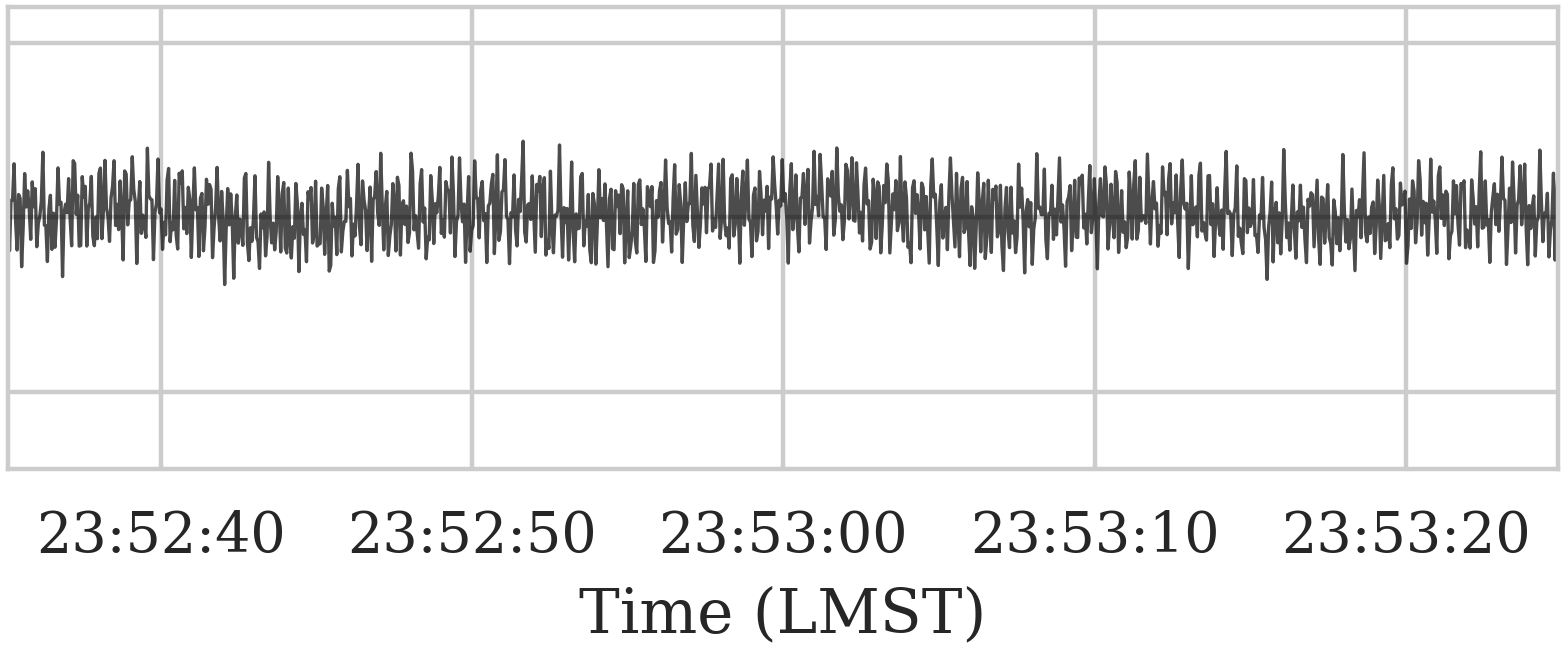}
      \caption{No spurious separation}
      \label{fig:source_separation_glitch_source_zoom_38}
  \end{subfigure}

  \caption{Separating glitches from a waveform extracted from cluster
  $4$ from the $54.6$-minute timescale. The cluster from which the raw
  waveform is selected is mostly concentrated during the Martian daytime
  waveform, we select as prior information $300$ samples from cluster
  $5$ in the $51.2$-second timescale. This cluster has occurrence time
  histogram mostly concentrated during Martian night and its
  characteristic waveforms contain no glitches. As such, samples from
  this cluster would be good candidates for ``backgrounds'' signals that
  could allow for separating glitches. For comparison with the
  state-of-practice baseline glitch separation algorithm, see
  \cref{fig:source_separation_nasa_glitch} in
  \cref{app:baseline_glitch_separation} of the Appendix. Refer to
  \cref{fig:source_separation_glitch_ica} in \cref{app:ica_comparison}
  of the Appendix for a comparison with the ICA-based source separation
  method.}
  \label{fig:source_separation_glitch}
\end{figure*}

Upon visual inspection, we can confirm that the characteristic waveforms
for all clusters at different time scales maintained the same structure
as illustrated in \cref{fig:clusters_short,fig:clusters_long}. This
ensures the nature of the clusters is unchanged across time, making
comparing their histograms relevant. We make the following observations.
While we acknowledge that not all variations observed in the time
histograms can be solely attributed to seasonal changes (e.g.,
operational changes of the lander and having access to less data during
the winter), we identified that clusters with Martian daytime
occurrences exhibited a tendency to shift in their peaks. Furthermore,
we also observe that some clusters exhibit slight changes in their
distribution, mainly involving a reduction or increase of their peaks.
Notable clusters involve the glitch clusters at the finest scale (see
\cref{fig:cluster-6_scale-1_season,fig:cluster-7_scale-1_season}).
Specifically, we can observe that we have detected more waveforms during
Martian daytime over the winter and fall seasons, which might be related
to less ambient seismic noise due to lower temperatures, allowing our
approach to identify more glitches during the day. In addition, as
mentioned before, glitches tend to occur more near sunset, which we
still can observe from the peak in per-season time histograms with the
peak time clearly shifting with the average sunset time per season.

While we observe strong correlations between the occurrence time
histograms and atmospheric conditions, it is worth mentioning that there
are seismic signals in each cluster unaffected by seasonal changes.
However, the heavy imprint of atmosphere-surface interactions skews the
histograms nonetheless. In the next sections, we will indicate that
indeed certain seismic events are concentrated in a few clusters, even
though they are not the main characterizing feature of those clusters.

\subsection{Unsupervised separation of prominent sources}

\begin{figure*}[t]
  \centering

  \begin{subfigure}[t]{0.325\textwidth}
      \includegraphics[width=\textwidth]{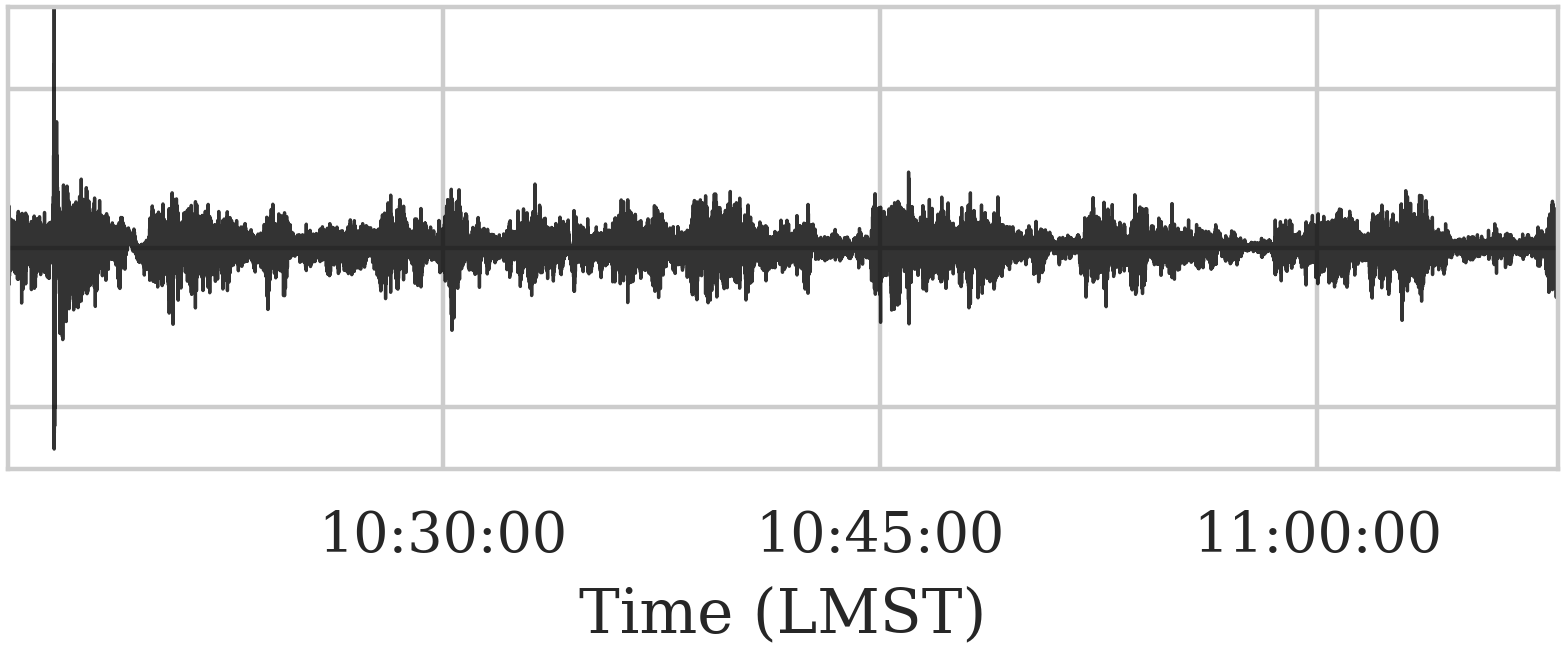}
      \caption{Raw signal with wind bursts}
      \label{fig:source_separation_wind_real}
  \end{subfigure}\hspace{0em}
  \begin{subfigure}[t]{0.325\textwidth}
      \includegraphics[width=\textwidth]{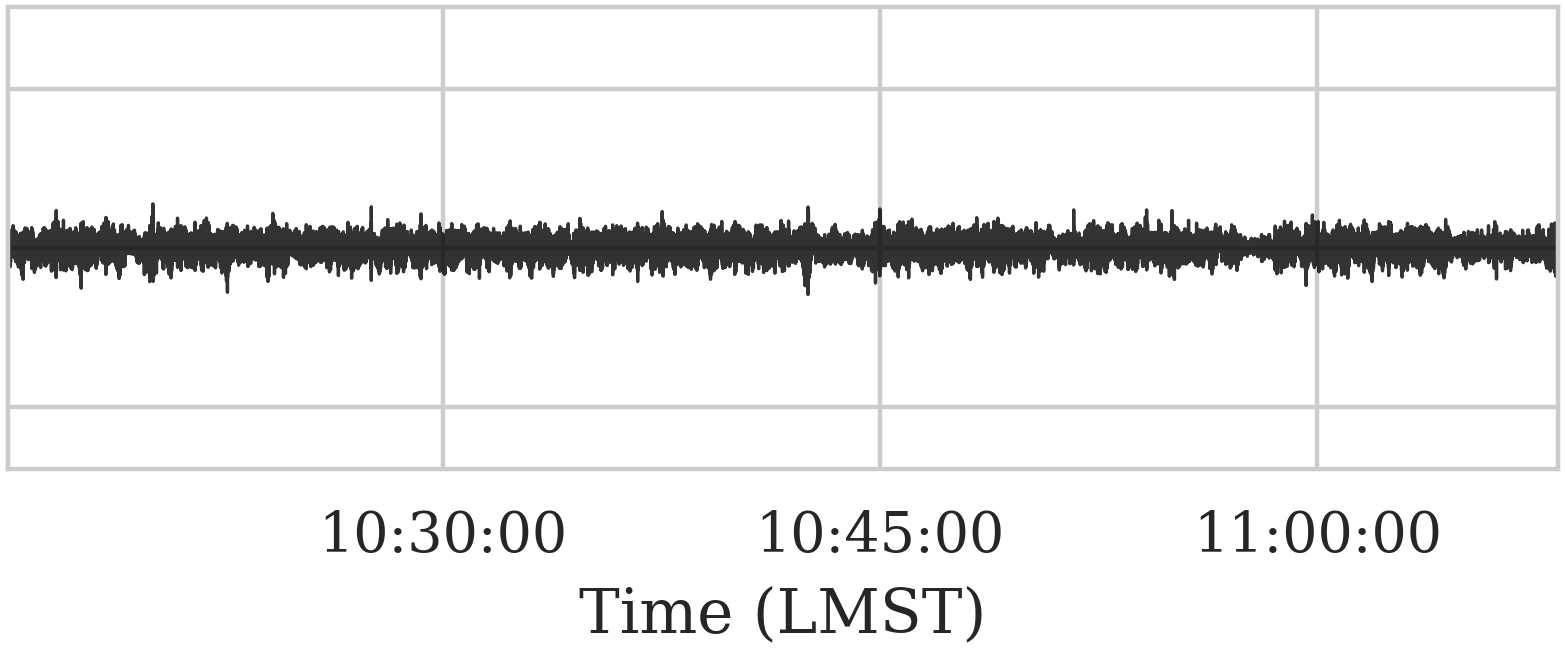}
      \caption{Background after wind separation}
      \label{fig:source_separation_wind_background}
  \end{subfigure}\hspace{0em}
  \begin{subfigure}[t]{0.325\textwidth}
      \includegraphics[width=\textwidth]{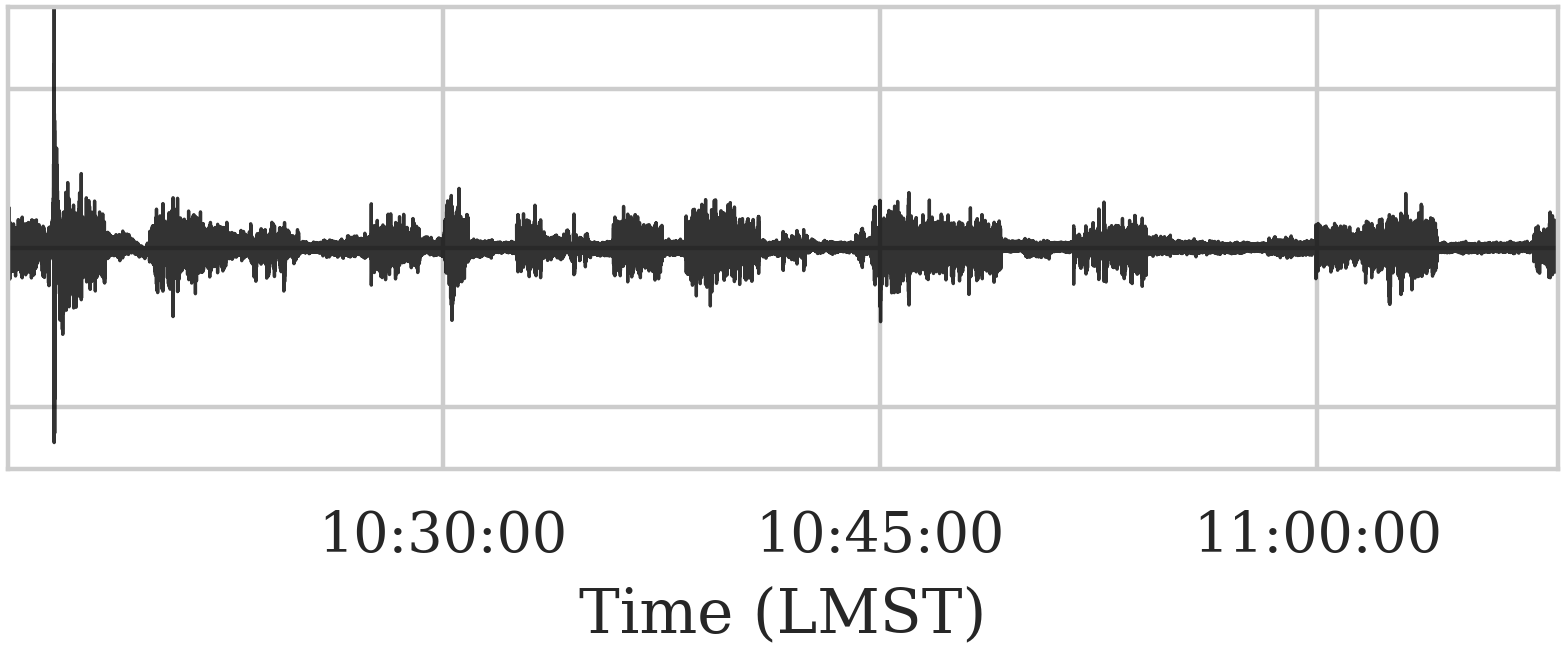}
      \caption{Extracted wind burst signals}
      \label{fig:source_separation_wind_source}
  \end{subfigure}

  \begin{subfigure}[t]{0.325\textwidth}
      \includegraphics[width=\textwidth]{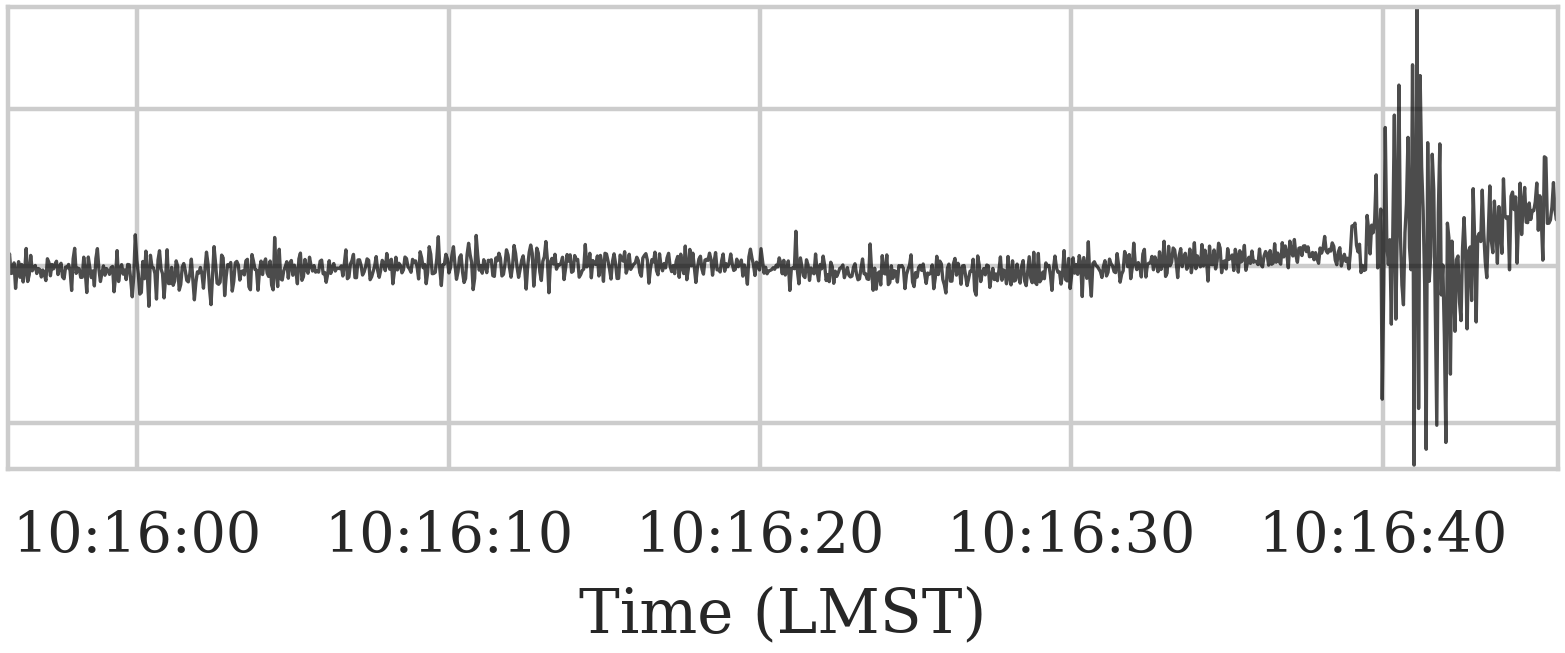}
      \caption{Wind burst with ringing (raw)}
      \label{fig:source_separation_wind_real_zoom_1}
  \end{subfigure}\hspace{0em}
  \begin{subfigure}[t]{0.325\textwidth}
      \includegraphics[width=\textwidth]{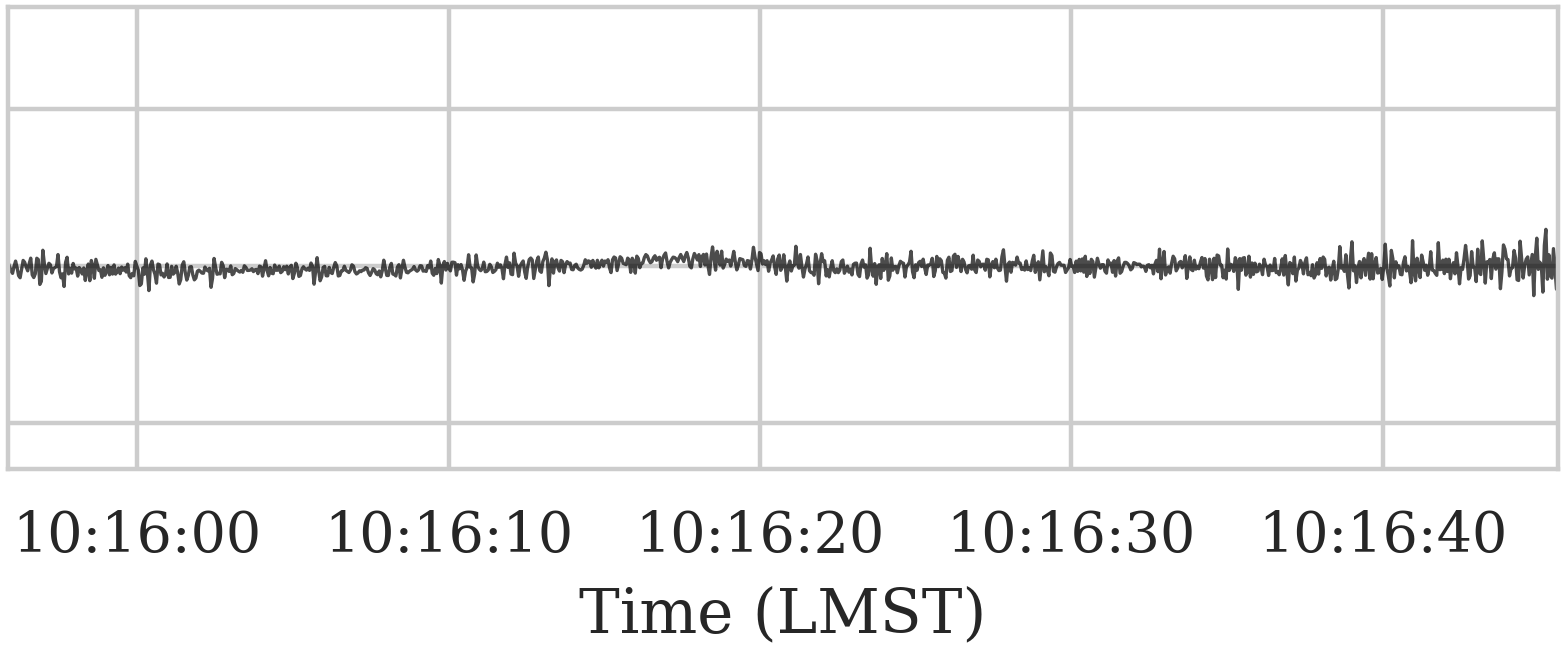}
      \caption{Background after separation}
      \label{fig:source_separation_wind_background_zoom_1}
  \end{subfigure}\hspace{0em}
  \begin{subfigure}[t]{0.325\textwidth}
      \includegraphics[width=\textwidth]{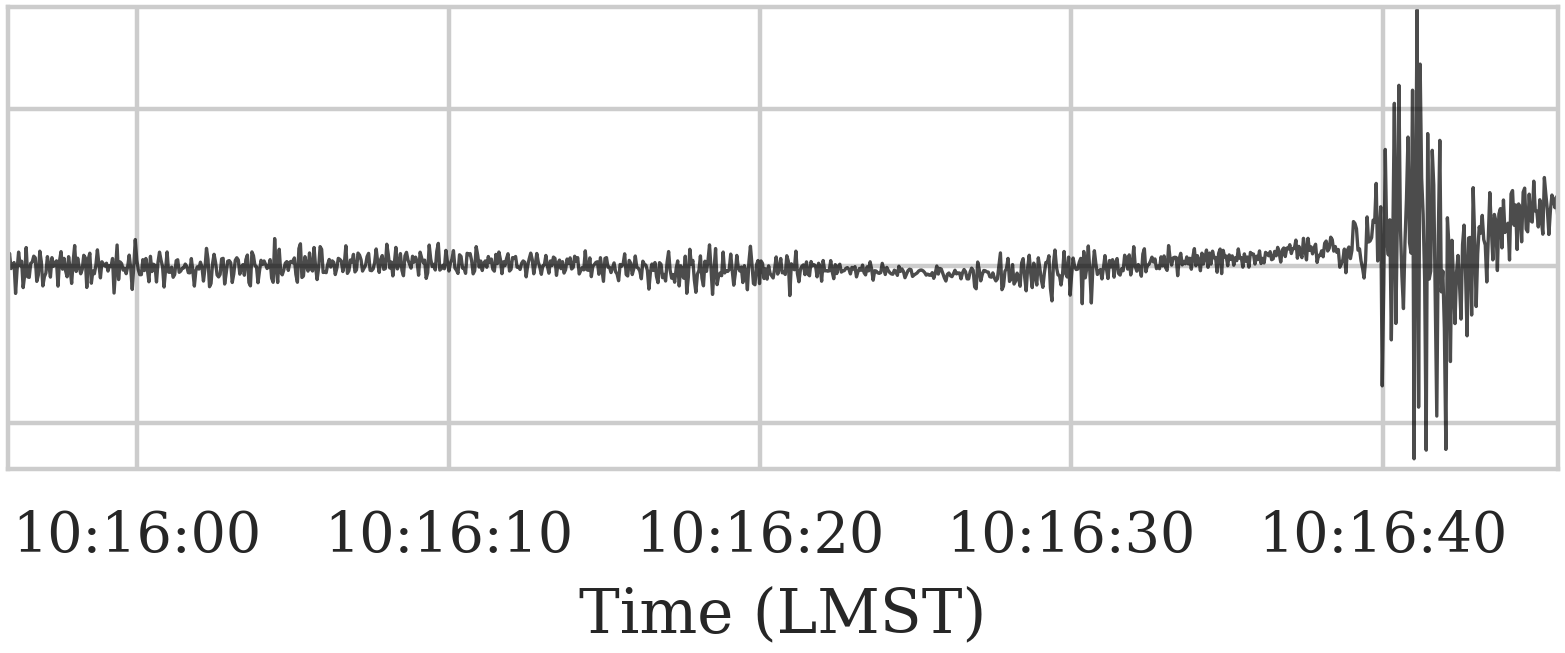}
      \caption{Separated wind burst}
      \label{fig:source_separation_wind_source_zoom_1}
  \end{subfigure}

  \begin{subfigure}[t]{0.325\textwidth}
      \includegraphics[width=\textwidth]{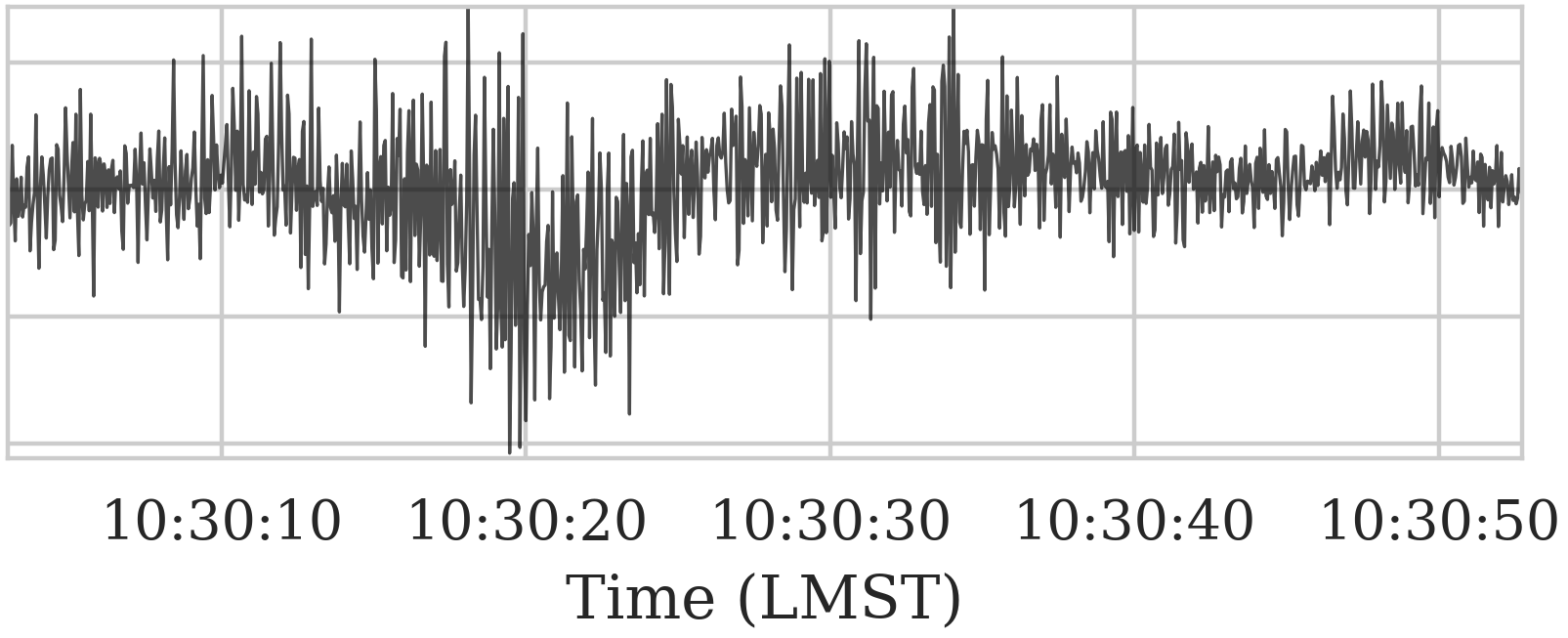}
      \caption{Sharp wind onset (raw)}
      \label{fig:source_separation_wind_real_zoom_18}
  \end{subfigure}\hspace{0em}
  \begin{subfigure}[t]{0.325\textwidth}
      \includegraphics[width=\textwidth]{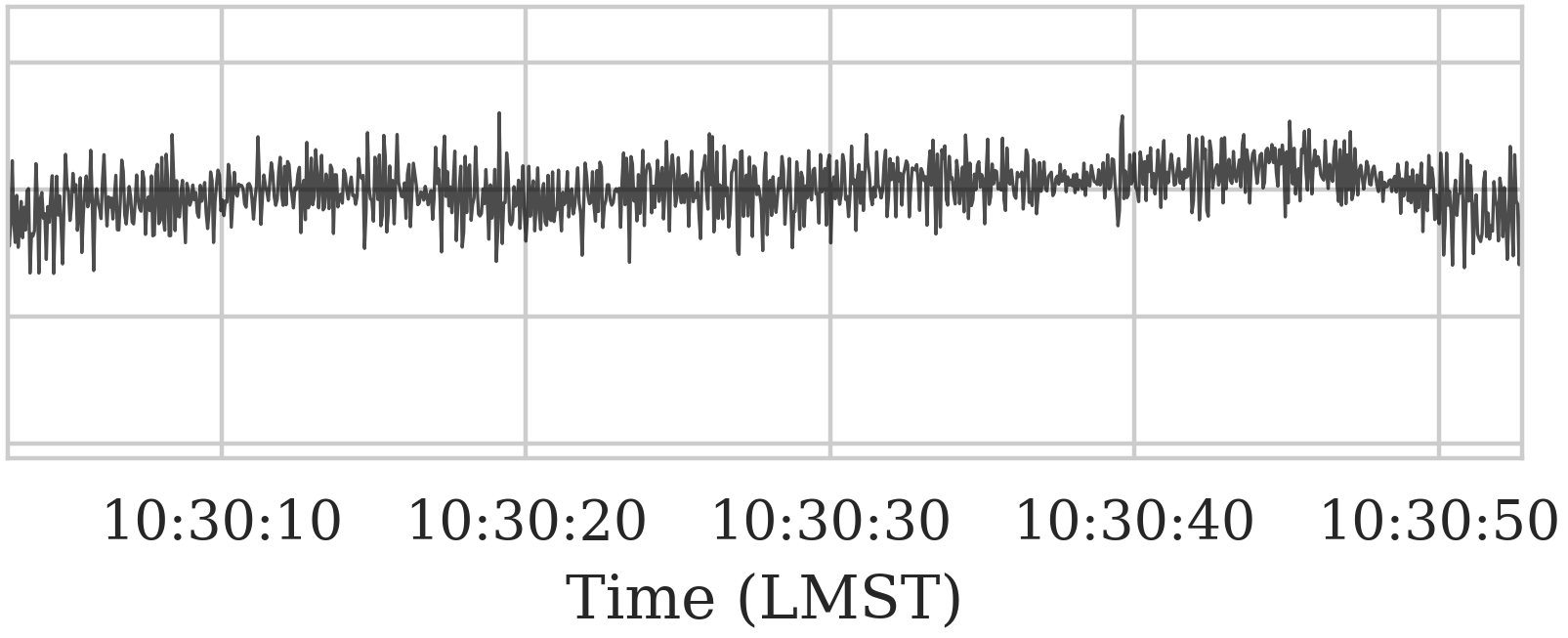}
      \caption{Background after separation}
      \label{fig:source_separation_wind_background_zoom_18}
  \end{subfigure}\hspace{0em}
  \begin{subfigure}[t]{0.325\textwidth}
      \includegraphics[width=\textwidth]{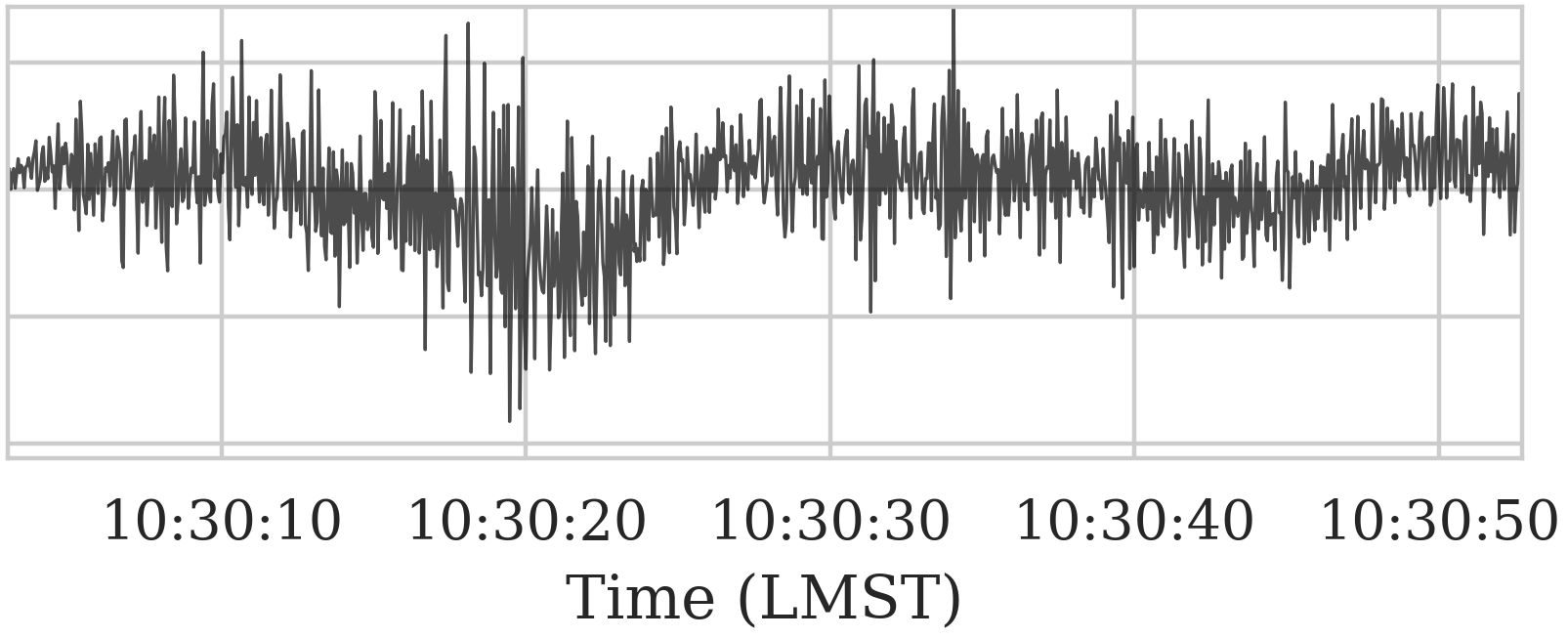}
      \caption{Separated wind signature}
      \label{fig:source_separation_wind_source_zoom_18}
  \end{subfigure}

  \begin{subfigure}[t]{0.325\textwidth}
      \includegraphics[width=\textwidth]{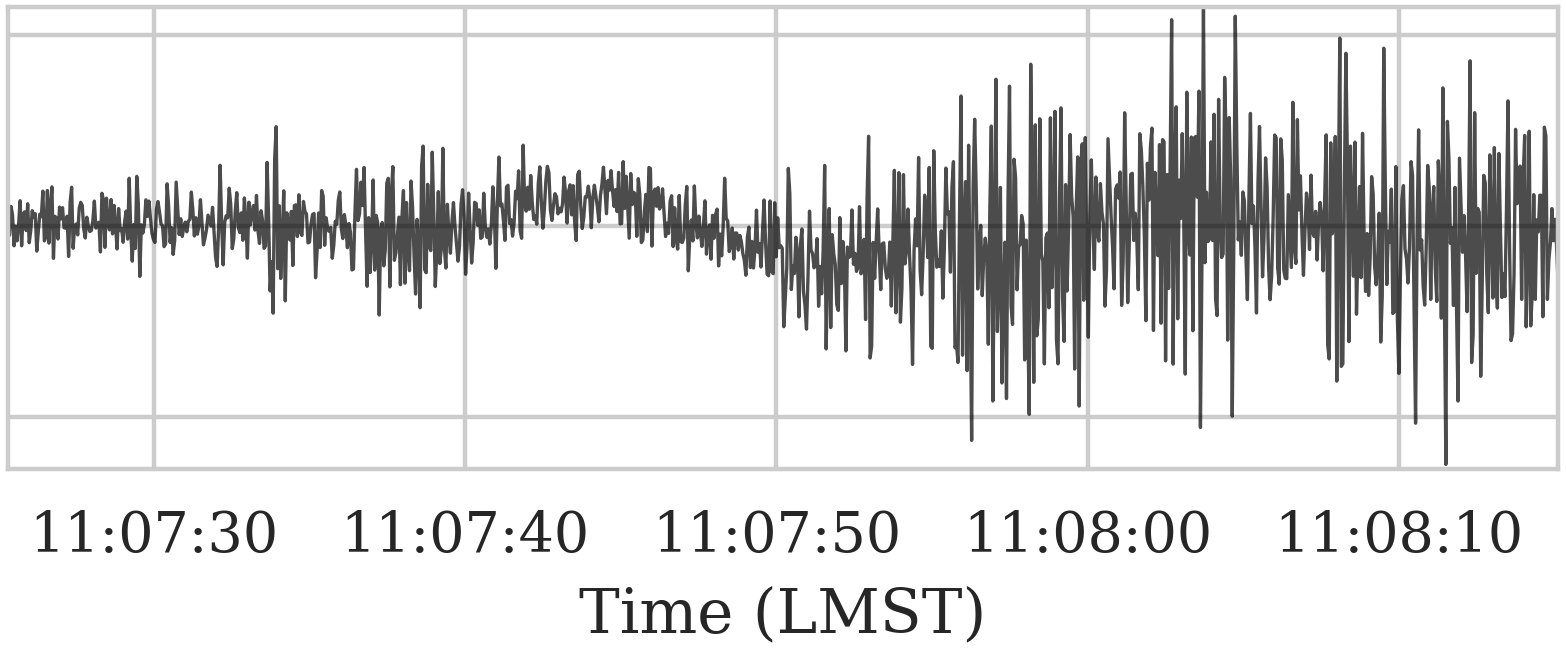}
      \caption{Complex wind pattern (raw)}
      \label{fig:source_separation_wind_real_zoom_63}
  \end{subfigure}\hspace{0em}
  \begin{subfigure}[t]{0.325\textwidth}
      \includegraphics[width=\textwidth]{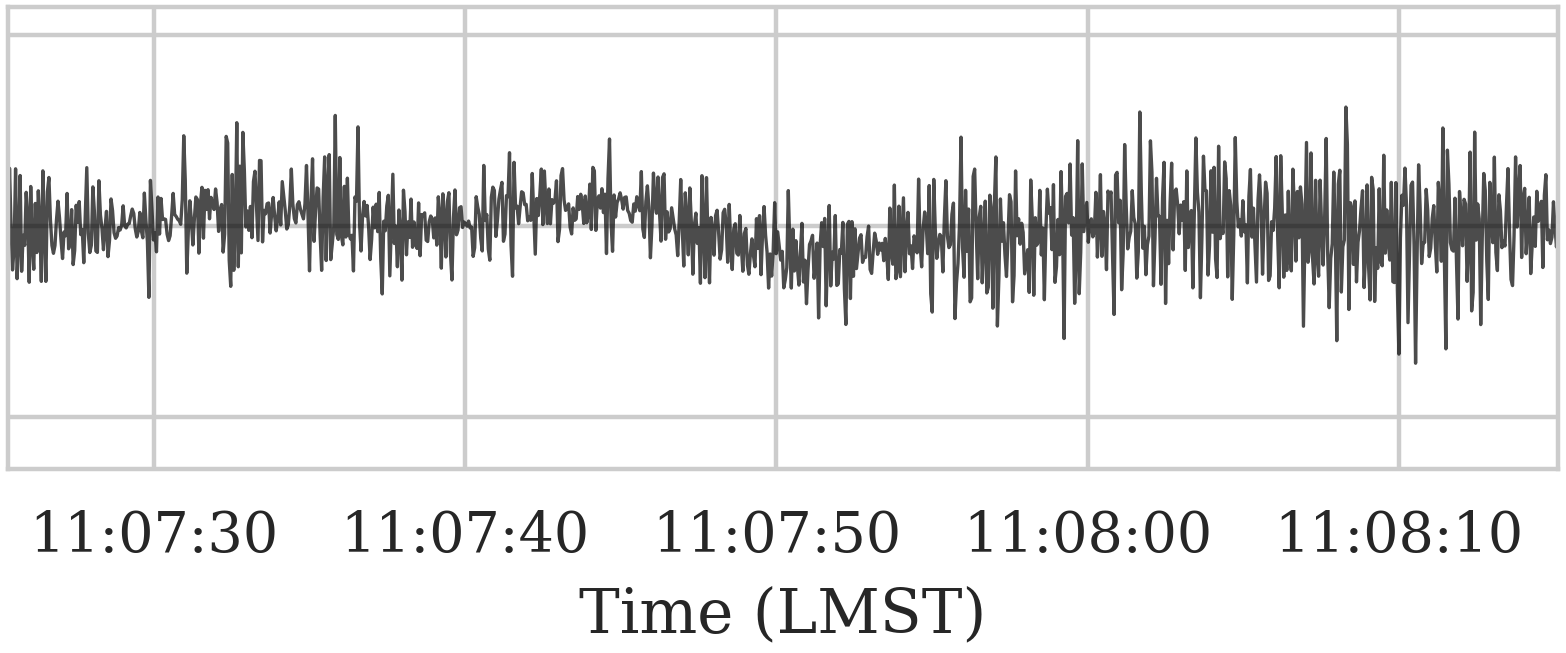}
      \caption{Background after separation}
      \label{fig:source_separation_wind_background_zoom_63}
  \end{subfigure}\hspace{0em}
  \begin{subfigure}[t]{0.325\textwidth}
      \includegraphics[width=\textwidth]{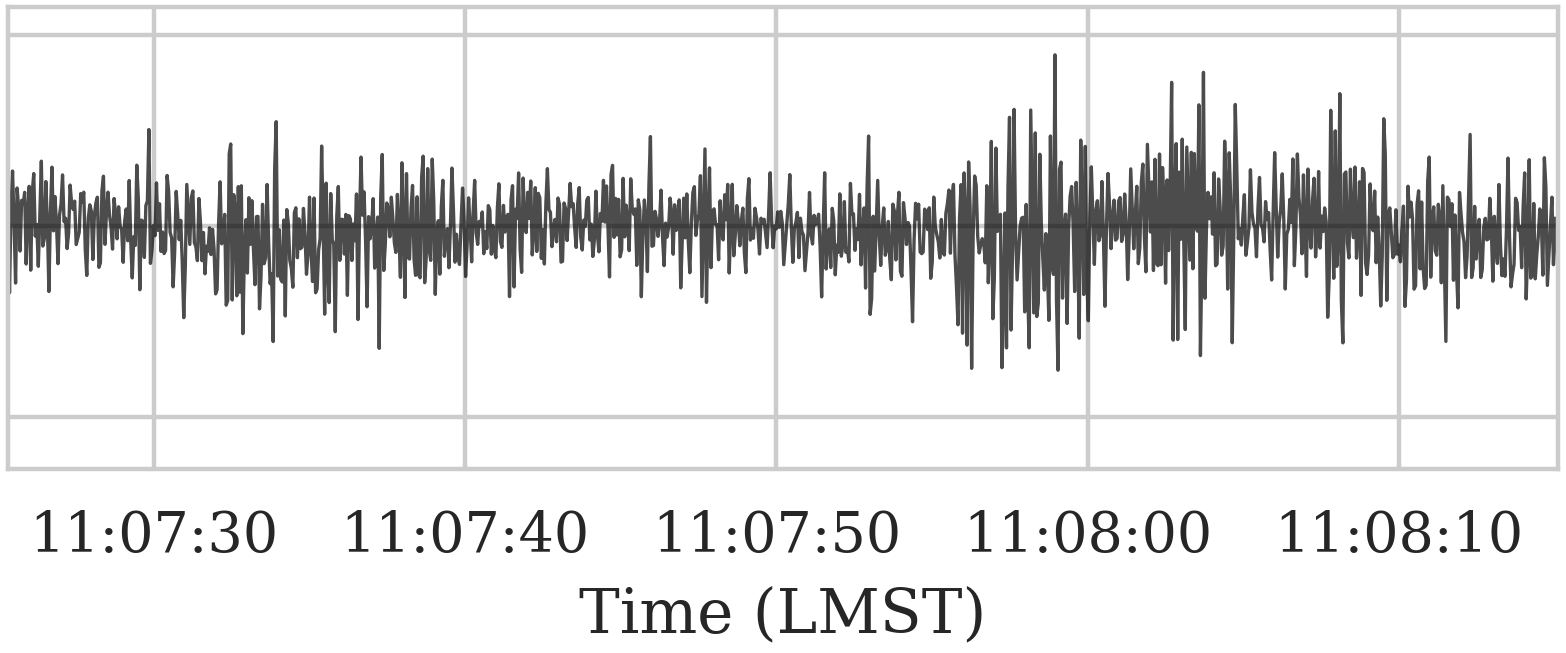}
      \caption{Separated wind pattern}
      \label{fig:source_separation_wind_source_zoom_63}
  \end{subfigure}

  \caption{Separating imprints of the wind from a waveform extracted
  from cluster $3$ from the $54.6$-minute timescale. The cluster from
  which the raw waveform is selected is mostly concentrated during the
  Martian daytime and in order to remove glitches from this daytime
  waveform, we select as prior information $300$ samples from clusters
  $1$ and $6$ in the $51.2$-second timescale. These clusters have
  occurrence time histograms mostly concentrated during the day and
  their characteristic waveform exhibit no wind burst signatures. As
  such, samples from these clusters would be good candidates for
  ``backgrounds'' signals that could allow for separating wind burst
  related noises. Refer to \cref{fig:source_separation_wind_ica} in
  \cref{app:ica_comparison} of the Appendix for a
  comparison with the ICA-based source separation method.}
  \label{fig:source_separation_wind}
\end{figure*}

\noindent Given access to clusters of glitches of wind-burst noises, in
this section, we show that using samples from these clusters, we can
perform source separation using the method described in
\cref{sec:src-sep}. In this method, a particular source of interest can
be separated from a given time series provided that we have access to
several data snippets that does not contain that source of interest.
Here explore separating glitches and the imprint of wind bursts.
\Cref{fig:source_separation_glitch,fig:source_separation_wind}
summarizes the results, and we provide details and observations for each
experiment below. We also provide an example for separating background
noise and glitches from a marsquake (cf. \cref{app:marsquake_separation}
of the Appendix), which extends the application of our method to sources
that are less prominent in the dataset. Since our approach allows source
separation to be performed at a particular time scale, this optimization
problem often involves solving multiple independent optimization
problems (one for each window of data). To utilize this embarrassingly
parallel structure, we run the optimization in parallel on four Tesla
V10 devices, for 1000 iterations of the L-BFGS optimization algorithm
\cite{Nocedal_1980} and a relative termination tolerance of 0.1. The
results of glitch separation using the state-of-practice baseline method
\cite{ScholzWidmer_SchnidrigDavisEtAl2020} are presented in
\cref{app:baseline_glitch_separation} of the Appendix. We also
provide a comparison with the ICA-based source separation method in
\cref{app:ica_comparison} of the Appendix.

\paragraph{Glitch separation} For the glitch separation experiment, we
select a $54.6$-minute long waveform during Martian nighttime with
several glitches (cf. \cref{fig:source_separation_glitch_real}). We use
$300$ samples from cluster $5$ in the $51.2$-second timescale as its
occurrence time histogram mostly concentrated during Martian night and
its characteristic waveforms contain no glitches. The optimization took
104 minutes to complete.
\Cref{fig:source_separation_glitch_background,fig:source_separation_glitch_source}
depicts the waveform after separation of glitches and the separated
glitches, respectively. The last four rows of
\cref{fig:source_separation_glitch} illustrate the zoomed-in views of
the results with columns from left to right being raw waveform, waveform
after separation of glitches, and the separated glitch, respectively.
These figures indicate the successful separation of glitches as the
one-sided pulses are removed with minimal changes to the rest of the
waveform. Specifically, the last row of
\cref{fig:source_separation_glitch_real_zoom_38} indicates a zoomed-in
view of a portion of the raw waveform that contains no glitches, and our
method does not separate any coherent signal from the waveform, which is
a crucial property of our method.

\begin{figure*}[!t]
  \centering
  \vspace*{-2em}
  \begin{subfigure}[t]{0.4\textwidth}
      \includegraphics[width=\textwidth]{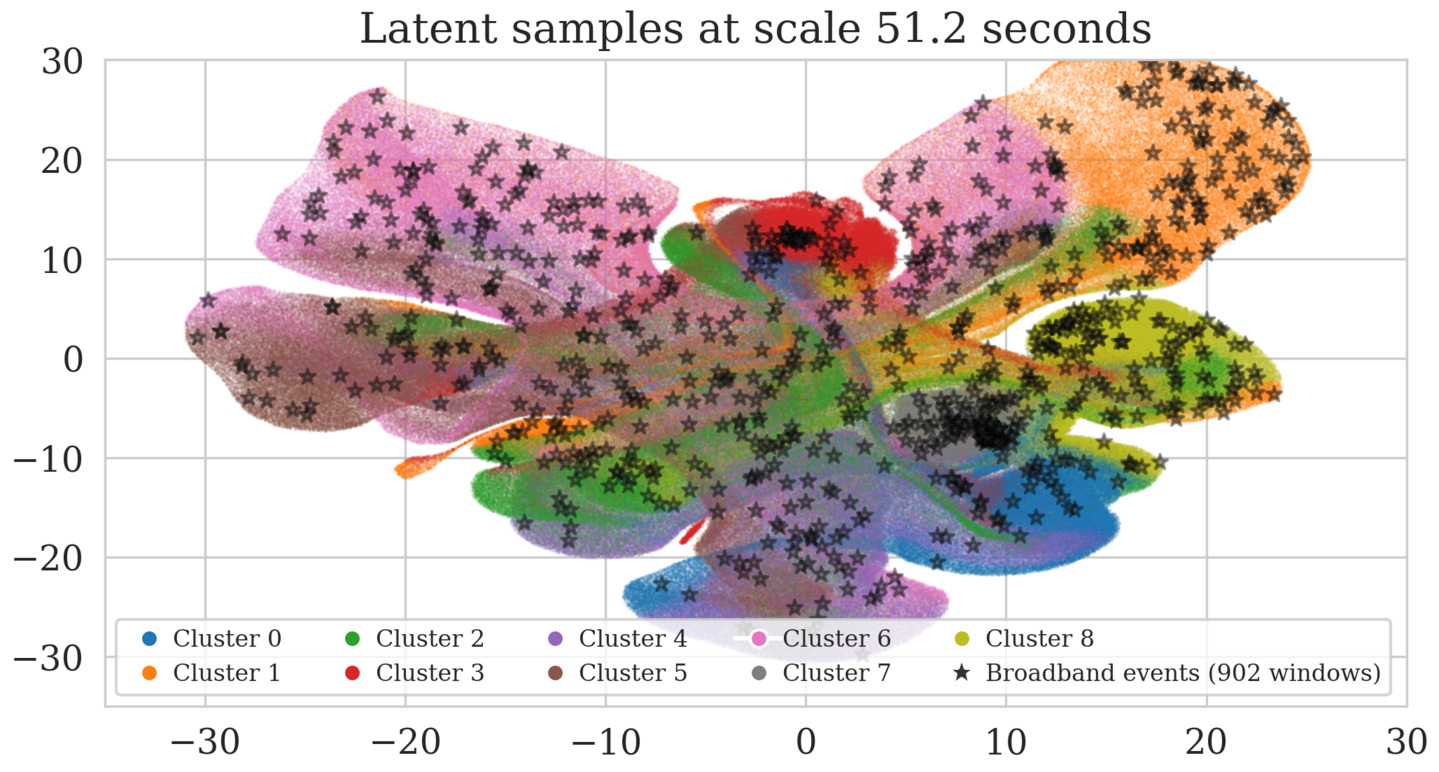}
      \caption{Broadband events (quality A) --- finest scale}
      \label{fig:scale-1_broadband}
  \end{subfigure}\hspace{2em}
  \begin{subfigure}[t]{0.4\textwidth}
      \includegraphics[width=\textwidth]{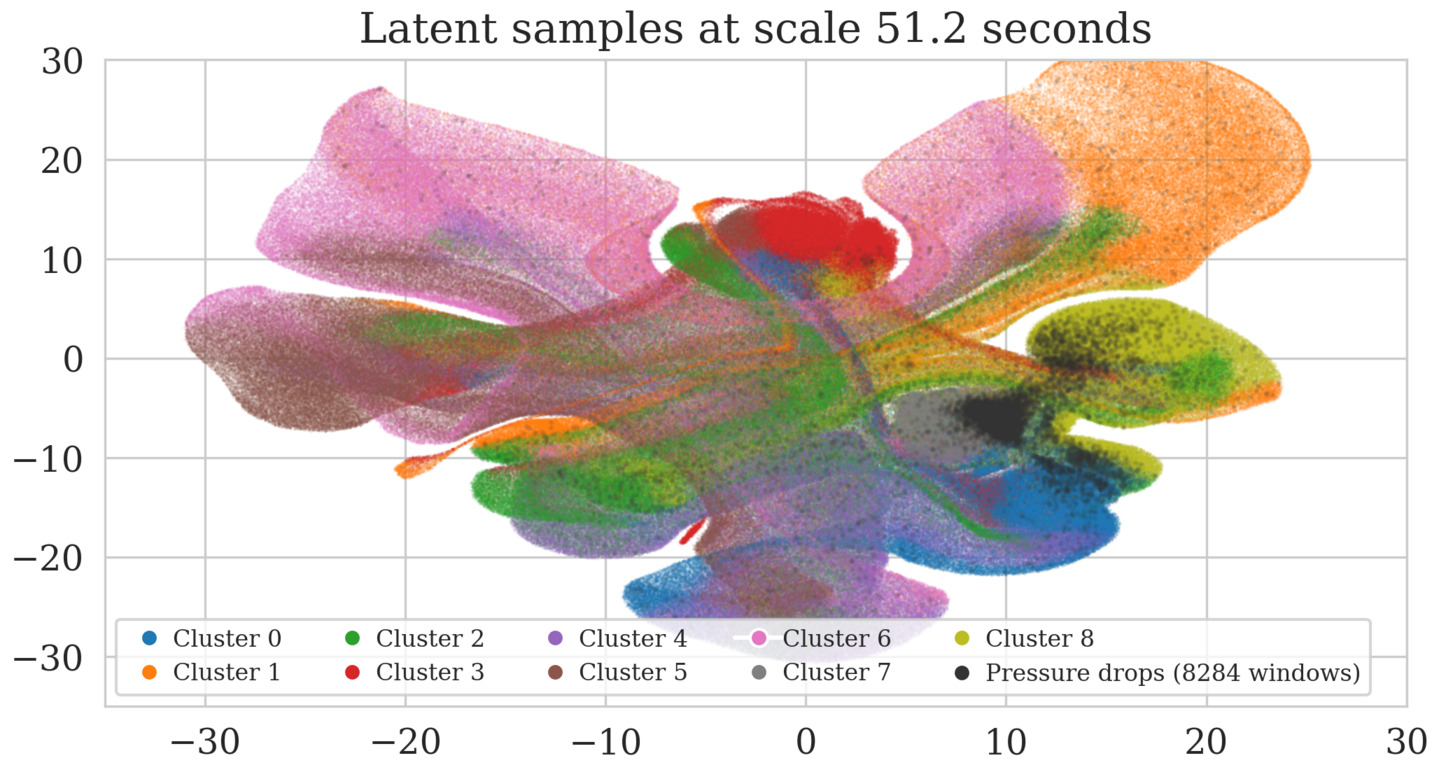}
      \caption{Atmospheric pressure drops --- finest scale}
      \label{fig:scale-1_drop}
  \end{subfigure}

  \begin{subfigure}[t]{0.4\textwidth}
      \includegraphics[width=\textwidth]{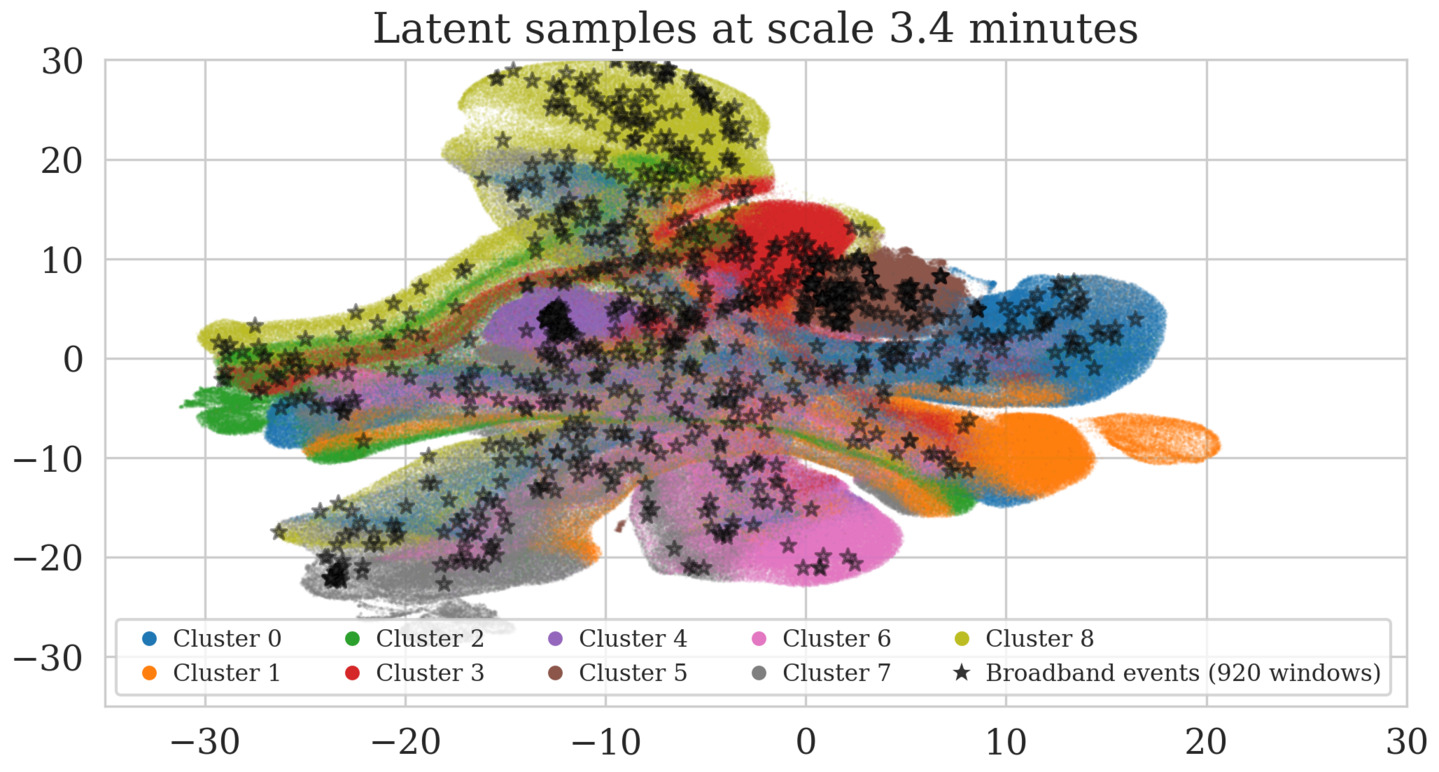}
      \caption{Broadband events (quality A) --- intermediate scale}
      \label{fig:scale-2_broadband}
  \end{subfigure}\hspace{2em}
  \begin{subfigure}[t]{0.4\textwidth}
      \includegraphics[width=\textwidth]{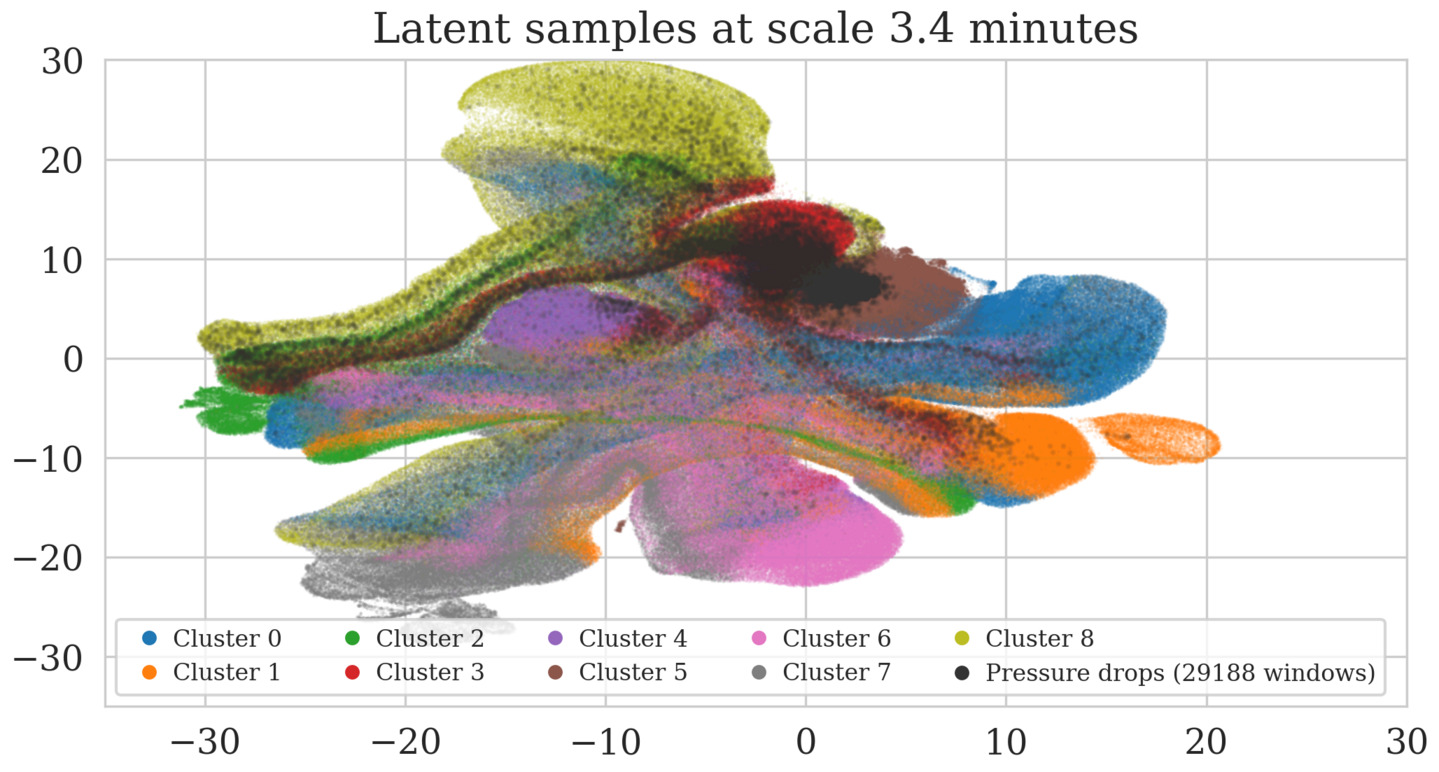}
      \caption{Atmospheric pressure drops --- intermediate scale}
      \label{fig:scale-2_drop}
  \end{subfigure}

  \begin{subfigure}[t]{0.4\textwidth}
      \includegraphics[width=\textwidth]{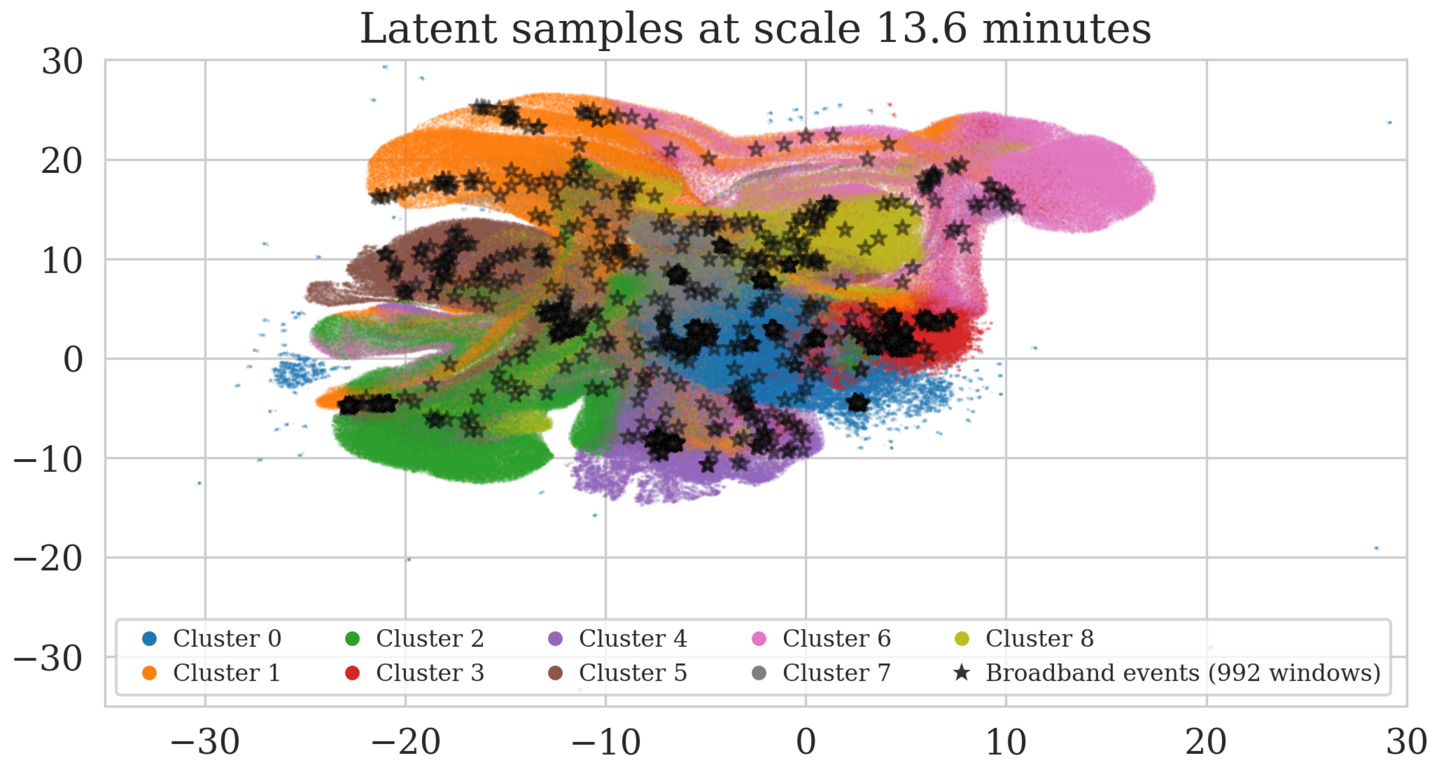}
      \caption{Broadband events (quality A) --- coarse scale}
      \label{fig:scale-3_broadband}
  \end{subfigure}\hspace{2em}
  \begin{subfigure}[t]{0.4\textwidth}
      \includegraphics[width=\textwidth]{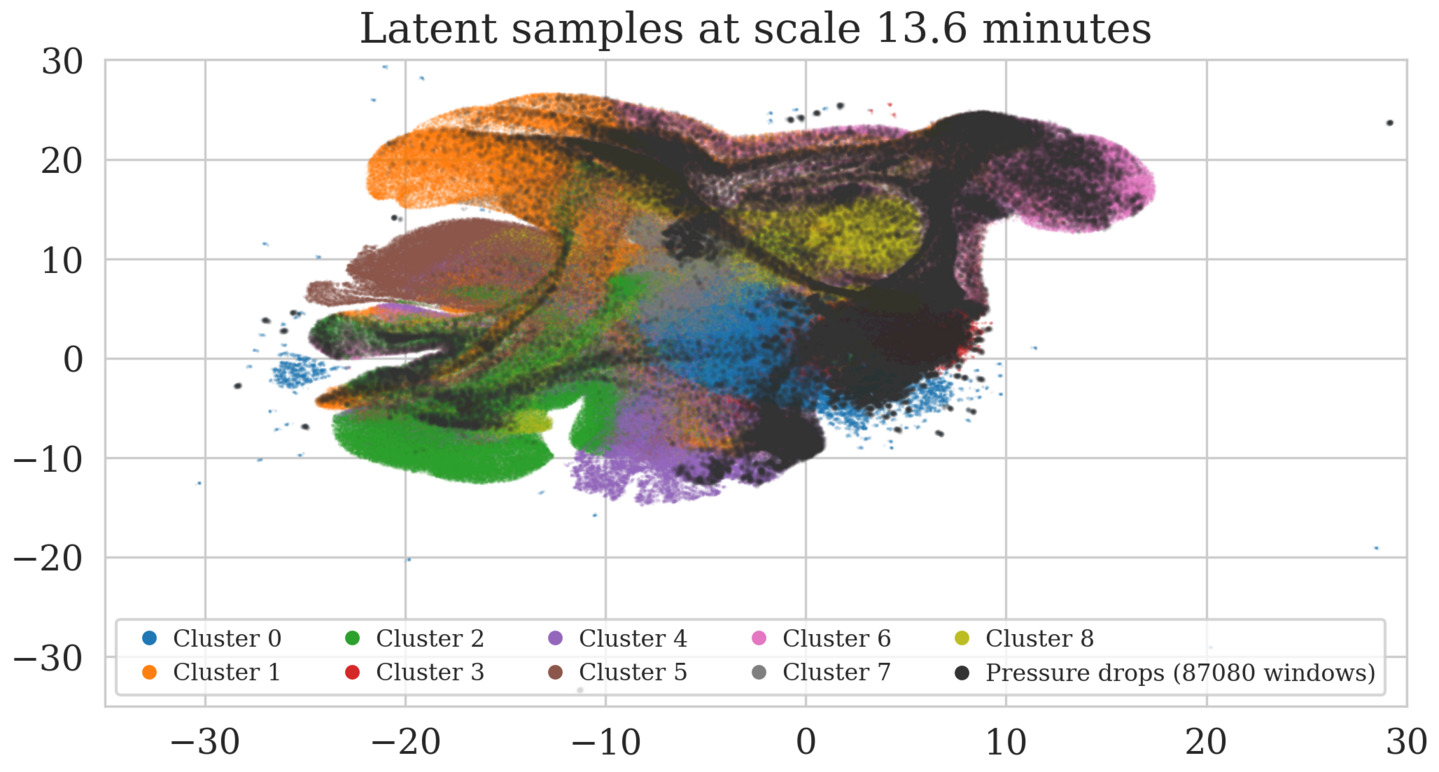}
      \caption{Atmospheric pressure drops --- coarse scale}
      \label{fig:scale-3_drop}
  \end{subfigure}

  \begin{subfigure}[t]{0.4\textwidth}
      \includegraphics[width=\textwidth]{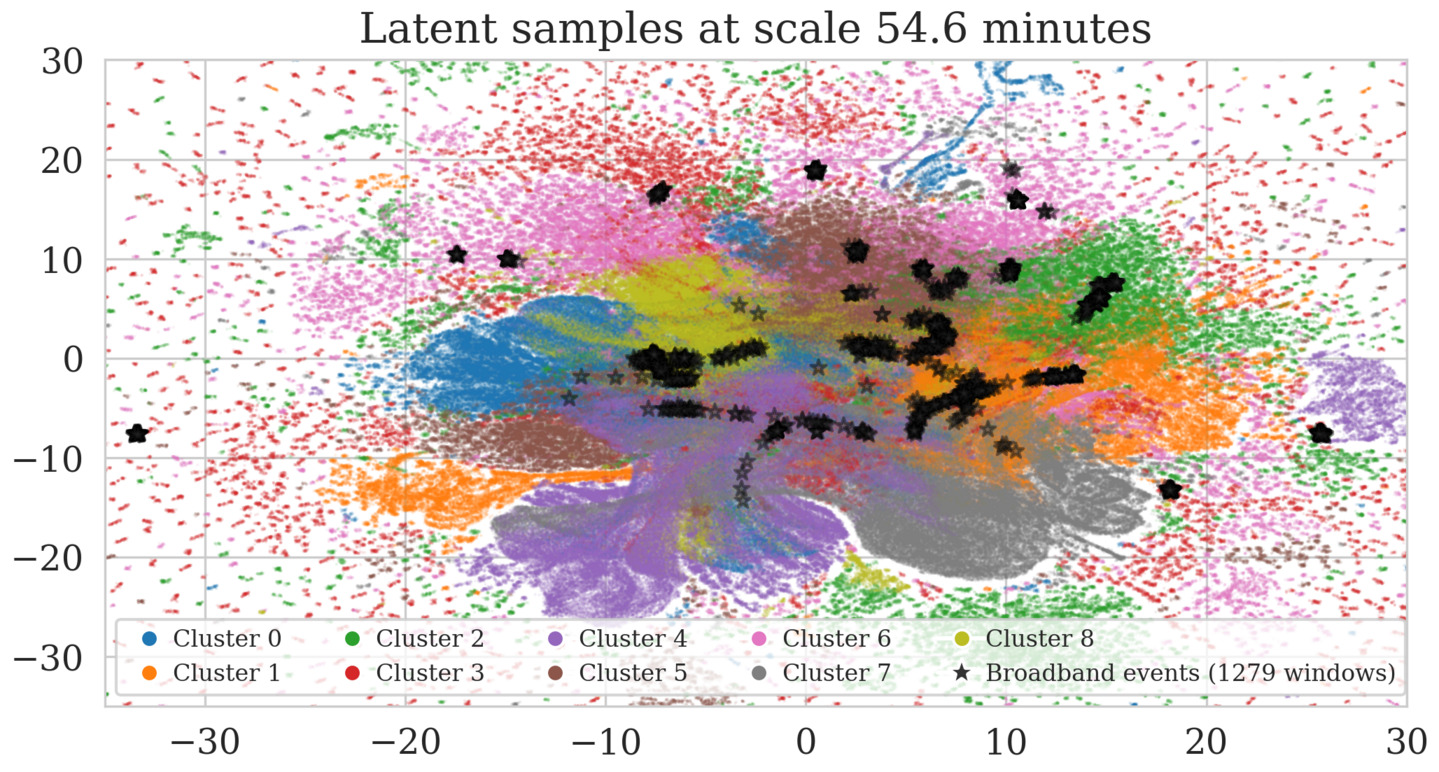}
      \caption{Broadband events (quality A) --- coarsest scale}
      \label{fig:scale-4_broadband}
  \end{subfigure}\hspace{2em}
  \begin{subfigure}[t]{0.4\textwidth}
      \includegraphics[width=\textwidth]{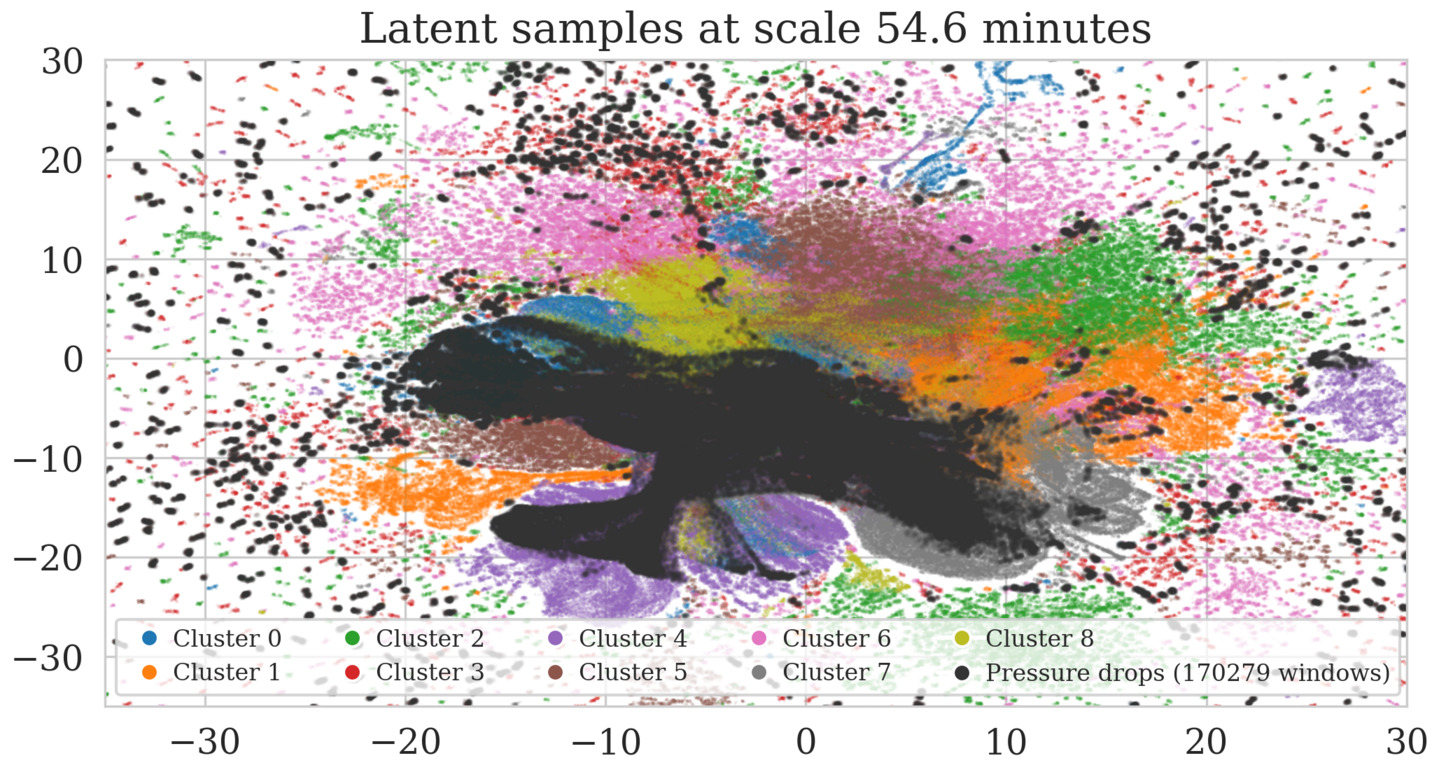}
      \caption{Atmospheric pressure drops --- coarsest scale}
      \label{fig:scale-4_drop}
  \end{subfigure}

  \caption{fVAE latent space low-dimensional visualization, via UMAP \cite{McinnesEtAl_2018}, across four timescales overlaid with quality ``A'' broadband events or pressure drops. The left column overlays quality ``A'' broadband events with star symbols, and the right column indicates pressure drops with black circles. The timescale of the plots increase from top to bottom. Due to our pyramidal scattering spectra construction mechanism, we have the same total number of latent samples in each scale while the number of samples identified as quality A events or pressure drops increases for larger timescales. Nonetheless, we observe broadband events tend to concentrate better in the largest timescale, which corresponds to the timescale typically exhibited by marsquakes. Similarly, pressure drops that are very short in timescale concentrate very well in the shortest timescale.}
  \label{fig:latent_visualization}
\end{figure*}

\paragraph{Wind burst imprint separation} For the wind burst imprint
separation we select a Martian daytime $54.6$-minute long waveform (cf.
\cref{fig:source_separation_wind_real}). We use $300$ samples from
cluster $1$ in the $51.2$-second timescale as its occurrence time
histogram mostly concentrated during Martian day and its characteristic
waveforms contain no visible wind bursts. The optimization took 106
minutes to complete.
\Cref{fig:source_separation_wind_background,fig:source_separation_wind_source}
depicts the waveform after separation of glitches and the separated
glitches, respectively. The last four rows of
\cref{fig:source_separation_wind} illustrate the zoomed-in views of the
results with columns from left to right being raw waveform, waveform
after separation of the wind burst imprint, and the separated wave burst
imprint, respectively. These figures indicate the successful separation
of wind burst imprints as the wind burst imprint (sharp onset followed
by ringing oscillations) is removed with minimal changes to the rest of
the waveform.

\subsection{Latent space exploration of clusters}

\noindent Towards better exploring the learned representation of fVAE, i.e., the per-scale representation obtained after the fVAE encoder, we visualize a low-dimensional representation of the entire dataset. We obtain this low-dimensional representation of the latent space via the UMAP algorithm \cite{McinnesEtAl_2018}, where we reduce the dimensionality of latent vectors from 32 (see \cref{sec:architecture} for a more detailed description of the architecture) to a two-dimensional representation. We obtain this two-dimensional representation independently for each scale and identify the most likely cluster that each point belongs to (computed through the fVAE) via colors. \Cref{fig:latent_visualization} contains these visualizations where each row corresponds to a different timescale with timescales increasing from top to bottom. To illustrate the importance of having a multi-scale representation, we overlay quality ``A'' broadband events (left column, with black star symbols) and the pressure-drop events for each timescale (right column, with black circles). Note that due to our pyramidal scattering spectra construction mechanism, we have the same total number of latent samples in each scale while the number of samples identified as quality A events or pressure drops increases for larger timescales. Broadband events, i.e., marsquakes, typically have a length of 30–60 minutes and as such, even though we have observed very few of them during the mission and we cannot expect to have a cluster associated with them, we expect a good representation of data to concentrate them in the latent space. Similarly, we expect pressure drops, which have a timescale of less than a minute to be well concentrated in the smaller timescale representations.

We make the following observations. The clusters in the $3.4$-minute and $13.6$-minute timescales (second and third rows of \cref{fig:latent_visualization}), compared to the finest and coarsest timescales, have clearer boundaries, which suggest that there is more structure in the data in these timescales. The $51.2$-second timescale clusters are also relatively well separated, except for the area in the middle of \cref{fig:scale-1_broadband,fig:scale-1_drop} that contains mixed samples from several clusters. The $54.6$-minute timescale cluster visualization, however, shows less cohesive structures and the points from several clusters are spread throughout the low-dimensional space. This might indicate that there is a less effective number of clusters at this scale. Nonetheless, we importantly observe that the quality ``A'' broadband events are much more concentrated in the coarsest time scale (see \cref{fig:scale-4_broadband}) compared to the others, with the concentration deteriorating as the timescale decreases. On the other hand, the pressure drops, which have a very short timescale, are very compactly situated in the finest timescale latent space while being spread across multiple clusters in the coarser timescales. These observations regarding the broadband and pressure-drop events indicate that the learned multi-scale representation appears to be meaningful and to yield localization in the latent space.

\section{Discussion and Conclusion}

\noindent In order to attain accurate source separation, having prior
understanding of the sources is essential. Unsupervised source
separation methods offer a solution when expert domain knowledge on the
existing source signals is limited by learning to extract sources solely
from a dataset of source mixtures. However, when dealing with data
containing sources of highly varied timescales, an architecture with
appropriate inductive biases is needed to enable multi-scale treatment
of source separation. In our work, we propose an approach using
factorial variational autoencoders (fVAE) nested within the pyramidal
wavelet scattering spectra representation. The intricate treatment of
the multi-scale nature of the data---via pyramidal scattering
spectra---can be also integrated in other clustering and source
separation methods and its usage is not limited to the proposed fVAE
architecture. Our results on data from NASA's InSight mission
demonstrate that the multi-scale fVAE approach successfully identifies
different non-seismic phenomena in the InSight data. While small
timescale detect more of the transient signals such as glitches, the
large timescale captures more of the global characteristics of the
background noise, which is more likely related regional change in
atmospheric conditions. The fVAE in turn enables unsupervised source
separation by leveraging prior knowledge from the clusters. This
approach makes minimal assumptions about the sources and provides a
truly unsupervised method for source separation in non-stationary
time-series with multi-scale sources.

Our proposed unsupervised, multi-scale clustering and source separation
approach demonstrated on Mars InSight data is highly applicable to
future planetary missions, including those to Europa and Titan, as these
environments present a profound lack of prior knowledge about the types,
origins, and timescales of signals that might be encountered. In such
unexplored settings, it is impossible to define realistic priors or
obtain labeled datasets in advance, which makes unsupervised methods not
just advantageous but essential; they can identify and disentangle
unknown and potentially novel sources without introducing bias or
unrealistic assumptions. This is especially relevant as NASA’s Farside
Seismic Suite (FSS) mission will soon deploy a modified version of the
SEIS seismometer, originally developed for the InSight mission, to the
Moon’s far side where seismic activity remains largely uncharacterized.

While our unsupervised multi-scale clustering and source separation
framework is robust to a range of real-world complexities, several fault
cases may affect its performance. When multiple sources overlap
substantially in both time and frequency, especially if they share
similar statistical or spectral characteristics, the clustering and
separation steps may yield ambiguous or mixed representations, as the
model relies on distinct multi-scale features for differentiation. High
levels of additive or non-stationary noise can further obscure
source-specific structure, potentially resulting in spurious clusters or
reduced separation quality, particularly at finer timescales.
Additionally, if the number of clusters per timescale is not
well-matched to the true diversity of sources, the model may
inadvertently merge distinct sources or fragment single sources across
multiple clusters. Such fault cases underscore the importance of careful
interpretation and iterative analysis. Importantly, our framework is
well-suited for exploratory data analysis: it enables users to
iteratively refine clustering and separation, unraveling the underlying
sources even in complex or ambiguous scenarios. This iterative process
can help identify limitations, guide parameter tuning, and ultimately
improve source characterization.

Future work involves letting the mother wavelet in the scattering
network to be learned, e.g., using \cite{BalestrieroEtAl_2018}. In that
setting, the mother wavelet is no longer specified a priori, but is
rather parametrized by a few learnable parameters. Those parameters can
then be learned to adapt the shape of the mother wavelet--and thus of
all the signal statistics that the scattering network computes--to
better minimize the loss at hand. Because the parametrization only adds
a few parameters, overfitting is not an issue, even in noisy settings,
as demonstrated in \cite{SeydouxEtAl_2020}. Additionally, training the
fVAE in the embedding space of pretrained seismic foundation models
\cite{LiuMnchmeyerLaurentiEtAl2024} could leverage rich representations from
large-scale seismic datasets to further enhance performance.

\bibliographystyle{IEEEtran}
\bibliography{refs}

\clearpage

\begin{appendices}

\section{Algorithm overview: Step-by-step process}
\label{app:algorithm-overview}

Below we provide a detailed walkthrough of the proposed unsupervised
multi-scale clustering and source separation framework, following the
flow of \cref{fig:schematic} from left to right. This description
enumerates the main steps required to apply our approach.

\begin{enumerate}
    \item \textbf{Pyramidal wavelet scattering spectra computation}
    \begin{itemize}
        \item Begin with the raw multichannel time series dataset,
        denoted as $\B{x}(t)$.
        \item Transform the data into the pyramidal wavelet scattering
        spectra representation, $\Psi(\B{x})$, as described in
        \cref{sec:scattering} and
        \cref{eq:pyramidal-scattering-cov}. This step captures the
        multi-scale, non-Gaussian characteristics of the data by
        computing scattering features over multiple window sizes
        (pyramidal averaging).
        \item The result is a set of multi-scale feature vectors that
        summarize the signal’s structure across a wide range of
        timescales (see multi-scale feature extractor in
        \cref{fig:schematic}).
    \end{itemize}

    \item \textbf{Probabilistic multi-scale clustering}
    \begin{itemize}
        \item Feed the multi-scale representations into the fVAE, which
        is trained to model the joint distribution of features at all
        timescales (see \cref{sec:fvae} and
        \cref{eq:reverse-kl-2}).
        \item The fVAE simultaneously performs probabilistic clustering at each timescale, assigning each data window to a cluster that represents a prominent source or process in the dataset.
        \item The model learns both the cluster assignments and the
        underlying distribution of features for each cluster, which will
        serve as prior information for separation (see fVAE module in
        \cref{fig:schematic}).
    \end{itemize}

    \item \textbf{Identification of prominent source clusters}
    \begin{itemize}
        \item Analyze the clustering results to identify clusters
        corresponding to sources of interest (e.g., glitches, wind
        bursts, background noise).
        \item This step may involve inspecting cluster occurrence
        histograms and representative waveforms, as shown in
        \cref{sec:clusters}.
    \end{itemize}

    \item \textbf{Formulation of the source separation problem}
    \begin{itemize}
        \item For a target source (cluster), collect representative
        samples from the identified cluster(s) to serve as prior
        information.
        \item Formulate the source separation as an optimization problem
        in the scattering spectra space, as described in
        \cref{sec:src-sep}. The goal is to
        recover the target source. The objective combines three loss
        terms:
        \begin{itemize}
            \item \textbf{Prior consistency}
            ($\mathcal{L}_{\text{prior}}$), ensuring the separated
            source matches the cluster statistics
            (cf. \cref{eq:loss-prior}),
            \item \textbf{Cross-independence}
            ($\mathcal{L}_{\text{cross}}$), promoting independence
            between the separated source and residual
            (cf. \cref{eq:loss-cross}),
            \item \textbf{Data fidelity} ($\mathcal{L}_{\text{data}}$),
            ensuring the sum of separated sources reconstructs the
            observed mixture (cf. \cref{eq:loss-data}).
        \end{itemize}
        \item See the source separation module in \cref{fig:schematic} for a high-level depiction of this step.
    \end{itemize}

    \item \textbf{Optimization and recovery of source signals}
    \begin{itemize}
        \item Solve the combined optimization problem, as presented in
        \cref{eq:ss_optimization}, to recover the target source in the
        time domain, using the learned priors and constraints from the
        previous step.
        \item The optimization is performed for each window or segment,
        and can be parallelized (final stage in \cref{fig:schematic}).
    \end{itemize}
\end{enumerate}

\section{Derivation of the fVAE loss function}\label{app:loss_derivation}

To overcome the intractability of minimizing the negative
log-likelihood, i.e., minimizing the expected value of $-\log
p_{\theta}(\B{u})$, we instead minimize the negative lower bound (ELBO).
Here, we detail the derivation of the objective function presented in
\cref{eq:reverse-kl-2}.
The derivation proceeds as follows:
\begin{itemize}
\item The derivation begins in~\cref{eq:reverse-kl-2-a} with the original objective of minimizing the expected negative log-likelihood of the data under the model. In~\cref{eq:reverse-kl-2-b}, latent variables $\B{z}, \B{y}$ are introduced via a variational distribution $q_{\phi}(\B{z}, \B{y}|\B{u})$, enabling a tractable approximation through sampling.

\item Line~\cref{eq:reverse-kl-2-c} rewrites the marginal likelihood using Bayes' rule, expressing $-\log p_{\theta}(\B{u})$ as a ratio of the joint and conditional distributions. In~\cref{eq:reverse-kl-2-d}, an identity involving the variational distribution is inserted by multiplying and dividing by $q_{\phi}(\B{z}, \B{y}|\B{u})$, allowing for further decomposition. This is rearranged in~\cref{eq:reverse-kl-2-e} to separate model and variational terms. \cref{eq:reverse-kl-2-f} then expands the log expression, revealing the reverse KL divergence between $q_{\phi}(\B{z}, \B{y} | \B{u})$ and $p_{\theta}(\B{z}, \B{y} | \B{u})$. The inequality in~\cref{eq:reverse-kl-2-g} follows by dropping this non-negative KL term, yielding a variational lower bound (ELBO) on the negative log-likelihood.

\item In~\cref{eq:reverse-kl-2-h-new}, we substitute the generative model factorization from \cref{eq:generative} and the inference model factorization from \cref{eq:inference} into the joint and variational distributions.

\item Line~\cref{eq:reverse-kl-2-j-new} converts the logarithm of the product into a sum of logarithms. In~\cref{eq:reverse-kl-2-i-new}, we interchange the expectation and summation operators, and factor the expectation over the categorical and Gaussian latent variables.

\item Finally,~\cref{eq:reverse-kl-2-h} through~\cref{eq:reverse-kl-2-j} decompose this lower bound into three interpretable components by expanding the logarithmic expression and regrouping terms: a per-scale reconstruction loss (ensuring the decoder can reconstruct the input scattering spectra), a categorical KL divergence for $y_i$ (regularizing the cluster assignment probabilities), and a Gaussian mixture KL divergence for $z_i$ (regularizing the latent representations within each cluster), resulting in the final form of the objective function presented in the main text.
\end{itemize}
\begin{subequations}
\label{eq:reverse-kl-2-derivation}
\begin{align}
& \min_{\Bs{\theta}}  \Exp_{\B{u} \sim  p(\B{u}) }\,\big[-\log p_{\theta}(\B{u}) \big]
\label{eq:reverse-kl-2-a} \\
= & \min_{\Bs{\theta}, \Bs{\phi}}  \Exp_{
\substack{\B{u} \sim  p(\B{u})\\[-1pt]
\B{z}, \B{y} \sim  q_{\phi}(\B{z}, \B{y} | \B{u})}}  \big[-\log  p_{\theta}(\B{u}) \big]
\label{eq:reverse-kl-2-b} \\
= & \min_{\Bs{\theta}, \Bs{\phi}}  \Exp_{
\substack{\B{u} \sim  p(\B{u})\\[-1pt]
\B{z}, \B{y} \sim  q_{\phi}(\B{z}, \B{y} | \B{u})}} \Big[-\log  \frac{p_{\theta}(\B{u}, \B{y}, \B{z})}{p_{\theta}(\B{z}, \B{y} | \B{u})} \Big]
\label{eq:reverse-kl-2-c} \\
= & \min_{\Bs{\theta}, \Bs{\phi}}  \Exp_{
\substack{\B{u} \sim  p(\B{u})\\[-1pt]
\B{z}, \B{y} \sim  q_{\phi}(\B{z}, \B{y} | \B{u})}} \Big[-\log   \frac{q_{\phi}(\B{z}, \B{y} | \B{u})}{q_{\phi}(\B{z}, \B{y} | \B{u})} \frac{p_{\theta}(\B{u}, \B{y}, \B{z})}{p_{\theta}(\B{z}, \B{y} | \B{u})}\Big]
\label{eq:reverse-kl-2-d} \\
= & \min_{\Bs{\theta}, \Bs{\phi}}  \Exp_{
\substack{\B{u} \sim  p(\B{u})\\[-1pt]
\B{z}, \B{y} \sim  q_{\phi}(\B{z}, \B{y} | \B{u})}} \Big[-\log  \frac{q_{\phi}(\B{z}, \B{y} | \B{u})}{p_{\theta}(\B{z}, \B{y} | \B{u})} \frac{p_{\theta}(\B{u}, \B{y}, \B{z})}{q_{\phi}(\B{z}, \B{y} | \B{u})} \Big]
\label{eq:reverse-kl-2-e} \\
= & \min_{\Bs{\theta}, \Bs{\phi}}  \Exp_{
\substack{\B{u} \sim  p(\B{u})\\[-1pt]
\B{z}, \B{y} \sim  q_{\phi}(\B{z}, \B{y} | \B{u})}} \Big[\log  \frac{p_{\theta}(\B{z}, \B{y} | \B{u})}{q_{\phi}(\B{z}, \B{y} | \B{u})} -\log  \frac{p_{\theta}(\B{u}, \B{y}, \B{z})}{q_{\phi}(\B{z}, \B{y} | \B{u})}\Big]
\label{eq:reverse-kl-2-f} \\
\leq & \min_{\Bs{\theta}, \Bs{\phi}}  \Exp_{
\substack{\B{u} \sim  p(\B{u})\\[-1pt]
\B{z}, \B{y} \sim  q_{\phi}(\B{z}, \B{y} | \B{u})}} \Big[-\log  \frac{p_{\theta}(\B{u}, \B{y}, \B{z})}{q_{\phi}(\B{z}, \B{y} | \B{u})} \Big]
\label{eq:reverse-kl-2-g} \\
= & \min_{\Bs{\theta}, \Bs{\phi}}  \Exp_{
\substack{\B{u} \sim  p(\B{u})\\[-1pt]
\B{z}, \B{y} \sim  q_{\phi}(\B{z}, \B{y} | \B{u})}} \Big[-\log  \frac{\prod_{i=0}^{s-1} p_{\theta}(\B{u}_i | \B{z}_i)\,p_{\theta}(\B{z}_i | y_i)\,p_{\theta}(y_i)}{\prod_{i=0}^{s-1} q_{\phi}(\B{z}_i | y_i, \B{u})\,q_{\phi}(y_i | \B{u})} \Big]
\label{eq:reverse-kl-2-h-new} \\
= & \min_{\Bs{\theta}, \Bs{\phi}}  \Exp_{
\substack{\B{u} \sim  p(\B{u})\\[-1pt]
\B{z}, \B{y} \sim  q_{\phi}(\B{z}, \B{y} | \B{u})}} \Big[-\sum_{i=0}^{s-1} \log  \frac{p_{\theta}(\B{u}_i | \B{z}_i)\,p_{\theta}(\B{z}_i | y_i)\,p_{\theta}(y_i)}{q_{\phi}(\B{z}_i | y_i, \B{u})\,q_{\phi}(y_i | \B{u})} \Big]
\label{eq:reverse-kl-2-j-new} \\
= & \min_{\Bs{\theta}, \Bs{\phi}}  \sum_{i=0}^{s-1} \Exp_{\B{u} \sim  p(\B{u}) }\, \Big[ \Exp_{
\substack{y_i \sim  q_{\phi}(y_i | \B{u})\\[-1pt]
\B{z}_i \sim  q_{\phi}(\B{z}_i | y_i, \B{u})}} \\
&\quad \Big[ -\log  \frac{p_{\theta}(\B{u}_i | \B{z}_i)\,p_{\theta}(\B{z}_i | y_i)\,p_{\theta}(y_i)}{q_{\phi}(\B{z}_i | y_i, \B{u})\,q_{\phi}(y_i | \B{u})} \Big] \Big]
\label{eq:reverse-kl-2-i-new} \\
= & \min_{\Bs{\theta}, \Bs{\phi}}  \sum_{i=0}^{s-1} \Exp_{\B{u} \sim  p(\B{u}) }\, \Big[ \Exp_{ \B{z}_i \sim  q_{\phi}(\B{z}_i | y_i, \B{u})}\, \underbrace{\big[ -\log p_\theta(\B{u}_i | \B{z}_i) \big] }_{\substack{
    \text{Per-scale reconstruction loss}}}
\label{eq:reverse-kl-2-h} \\
& \quad + \underbrace{\KL\,\big( q_{\phi}(y_i | \B{u}) \,||\, p_{\theta}(y_i) \big)}_{\substack{
    \text{Categorical prior on } y_i}}
\label{eq:reverse-kl-2-i} \\
& \quad +  \Exp_{ y_i \sim  q_{\phi}(y_i | \B{u}) }\, \underbrace{\big[ \KL\,\big( q_{\phi}(\B{z}_i | y_i, \B{u}) \,||\,  p_\theta(\B{z}_i | y_i)  \big) \big] }_{\substack{
    \text{Gaussian mixture prior on } \B{z}_i}} \Big]
\label{eq:reverse-kl-2-j}
\end{align}
\end{subequations}

\section{Optimal value of source separation objective function}
\label{app:optimal_value}

As a quantitative measure of the source separation performance, we derive the optimal value of the objective function in \cref{eq:ss_optimization}. Assuming that the algorithm successfully separated the source $s_1$, one can derive a typical value for the loss $\mathcal{L}=\mathcal{L}_\text{prior} + \mathcal{L}_\text{cross} + \mathcal{L}_\text{data}$.
This value can be used to quantify the separation in our algorithm.

Working term by term, the first term writes
\begin{equation*}
\mathcal{L}_\text{prior}(\B{s}_1) = \sum_{i=1}^N
\frac{\Big\|\Psi_k\big(\B{s}_1\big) - \Psi_k\big(\B{s}^i_1\big)\Big\|_2^2}{\sigma^2\big(\Psi_k\big(\B{s}^i_1\big)\big)}.
\end{equation*}
In the best case, our retrieved statistics $\Psi(s_1)$ have the same mean and the same variance as the observed population of statistics $\{\Psi(s^i_1)\}_{i=1}^N$. Furthermore, we assume that the population of statistics $\{\Psi_k(s^{i}_1)\}_{i=1}^N$ follows a multivariate independent Gaussian that is independent from $\Psi(s_1)$.
% \ram{Mention the analogy with \chi^2}
In that case, each summed term in $\mathcal{L}_\text{prior}(\B{s}_1)$ follows twice a $\chi^2$ distribution with $r$ degrees of freedom, where $r$ is the number of coefficients in $\Psi$. The term $\mathcal{L}_\text{prior}(\B{s}_1)$ is thus equal to $2Nr$ in expectation (or $2N$ if the $\|\cdot\|_2$ is normalized by $r$).

Regarding the third term
\begin{equation*}
\mathcal{L}_\text{data}(\B{s}_1) =  \sum_{i=1}^N \frac{\Big\| \Psi_k\left(\B{x} - \B{s}_1 + \B{s}^i_1 \right) - \Psi_k\left(\B{x}\right)\Big\|^2_2}{\sigma^2\big(\Psi_k\big(\B{x}+\B{s}^i_1\big)\big)}
\end{equation*}
if we assume that the variance of the coefficients $\sigma^2\big(\Psi_k\big(\B{x}+\B{s}^i_1\big)$ is close to the variance of the same coefficients evaluated on the mixture signal $\sigma^2\big(\Psi_k\big(\B{x}\big)\big)$, then, one can derive similarly a typical value of $2Nr$ in case of successful separation.

The second term
\begin{equation*}
\mathcal{L}_\text{cross}(\B{s}_1) =  \sum_{i=1}^N \frac{\Big\| \Psi_k\left(\B{s}^i_1, \B{x}-\B{s}_1 \right)\Big\|^2_2}{\sigma^2\big(\Psi_k\big(\B{s}^i_1,\B{x}\big)\big)},
\end{equation*}
can be assumed to be negligible against the other ones in case of successful separation, where the terms $s_1^i$ and $x-s_1$ should be decorrelated.

In the case of successful separation, the loss $\mathcal{L}$ should thus be equal to $4Nr$ in expectation.

\section{All identified clusters and their representative waveforms}
\label{app:all_results}

\begin{figure*}[p]
    \centering

    \begin{subfigure}[t]{0.4\textwidth}
        \centering
        \includegraphics[width=\textwidth]{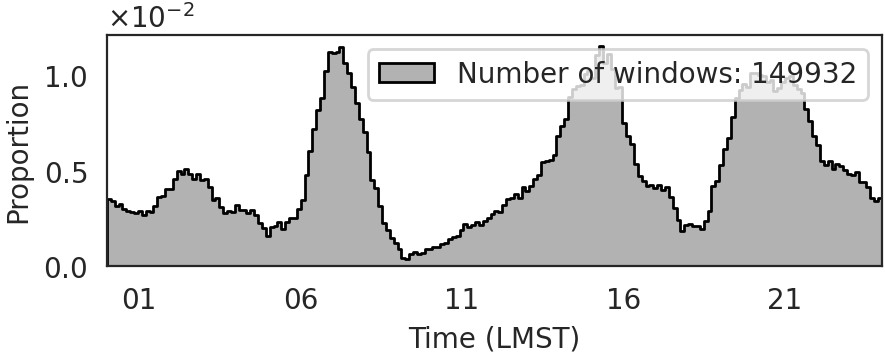}
        \caption{}
        \label{fig:app_histogram_cluster-0_scale-1}
        \end{subfigure}\hspace{1em}
    \begin{subfigure}[t]{0.4\textwidth}
        \centering
        \includegraphics[width=\textwidth]{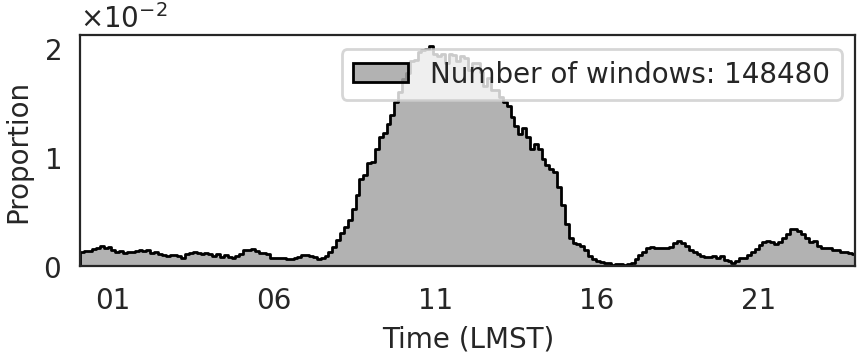}
        \caption{}
        \label{fig:app_histogram_cluster-1_scale-1}
    \end{subfigure}

    \begin{subfigure}[t]{0.4\textwidth}
        \centering
        \includegraphics[width=\textwidth]{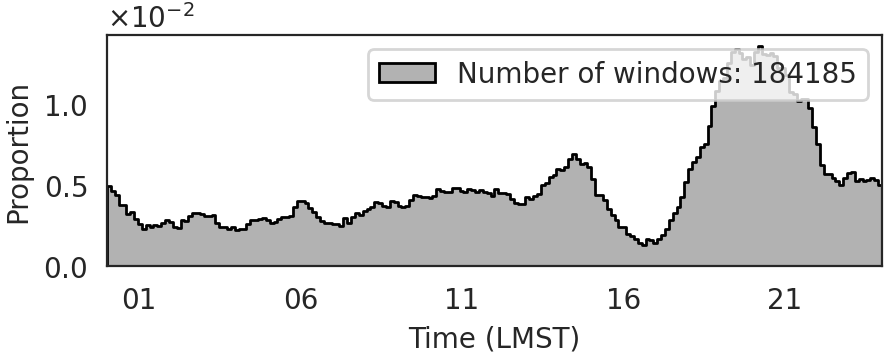}
        \caption{}
        \label{fig:app_histogram_cluster-2_scale-1}
        \end{subfigure}\hspace{1em}
    \begin{subfigure}[t]{0.4\textwidth}
        \centering
        \includegraphics[width=\textwidth]{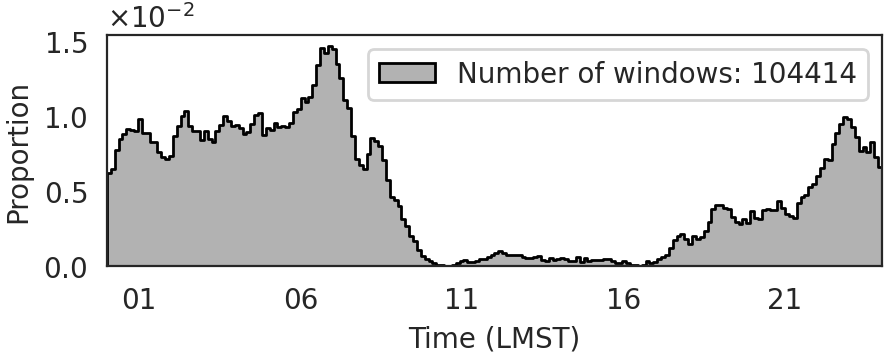}
        \caption{}
        \label{fig:app_histogram_cluster-3_scale-1}
    \end{subfigure}

    \begin{subfigure}[t]{0.4\textwidth}
        \centering
        \includegraphics[width=\textwidth]{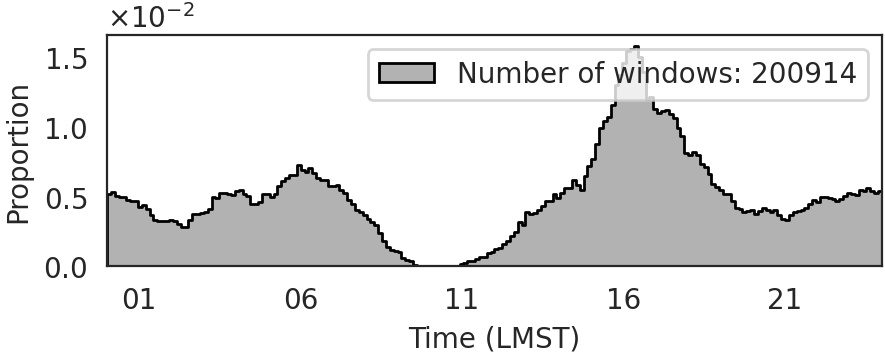}
        \caption{}
        \label{fig:app_histogram_cluster-4_scale-1}
        \end{subfigure}\hspace{1em}
    \begin{subfigure}[t]{0.4\textwidth}
        \centering
        \includegraphics[width=\textwidth]{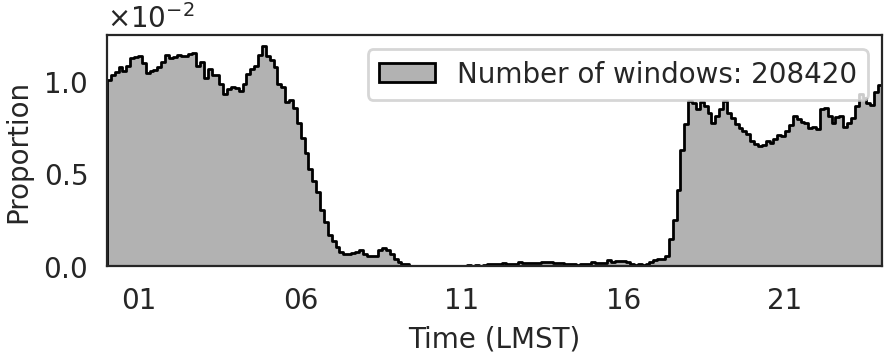}
        \caption{}
        \label{fig:app_histogram_cluster-5_scale-1}
    \end{subfigure}

    \begin{subfigure}[t]{0.4\textwidth}
        \centering
        \includegraphics[width=\textwidth]{figs/scale_1024/time_histograms/time_histogram_cluster-5.png}
        \caption{}
        \label{fig:app_histogram_cluster-6_scale-1}
        \end{subfigure}\hspace{1em}
    \begin{subfigure}[t]{0.4\textwidth}
        \centering
        \includegraphics[width=\textwidth]{figs/scale_1024/time_histograms/time_histogram_cluster-7.png}
        \caption{}
        \label{fig:app_histogram_cluster-7_scale-1}
    \end{subfigure}

    \begin{subfigure}[t]{0.4\textwidth}
        \centering
        \includegraphics[width=\textwidth]{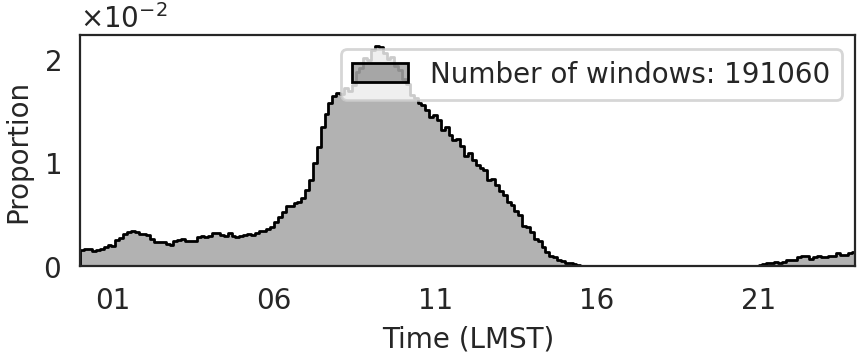}
        \caption{}
        \label{fig:app_histogram_cluster-8_scale-1}
        \end{subfigure}

    \caption{The occurrence time histogram of all nine identified clusters within the $51.2$-second timescale. The horizontal axis of the histograms represents the local mean solar time (LMST). Clusters 0--9 are shown in \cref{fig:app_histogram_cluster-0_scale-1,fig:app_histogram_cluster-1_scale-1,fig:app_histogram_cluster-2_scale-1,fig:app_histogram_cluster-3_scale-1,fig:app_histogram_cluster-4_scale-1,fig:app_histogram_cluster-5_scale-1,fig:app_histogram_cluster-6_scale-1,fig:app_histogram_cluster-7_scale-1,fig:app_histogram_cluster-8_scale-1}, respectively.}
    \label{fig:app_histogram_clusters_scale-1}
\end{figure*}

\begin{figure*}[p]
    \centering

    \begin{subfigure}[t]{0.495\textwidth}
        \centering
        \includegraphics[width=\textwidth]{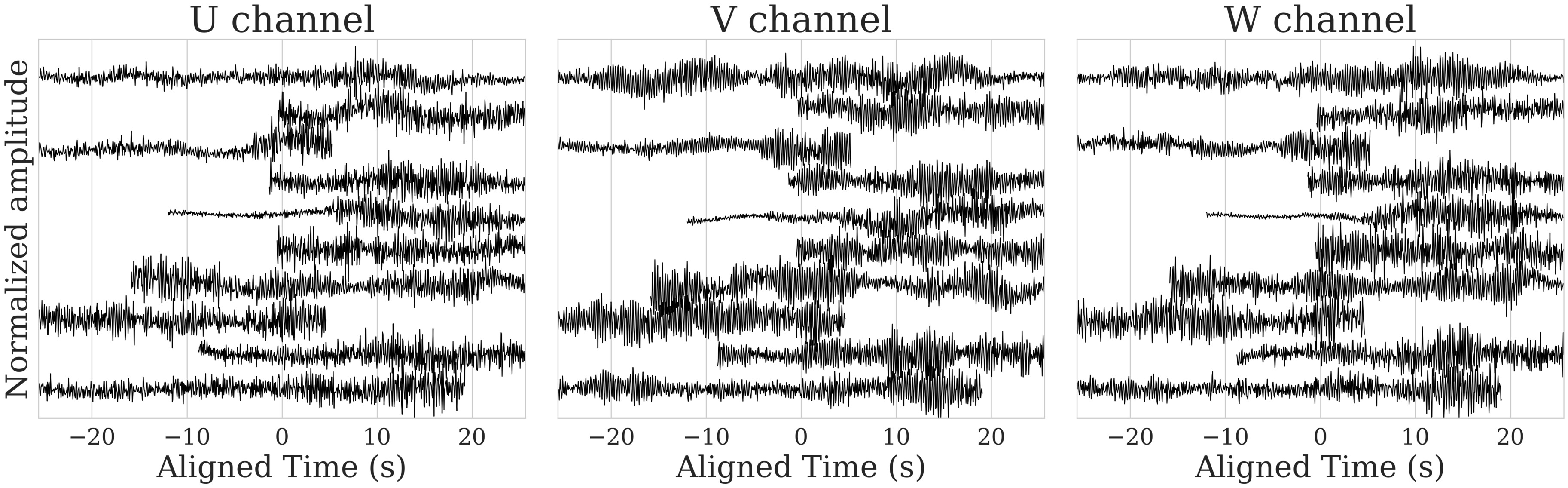}
        \caption{}
        \label{fig:app_cluster-0_scale-1}
        \end{subfigure}\hspace{0.1em}
    \begin{subfigure}[t]{0.495\textwidth}
        \centering
        \includegraphics[width=\textwidth]{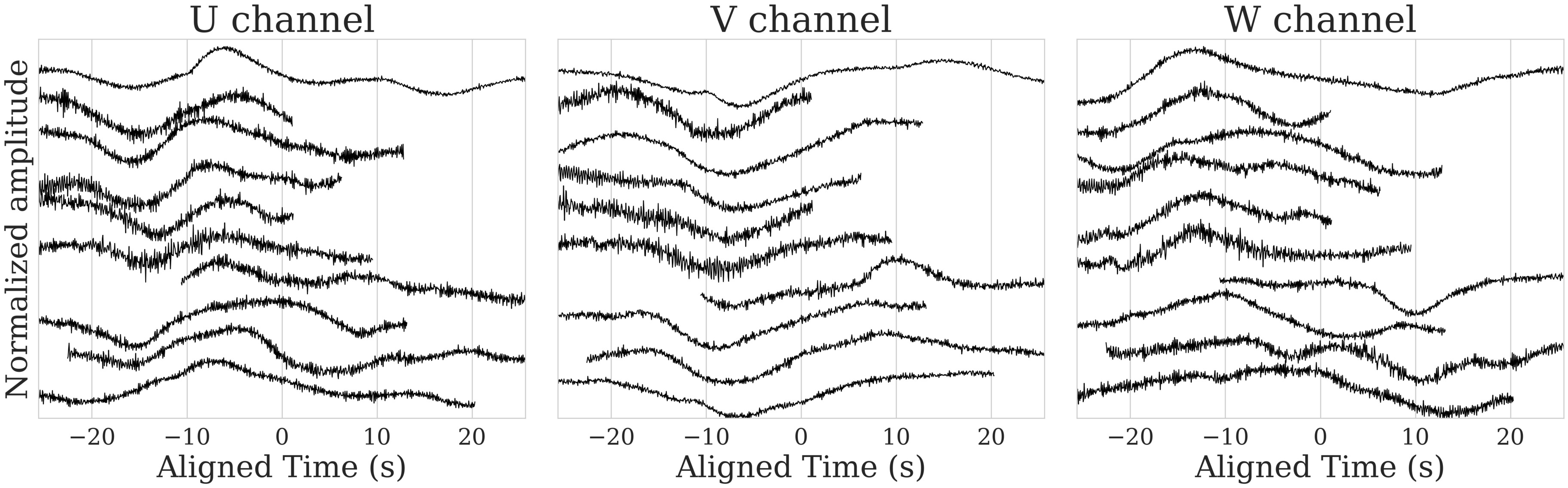}
        \caption{}
        \label{fig:app_cluster-1_scale-1}
    \end{subfigure}

    \begin{subfigure}[t]{0.495\textwidth}
        \centering
        \includegraphics[width=\textwidth]{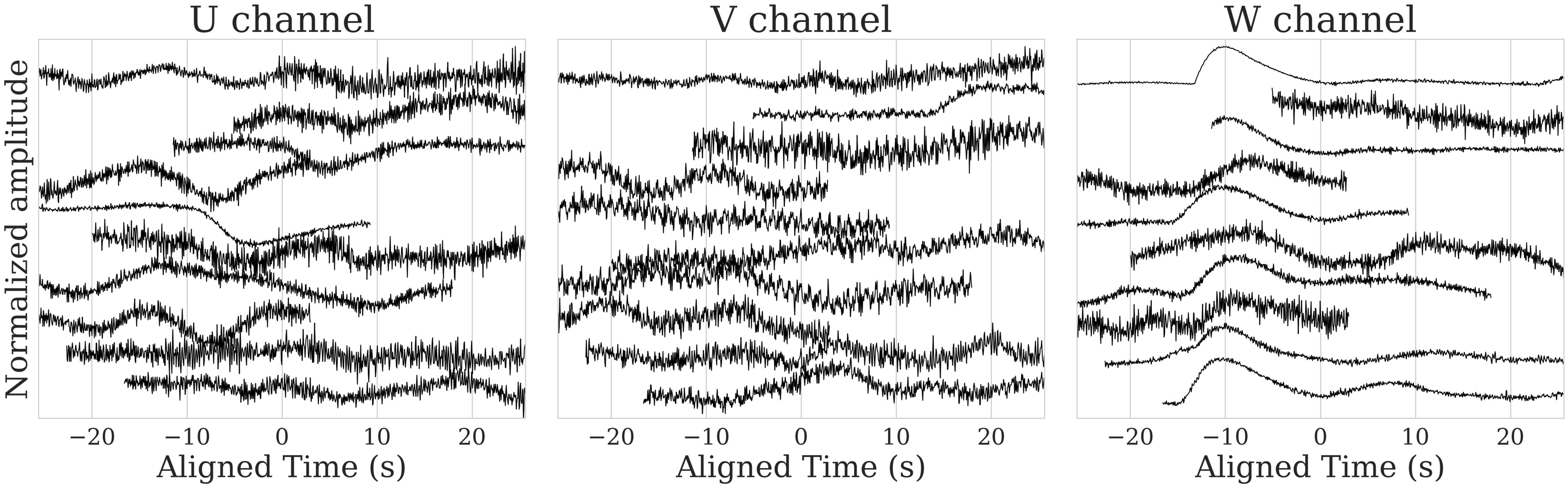}
        \caption{}
        \label{fig:app_cluster-2_scale-1}
      \end{subfigure}\hspace{0.1em}
    \begin{subfigure}[t]{0.495\textwidth}
        \centering
        \includegraphics[width=\textwidth]{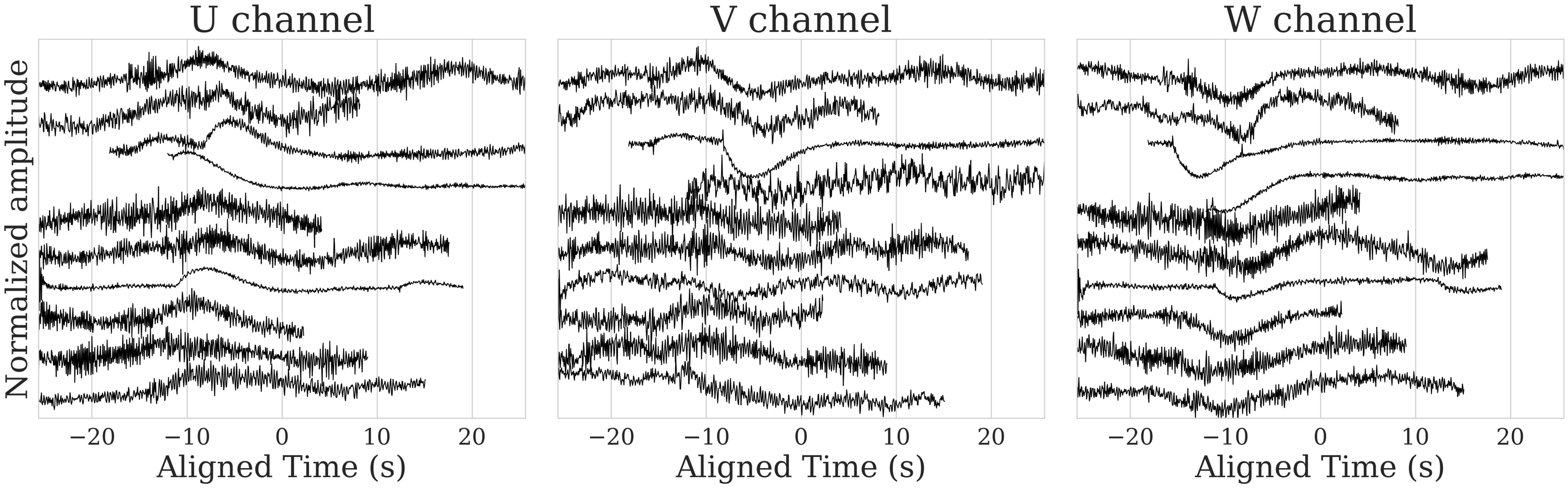}
        \caption{}
        \label{fig:app_cluster-3_scale-1}
      \end{subfigure}

    \begin{subfigure}[t]{0.495\textwidth}
        \centering
        \includegraphics[width=\textwidth]{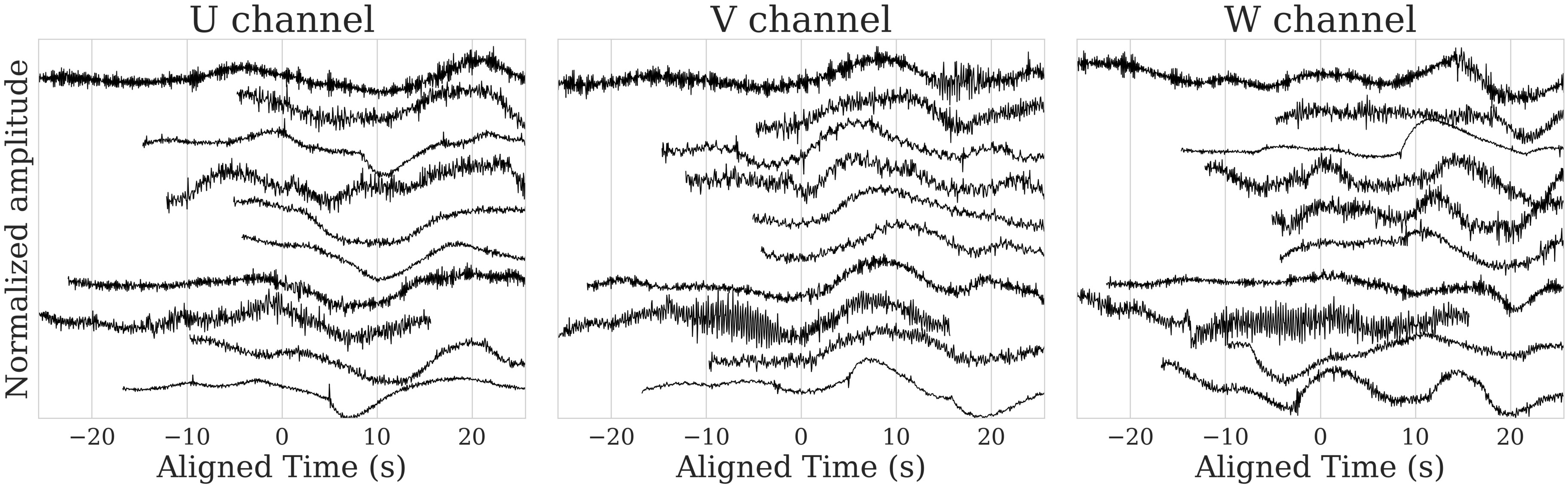}
        \caption{}
        \label{fig:app_cluster-4_scale-1}
      \end{subfigure}\hspace{0.1em}
    \begin{subfigure}[t]{0.495\textwidth}
        \centering
        \includegraphics[width=\textwidth]{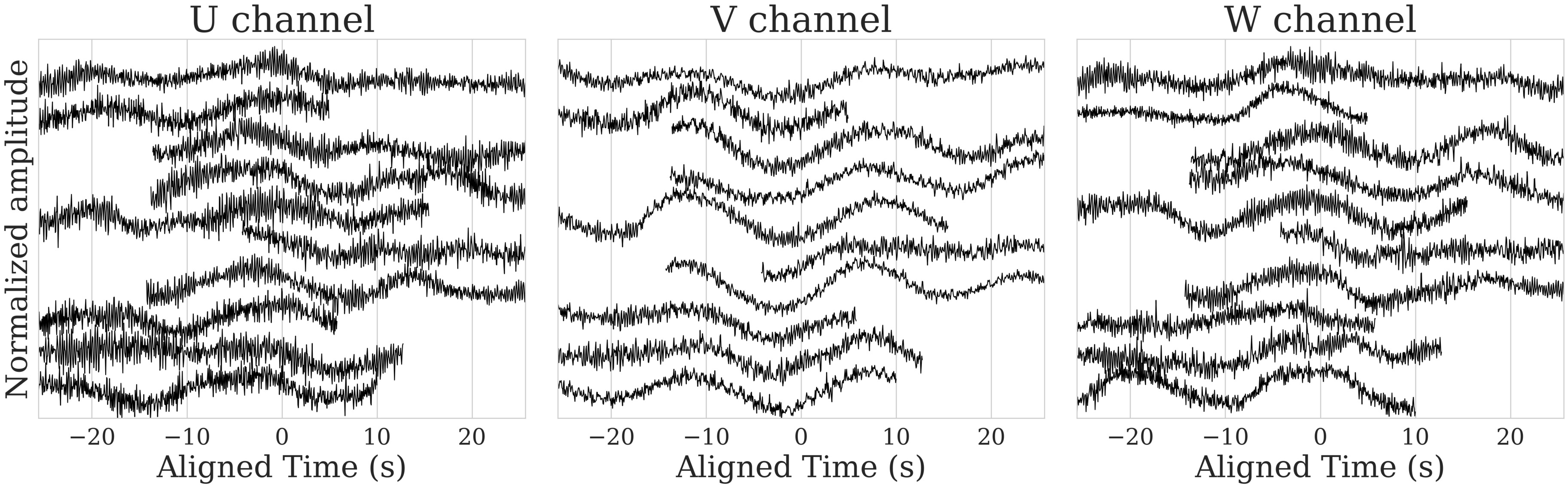}
        \caption{}
        \label{fig:app_cluster-5_scale-1}
      \end{subfigure}

    \begin{subfigure}[t]{0.495\textwidth}
        \centering
        \includegraphics[width=\textwidth]{figs/scale_1024/aligned_waveforms/aligned_waveforms_cluster_5.jpg}
        \caption{}
        \label{fig:app_cluster-6_scale-1}
      \end{subfigure}\hspace{0.1em}
    \begin{subfigure}[t]{0.495\textwidth}
        \centering
        \includegraphics[width=\textwidth]{figs/scale_1024/aligned_waveforms/aligned_waveforms_cluster_7.jpg}
        \caption{}
        \label{fig:app_cluster-7_scale-1}
      \end{subfigure}

    \begin{subfigure}[t]{0.495\textwidth}
        \centering
        \includegraphics[width=\textwidth]{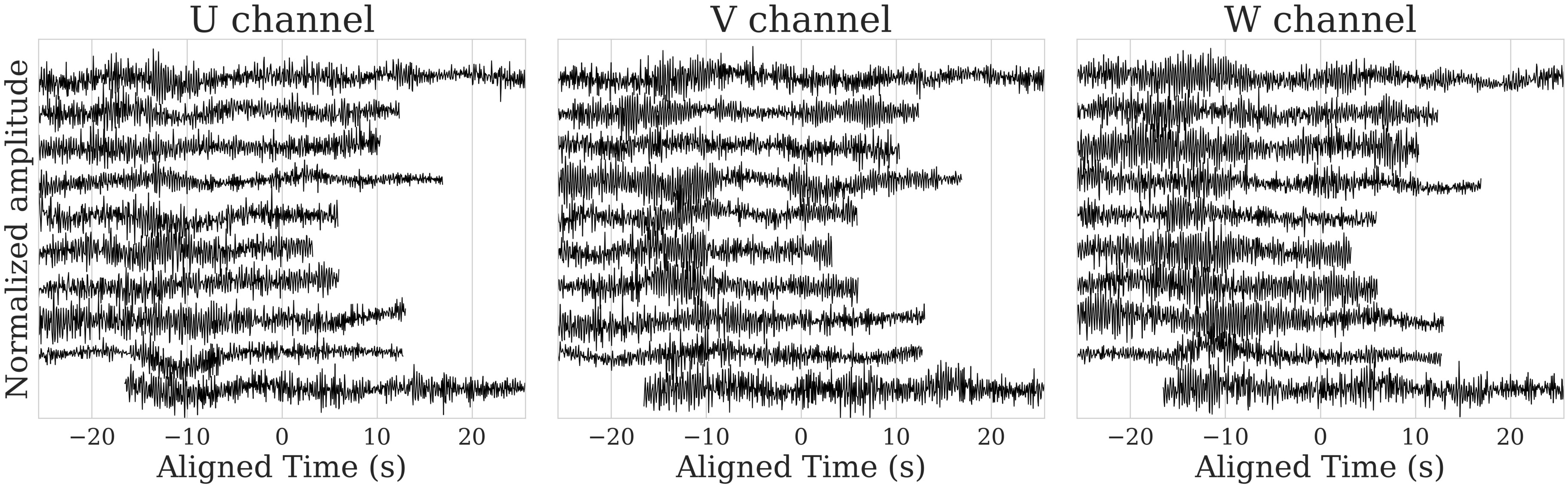}
        \caption{}
        \label{fig:app_cluster-8_scale-1}
      \end{subfigure}

    \caption{The aligned waveforms of all nine identified clusters within the $51.2$-second timescale. Clusters 0--9 are shown in \cref{fig:app_cluster-0_scale-1,fig:app_cluster-1_scale-1,fig:app_cluster-2_scale-1,fig:app_cluster-3_scale-1,fig:app_cluster-4_scale-1,fig:app_cluster-5_scale-1,fig:app_cluster-6_scale-1,fig:app_cluster-7_scale-1,fig:app_cluster-8_scale-1}, respectively.}
    \label{fig:app_clusters_scale-1}
\end{figure*}

\begin{figure*}[p]
    \centering

    \begin{subfigure}[t]{0.4\textwidth}
        \centering
        \includegraphics[width=\textwidth]{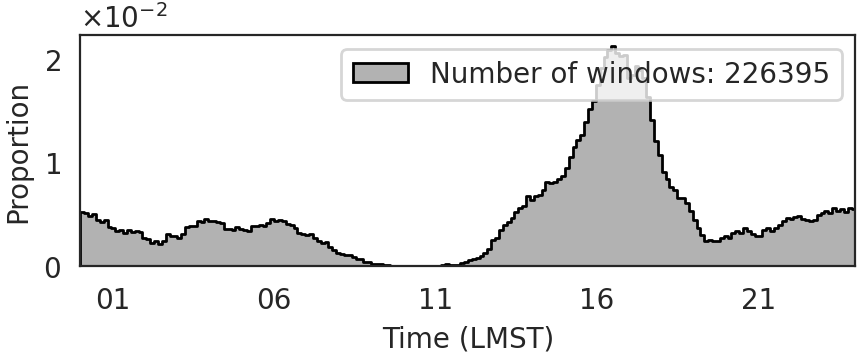}
        \caption{}
        \label{fig:app_histogram_cluster-0_scale-2}
      \end{subfigure}\hspace{1em}
    \begin{subfigure}[t]{0.4\textwidth}
        \centering
        \includegraphics[width=\textwidth]{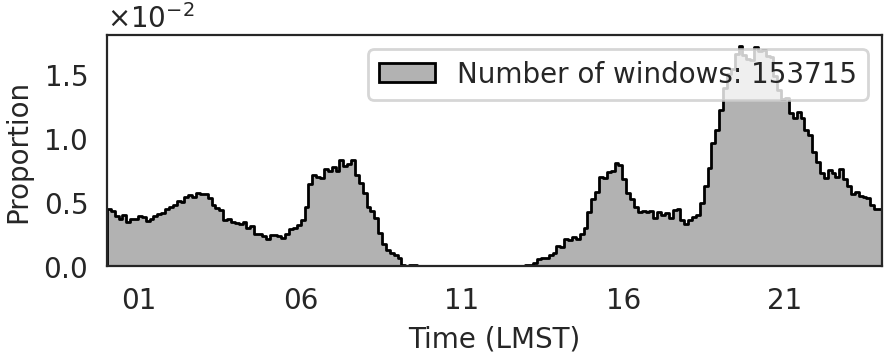}
        \caption{}
        \label{fig:app_histogram_cluster-1_scale-2}
      \end{subfigure}

    \begin{subfigure}[t]{0.4\textwidth}
        \centering
        \includegraphics[width=\textwidth]{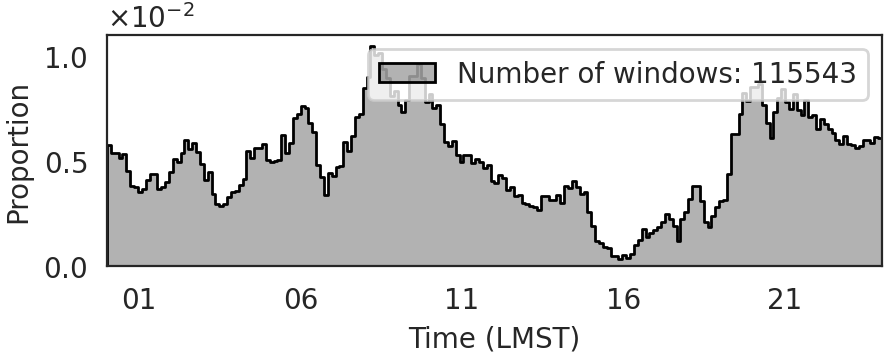}
        \caption{}
        \label{fig:app_histogram_cluster-2_scale-2}
      \end{subfigure}\hspace{1em}
    \begin{subfigure}[t]{0.4\textwidth}
        \centering
        \includegraphics[width=\textwidth]{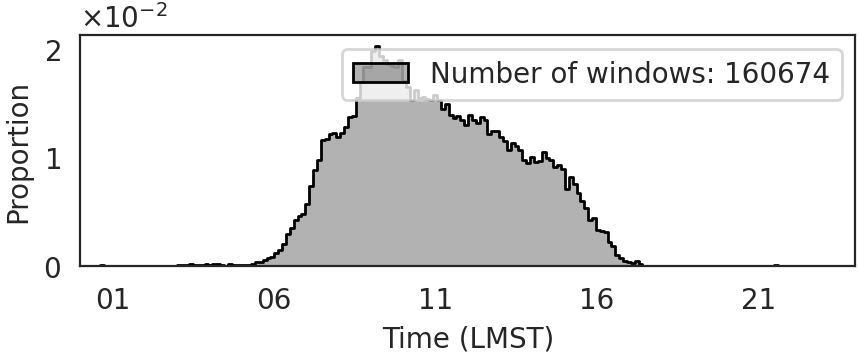}
        \caption{}
        \label{fig:app_histogram_cluster-3_scale-2}
      \end{subfigure}

    \begin{subfigure}[t]{0.4\textwidth}
        \centering
        \includegraphics[width=\textwidth]{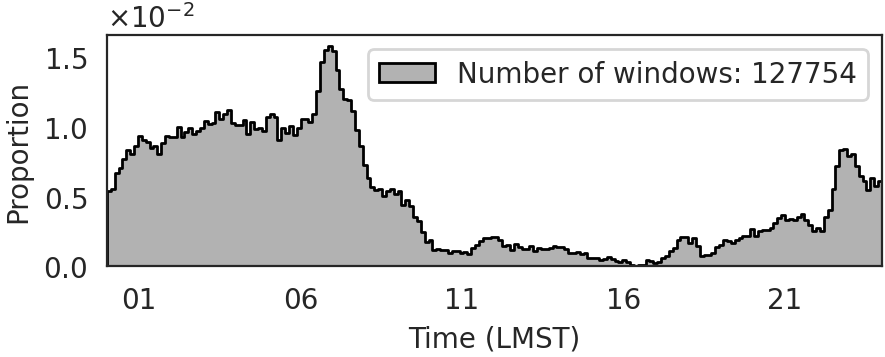}
        \caption{}
        \label{fig:app_histogram_cluster-4_scale-2}
      \end{subfigure}\hspace{1em}
    \begin{subfigure}[t]{0.4\textwidth}
        \centering
        \includegraphics[width=\textwidth]{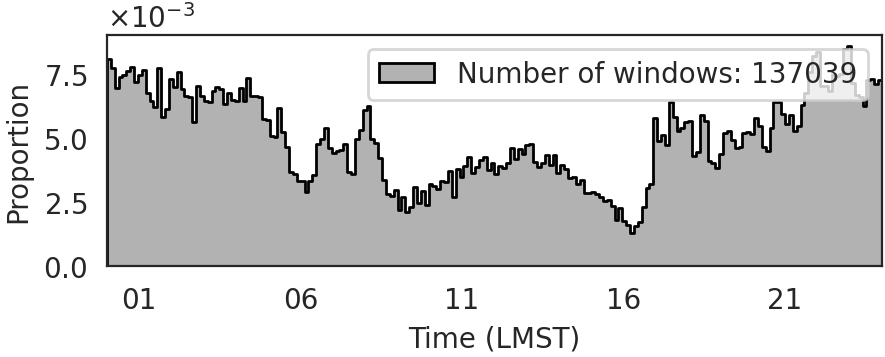}
        \caption{}
        \label{fig:app_histogram_cluster-5_scale-2}
      \end{subfigure}

    \begin{subfigure}[t]{0.4\textwidth}
        \centering
        \includegraphics[width=\textwidth]{figs/scale_4096/time_histograms/time_histogram_cluster-5.png}
        \caption{}
        \label{fig:app_histogram_cluster-6_scale-2}
      \end{subfigure}\hspace{1em}
    \begin{subfigure}[t]{0.4\textwidth}
        \centering
        \includegraphics[width=\textwidth]{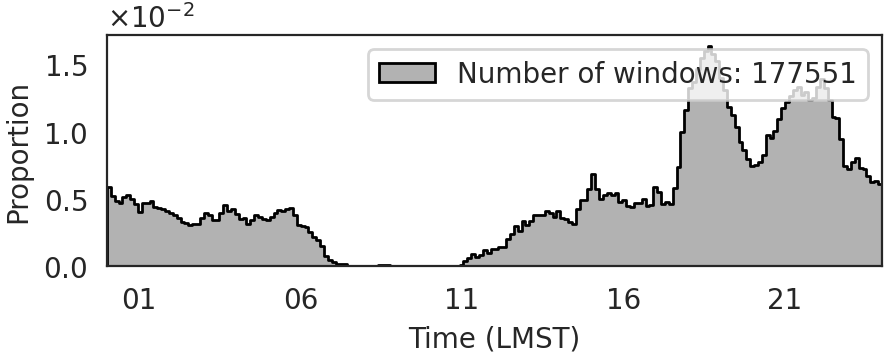}
        \caption{}
        \label{fig:app_histogram_cluster-7_scale-2}
      \end{subfigure}

    \begin{subfigure}[t]{0.4\textwidth}
        \centering
        \includegraphics[width=\textwidth]{figs/scale_4096/time_histograms/time_histogram_cluster-8.png}
        \caption{}
        \label{fig:app_histogram_cluster-8_scale-2}
      \end{subfigure}

    \caption{The occurrence time histogram of all nine identified clusters within the $3.4$-minute timescale. The horizontal axis of the histograms represents the local mean solar time (LMST). Clusters 0--9 are shown in \cref{fig:app_histogram_cluster-0_scale-2,fig:app_histogram_cluster-1_scale-2,fig:app_histogram_cluster-2_scale-2,fig:app_histogram_cluster-3_scale-2,fig:app_histogram_cluster-4_scale-2,fig:app_histogram_cluster-5_scale-2,fig:app_histogram_cluster-6_scale-2,fig:app_histogram_cluster-7_scale-2,fig:app_histogram_cluster-8_scale-2}, respectively.}
    \label{fig:app_histogram_clusters_scale-2}
\end{figure*}

\begin{figure*}[p]
    \centering

    \begin{subfigure}[t]{0.495\textwidth}
        \centering
        \includegraphics[width=\textwidth]{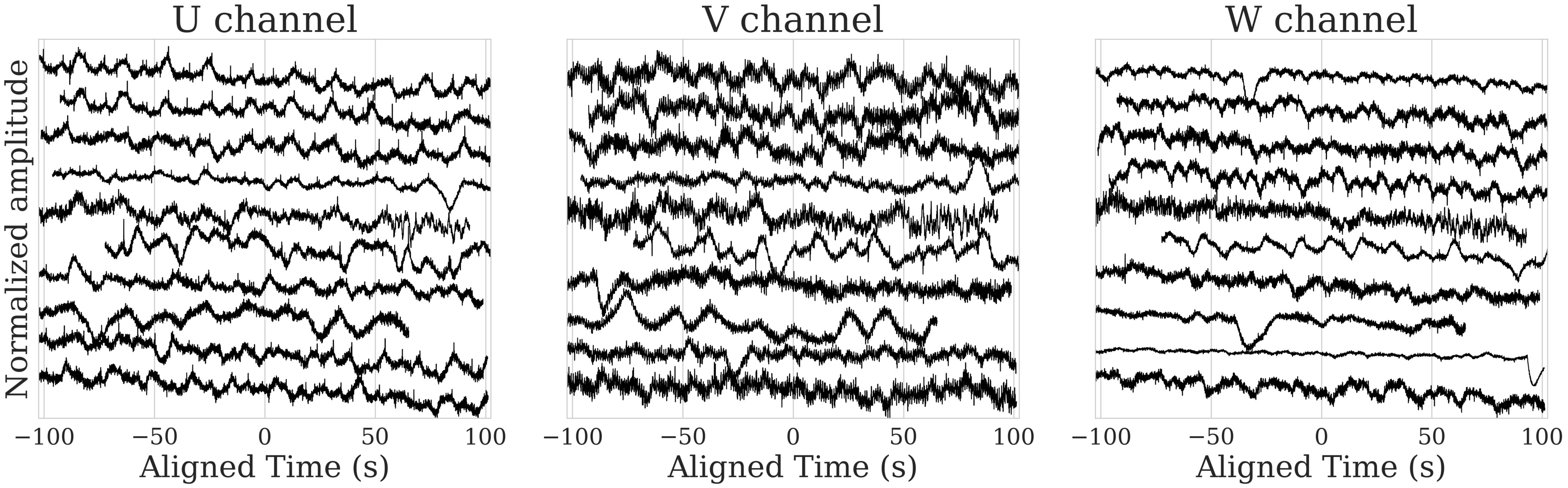}
        \caption{}
        \label{fig:app_cluster-0_scale-2}
      \end{subfigure}\hspace{0.1em}
    \begin{subfigure}[t]{0.495\textwidth}
        \centering
        \includegraphics[width=\textwidth]{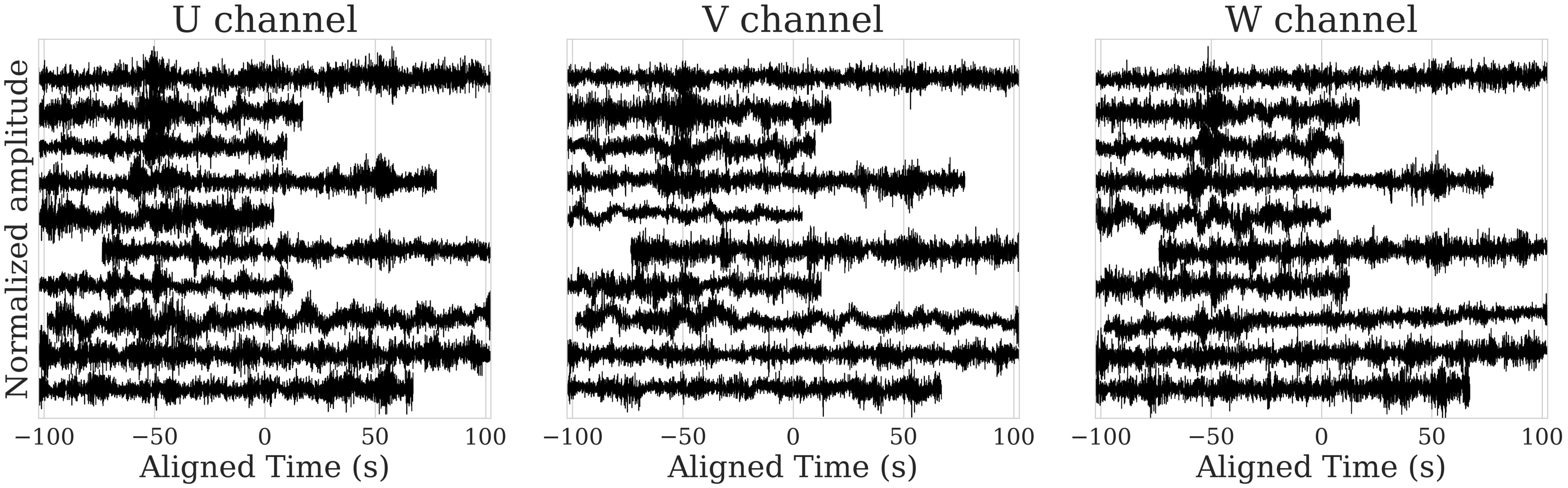}
        \caption{}
        \label{fig:app_cluster-1_scale-2}
      \end{subfigure}

    \begin{subfigure}[t]{0.495\textwidth}
        \centering
        \includegraphics[width=\textwidth]{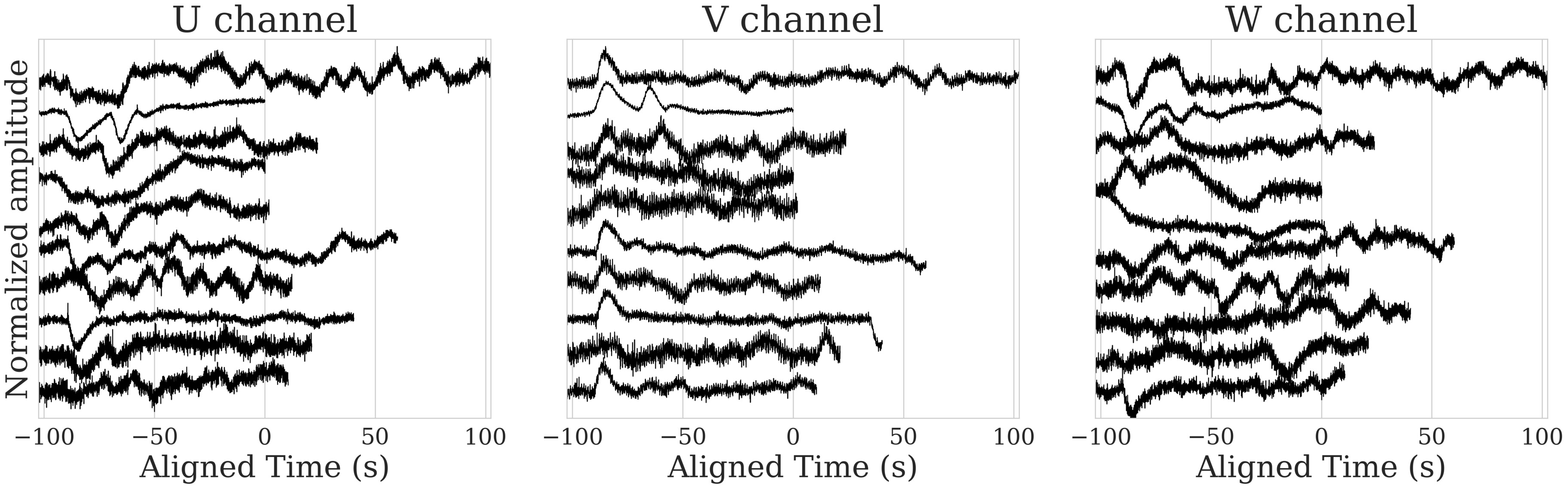}
        \caption{}
        \label{fig:app_cluster-2_scale-2}
      \end{subfigure}\hspace{0.1em}
    \begin{subfigure}[t]{0.495\textwidth}
        \centering
        \includegraphics[width=\textwidth]{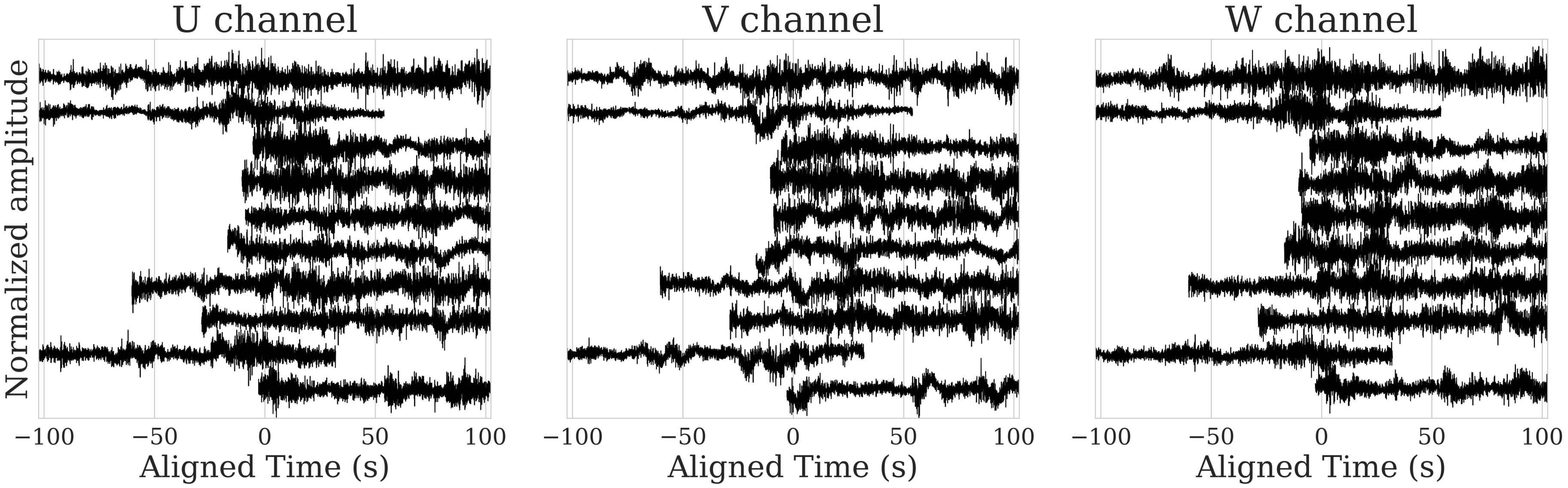}
        \caption{}
        \label{fig:app_cluster-3_scale-2}
      \end{subfigure}

    \begin{subfigure}[t]{0.495\textwidth}
        \centering
        \includegraphics[width=\textwidth]{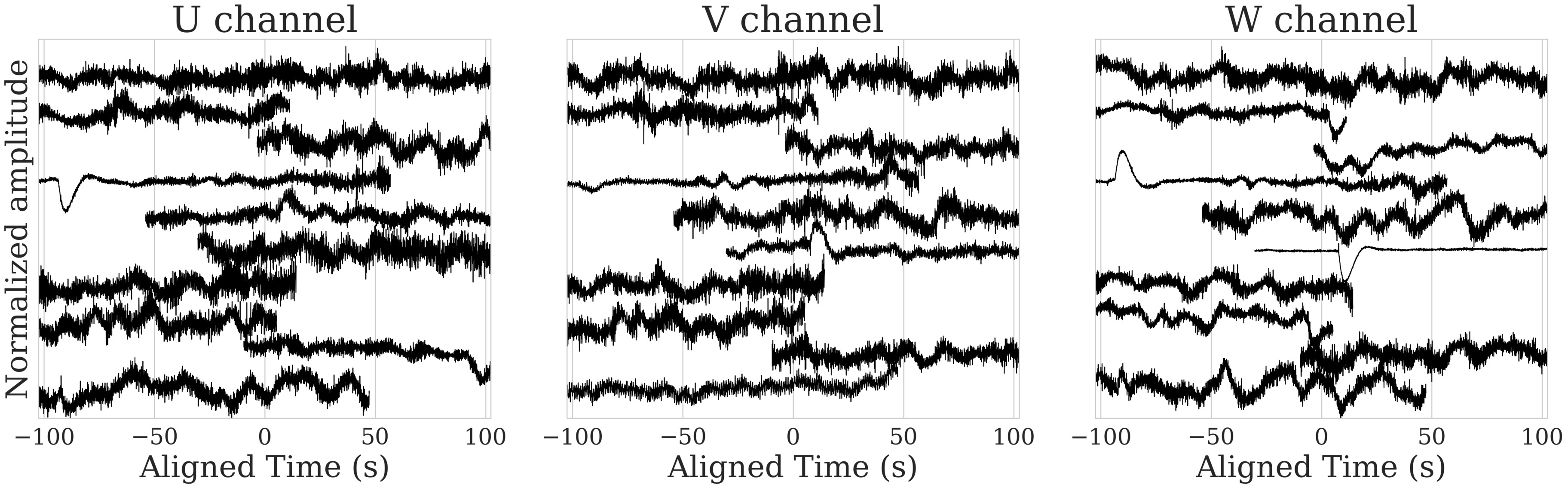}
        \caption{}
        \label{fig:app_cluster-4_scale-2}
      \end{subfigure}\hspace{0.1em}
    \begin{subfigure}[t]{0.495\textwidth}
        \centering
        \includegraphics[width=\textwidth]{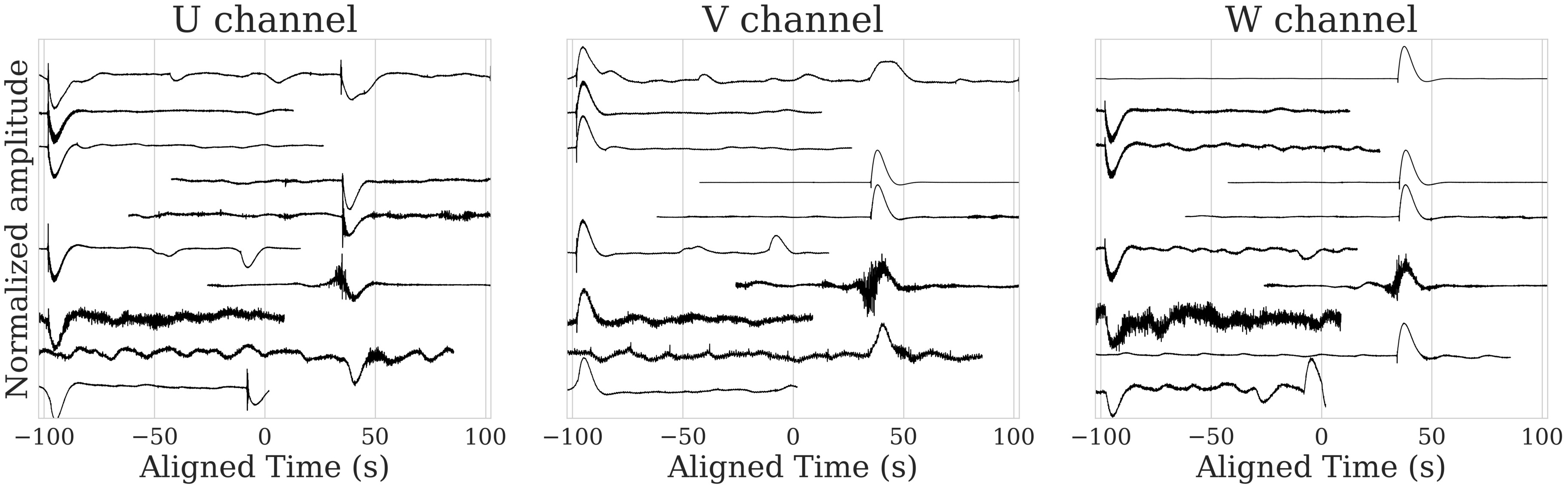}
        \caption{}
        \label{fig:app_cluster-5_scale-2}
      \end{subfigure}

    \begin{subfigure}[t]{0.495\textwidth}
        \centering
        \includegraphics[width=\textwidth]{figs/scale_4096/aligned_waveforms/aligned_waveforms_cluster_5.jpg}
        \caption{}
        \label{fig:app_cluster-6_scale-2}
      \end{subfigure}\hspace{0.1em}
    \begin{subfigure}[t]{0.495\textwidth}
        \centering
        \includegraphics[width=\textwidth]{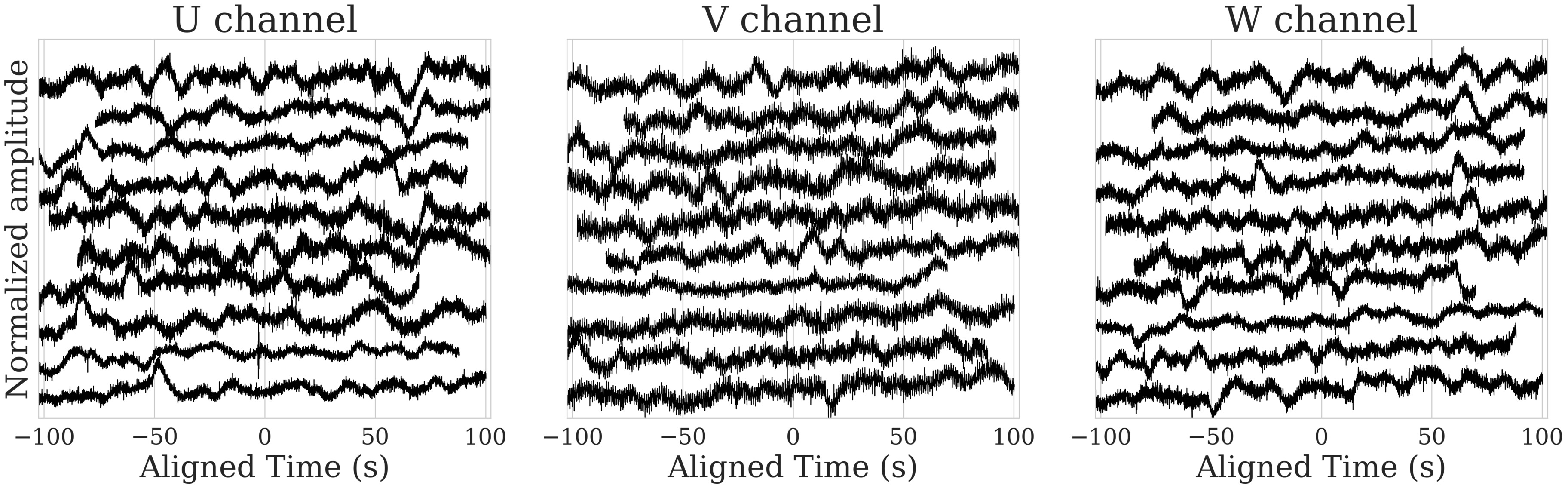}
        \caption{}
        \label{fig:app_cluster-7_scale-2}
      \end{subfigure}

    \begin{subfigure}[t]{0.495\textwidth}
        \centering
        \includegraphics[width=\textwidth]{figs/scale_4096/aligned_waveforms/aligned_waveforms_cluster_8.jpg}
        \caption{}
        \label{fig:app_cluster-8_scale-2}
      \end{subfigure}

    \caption{The aligned waveforms of all nine identified clusters within the $3.4$-minute timescale. Clusters 0--9 are shown in \cref{fig:app_cluster-0_scale-2,fig:app_cluster-1_scale-2,fig:app_cluster-2_scale-2,fig:app_cluster-3_scale-2,fig:app_cluster-4_scale-2,fig:app_cluster-5_scale-2,fig:app_cluster-6_scale-2,fig:app_cluster-7_scale-2,fig:app_cluster-8_scale-2}, respectively.}
    \label{fig:app_clusters_scale-2}
\end{figure*}

\begin{figure*}[p]
    \centering

    \begin{subfigure}[t]{0.4\textwidth}
        \centering
        \includegraphics[width=\textwidth]{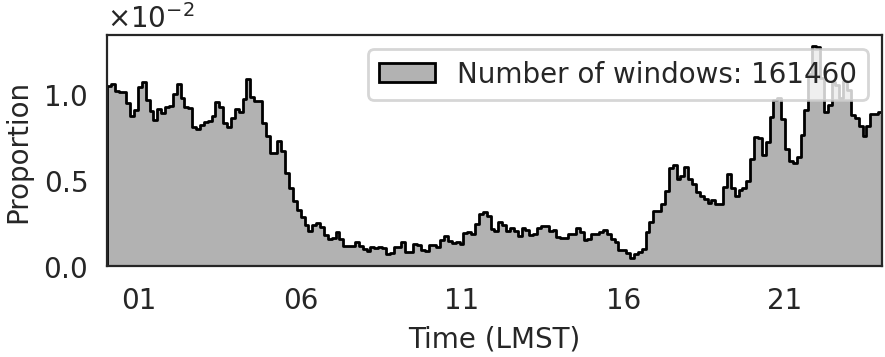}
        \caption{}
        \label{fig:app_histogram_cluster-0_scale-3}
      \end{subfigure}\hspace{1em}
    \begin{subfigure}[t]{0.4\textwidth}
        \centering
        \includegraphics[width=\textwidth]{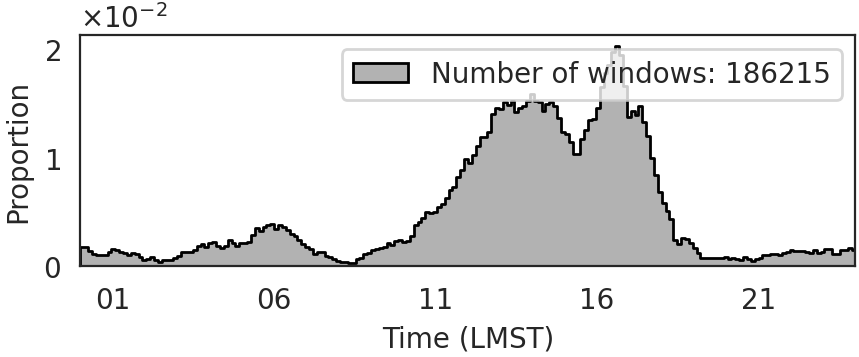}
        \caption{}
        \label{fig:app_histogram_cluster-1_scale-3}
      \end{subfigure}

    \begin{subfigure}[t]{0.4\textwidth}
        \centering
        \includegraphics[width=\textwidth]{figs/scale_16384/time_histograms/time_histogram_cluster-2.png}
        \caption{}
        \label{fig:app_histogram_cluster-2_scale-3}
      \end{subfigure}\hspace{1em}
    \begin{subfigure}[t]{0.4\textwidth}
        \centering
        \includegraphics[width=\textwidth]{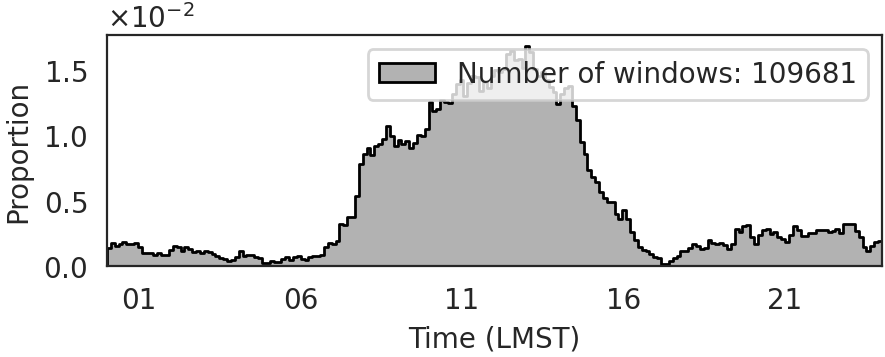}
        \caption{}
        \label{fig:app_histogram_cluster-3_scale-3}
      \end{subfigure}

    \begin{subfigure}[t]{0.4\textwidth}
        \centering
        \includegraphics[width=\textwidth]{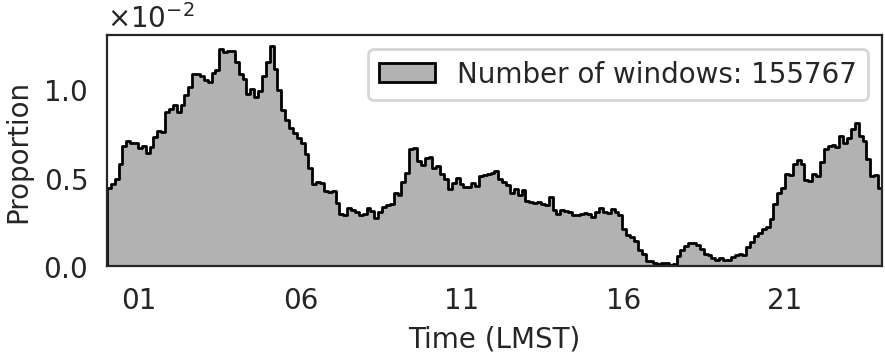}
        \caption{}
        \label{fig:app_histogram_cluster-4_scale-3}
      \end{subfigure}\hspace{1em}
    \begin{subfigure}[t]{0.4\textwidth}
        \centering
        \includegraphics[width=\textwidth]{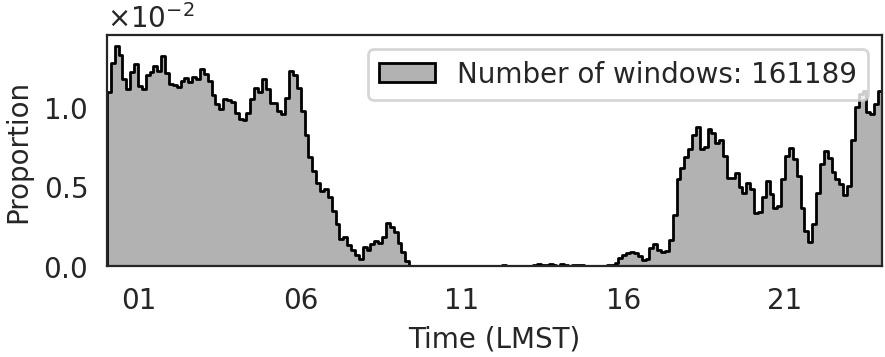}
        \caption{}
        \label{fig:app_histogram_cluster-5_scale-3}
      \end{subfigure}

    \begin{subfigure}[t]{0.4\textwidth}
        \centering
        \includegraphics[width=\textwidth]{figs/scale_16384/time_histograms/time_histogram_cluster-5.png}
        \caption{}
        \label{fig:app_histogram_cluster-6_scale-3}
      \end{subfigure}\hspace{1em}
    \begin{subfigure}[t]{0.4\textwidth}
        \centering
        \includegraphics[width=\textwidth]{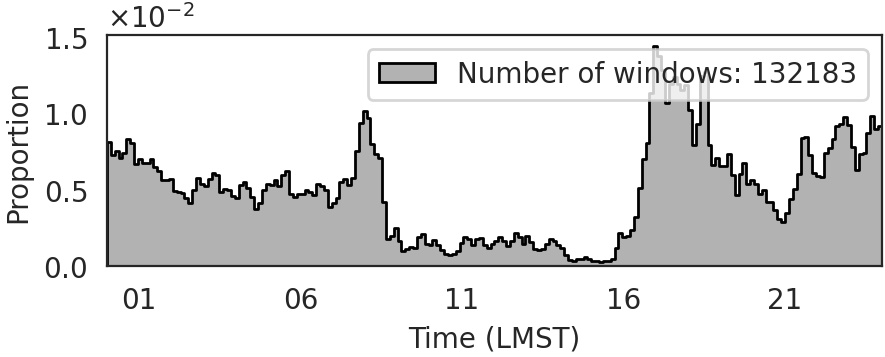}
        \caption{}
        \label{fig:app_histogram_cluster-7_scale-3}
      \end{subfigure}

    \begin{subfigure}[t]{0.4\textwidth}
        \centering
        \includegraphics[width=\textwidth]{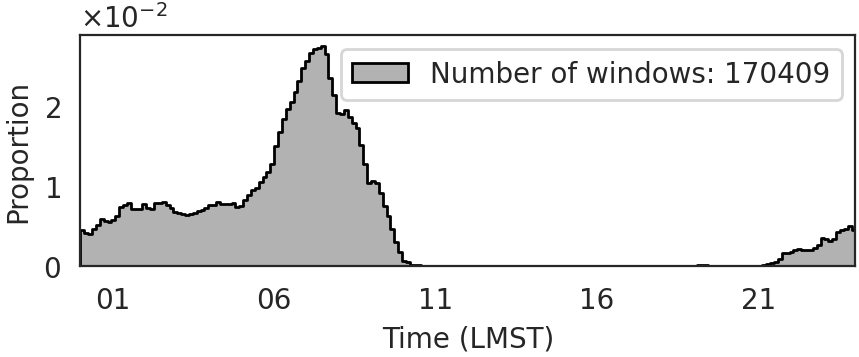}
        \caption{}
        \label{fig:app_histogram_cluster-8_scale-3}
      \end{subfigure}

    \caption{The occurrence time histogram of all nine identified clusters within the $13.6$-minute timescale. The horizontal axis of the histograms represents the local mean solar time (LMST). Clusters 0--9 are shown in \cref{fig:app_histogram_cluster-0_scale-3,fig:app_histogram_cluster-1_scale-3,fig:app_histogram_cluster-2_scale-3,fig:app_histogram_cluster-3_scale-3,fig:app_histogram_cluster-4_scale-3,fig:app_histogram_cluster-5_scale-3,fig:app_histogram_cluster-6_scale-3,fig:app_histogram_cluster-7_scale-3,fig:app_histogram_cluster-8_scale-3}, respectively.}
    \label{fig:app_histogram_clusters_scale-3}
\end{figure*}

\begin{figure*}[p]
    \centering

    \begin{subfigure}[t]{0.495\textwidth}
        \centering
        \includegraphics[width=\textwidth]{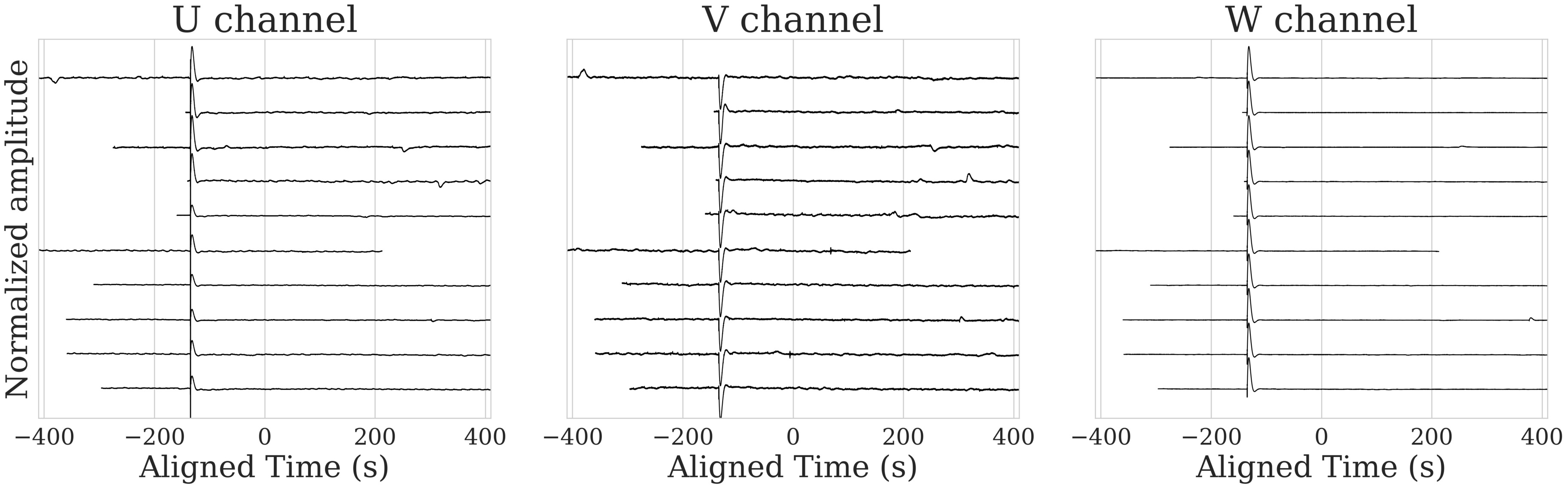}
        \caption{}
        \label{fig:app_cluster-0_scale-3}
      \end{subfigure}\hspace{0.1em}
    \begin{subfigure}[t]{0.495\textwidth}
        \centering
        \includegraphics[width=\textwidth]{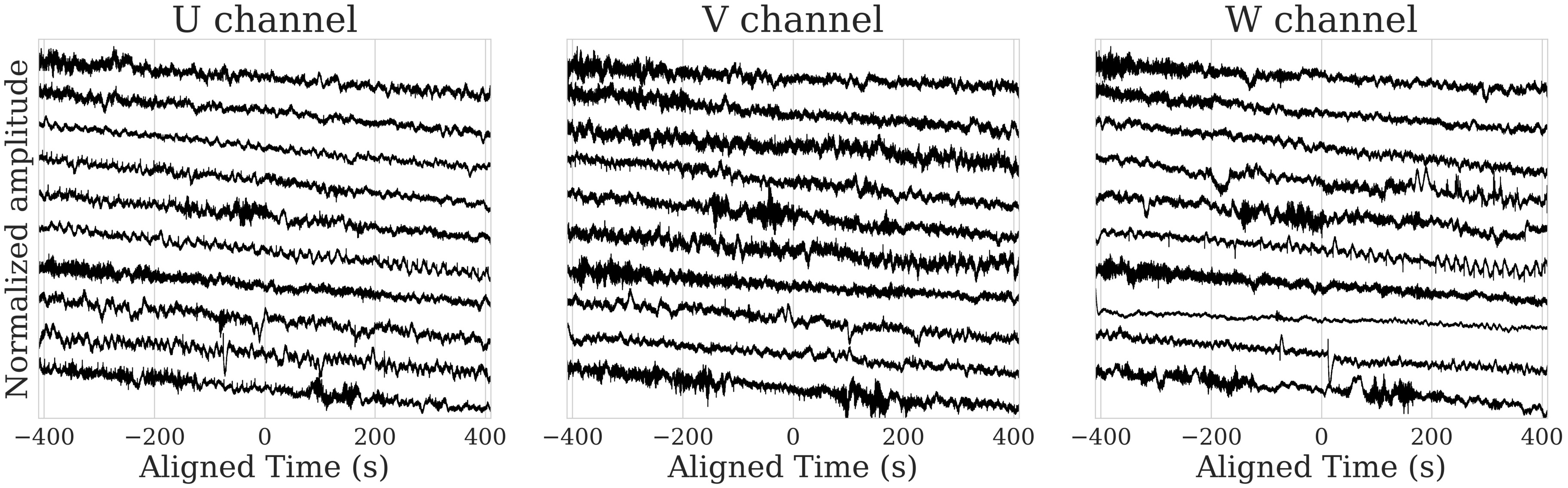}
        \caption{}
        \label{fig:app_cluster-1_scale-3}
      \end{subfigure}

    \begin{subfigure}[t]{0.495\textwidth}
        \centering
        \includegraphics[width=\textwidth]{figs/scale_16384/aligned_waveforms/aligned_waveforms_cluster_2.jpg}
        \caption{}
        \label{fig:app_cluster-2_scale-3}
      \end{subfigure}\hspace{0.1em}
    \begin{subfigure}[t]{0.495\textwidth}
        \centering
        \includegraphics[width=\textwidth]{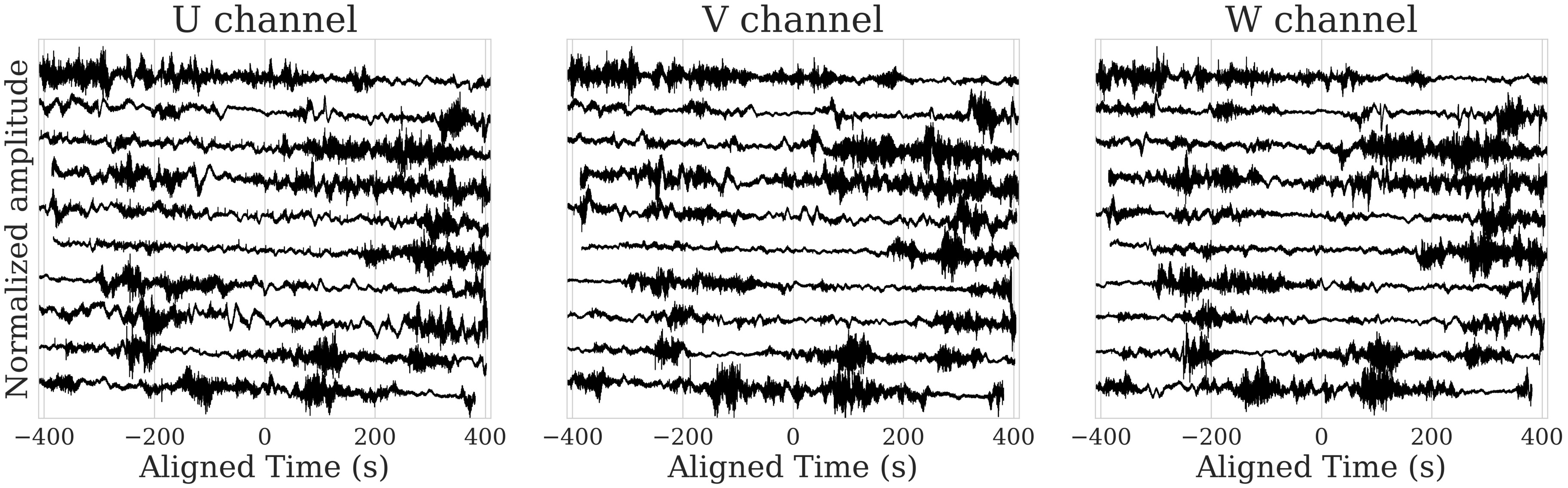}
        \caption{}
        \label{fig:app_cluster-3_scale-3}
      \end{subfigure}

    \begin{subfigure}[t]{0.495\textwidth}
        \centering
        \includegraphics[width=\textwidth]{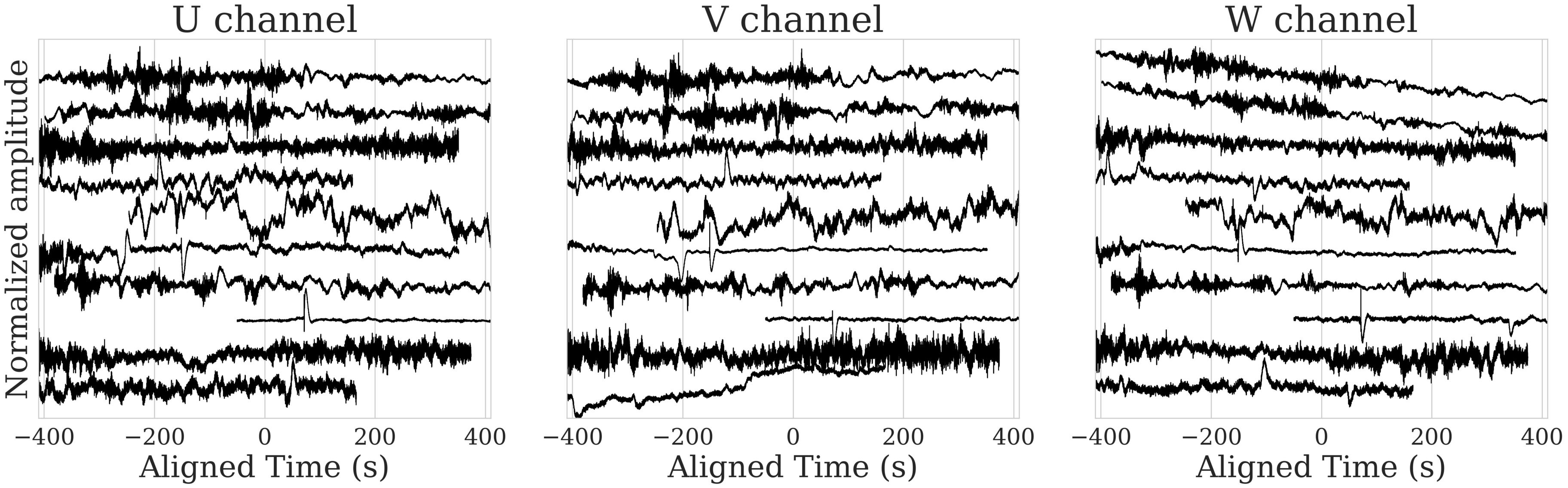}
        \caption{}
        \label{fig:app_cluster-4_scale-3}
      \end{subfigure}\hspace{0.1em}
    \begin{subfigure}[t]{0.495\textwidth}
        \centering
        \includegraphics[width=\textwidth]{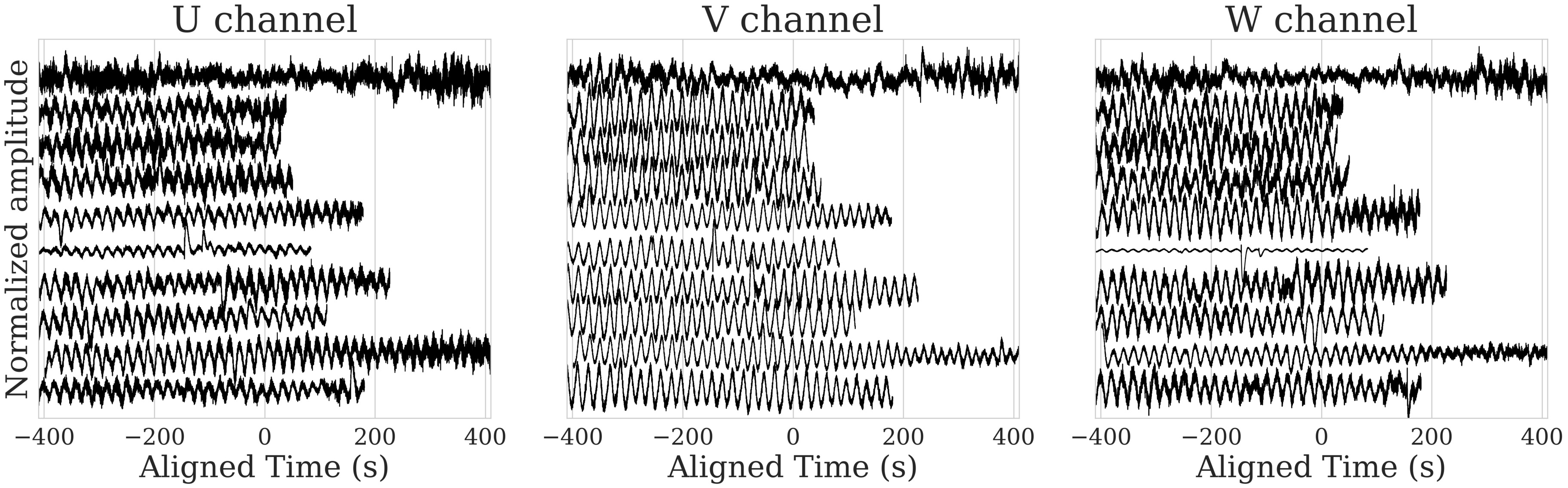}
        \caption{}
        \label{fig:app_cluster-5_scale-3}
      \end{subfigure}

    \begin{subfigure}[t]{0.495\textwidth}
        \centering
        \includegraphics[width=\textwidth]{figs/scale_16384/aligned_waveforms/aligned_waveforms_cluster_5.jpg}
        \caption{}
        \label{fig:app_cluster-6_scale-3}
      \end{subfigure}\hspace{0.1em}
    \begin{subfigure}[t]{0.495\textwidth}
        \centering
        \includegraphics[width=\textwidth]{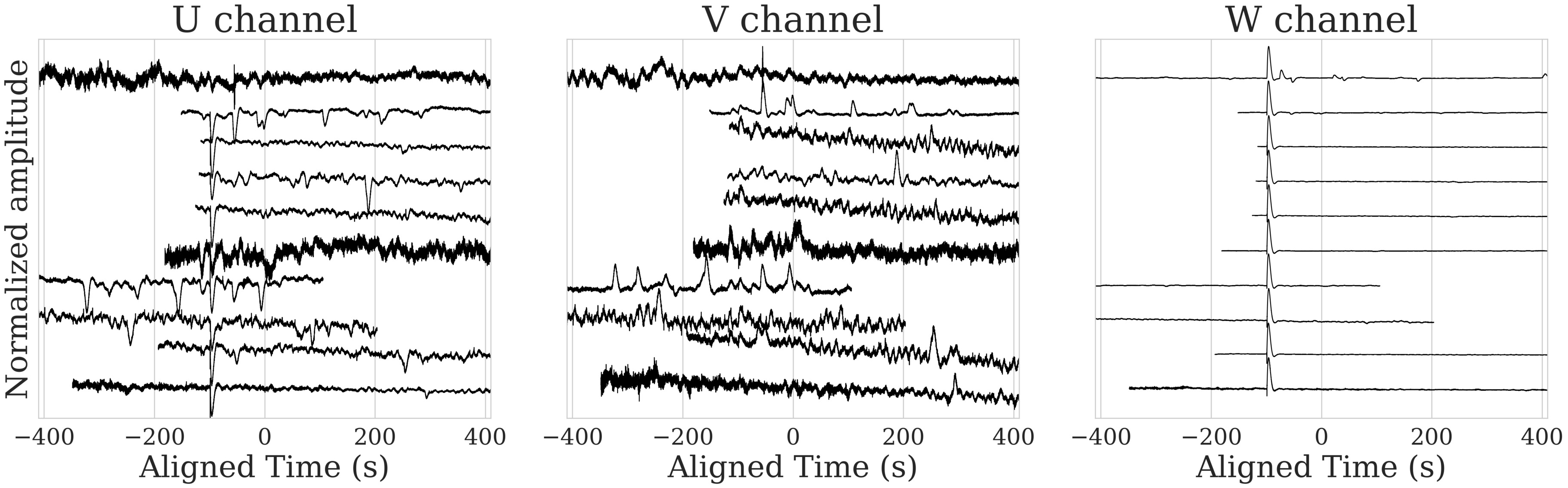}
        \caption{}
        \label{fig:app_cluster-7_scale-3}
      \end{subfigure}

    \begin{subfigure}[t]{0.495\textwidth}
        \centering
        \includegraphics[width=\textwidth]{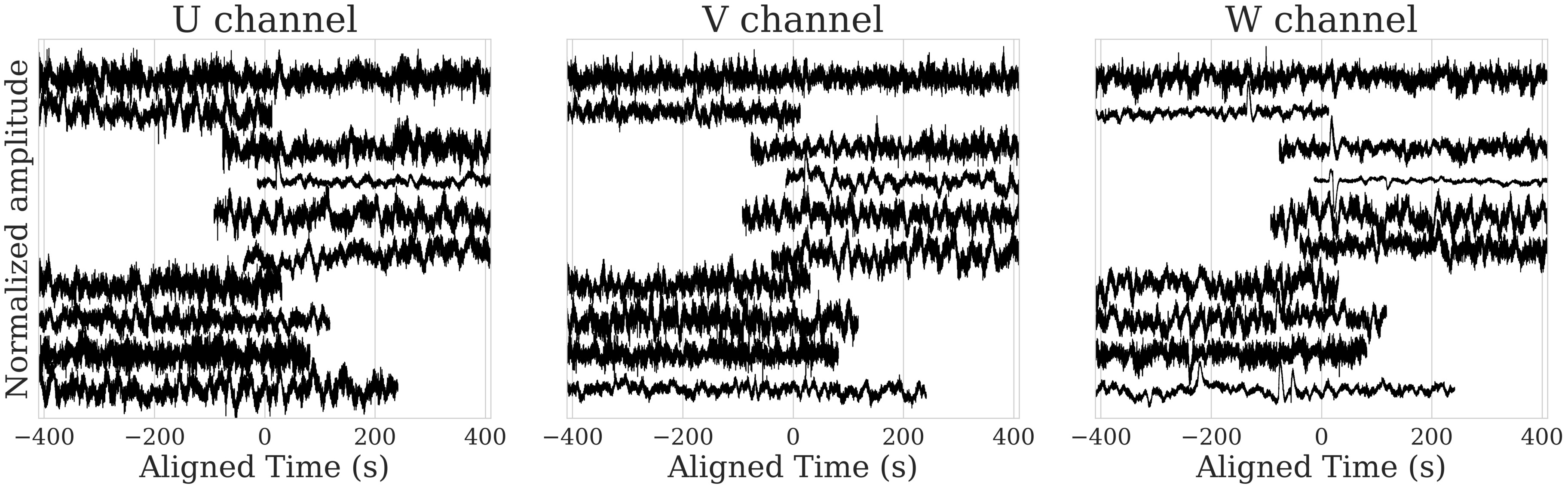}
        \caption{}
        \label{fig:app_cluster-8_scale-3}
      \end{subfigure}

    \caption{The aligned waveforms of all nine identified clusters within the $13.6$-minute timescale. Clusters 0--9 are shown in \cref{fig:app_cluster-0_scale-3,fig:app_cluster-1_scale-3,fig:app_cluster-2_scale-3,fig:app_cluster-3_scale-3,fig:app_cluster-4_scale-3,fig:app_cluster-5_scale-3,fig:app_cluster-6_scale-3,fig:app_cluster-7_scale-3,fig:app_cluster-8_scale-3}, respectively.}
    \label{fig:app_clusters_scale-3}
\end{figure*}

\begin{figure*}[p]
    \centering

    \begin{subfigure}[t]{0.4\textwidth}
        \centering
        \includegraphics[width=\textwidth]{figs/scale_65536/time_histograms/time_histogram_cluster-0.png}
        \caption{}
        \label{fig:app_histogram_cluster-0_scale-4}
      \end{subfigure}\hspace{1em}\hspace{1em}
    \begin{subfigure}[t]{0.4\textwidth}
        \centering
        \includegraphics[width=\textwidth]{figs/scale_65536/time_histograms/time_histogram_cluster-1.png}
        \caption{}
        \label{fig:app_histogram_cluster-1_scale-4}
      \end{subfigure}

    \begin{subfigure}[t]{0.4\textwidth}
        \centering
        \includegraphics[width=\textwidth]{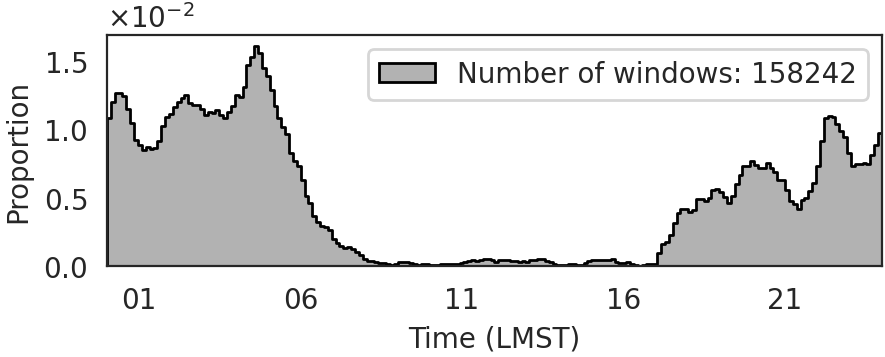}
        \caption{}
        \label{fig:app_histogram_cluster-2_scale-4}
      \end{subfigure}\hspace{1em}
    \begin{subfigure}[t]{0.4\textwidth}
        \centering
        \includegraphics[width=\textwidth]{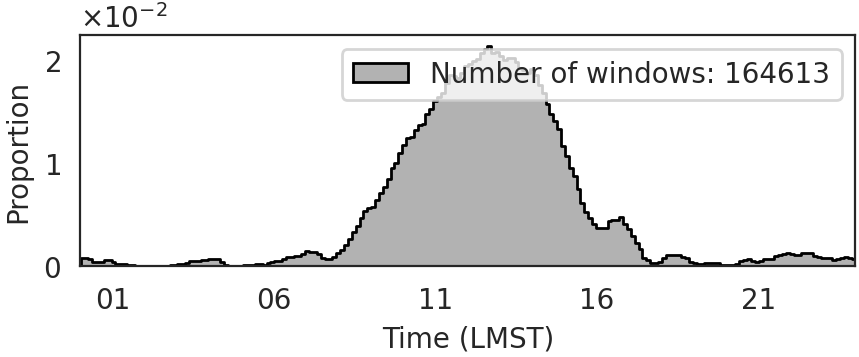}
        \caption{}
        \label{fig:app_histogram_cluster-3_scale-4}
      \end{subfigure}

    \begin{subfigure}[t]{0.4\textwidth}
        \centering
        \includegraphics[width=\textwidth]{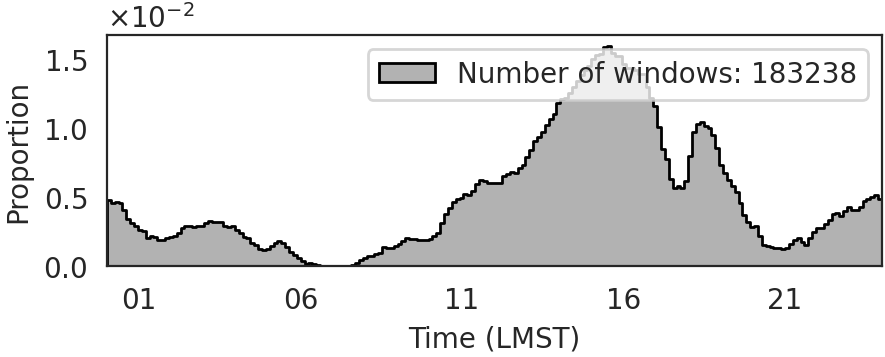}
        \caption{}
        \label{fig:app_histogram_cluster-4_scale-4}
      \end{subfigure}\hspace{1em}
    \begin{subfigure}[t]{0.4\textwidth}
        \centering
        \includegraphics[width=\textwidth]{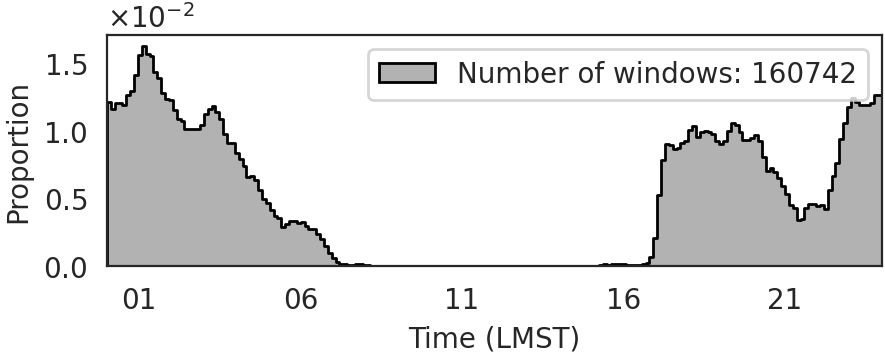}
        \caption{}
        \label{fig:app_histogram_cluster-5_scale-4}
      \end{subfigure}

    \begin{subfigure}[t]{0.4\textwidth}
        \centering
        \includegraphics[width=\textwidth]{figs/scale_65536/time_histograms/time_histogram_cluster-5.png}
        \caption{}
        \label{fig:app_histogram_cluster-6_scale-4}
      \end{subfigure}\hspace{1em}
    \begin{subfigure}[t]{0.4\textwidth}
        \centering
        \includegraphics[width=\textwidth]{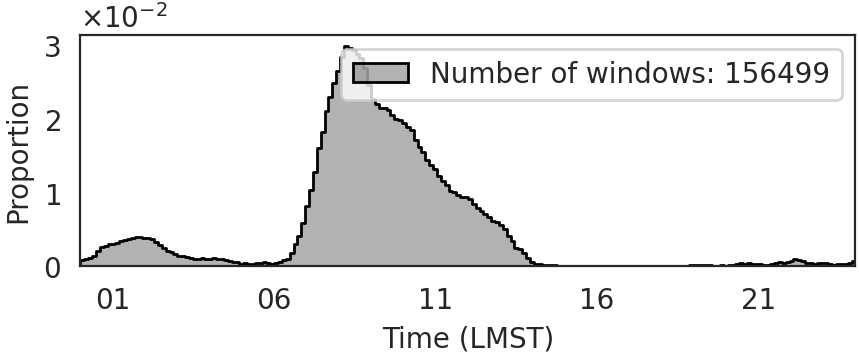}
        \caption{}
        \label{fig:app_histogram_cluster-7_scale-4}
      \end{subfigure}

    \begin{subfigure}[t]{0.4\textwidth}
        \centering
        \includegraphics[width=\textwidth]{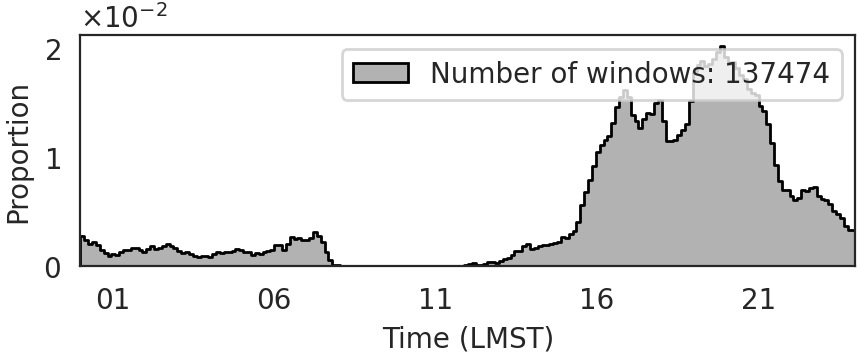}
        \caption{}
        \label{fig:app_histogram_cluster-8_scale-4}
      \end{subfigure}

    \caption{The occurrence time histogram of all nine identified clusters within the $54.6$-minute timescale. The horizontal axis of the histograms represents the local mean solar time (LMST). Clusters 0--9 are shown in \cref{fig:app_histogram_cluster-0_scale-4,fig:app_histogram_cluster-1_scale-4,fig:app_histogram_cluster-2_scale-4,fig:app_histogram_cluster-3_scale-4,fig:app_histogram_cluster-4_scale-4,fig:app_histogram_cluster-5_scale-4,fig:app_histogram_cluster-6_scale-4,fig:app_histogram_cluster-7_scale-4,fig:app_histogram_cluster-8_scale-4}, respectively.}
    \label{fig:app_histogram_clusters_scale-4}
\end{figure*}

\begin{figure*}[p]
    \centering

    \begin{subfigure}[t]{0.495\textwidth}
        \centering
        \includegraphics[width=\textwidth]{figs/scale_65536/aligned_waveforms/aligned_waveforms_cluster_0.jpg}
        \caption{}
        \label{fig:app_cluster-0_scale-4}
      \end{subfigure}\hspace{0.1em}
    \begin{subfigure}[t]{0.495\textwidth}
        \centering
        \includegraphics[width=\textwidth]{figs/scale_65536/aligned_waveforms/aligned_waveforms_cluster_1.jpg}
        \caption{}
        \label{fig:app_cluster-1_scale-4}
      \end{subfigure}

    \begin{subfigure}[t]{0.495\textwidth}
        \centering
        \includegraphics[width=\textwidth]{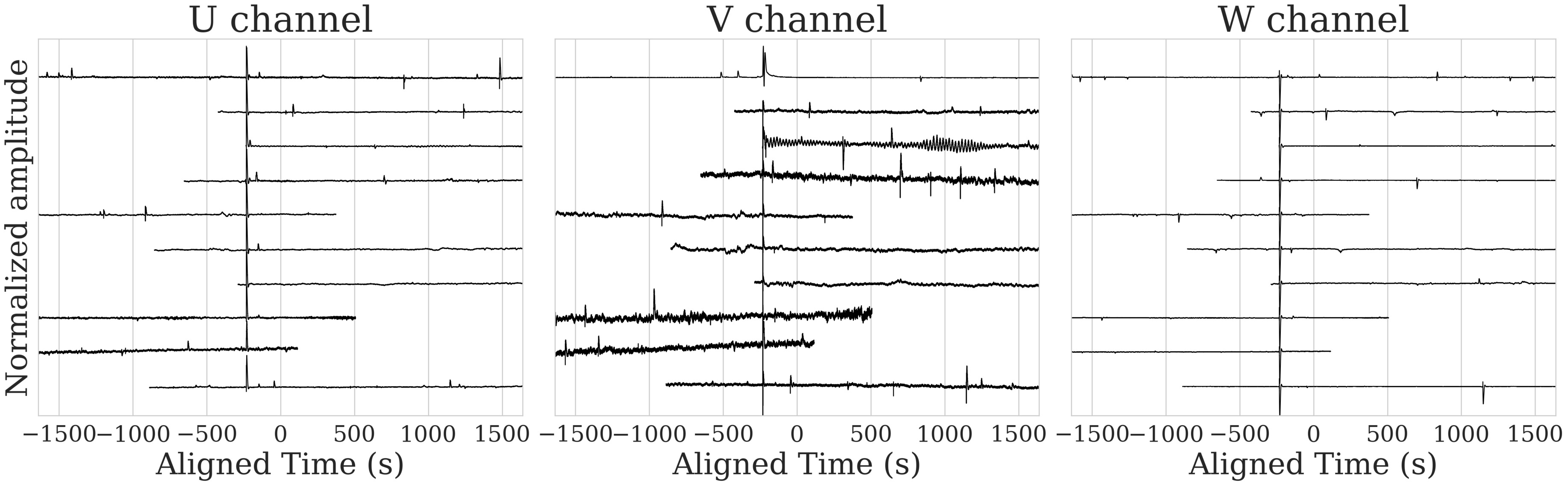}
        \caption{}
        \label{fig:app_cluster-2_scale-4}
      \end{subfigure}\hspace{0.1em}
    \begin{subfigure}[t]{0.495\textwidth}
        \centering
        \includegraphics[width=\textwidth]{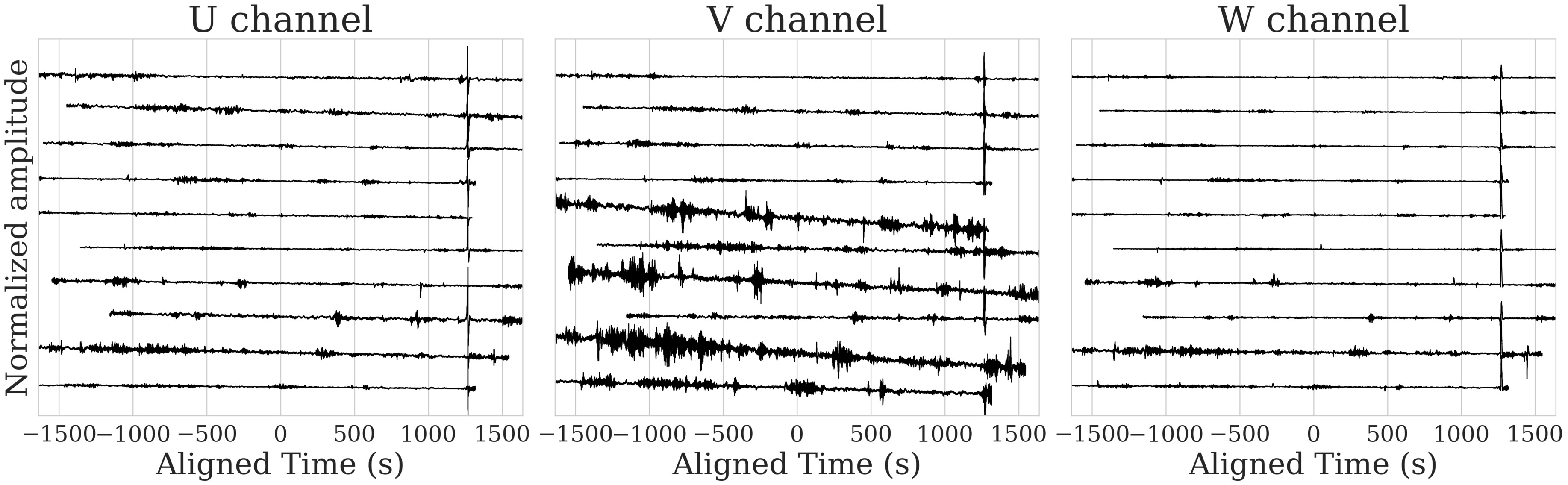}
        \caption{}
        \label{fig:app_cluster-3_scale-4}
      \end{subfigure}

    \begin{subfigure}[t]{0.495\textwidth}
        \centering
        \includegraphics[width=\textwidth]{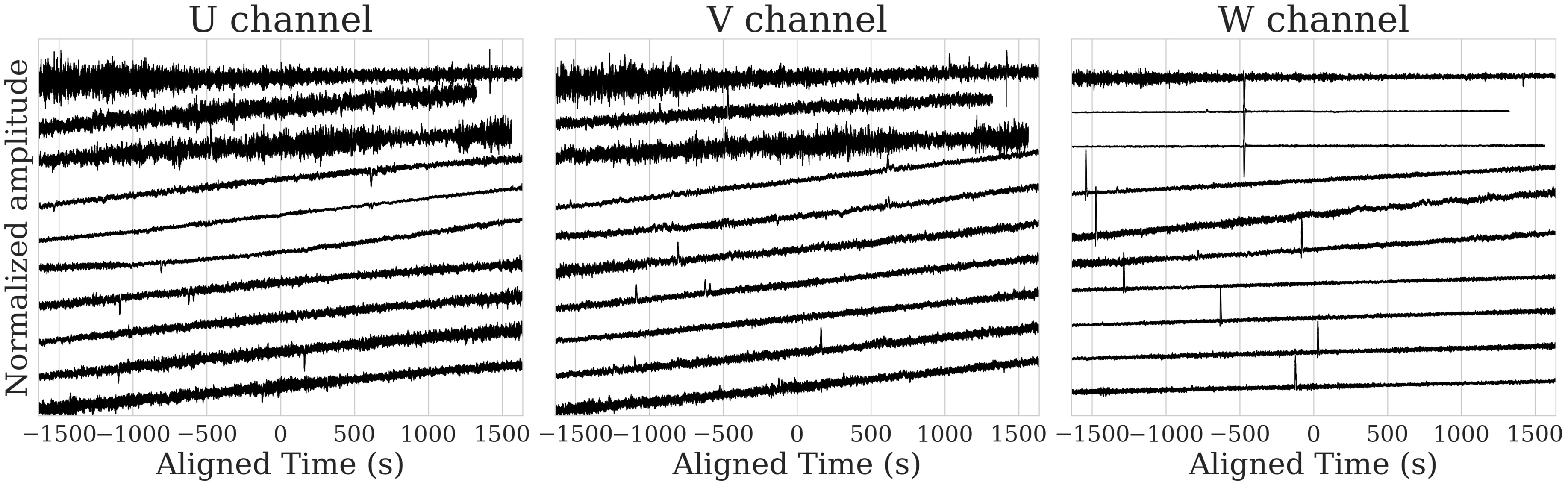}
        \caption{}
        \label{fig:app_cluster-4_scale-4}
      \end{subfigure}\hspace{0.1em}
    \begin{subfigure}[t]{0.495\textwidth}
        \centering
        \includegraphics[width=\textwidth]{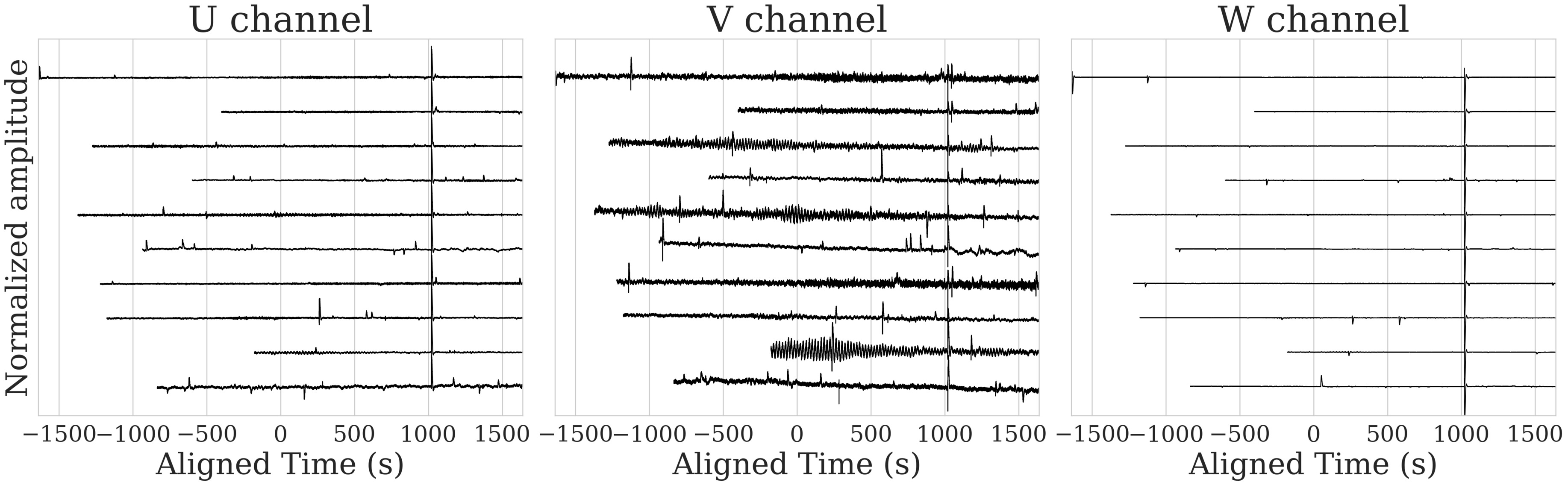}
        \caption{}
        \label{fig:app_cluster-5_scale-4}
      \end{subfigure}

    \begin{subfigure}[t]{0.495\textwidth}
        \centering
        \includegraphics[width=\textwidth]{figs/scale_65536/aligned_waveforms/aligned_waveforms_cluster_5.jpg}
        \caption{}
        \label{fig:app_cluster-6_scale-4}
      \end{subfigure}\hspace{0.1em}
    \begin{subfigure}[t]{0.495\textwidth}
        \centering
        \includegraphics[width=\textwidth]{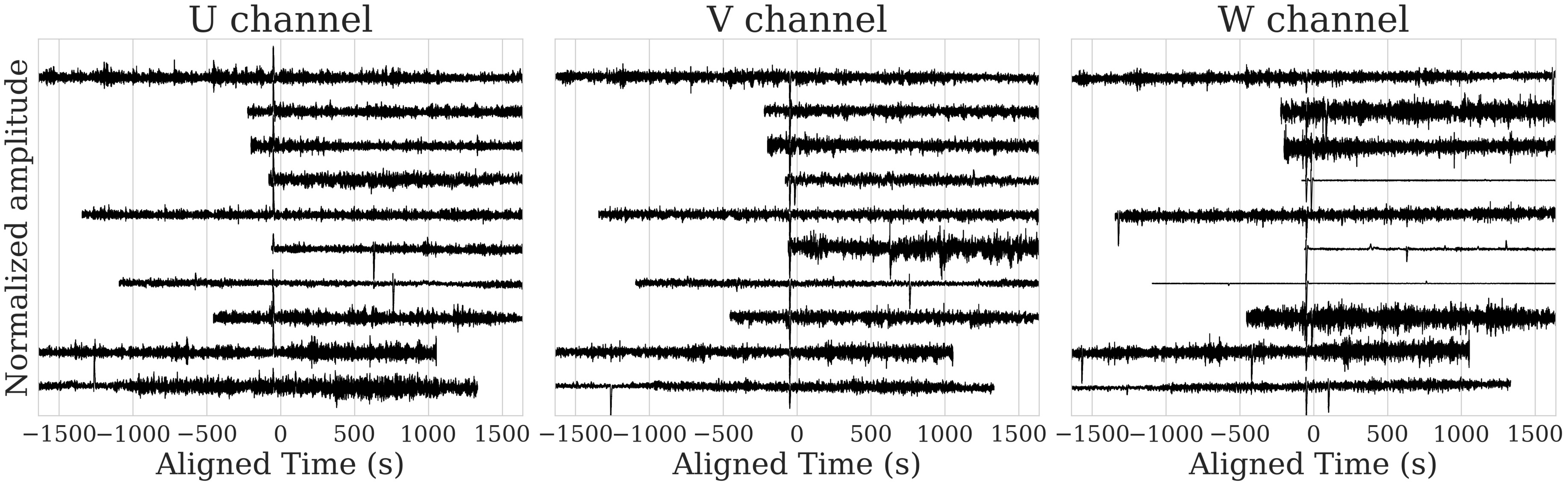}
        \caption{}
        \label{fig:app_cluster-7_scale-4}
      \end{subfigure}

    \begin{subfigure}[t]{0.495\textwidth}
        \centering
        \includegraphics[width=\textwidth]{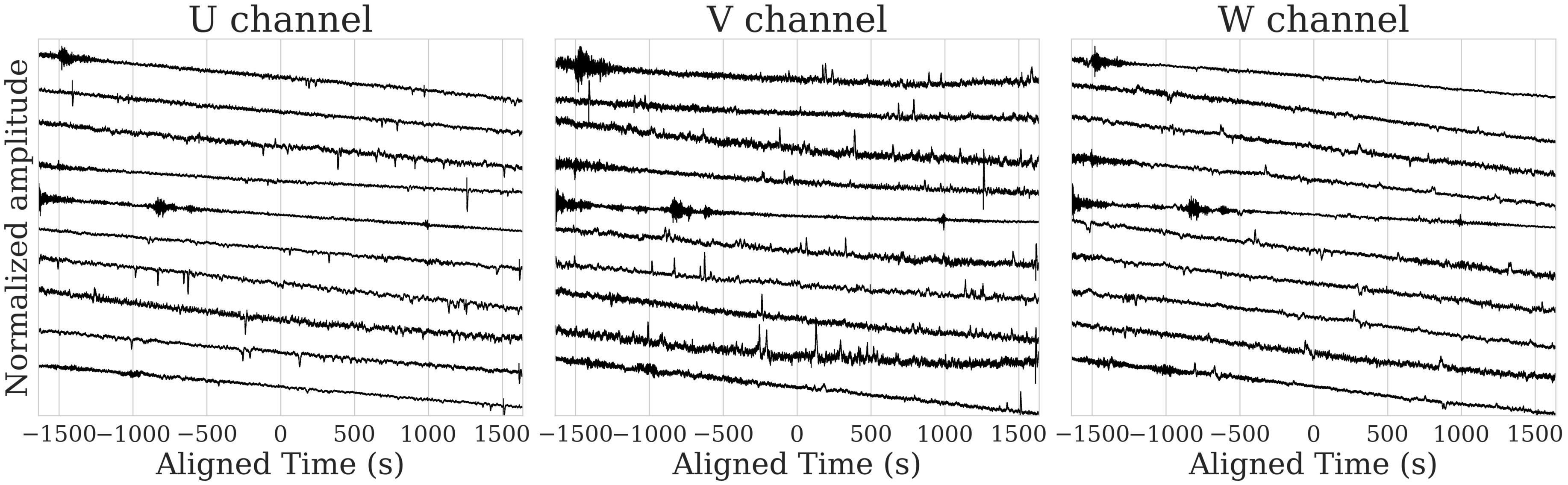}
        \caption{}
        \label{fig:app_cluster-8_scale-4}
      \end{subfigure}

    \caption{The aligned waveforms of all nine identified clusters within the $54.6$-minute timescale. Clusters 0--9 are shown in \cref{fig:app_cluster-0_scale-4,fig:app_cluster-1_scale-4,fig:app_cluster-2_scale-4,fig:app_cluster-3_scale-4,fig:app_cluster-4_scale-4,fig:app_cluster-5_scale-4,fig:app_cluster-6_scale-4,fig:app_cluster-7_scale-4,fig:app_cluster-8_scale-4}, respectively.}
    \label{fig:app_clusters_scale-4}
\end{figure*}

In this section, we present the complete set of nine clusters identified
by the fVAE across four different time scales.
\cref{fig:app_histogram_clusters_scale-1,fig:app_histogram_clusters_scale-2,fig:app_histogram_clusters_scale-3,fig:app_histogram_clusters_scale-4}
display the occurrence time histograms of these clusters for
$51.2$-second timescale, $3.4$-minute timescale, $13.6$-minute
timescale, and $54.6$-minute timescale, respectively, by aggregating
data from the entire mission. Additionally, we provide the
representative aligned waveforms for each cluster at all time scales.
\cref{fig:app_clusters_scale-1,fig:app_clusters_scale-2,fig:app_clusters_scale-3,fig:app_clusters_scale-4}
depict these waveforms for $51.2$-second timescale, $3.4$-minute
timescale, $13.6$-minute timescale, and $54.6$-minute timescale,
respectively.

In a similar study, \cite{BarkaouiEtAl_2021} also aimed at detecting and
clustering microevents and noise patterns in the seismic data recorded
during the InSight mission on Mars. The authors identified nine main
clusters of events and noise patterns in the seismic data when focusing
on $100\,\mathrm{s}$ time windows. Similar to our observations with the
finest timescale used in our results, \cite{BarkaouiEtAl_2021} mostly
identified transient events, such as pressure drops and various types of
glitches. Since their timescale was slightly longer than our finest
timescale, they also discovered repeating sequences of glitches with
relatively constant time offsets. This is analogous to the two clusters
we identify in the $3.4$-minute timescale presented in
\cref{fig:app_cluster-0_scale-2,fig:app_cluster-5_scale-2}, whose time
histograms also align with the typical occurrence time of glitches (see
\cref{fig:app_histogram_cluster-0_scale-2,fig:app_histogram_cluster-5_scale-2}).

Regarding pressure drops, \cite{BarkaouiEtAl_2021} reported that these
were mainly spread across three clusters. Although the aligned waveforms
of clusters 0, 7, and 8 identified by our approach (see
\cref{fig:app_cluster-0_scale-1,fig:app_cluster-7_scale-1,fig:app_cluster-8_scale-1})
do not indicate pressure drops, the latent space visualization of
pressure drops from the catalog at the finest scale (see
\cref{fig:scale-1_drop}) shows that these pressure drops are mostly
concentrated in cluster 7, with the rest primarily in clusters 0 and 8.
Additionally, \cite{BarkaouiEtAl_2021} characterized daily and seasonal
variations in event occurrence related to temperature changes, which we
similarly observed in \cref{fig:season_all_clusters}.

Beyond these observations, our multi-scale approach and the use of
scattering spectra, which are more capable at distinguishing between
different non-Gaussian stochastic processes than scattering coefficients
\cite{MorelEtAl_2022}, allowed us to identify more patterns in the
datasets, including events with much longer timescales, such as
atmospheric-surface interactions due to wind. Furthermore, we utilized
the entire seismic dataset to train our model, in contrast to the
approximately one week of data used in \cite{BarkaouiEtAl_2021},
providing a more comprehensive overview of Martian seismic activity and
arguably a more robust data representation. This enriched representation
and the multi-scale nature of our approach facilitated the localization
of rare marsquakes in the latent space. As noted in
\cite{BarkaouiEtAl_2021}, due to their rarity, marsquakes do not
significantly influence the clustering results. Finally, thanks to our
probabilistic multi-scale clustering, we were able to use the scattering
spectra representation of signals of interest to perform source
separation.

\section{fVAE decoder reconstruction quality}
\label{app:fvae_reconstruction}

\begin{figure*}[p]
    \centering

    \begin{subfigure}[t]{0.495\textwidth}
        \centering
        \includegraphics[width=\textwidth]{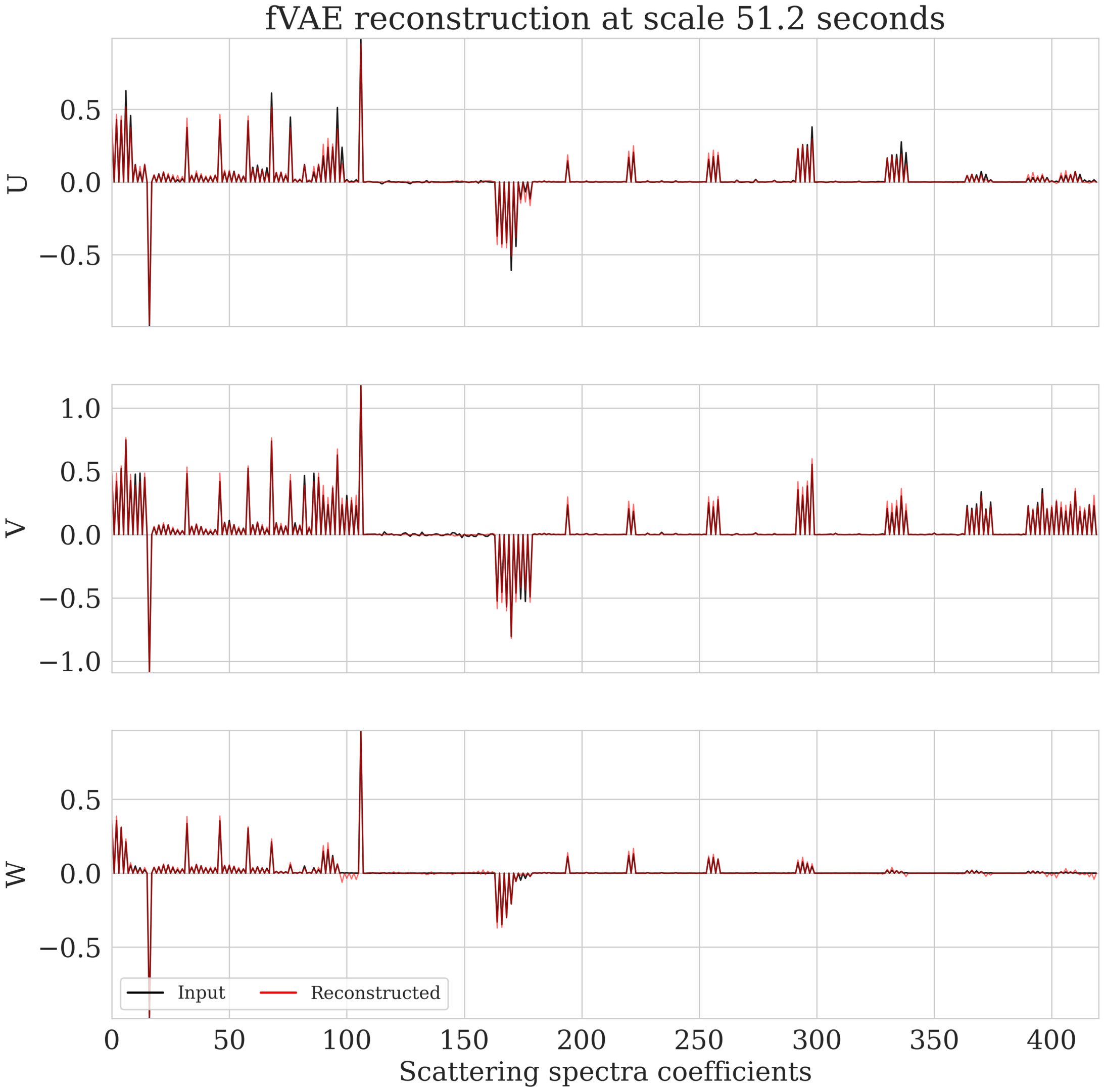}
        \caption{}
        \label{fig:app_decoder_scale-1-0}
      \end{subfigure}\hspace{0.1em}
    \begin{subfigure}[t]{0.495\textwidth}
        \centering
        \includegraphics[width=\textwidth]{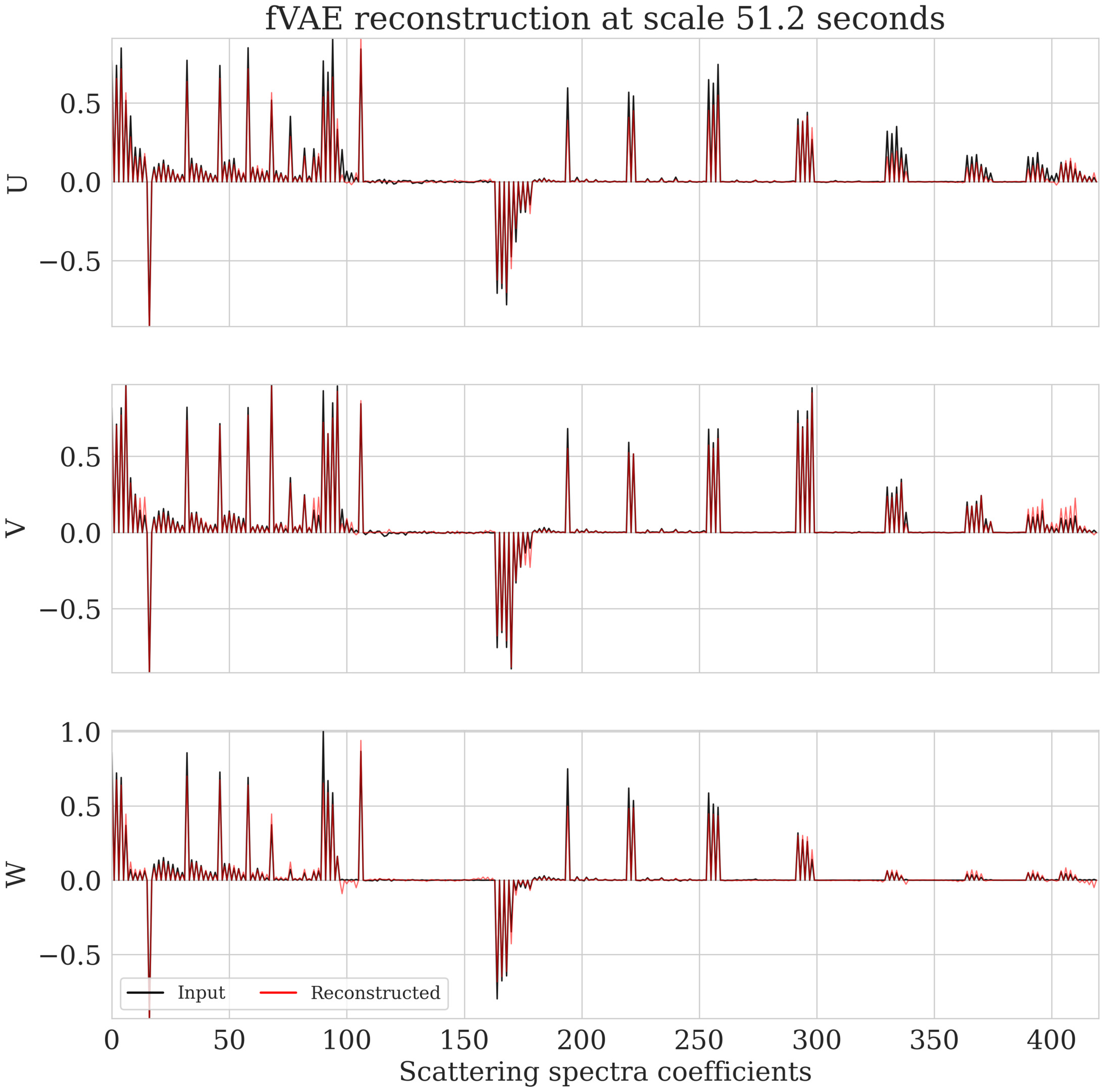}
        \caption{}
        \label{fig:app_decoder_scale-1-1}
      \end{subfigure}

    \begin{subfigure}[t]{0.495\textwidth}
        \centering
        \includegraphics[width=\textwidth]{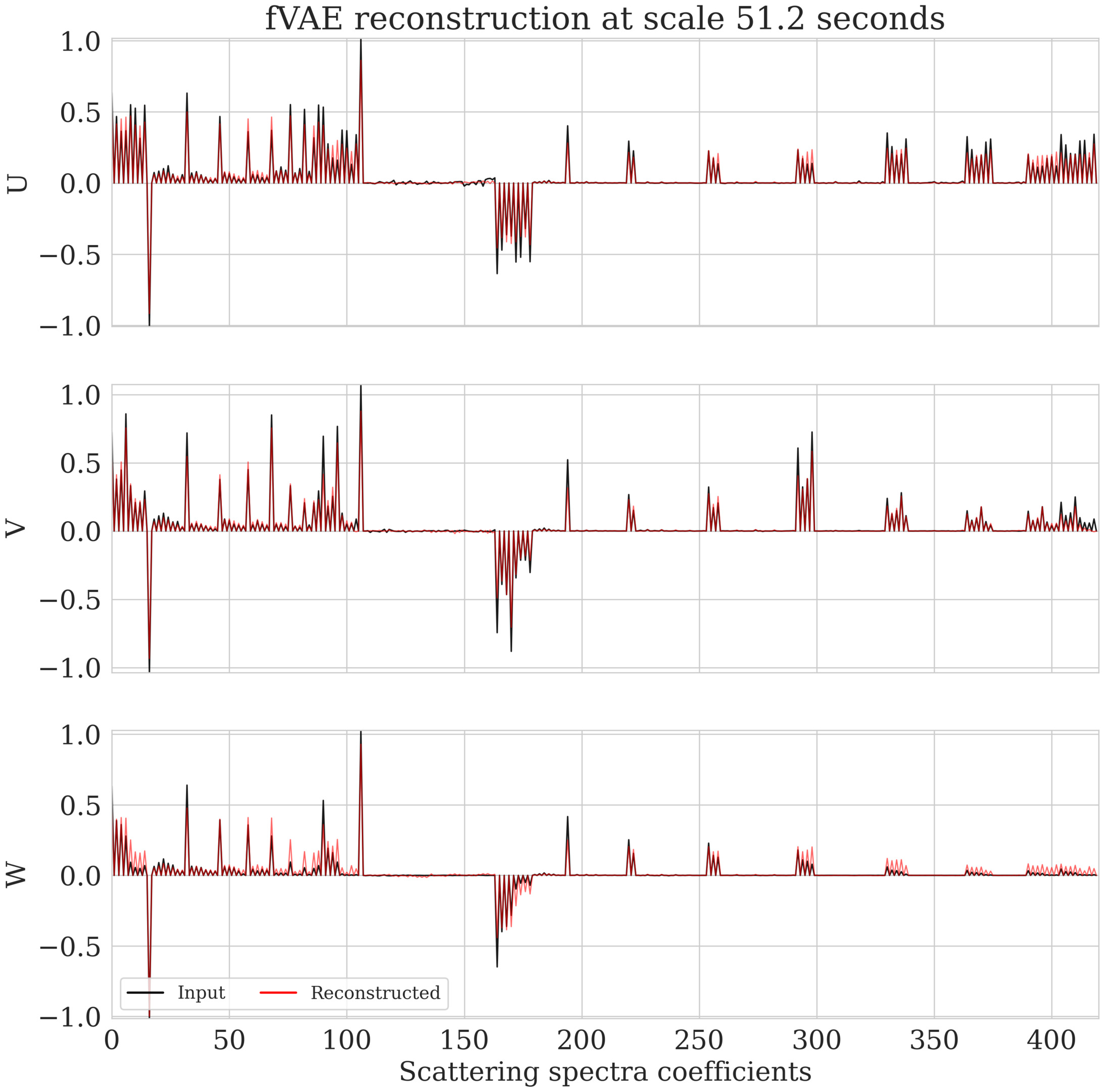}
        \caption{}
        \label{fig:app_decoder_scale-1-2}
      \end{subfigure}\hspace{0.1em}
    \begin{subfigure}[t]{0.495\textwidth}
        \centering
        \includegraphics[width=\textwidth]{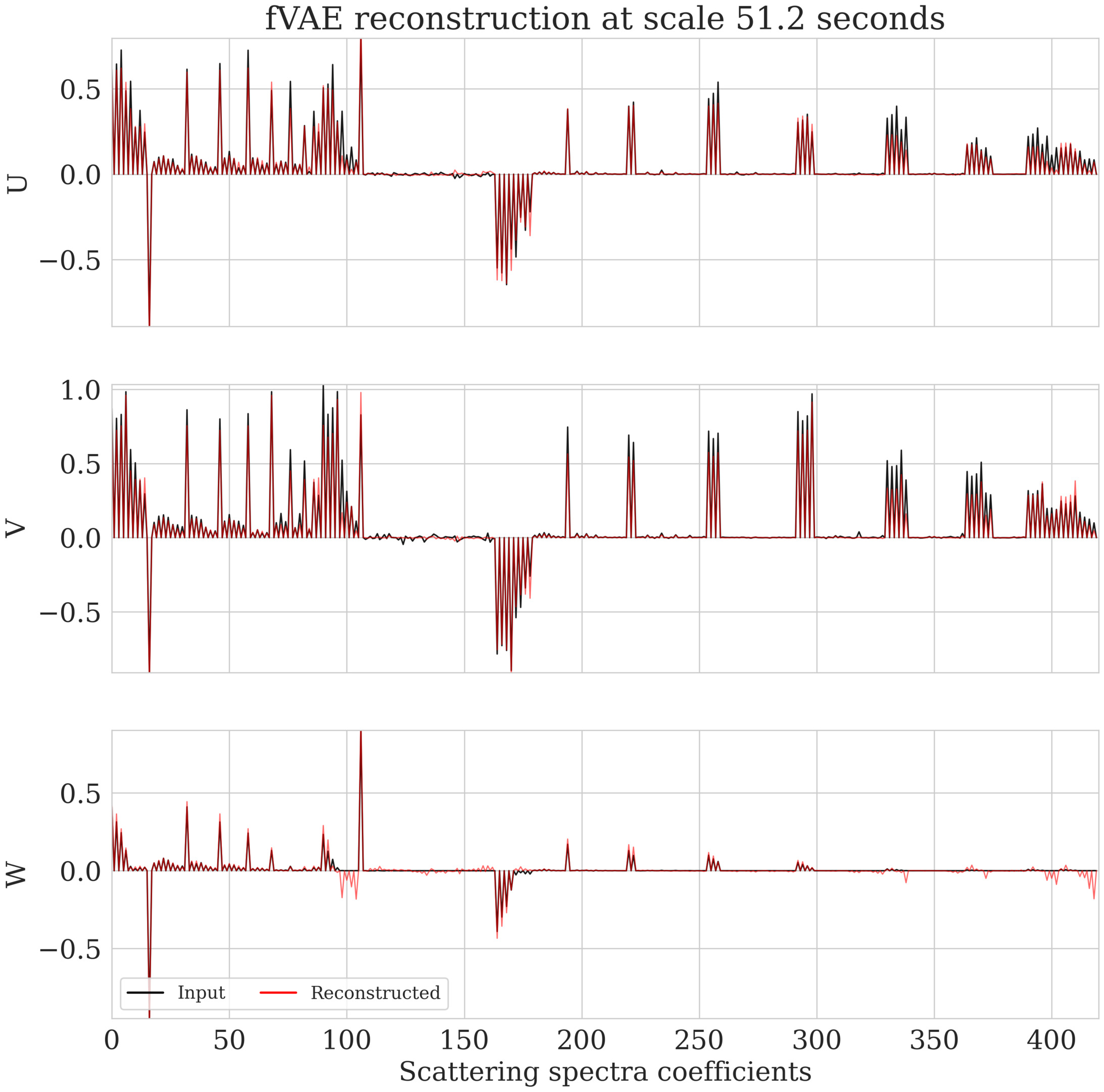}
        \caption{}
        \label{fig:app_decoder_scale-1-3}
      \end{subfigure}

    \caption{The input scattering spectra (black) and reconstruction via the fVAE decoder (red) for the U, V, and W components of four random windows from the $51.2$-second timescale.}
    \label{fig:app_decoder_scale-1}
\end{figure*}

\begin{figure*}[p]
    \centering

    \begin{subfigure}[t]{0.495\textwidth}
        \centering
        \includegraphics[width=\textwidth]{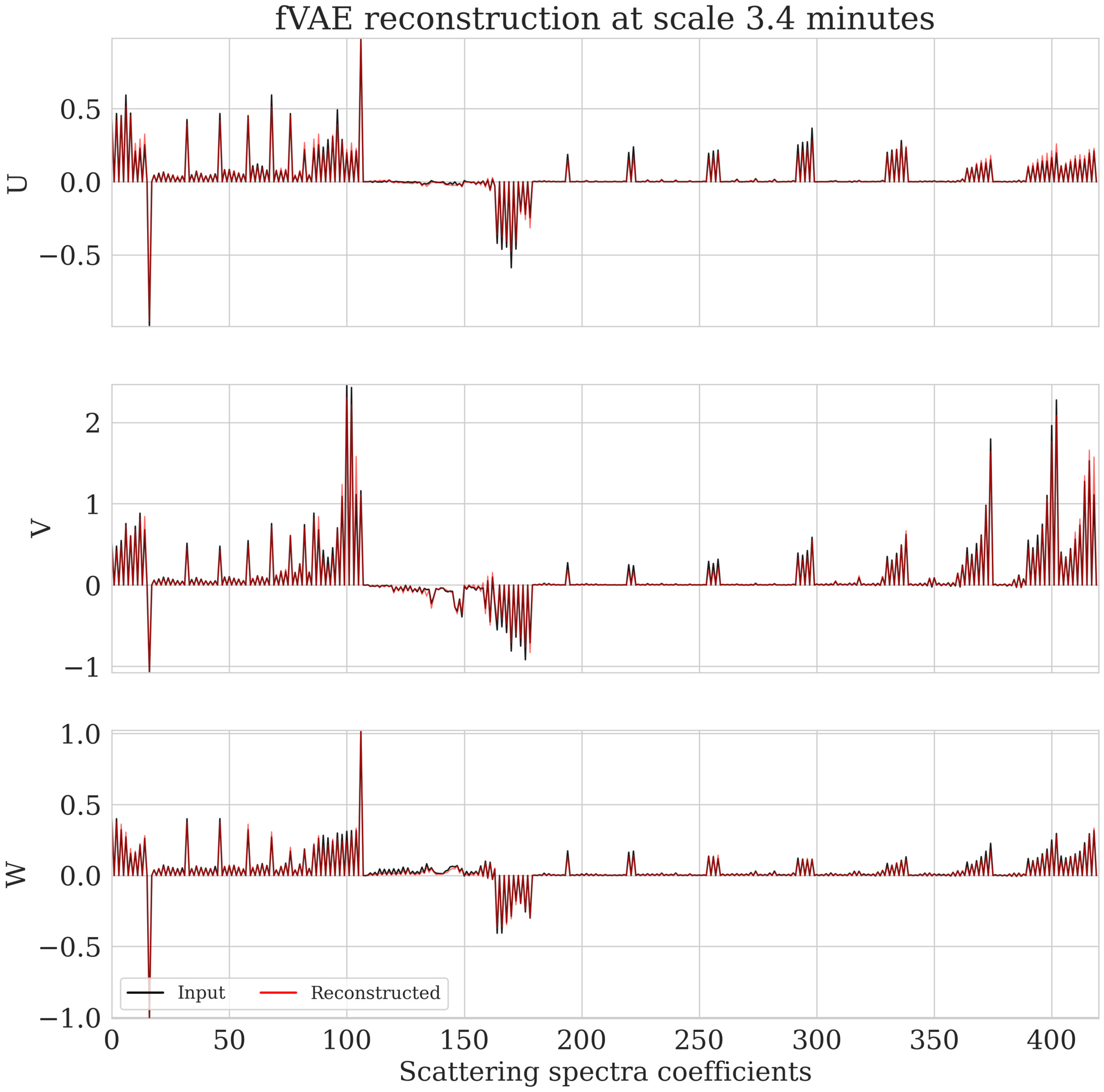}
        \caption{}
        \label{fig:app_decoder_scale-2-0}
      \end{subfigure}\hspace{0.1em}
    \begin{subfigure}[t]{0.495\textwidth}
        \centering
        \includegraphics[width=\textwidth]{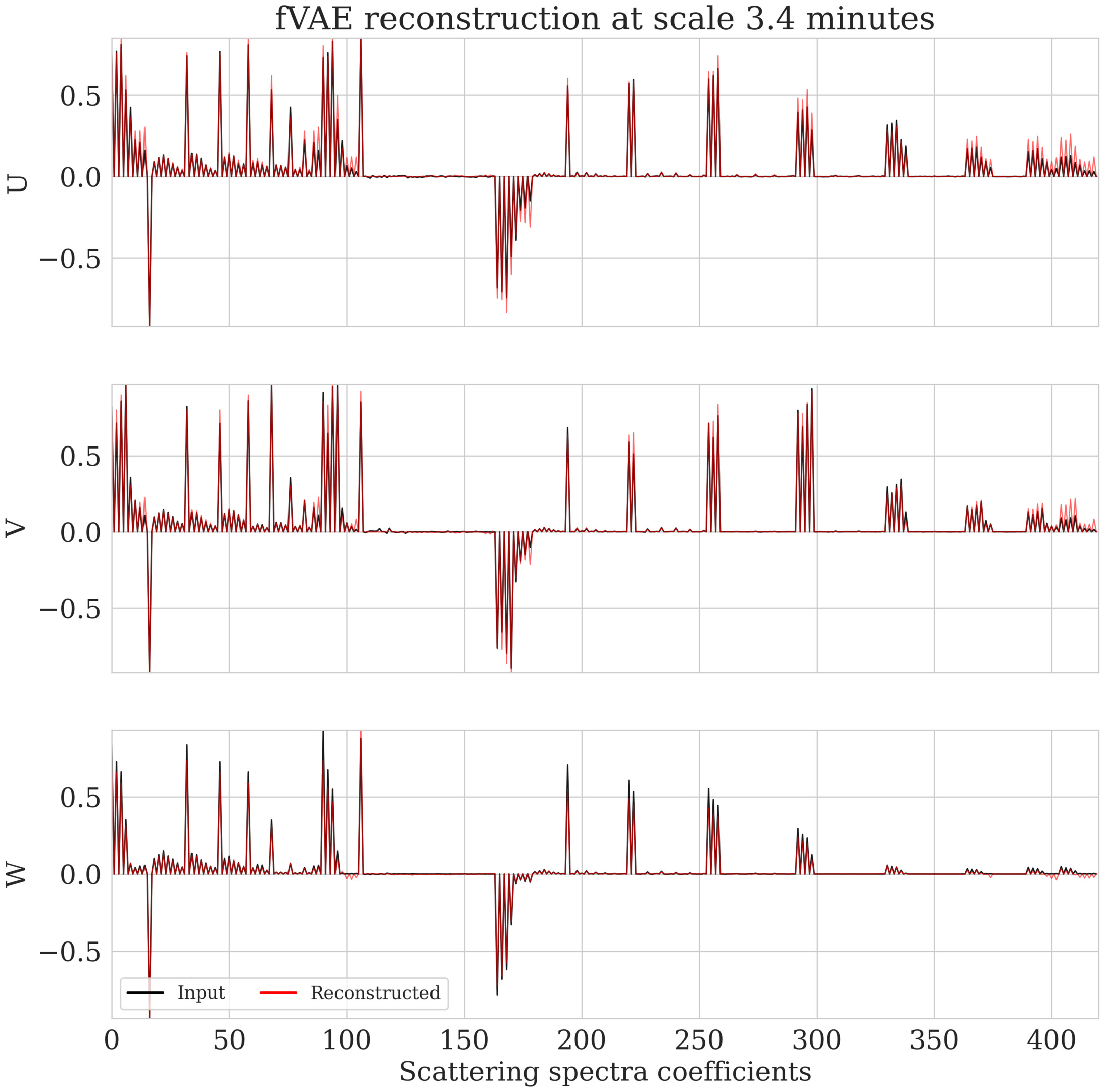}
        \caption{}
        \label{fig:app_decoder_scale-2-1}

      \end{subfigure}

    \begin{subfigure}[t]{0.495\textwidth}
        \centering
        \includegraphics[width=\textwidth]{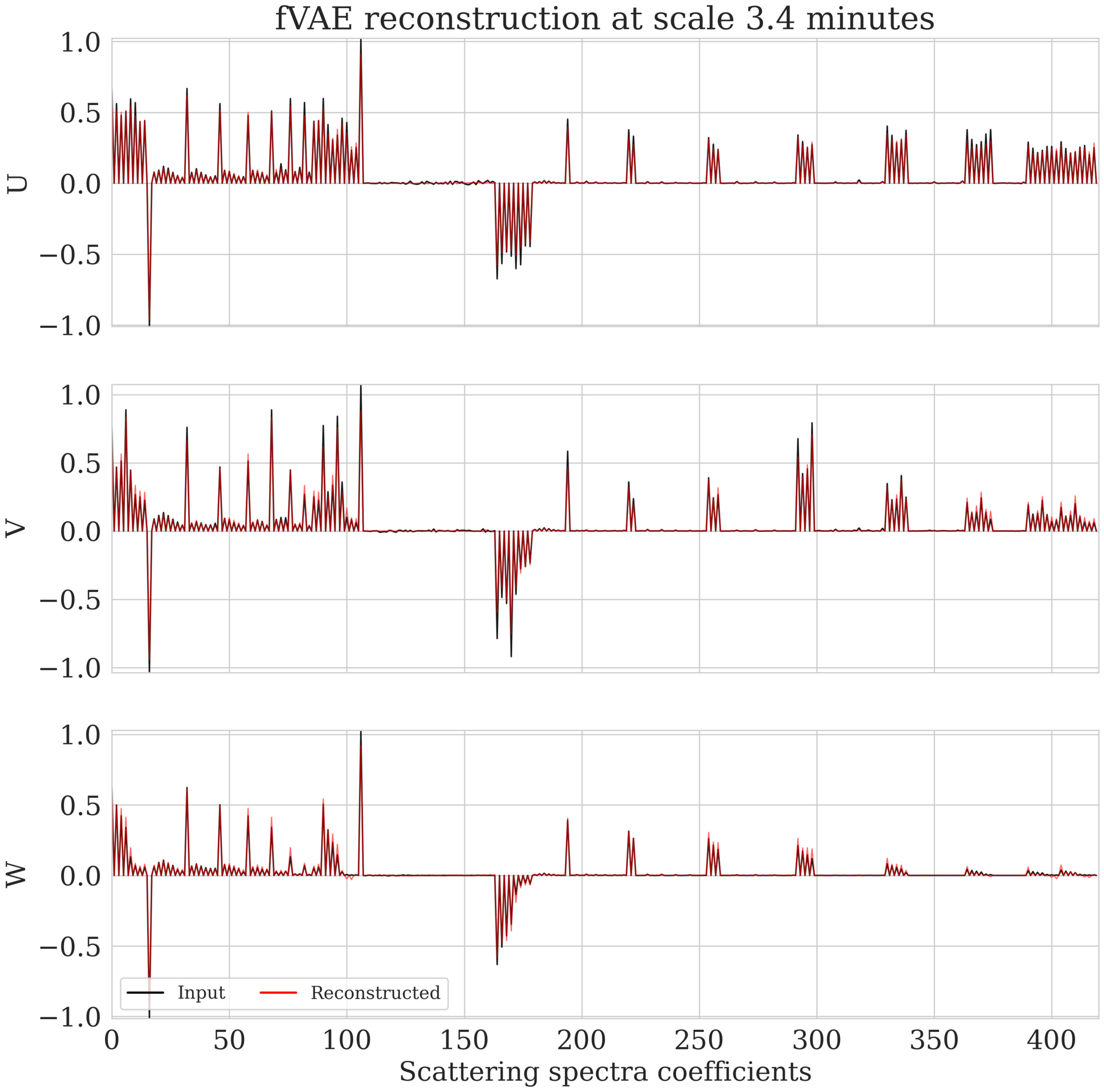}
        \caption{}
        \label{fig:app_decoder_scale-2-2}
      \end{subfigure}\hspace{0.1em}
    \begin{subfigure}[t]{0.495\textwidth}
        \centering
        \includegraphics[width=\textwidth]{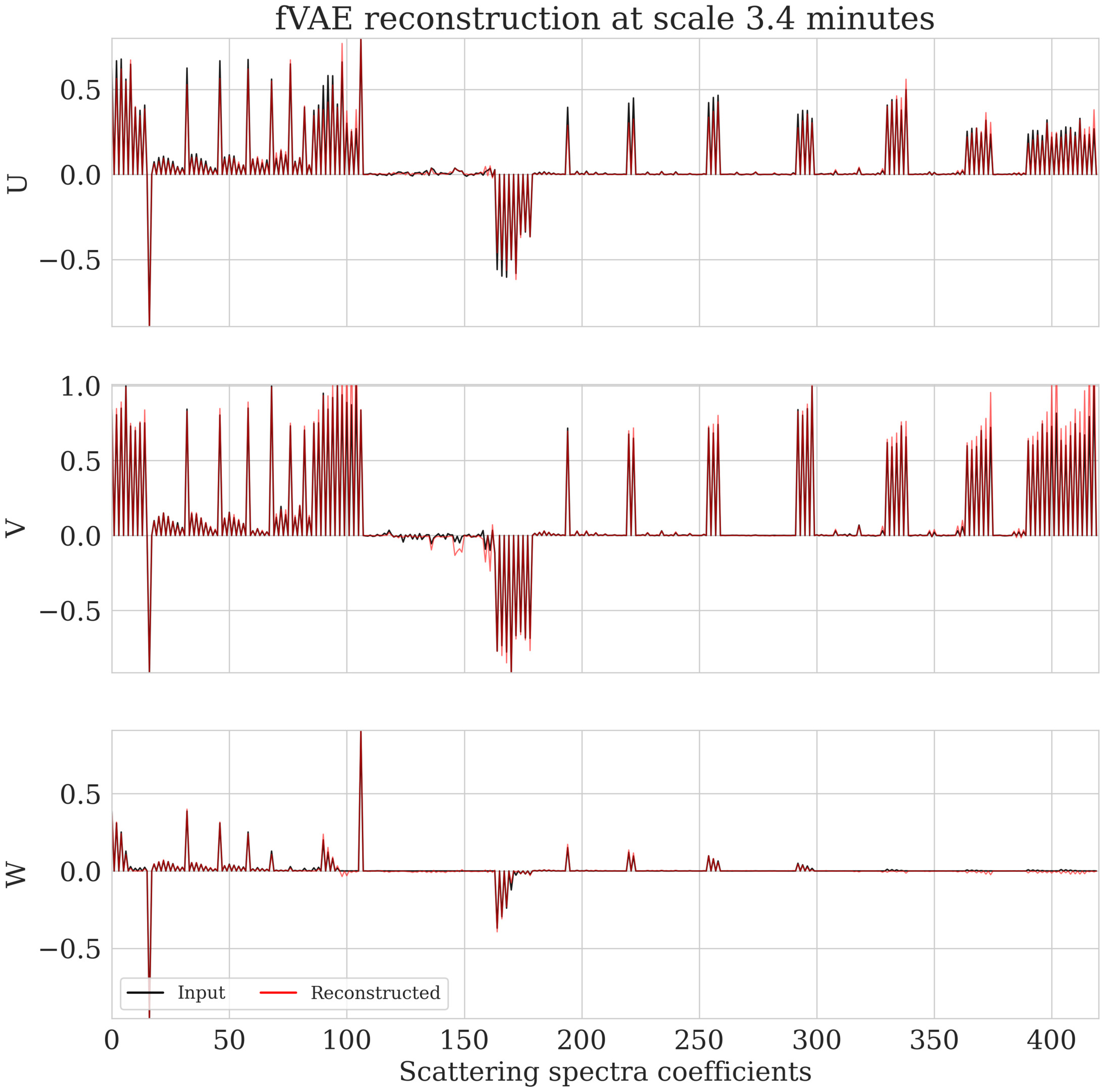}
        \caption{}
        \label{fig:app_decoder_scale-2-3}
      \end{subfigure}

    \caption{The input scattering spectra (black) and reconstruction via the fVAE decoder (red) for the U, V, and W components of four random windows from the $3.4$-minute timescale.}
    \label{fig:app_decoder_scale-2}
\end{figure*}

\begin{figure*}[p]
    \centering

    \begin{subfigure}[t]{0.495\textwidth}
        \centering
        \includegraphics[width=\textwidth]{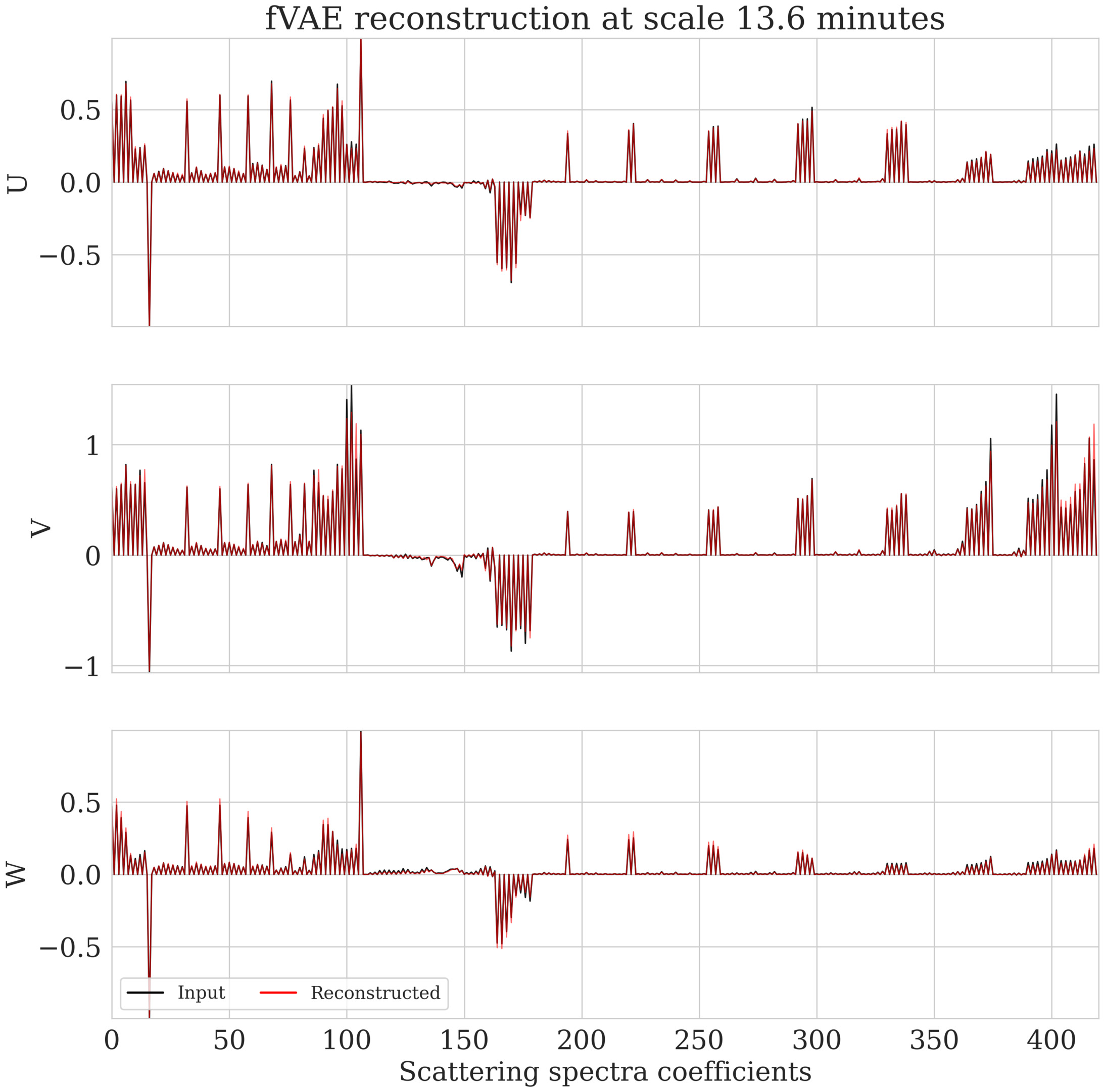}
        \caption{}
        \label{fig:app_decoder_scale-3-0}
      \end{subfigure}\hspace{0.1em}
    \begin{subfigure}[t]{0.495\textwidth}
        \centering
        \includegraphics[width=\textwidth]{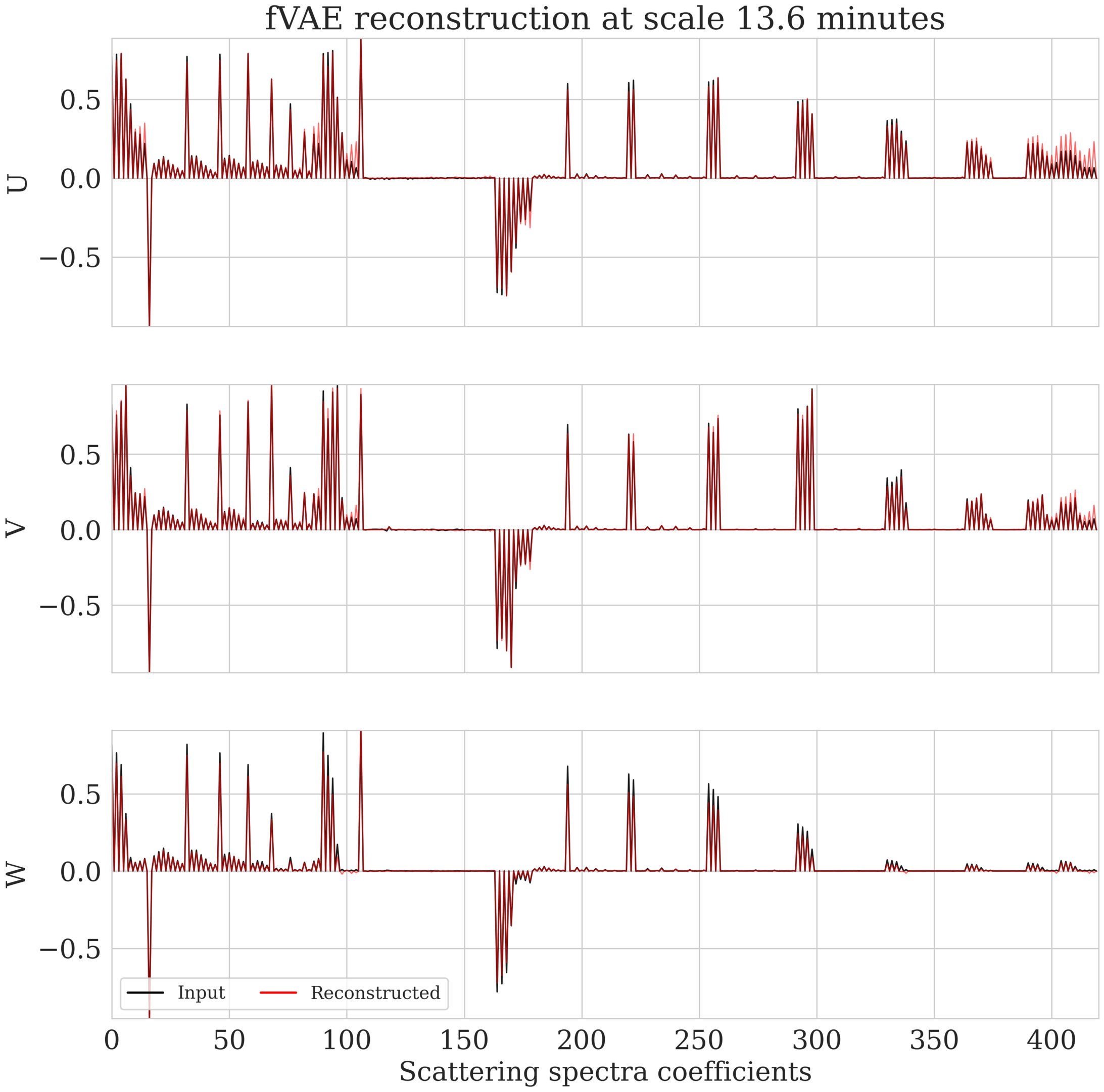}
        \caption{}
        \label{fig:app_decoder_scale-3-1}
      \end{subfigure}

    \begin{subfigure}[t]{0.495\textwidth}
        \centering
        \includegraphics[width=\textwidth]{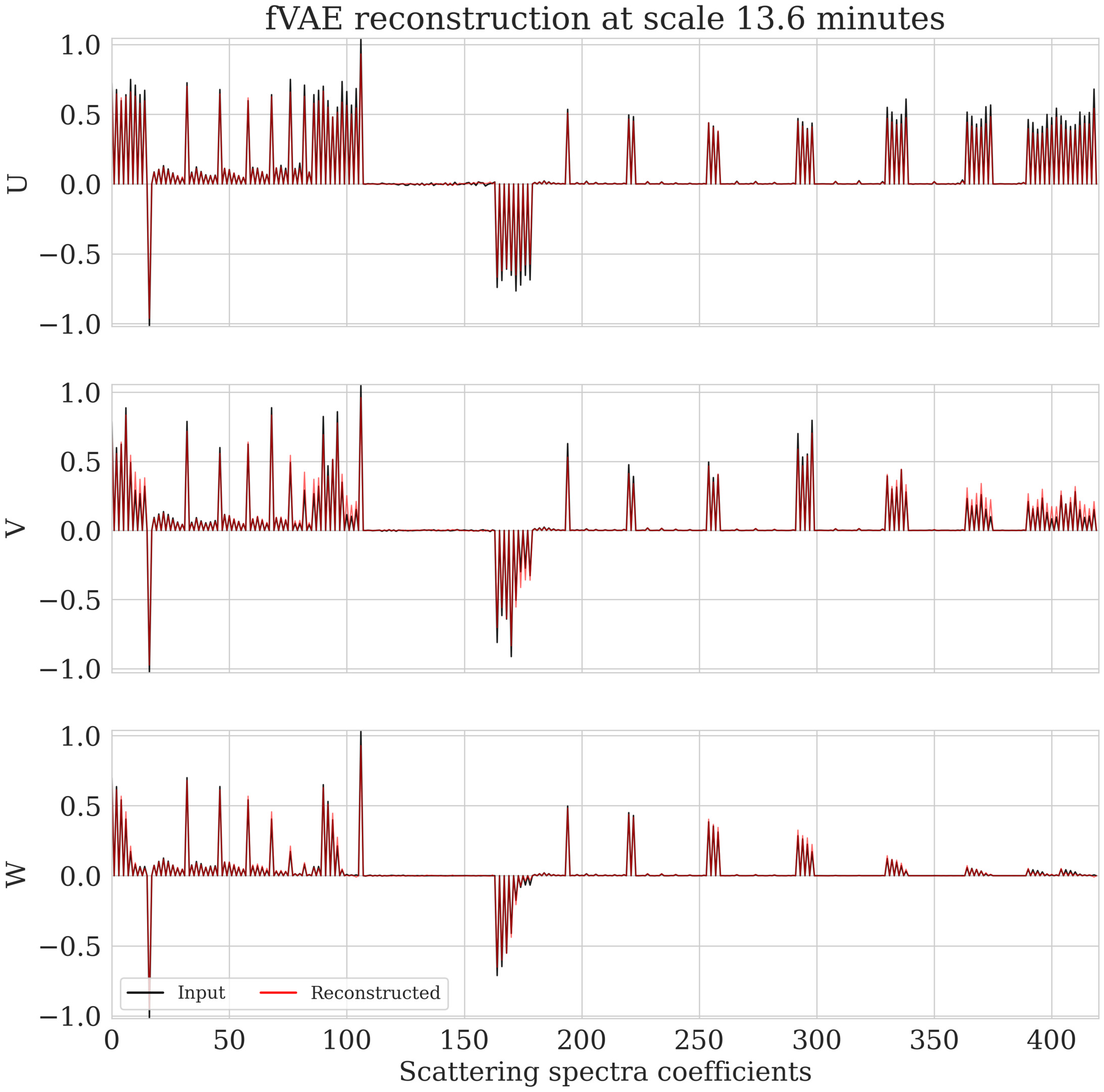}
        \caption{}
        \label{fig:app_decoder_scale-3-2}
      \end{subfigure}\hspace{0.1em}
    \begin{subfigure}[t]{0.495\textwidth}
        \centering
        \includegraphics[width=\textwidth]{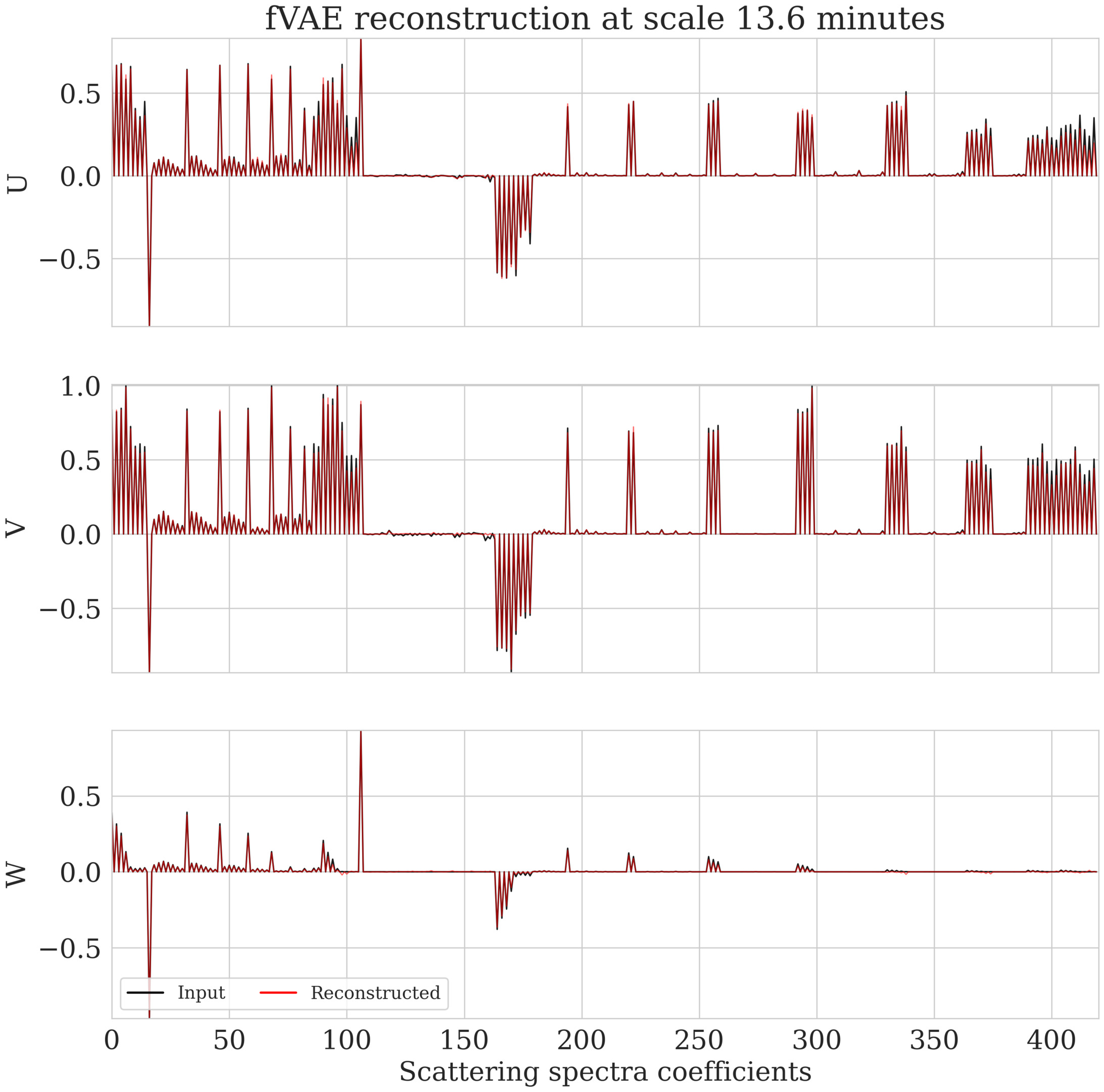}
        \caption{}
        \label{fig:app_decoder_scale-3-3}
      \end{subfigure}

    \caption{The input scattering spectra (black) and reconstruction via the fVAE decoder (red) for the U, V, and W components of four random windows from the $13.6$-minute timescale.}
    \label{fig:app_decoder_scale-3}
\end{figure*}

\begin{figure*}[p]
    \centering

    \begin{subfigure}[t]{0.495\textwidth}
        \centering
        \includegraphics[width=\textwidth]{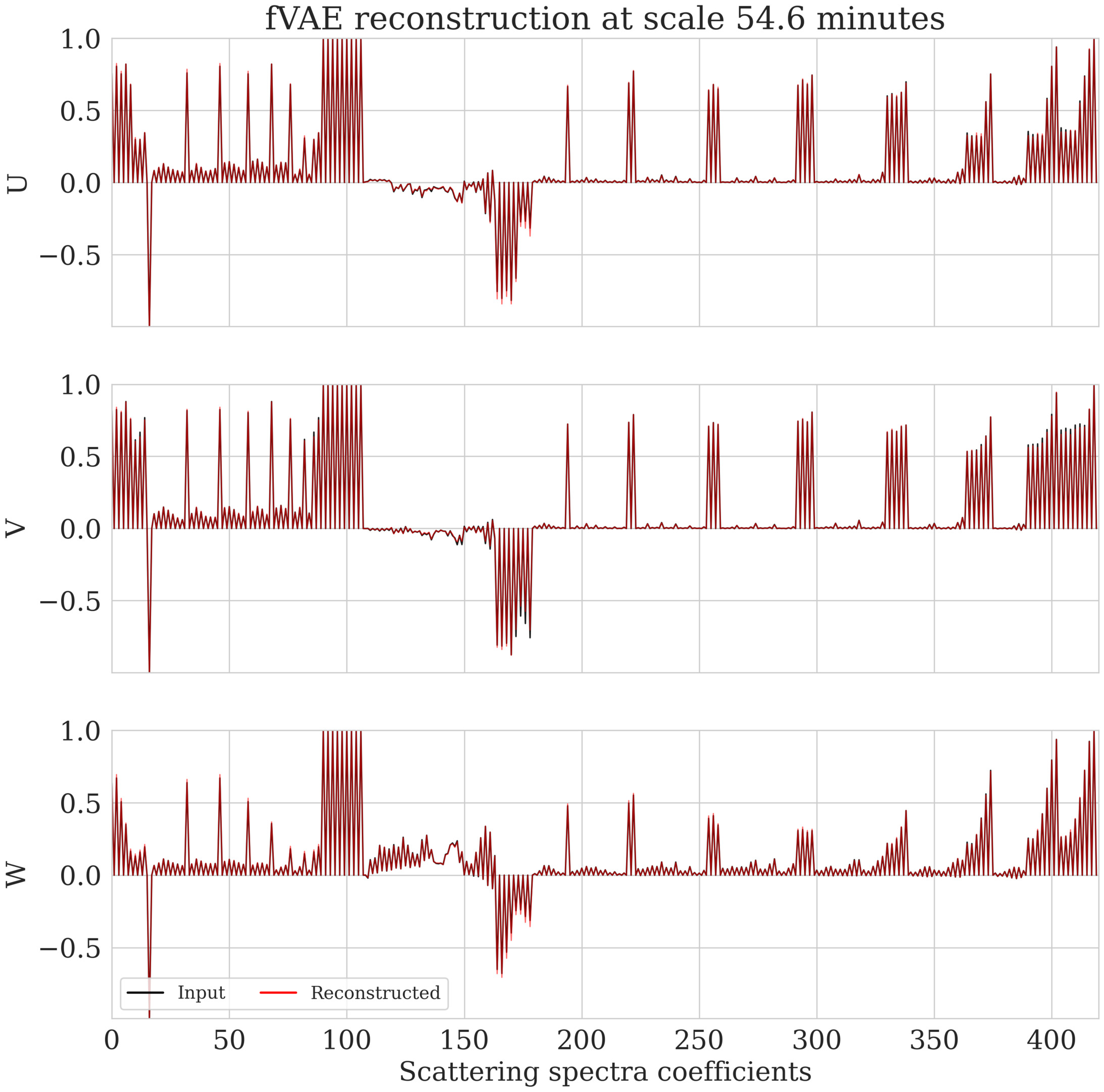}
        \caption{}
        \label{fig:app_decoder_scale-4-0}
      \end{subfigure}\hspace{0.1em}
    \begin{subfigure}[t]{0.495\textwidth}
        \centering
        \includegraphics[width=\textwidth]{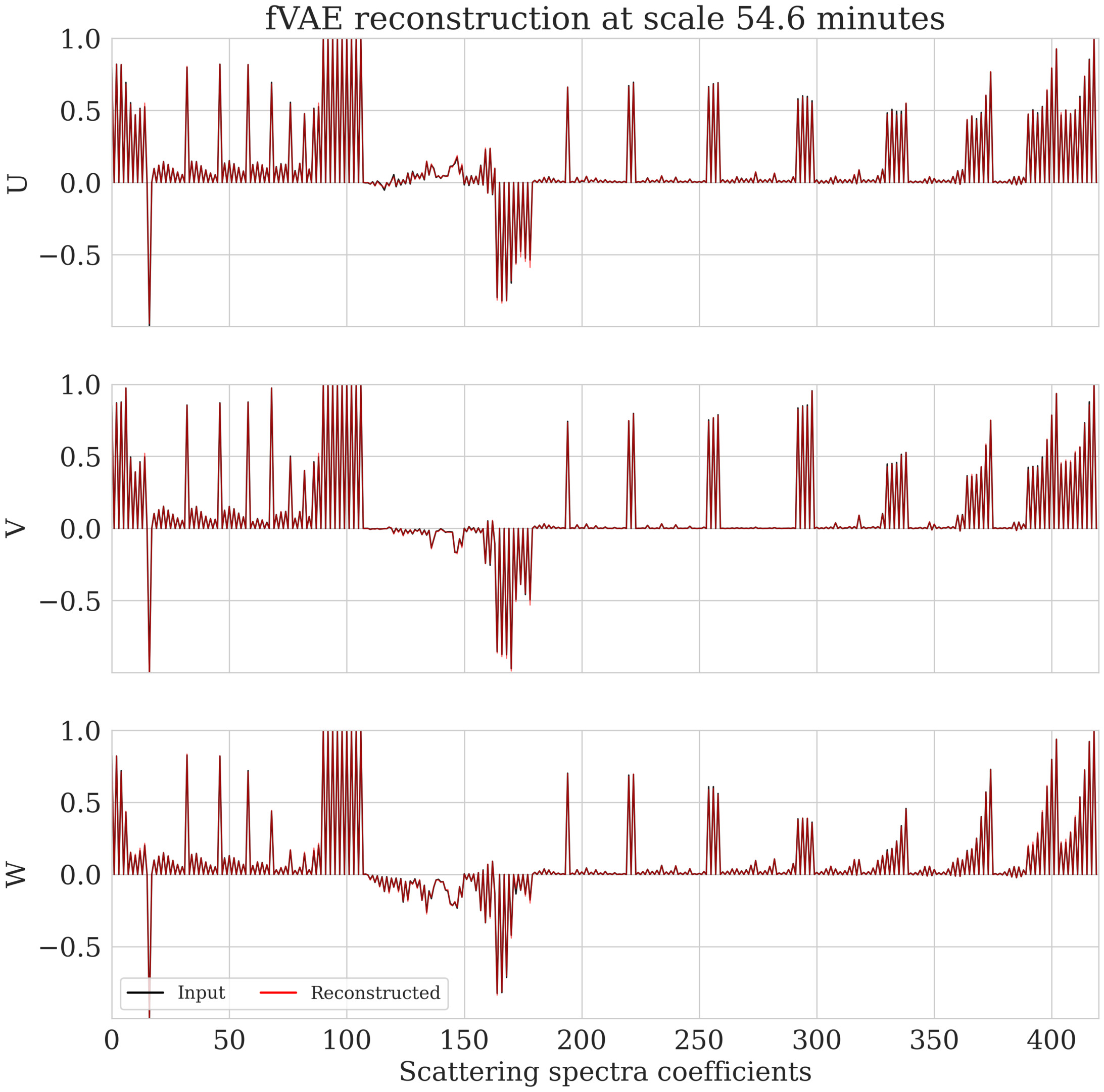}
        \caption{}
        \label{fig:app_decoder_scale-4-1}
      \end{subfigure}

    \begin{subfigure}[t]{0.495\textwidth}
        \centering
        \includegraphics[width=\textwidth]{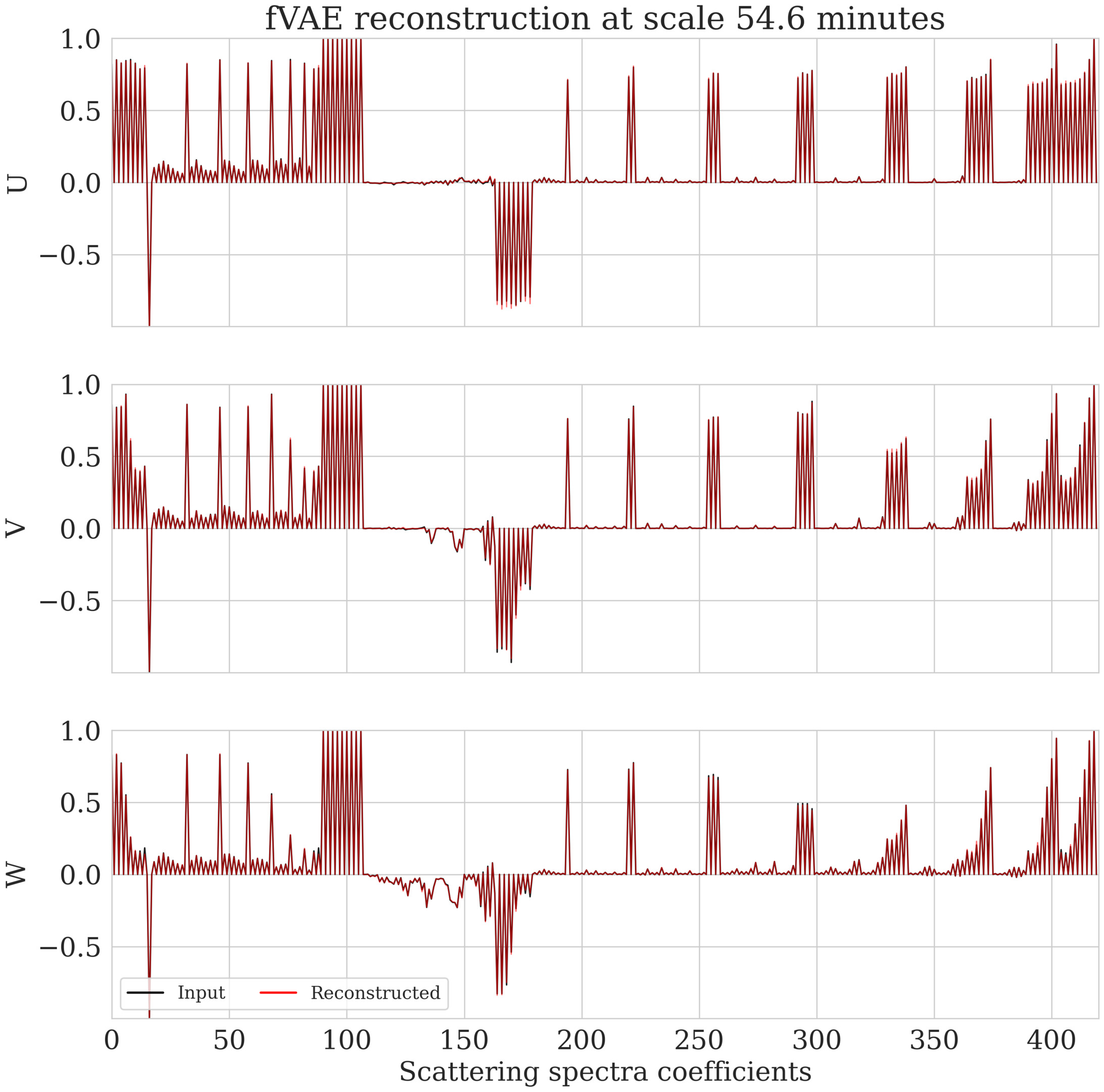}
        \caption{}
        \label{fig:app_decoder_scale-4-2}
      \end{subfigure}\hspace{0.1em}
    \begin{subfigure}[t]{0.495\textwidth}
        \centering
        \includegraphics[width=\textwidth]{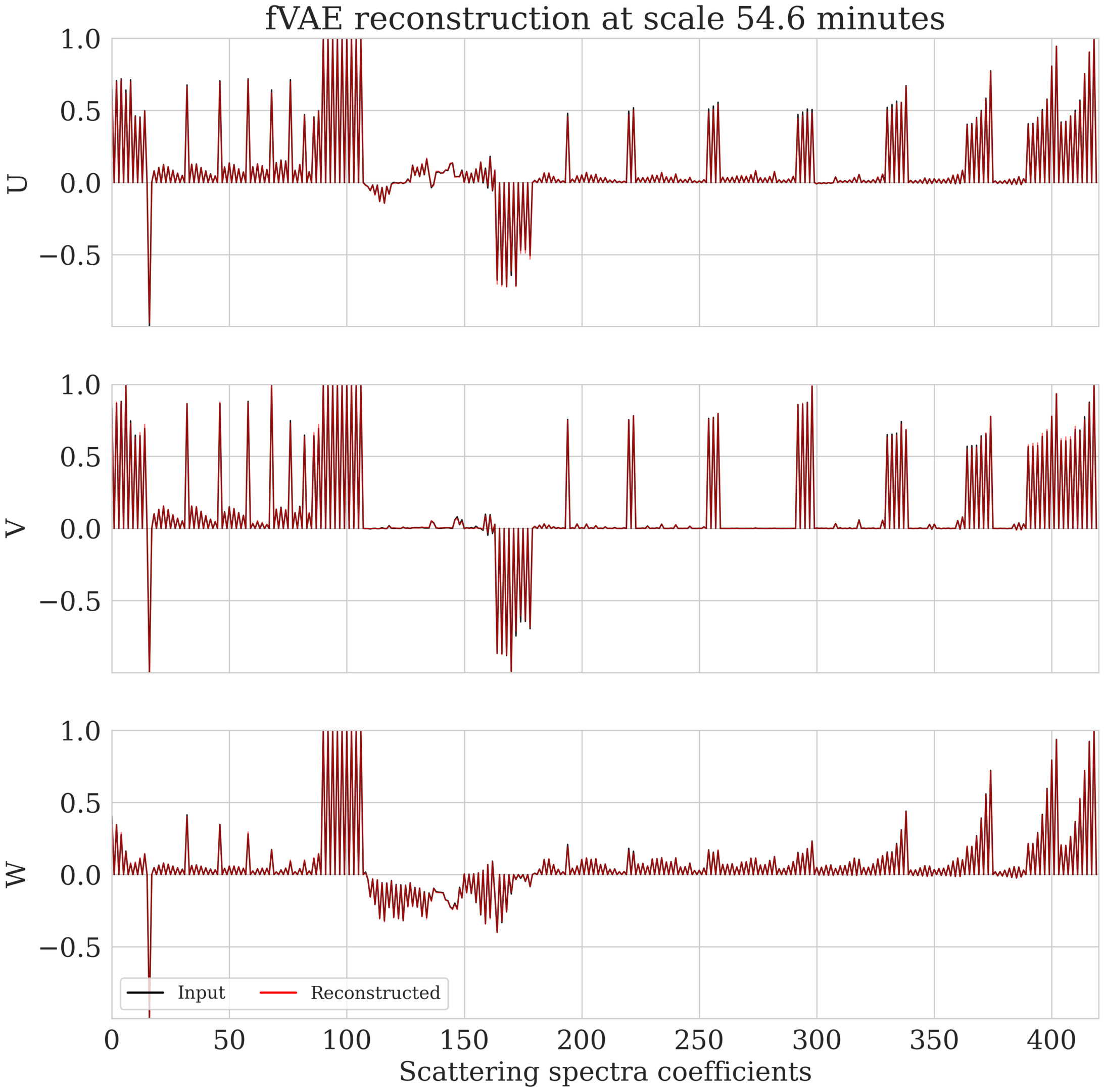}
        \caption{}
        \label{fig:app_decoder_scale-4-3}
      \end{subfigure}

    \caption{The input scattering spectra (black) and reconstruction via the fVAE decoder (red) for the U, V, and W components of four random windows from the $54.6$-minute timescale.}
    \label{fig:app_decoder_scale-4}
\end{figure*}

In this section, we present the reconstructed input scattering spectra
through the fVAE.
\cref{fig:app_decoder_scale-1,fig:app_decoder_scale-2,fig:app_decoder_scale-3,fig:app_decoder_scale-4}
summarize the results for  $51.2$-second timescale, $3.4$-minute
timescale, $13.6$-minute timescale, and $54.6$-minute timescale,
respectively. Each one of these figures shows the input scattering
spectra (in black) and the corresponding reconstructed input (in red)
for the U, V, and W components of four randomly selected waveforms.

\section{Baseline glitch separation comparison}
\label{app:baseline_glitch_separation}

\begin{figure*}[!t]
  \centering
  \begin{subfigure}[t]{0.325\textwidth}
      \includegraphics[width=\textwidth]{figs/glitch_removal/1/real-data.png}
      % \caption{}
      \caption{Raw waveform}
      \label{fig:source_separation_nasa_glitch_real}
  \end{subfigure}\hspace{0em}
  \begin{subfigure}[t]{0.325\textwidth}
      \includegraphics[width=\textwidth]{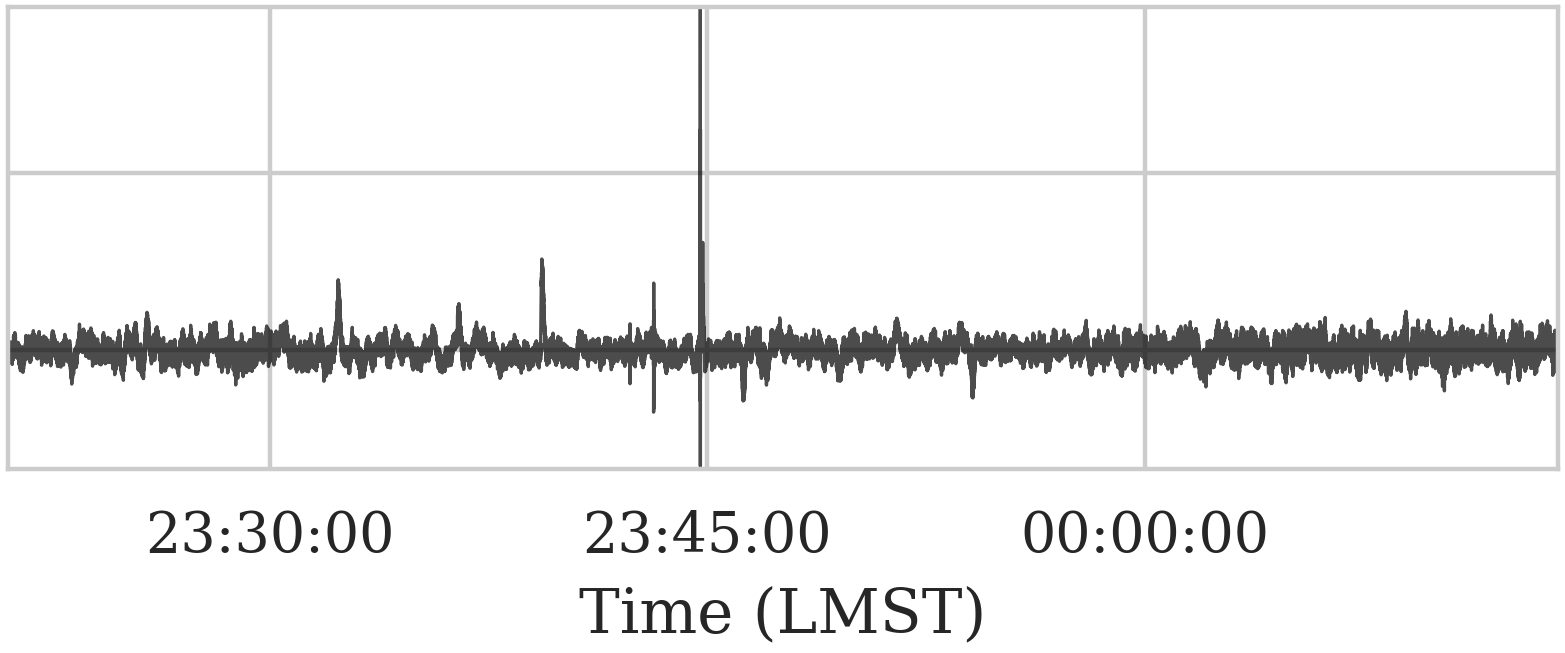}
      % \caption{}
      \caption{After separation}
      \label{fig:source_separation_nasa_glitch_background}
  \end{subfigure}\hspace{0em}
  \begin{subfigure}[t]{0.325\textwidth}
      \includegraphics[width=\textwidth]{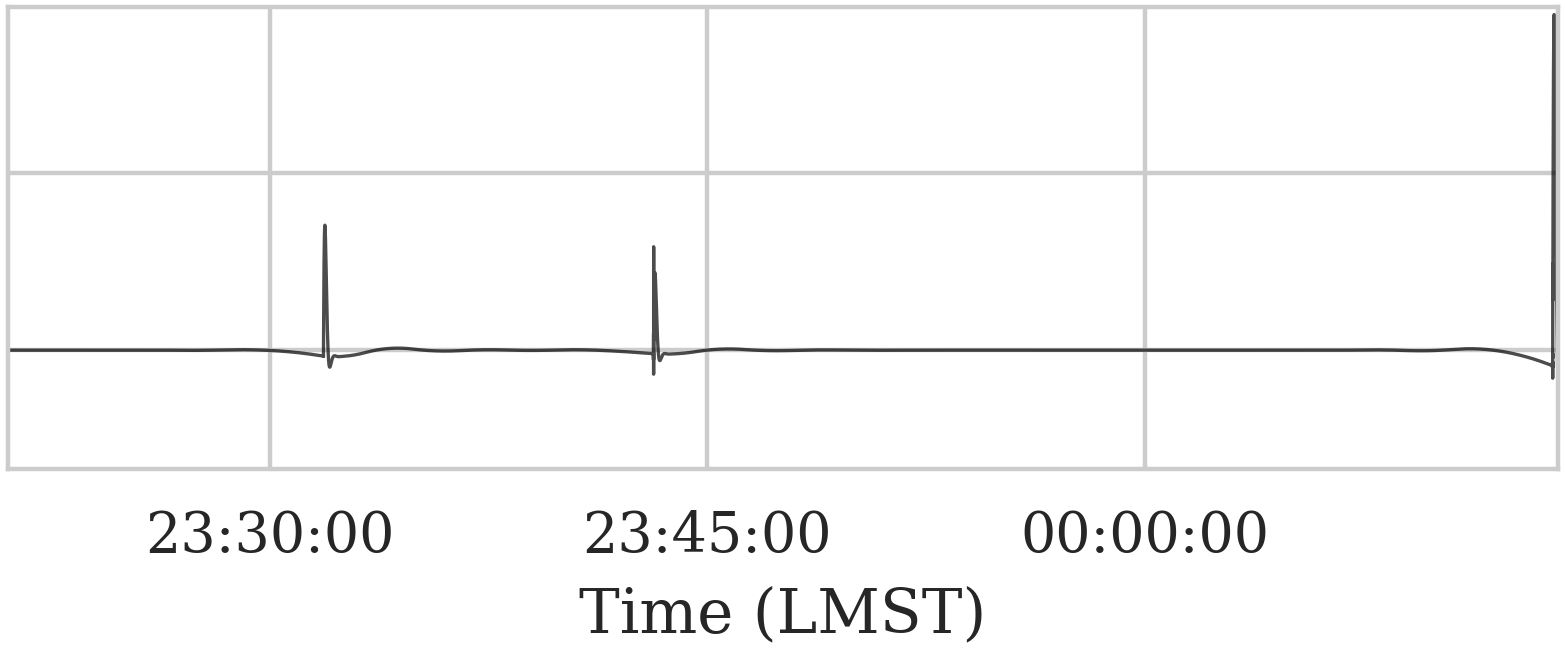}
      % \caption{}
      \caption{Separated signal}
      \label{fig:source_separation_nasa_glitch_source}
  \end{subfigure}

  \begin{subfigure}[t]{0.325\textwidth}
      \includegraphics[width=\textwidth]{figs/glitch_removal/1/zoom-13/x_obs_13.png}
      % \caption{}
      \caption{Raw waveform close-up}
      \label{fig:source_separation_nasa_glitch_real_zoom_13}
  \end{subfigure}\hspace{0em}
  \begin{subfigure}[t]{0.325\textwidth}
      \includegraphics[width=\textwidth]{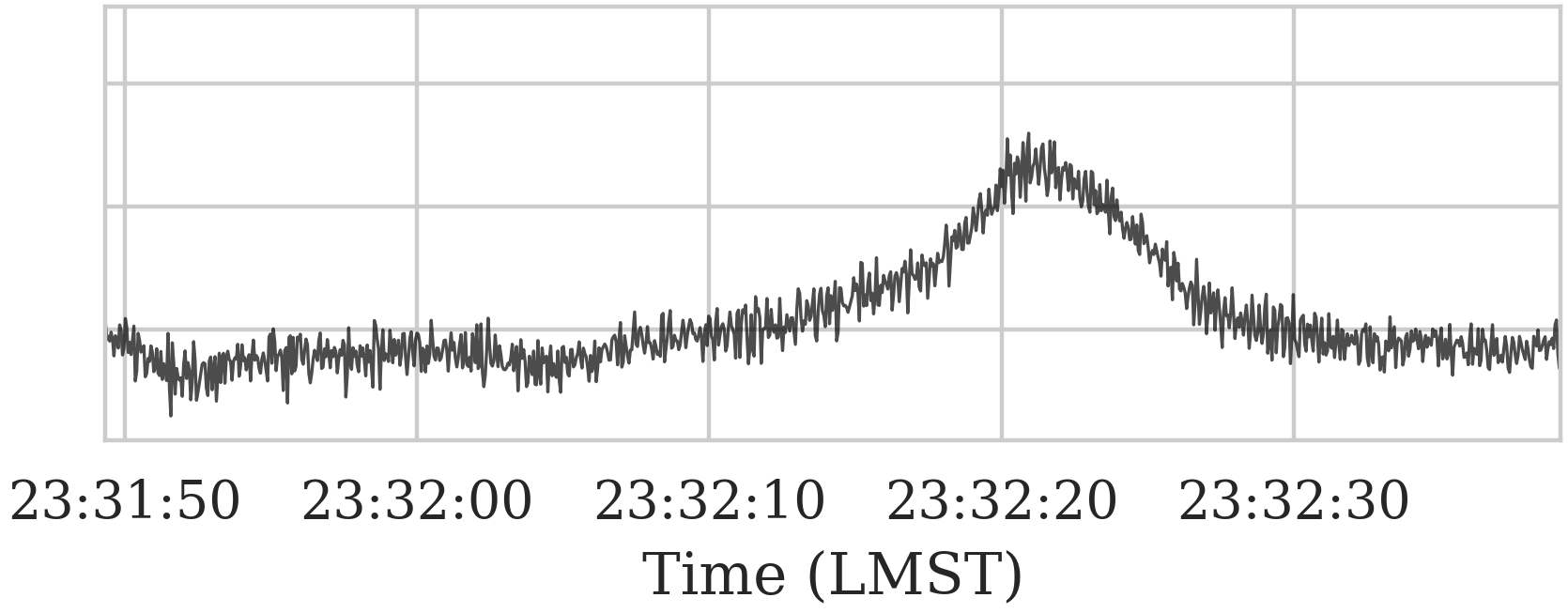}
      % \caption{}
      \caption{After separation close-up}
      \label{fig:source_separation_nasa_glitch_background_zoom_13}
  \end{subfigure}\hspace{0em}
  \begin{subfigure}[t]{0.325\textwidth}
      \includegraphics[width=\textwidth]{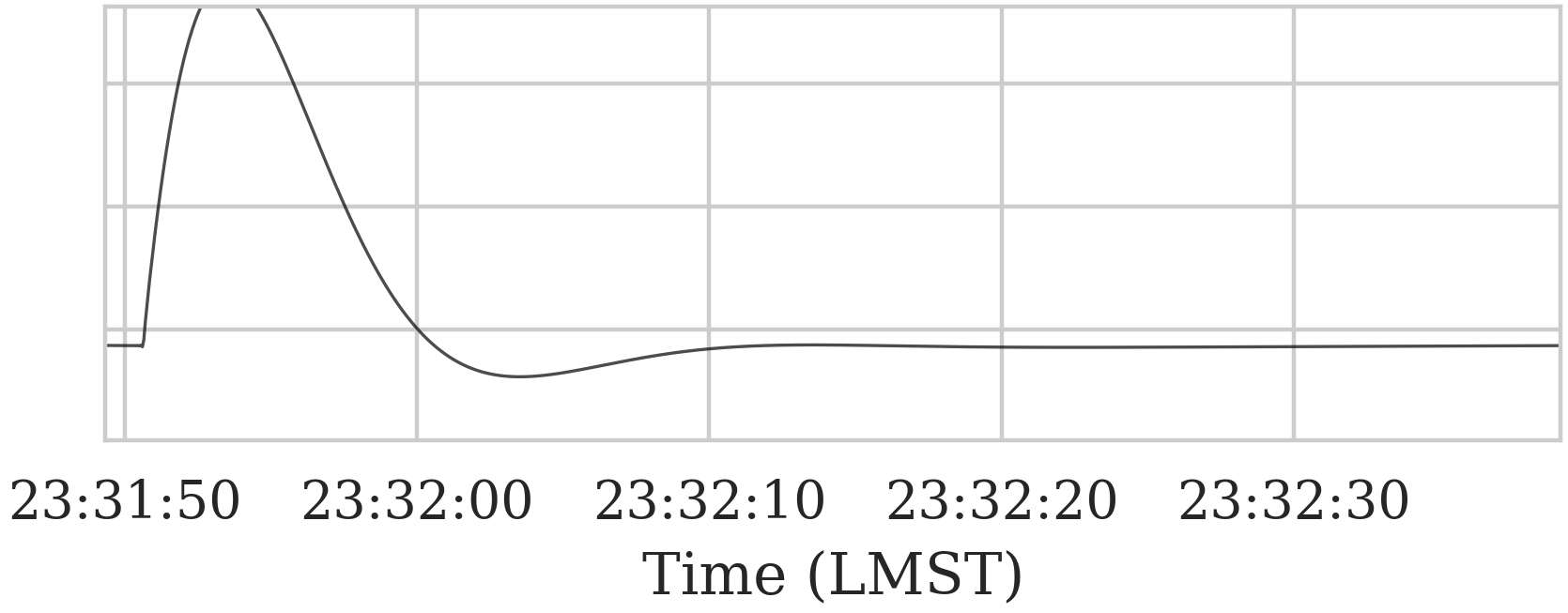}
      % \caption{}
      \caption{Separated signal close-up}
      \label{fig:source_separation_nasa_glitch_source_zoom_13}
  \end{subfigure}

  \begin{subfigure}[t]{0.325\textwidth}
      \includegraphics[width=\textwidth]{figs/glitch_removal/1/zoom-26/x_obs_26.png}
      % \caption{}
      \caption{Raw waveform close-up}
      \label{fig:source_separation_nasa_glitch_real_zoom_26}
  \end{subfigure}\hspace{0em}
  \begin{subfigure}[t]{0.325\textwidth}
      \includegraphics[width=\textwidth]{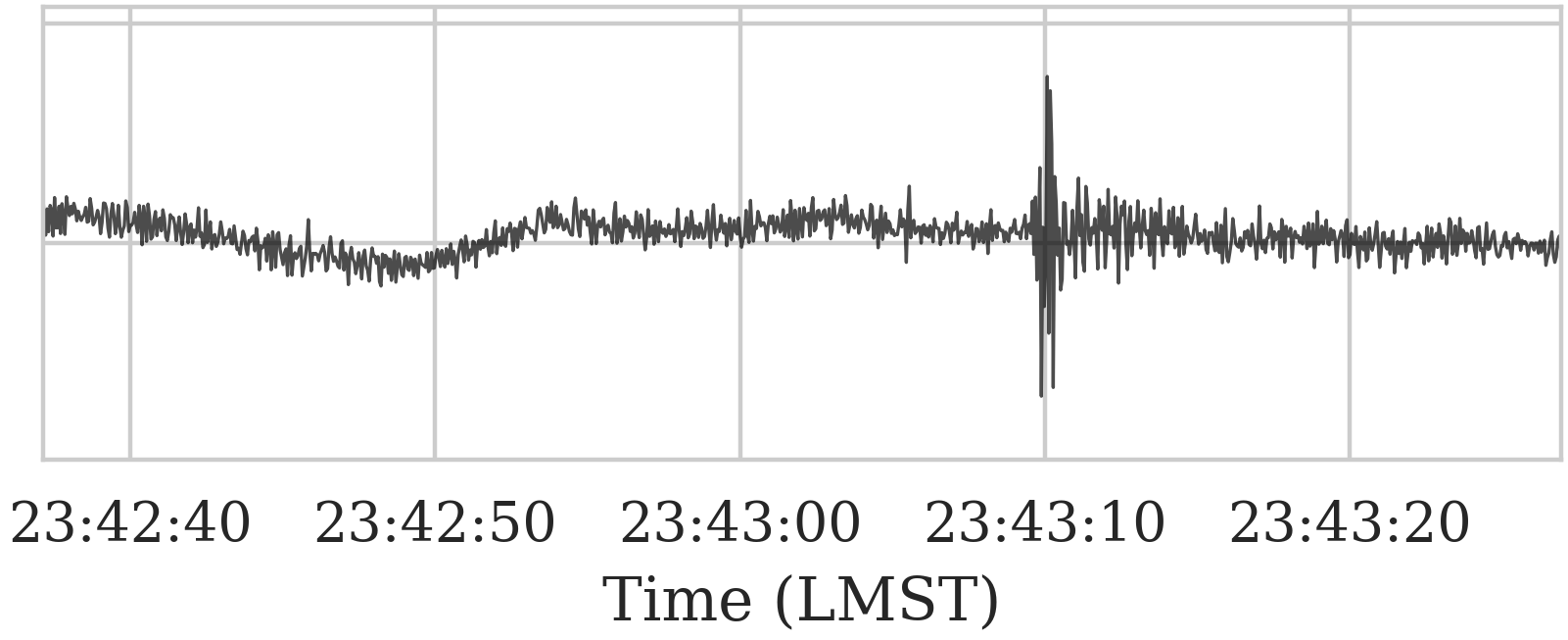}
      % \caption{}
      \caption{After separation close-up}
      \label{fig:source_separation_nasa_glitch_background_zoom_26}
  \end{subfigure}\hspace{0em}
  \begin{subfigure}[t]{0.325\textwidth}
      \includegraphics[width=\textwidth]{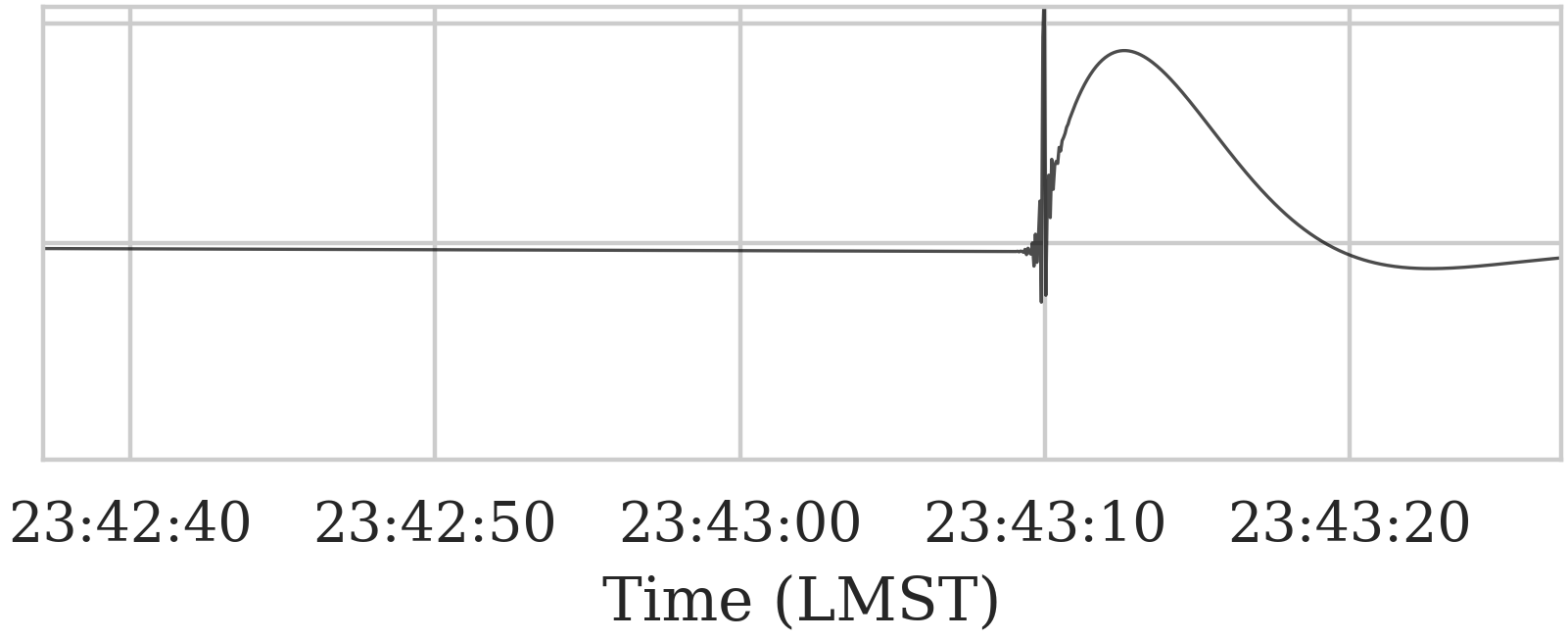}
      % \caption{}
      \caption{Separated signal close-up}
      \label{fig:source_separation_nasa_glitch_source_zoom_26}
  \end{subfigure}

  % \begin{subfigure}[t]{0.325\textwidth}
  %     \includegraphics[width=\textwidth]{figs/glitch_removal/1/zoom-28/x_obs_28.png}
  %     % \caption{}
  %     \caption{Raw waveform close-up}
  %     \label{fig:source_separation_nasa_glitch_real_zoom_28}
  % \end{subfigure}\hspace{0em}
  % \begin{subfigure}[t]{0.325\textwidth}
  %     \includegraphics[width=\textwidth]{figs/glitch_removal/1/zoom-28/x_hat_28.png}
  %     % \caption{}
  %     \caption{After separation close-up}
  %     \label{fig:source_separation_nasa_glitch_background_zoom_28}
  % \end{subfigure}\hspace{0em}
  % \begin{subfigure}[t]{0.325\textwidth}
  %     \includegraphics[width=\textwidth]{figs/glitch_removal/1/zoom-28/glitch_28.png}
  %     % \caption{}
  %     \caption{Separated signal close-up}
  %     \label{fig:source_separation_nasa_glitch_source_zoom_28}
  % \end{subfigure}

  \begin{subfigure}[t]{0.325\textwidth}
      \includegraphics[width=\textwidth]{figs/glitch_removal/1/zoom-38/x_obs_38.png}
      % \caption{}
      \caption{Raw waveform close-up}
      \label{fig:source_separation_nasa_glitch_real_zoom_38}
  \end{subfigure}\hspace{0em}
  \begin{subfigure}[t]{0.325\textwidth}
      \includegraphics[width=\textwidth]{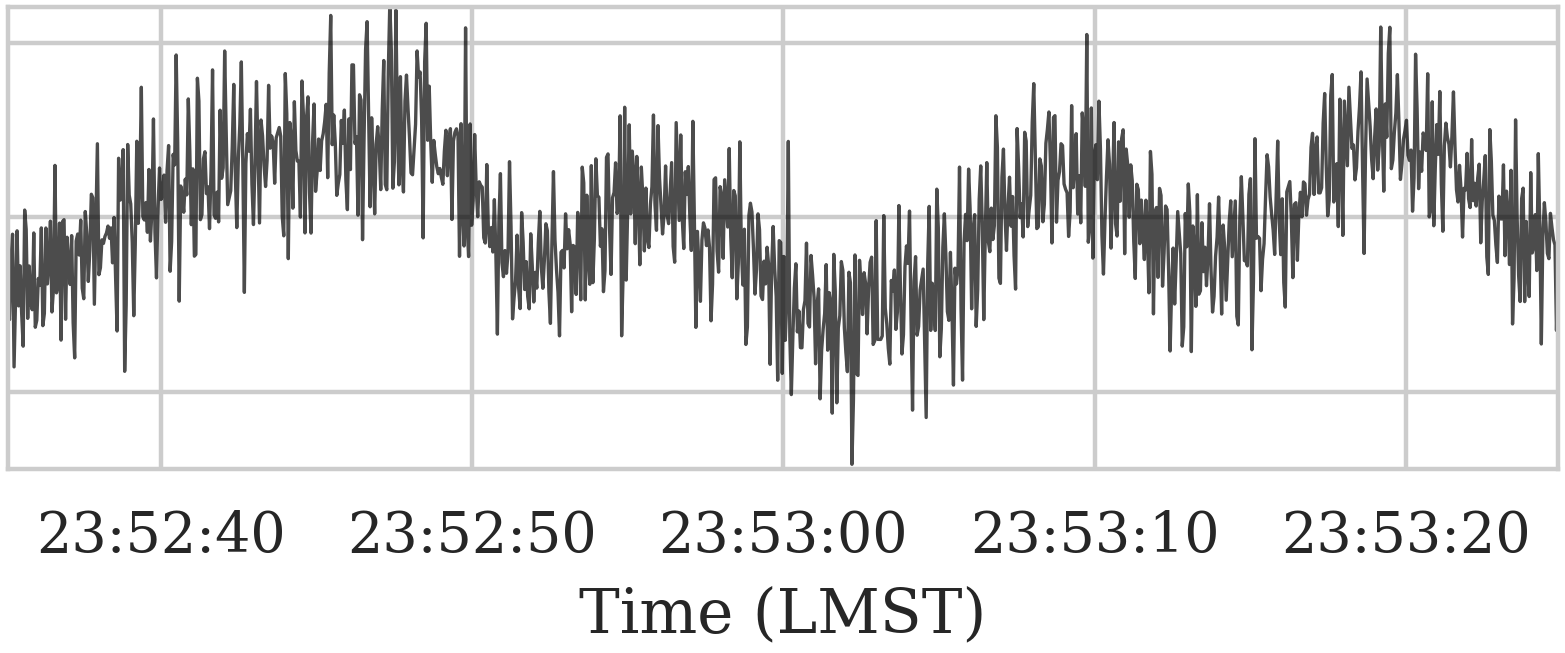}
      % \caption{}
      \caption{After separation close-up}
      \label{fig:source_separation_nasa_glitch_background_zoom_38}
  \end{subfigure}\hspace{0em}
  \begin{subfigure}[t]{0.325\textwidth}
      \includegraphics[width=\textwidth]{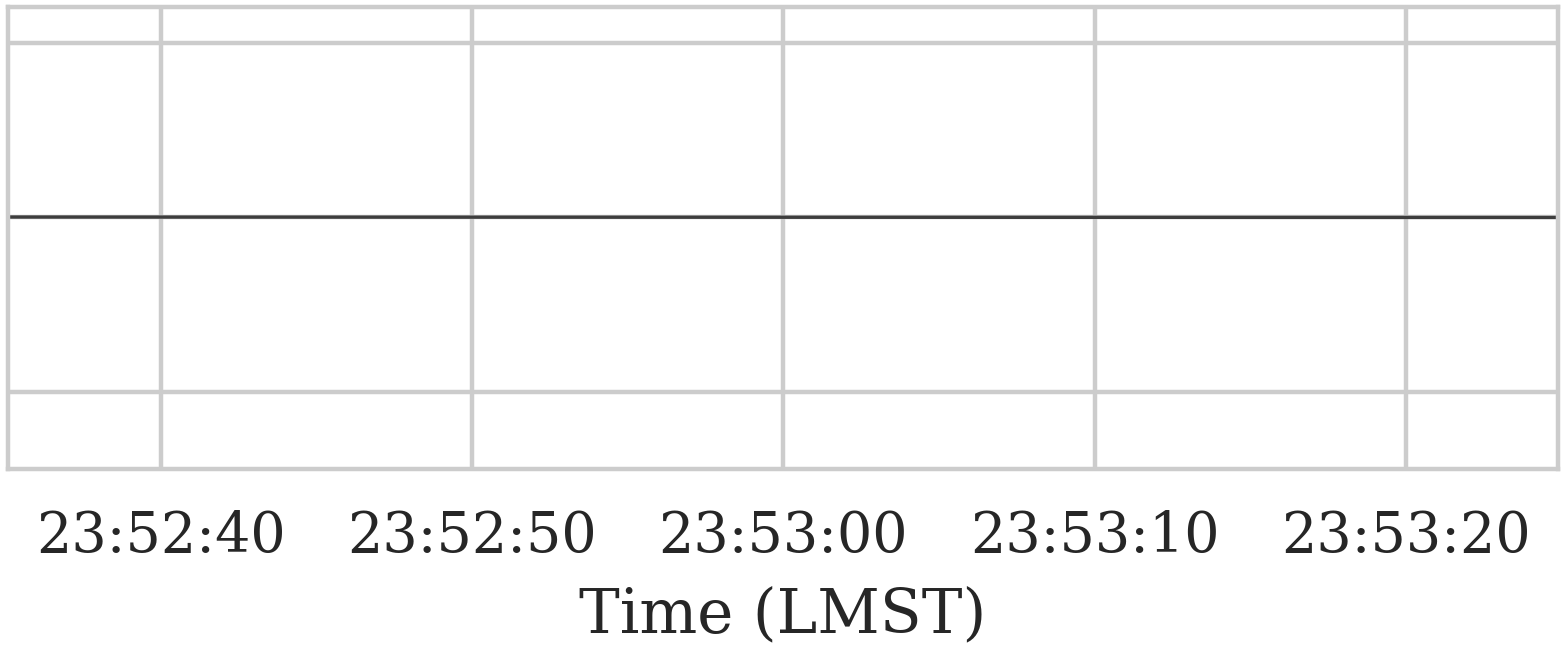}
      % \caption{}
      \caption{Separated signal close-up}
      \label{fig:source_separation_nasa_glitch_source_zoom_38}
  \end{subfigure}

  \vspace{0.5em}
  \caption{Separating glitches from a waveform extracted from cluster
  $4$ from the $54.6$-minute timescale using the baseline method
  \cite{ScholzWidmer_SchnidrigDavisEtAl2020} (juxtapose with \cref{fig:source_separation_glitch}).}
  \label{fig:source_separation_nasa_glitch}
\end{figure*}

We have included a baseline comparison with the NASA InSight glitch
separation algorithm \cite{ScholzWidmer_SchnidrigDavisEtAl2020}, which
represents the current state-of-practice for processing InSight seismic
data. We illustrate the results of this comparison in \cref{fig:source_separation_nasa_glitch}.

This method employs a multi-step pipeline including data decimation,
deconvolution with band-pass filtering, time derivative calculation to
convert acceleration steps into impulse-like signals, and
threshold-based detection followed by parametric modeling of glitch
signatures. The separation process assumes observed data as a linear
combination of glitch and background signal, then fits a three-parameter
model (amplitude, offset, and linear trend) through nonlinear least
squares optimization before subtracting the fitted glitch from the
original data. While this approach represents established practice, it
has fundamental limitations compared to our method. The NASA algorithm
relies on explicit parametric modeling of glitch functional forms and
assumes simple linear mixing, whereas our approach learns the
statistical structure of background noise through wavelet scattering
spectra without making assumptions about source morphology. As
demonstrated in \cref{fig:source_separation_nasa_glitch}, our method
achieves better separation quality by leveraging learned priors from
clustering rather than rigid parametric templates, highlighting the
advantage of data-driven approaches for complex planetary phenomena
where source characteristics are not well understood a priori.
Specifically, as confirmed by experts at the InSight team, our approach
removes multiple glitches that the baseline has ignored while also
performing better on glitches that have a spiking precursor, which the
baseline method struggles to fully separate (compare third row of
\cref{fig:source_separation_glitch} and
\cref{fig:source_separation_nasa_glitch}.)

\section{Comparison with ICA}
\label{app:ica_comparison}
To check whether the classical unsupervised source separation methods
are able to perform source separation in our setting, we implemented ICA
on our test cases including glitch separation and wind burst separation.
As anticipated from our problem formulation, ICA failed to produce
accurate separations for the desired sources. This failure confirms our
initial assessment that traditional methods designed for stationary
sources are inadequate for the complex temporal phenomena encountered in
planetary seismic data. ICA's poor performance stems from the
fundamental mismatch between its core assumptions and the
characteristics of Martian seismic data. The method operates on fixed
time windows with stationarity assumptions, while our sources span
vastly different timescales (51.2 seconds to 54.6 minutes) and exhibit
strong non-stationarity through diurnal and seasonal variations as
demonstrated in \cref{fig:season_all_clusters}. Additionally, ICA relies
solely on statistical independence without domain-specific knowledge,
whereas our successful separations depend critically on learned priors
from clustering similar source signatures. This comparison validates our
multi-scale approach and demonstrates why domain-aware methodology is
essential for analyzing complex planetary phenomena that systematically
violate the assumptions underlying traditional signal processing
methods.

\begin{figure*}[!t]
  \centering
  \begin{subfigure}[t]{0.325\textwidth}
      \includegraphics[width=\textwidth]{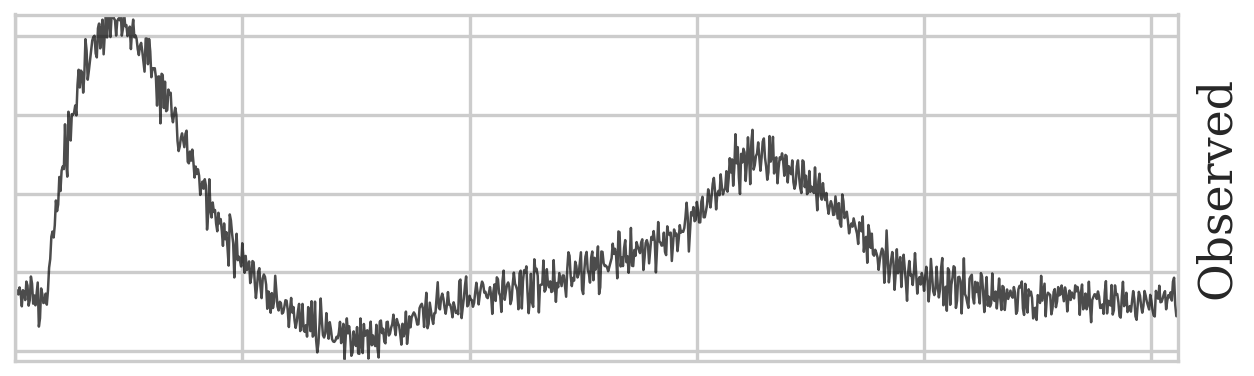}
      \caption{Multiple glitches (raw)}
      \label{fig:source_separation_glitch_real_13}
  \end{subfigure}\hspace{0em}
  \begin{subfigure}[t]{0.325\textwidth}
      \includegraphics[width=\textwidth]{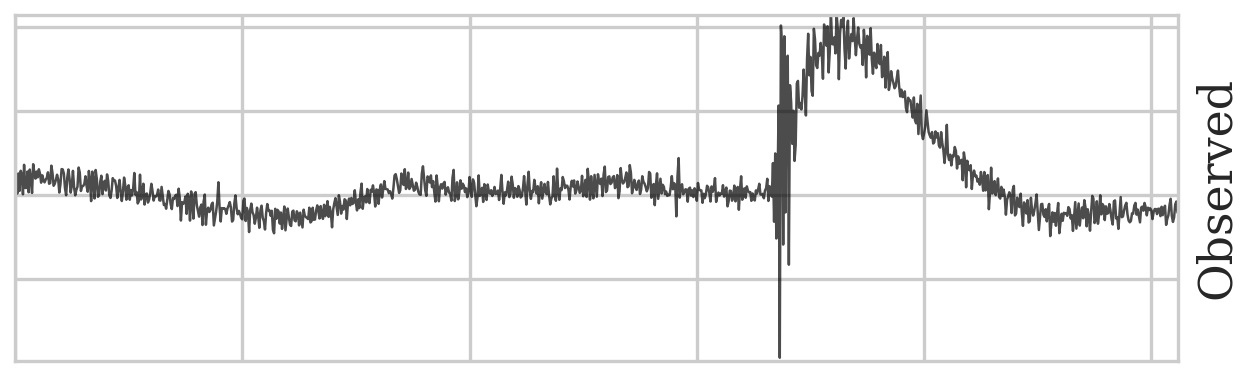}
      \caption{Glitch with precursor (raw)}
      \label{fig:source_separation_glitch_real_26}
  \end{subfigure}\hspace{0em}
  \begin{subfigure}[t]{0.325\textwidth}
      \includegraphics[width=\textwidth]{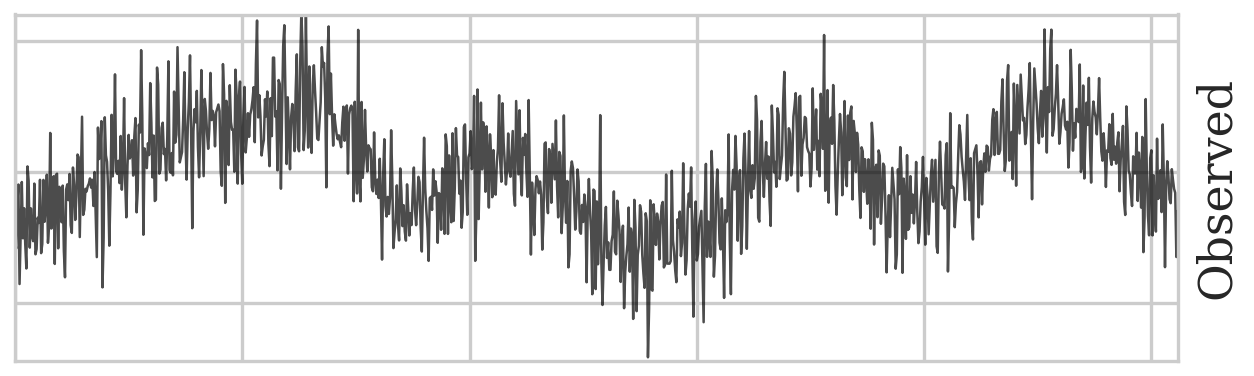}
      \caption{Clean background region (raw)}
      \label{fig:source_separation_glitch_real_38}
  \end{subfigure}

  \begin{subfigure}[t]{0.325\textwidth}
      \includegraphics[width=\textwidth]{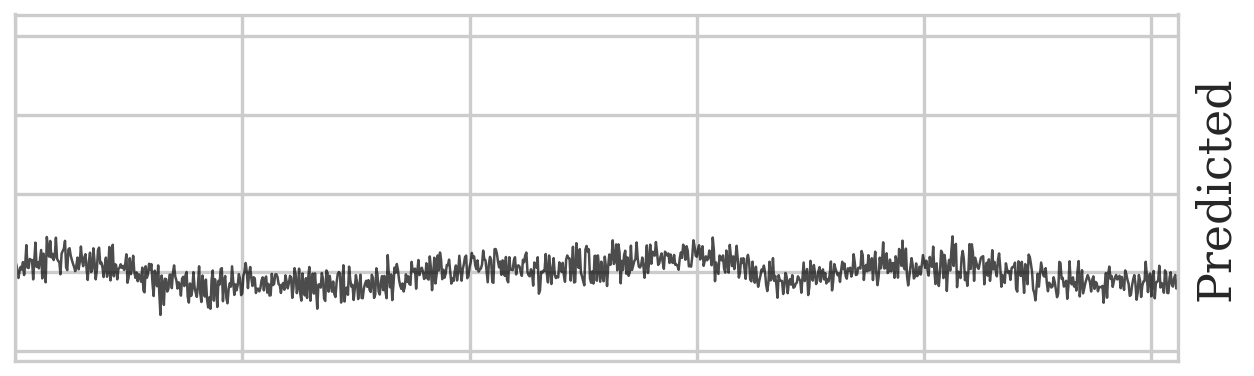}
      \caption{Independent component 1}
      \label{fig:source_separation_glitch_ica_13_1}
  \end{subfigure}\hspace{0em}
  \begin{subfigure}[t]{0.325\textwidth}
      \includegraphics[width=\textwidth]{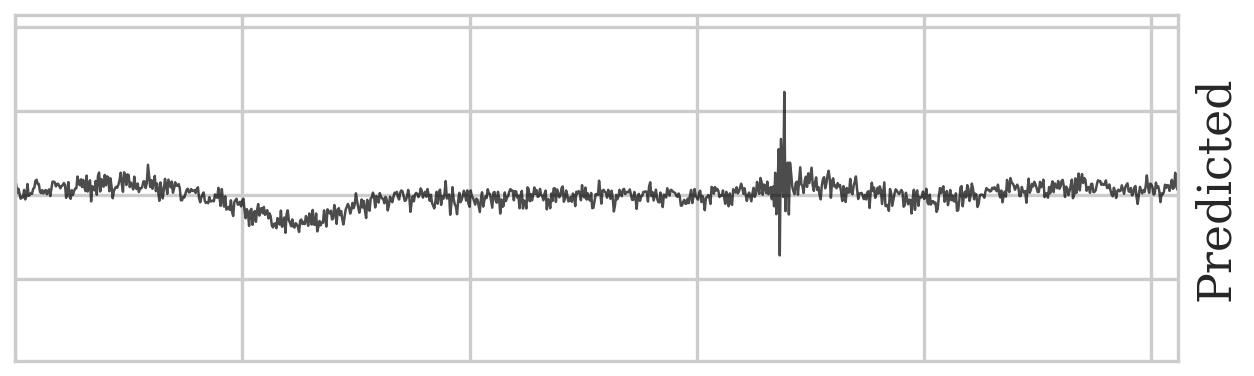}
      \caption{Independent component 1}
      \label{fig:source_separation_glitch_ica_26_1}
  \end{subfigure}\hspace{0em}
  \begin{subfigure}[t]{0.325\textwidth}
      \includegraphics[width=\textwidth]{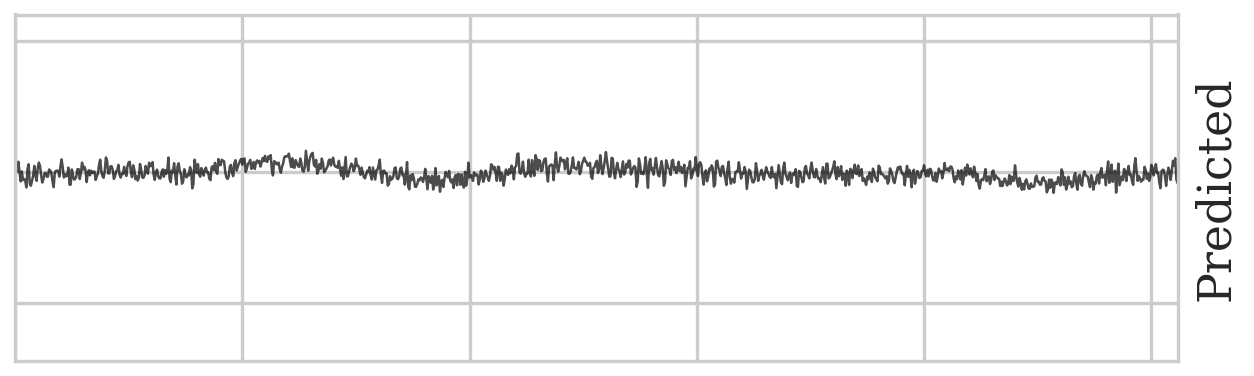}
      \caption{Independent component 1}
      \label{fig:source_separation_glitch_ica_38_1}
  \end{subfigure}

  \begin{subfigure}[t]{0.325\textwidth}
      \includegraphics[width=\textwidth]{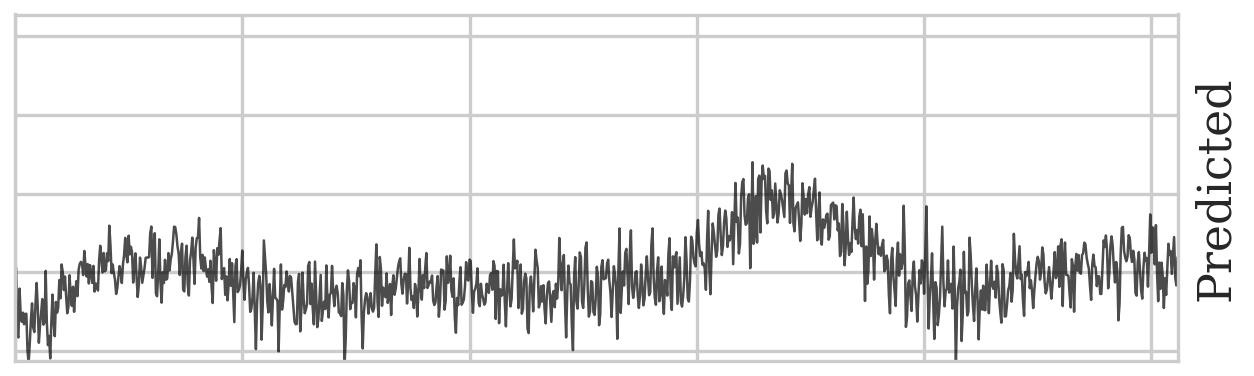}
      \caption{Independent component 2}
      \label{fig:source_separation_glitch_ica_13_2}
  \end{subfigure}\hspace{0em}
  \begin{subfigure}[t]{0.325\textwidth}
      \includegraphics[width=\textwidth]{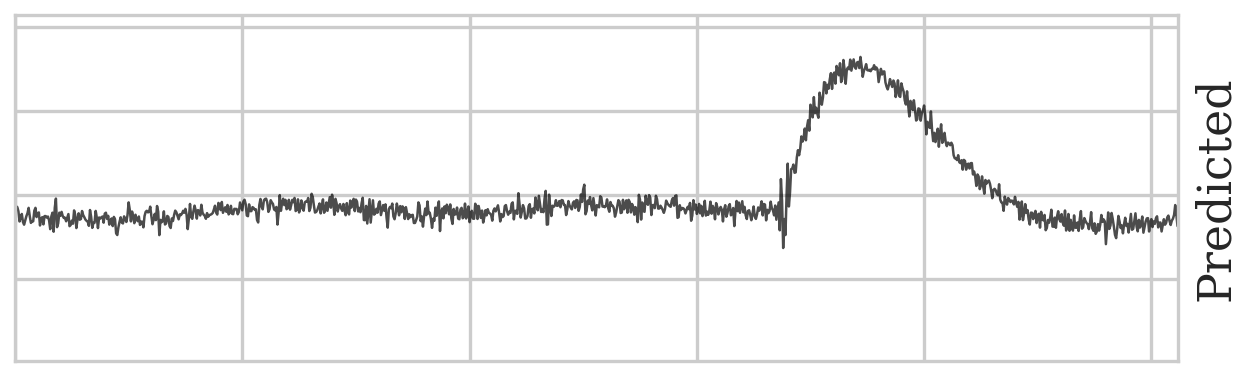}
      \caption{Independent component 2}
      \label{fig:source_separation_glitch_ica_26_2}
  \end{subfigure}\hspace{0em}
  \begin{subfigure}[t]{0.325\textwidth}
      \includegraphics[width=\textwidth]{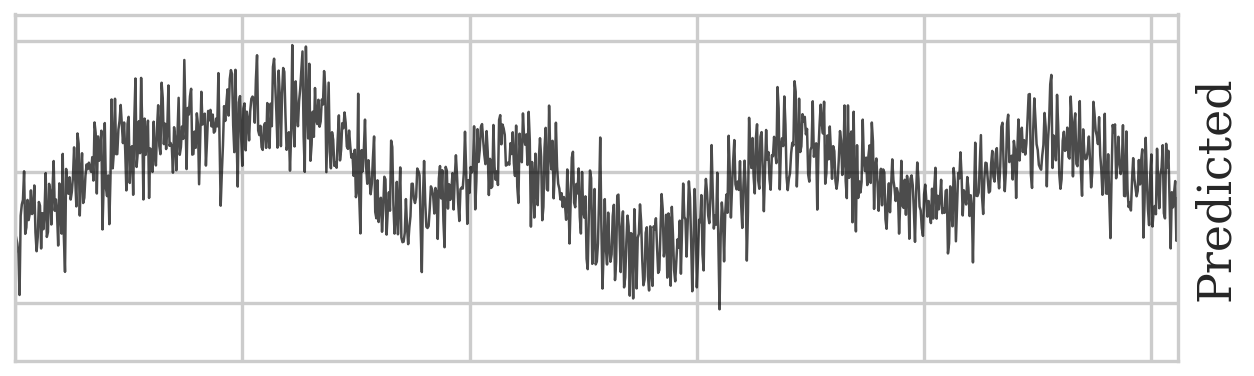}
      \caption{Independent component 2}
      \label{fig:source_separation_glitch_ica_38_2}
  \end{subfigure}

  \begin{subfigure}[t]{0.325\textwidth}
      \includegraphics[width=\textwidth]{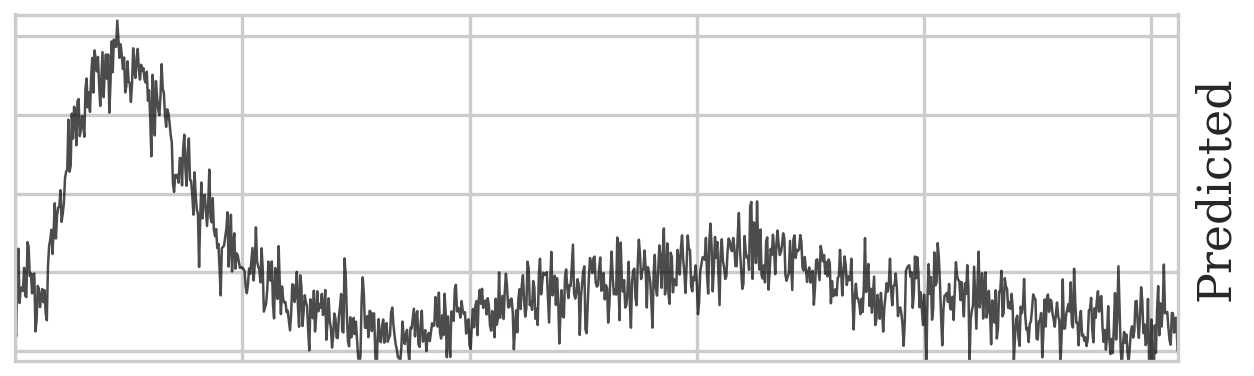}
      \caption{Independent component 3}
      \label{fig:source_separation_glitch_ica_13_3}
  \end{subfigure}\hspace{0em}
  \begin{subfigure}[t]{0.325\textwidth}
      \includegraphics[width=\textwidth]{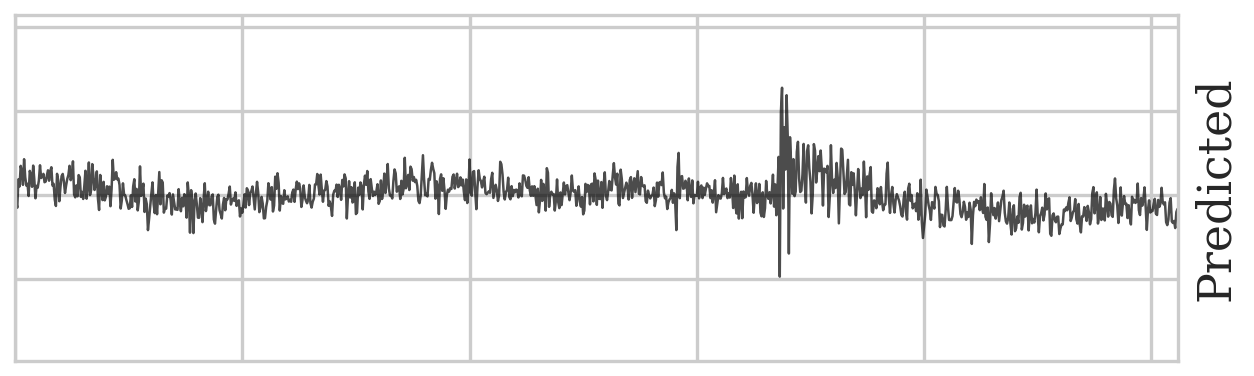}
      \caption{Independent component 3}
      \label{fig:source_separation_glitch_ica_26_3}
  \end{subfigure}\hspace{0em}
  \begin{subfigure}[t]{0.325\textwidth}
      \includegraphics[width=\textwidth]{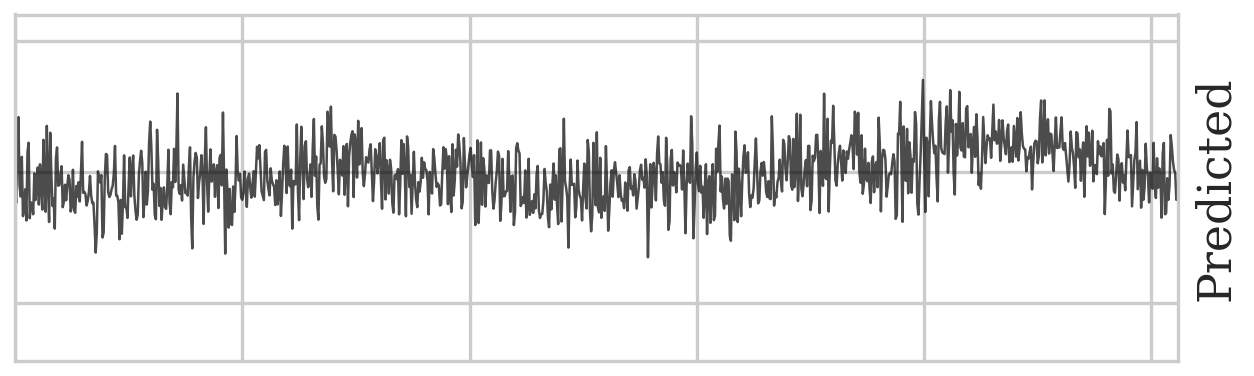}
      \caption{Independent component 3}
      \label{fig:source_separation_glitch_ica_38_3}
  \end{subfigure}

  \caption{ICA-based source separation applied to three-component seismic data (U, V, W channels) for comparison with our proposed method. FastICA decomposes each mixed signal into three independent components, shown in rows 2-4. Each column represents a different time window: multiple glitches, glitch with precursor, and clean background region. The three independent components sum to reconstruct the original U-component signal (top row).}
  \label{fig:source_separation_glitch_ica}
\end{figure*}

\begin{figure*}[t]
  \centering

  \begin{subfigure}[t]{0.325\textwidth}
      \includegraphics[width=\textwidth]{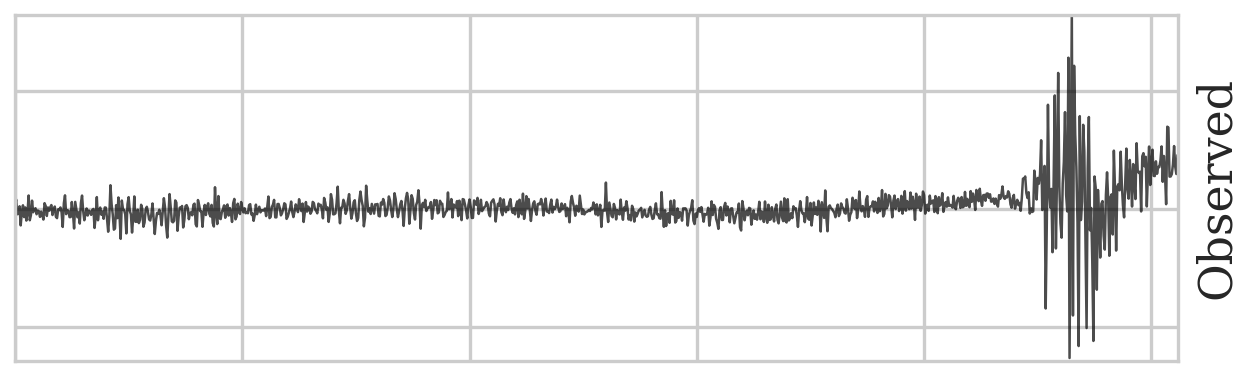}
      \caption{Wind burst with ringing (raw)}
      \label{fig:source_separation_wind_real_1}
  \end{subfigure}\hspace{0em}
  \begin{subfigure}[t]{0.325\textwidth}
      \includegraphics[width=\textwidth]{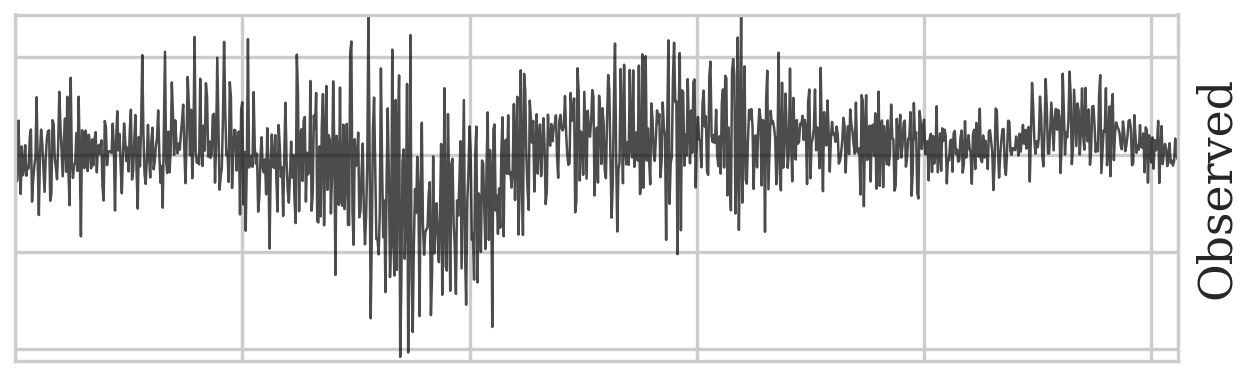}
      \caption{Sharp wind onset (raw)}
      \label{fig:source_separation_wind_real_18}
  \end{subfigure}\hspace{0em}
  \begin{subfigure}[t]{0.325\textwidth}
      \includegraphics[width=\textwidth]{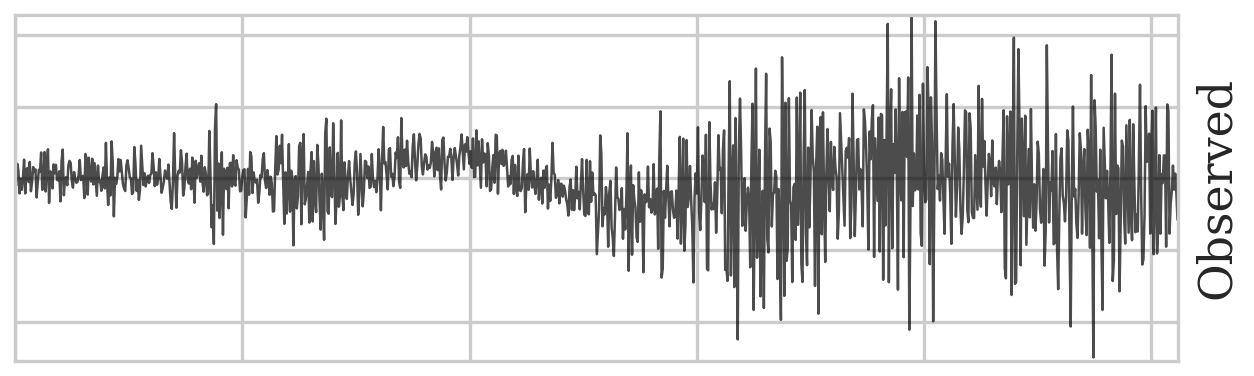}
      \caption{Complex wind pattern (raw)}
      \label{fig:source_separation_wind_real_63}
  \end{subfigure}

  \begin{subfigure}[t]{0.325\textwidth}
      \includegraphics[width=\textwidth]{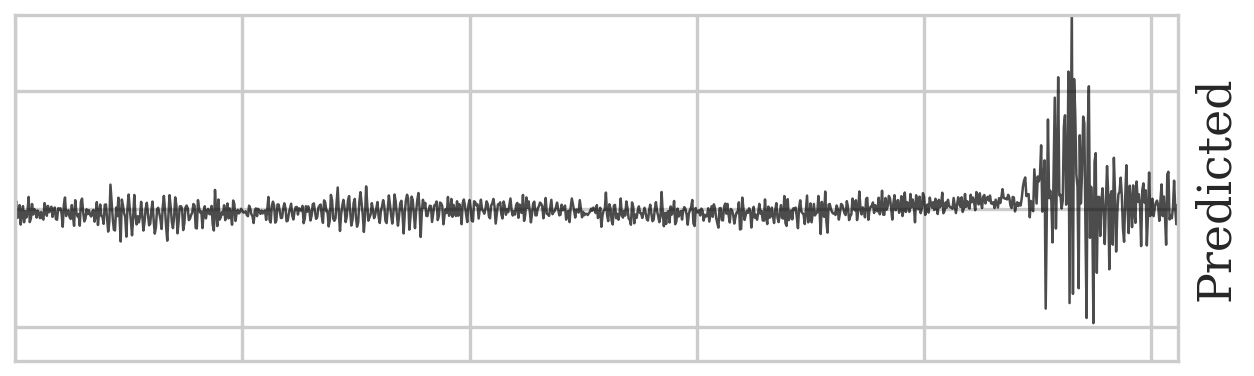}
      \caption{Independent component 1}
      \label{fig:source_separation_wind_ica_1_1}
  \end{subfigure}\hspace{0em}
  \begin{subfigure}[t]{0.325\textwidth}
      \includegraphics[width=\textwidth]{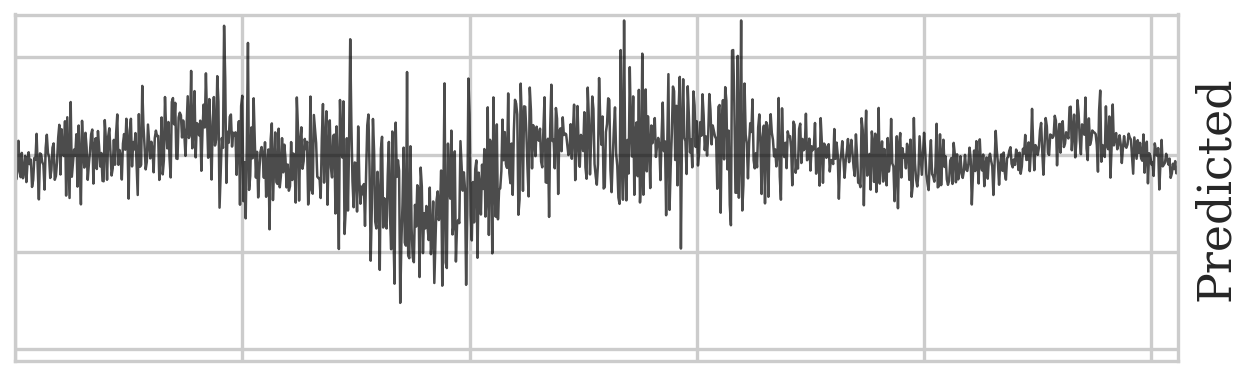}
      \caption{Independent component 1}
      \label{fig:source_separation_wind_ica_18_1}
  \end{subfigure}\hspace{0em}
  \begin{subfigure}[t]{0.325\textwidth}
      \includegraphics[width=\textwidth]{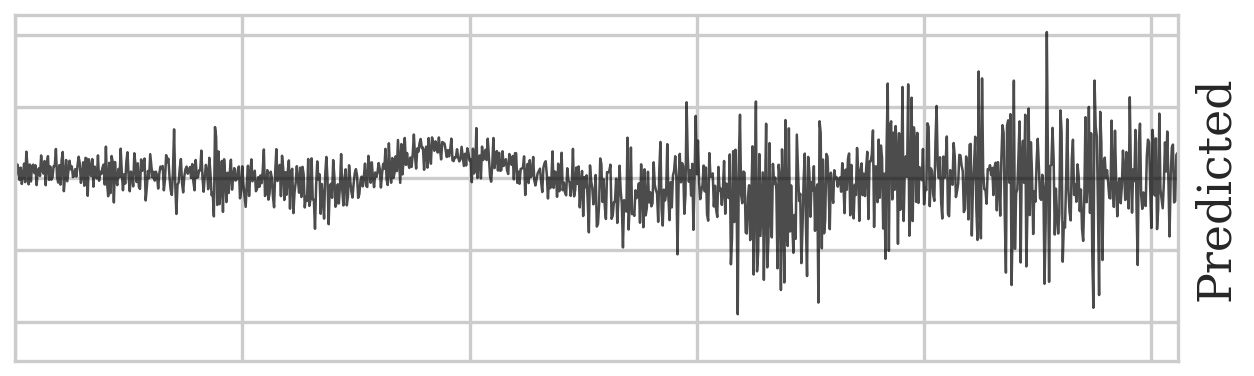}
      \caption{Independent component 1}
      \label{fig:source_separation_wind_ica_63_1}
  \end{subfigure}

  \begin{subfigure}[t]{0.325\textwidth}
      \includegraphics[width=\textwidth]{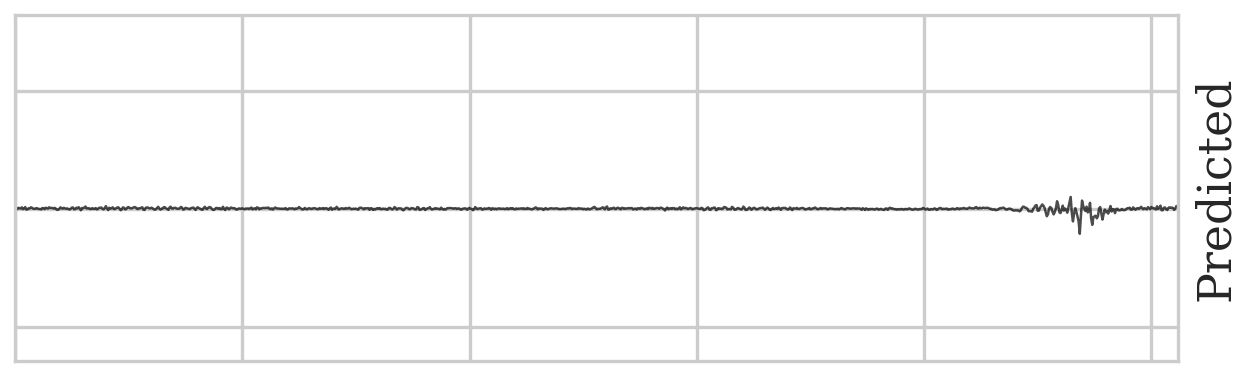}
      \caption{Independent component 2}
      \label{fig:source_separation_wind_ica_1_2}
  \end{subfigure}\hspace{0em}
  \begin{subfigure}[t]{0.325\textwidth}
      \includegraphics[width=\textwidth]{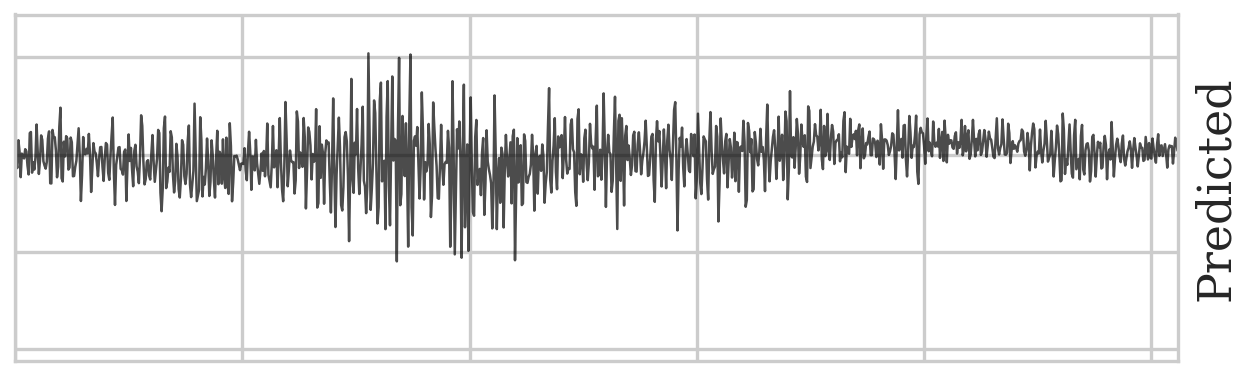}
      \caption{Independent component 2}
      \label{fig:source_separation_wind_ica_18_2}
  \end{subfigure}\hspace{0em}
  \begin{subfigure}[t]{0.325\textwidth}
      \includegraphics[width=\textwidth]{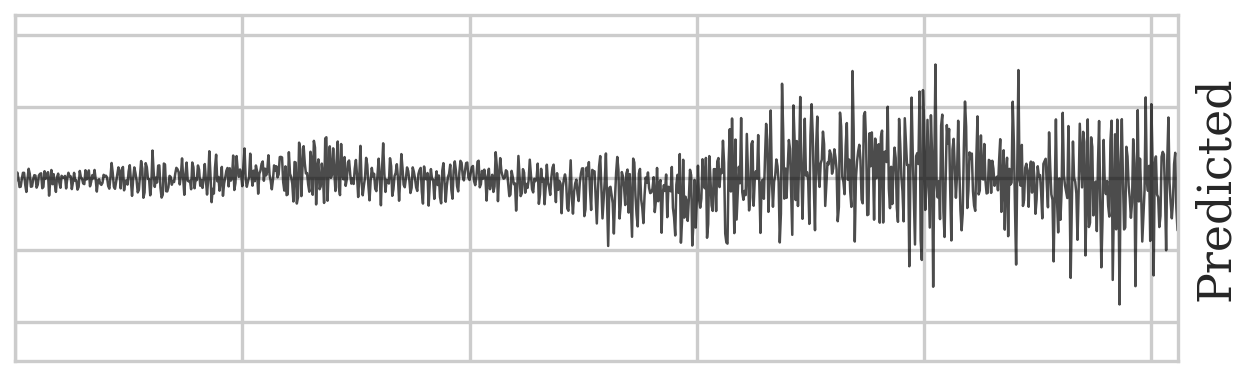}
      \caption{Independent component 2}
      \label{fig:source_separation_wind_ica_63_2}
  \end{subfigure}

  \begin{subfigure}[t]{0.325\textwidth}
      \includegraphics[width=\textwidth]{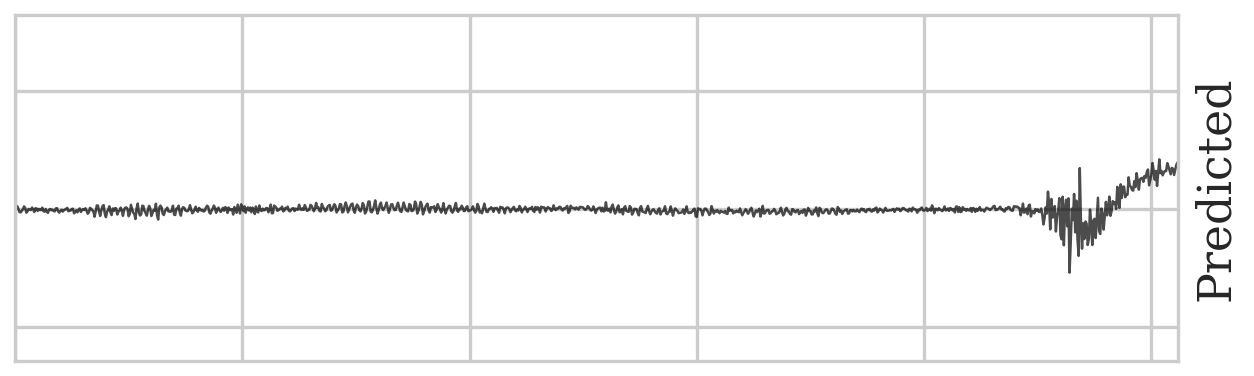}
      \caption{Independent component 3}
      \label{fig:source_separation_wind_ica_1_3}
  \end{subfigure}\hspace{0em}
  \begin{subfigure}[t]{0.325\textwidth}
      \includegraphics[width=\textwidth]{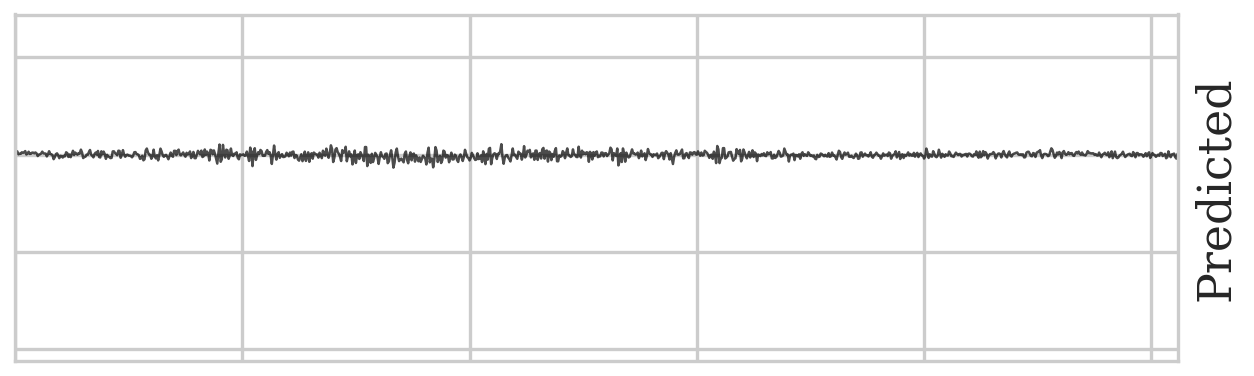}
      \caption{Independent component 3}
      \label{fig:source_separation_wind_ica_18_3}
  \end{subfigure}\hspace{0em}
  \begin{subfigure}[t]{0.325\textwidth}
      \includegraphics[width=\textwidth]{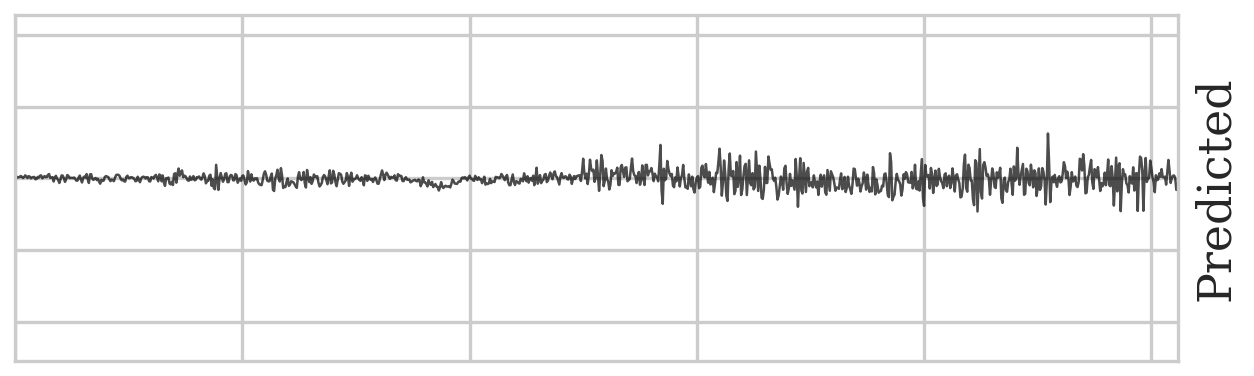}
      \caption{Independent component 3}
      \label{fig:source_separation_wind_ica_63_3}
  \end{subfigure}

  \caption{ICA-based source separation applied to three-component seismic data (U, V, W channels) for wind burst separation comparison. FastICA decomposes each mixed signal into three independent components, shown in rows 2-4. Each column represents a different wind pattern: wind burst with ringing, sharp wind onset, and complex wind pattern. The three independent components sum to reconstruct the original U-component signal (top row).}
  \label{fig:source_separation_wind_ica}
\end{figure*}

We demonstrate ICA's limitations through comprehensive comparison
figures. Figure \cref{fig:source_separation_glitch_ica} shows ICA
applied to glitch separation across three representative time windows.
While ICA successfully decomposes the seismic data into independent
components, none of the resulting components cleanly isolate the glitch
signals. Instead, the glitches are distributed across multiple
components mixed with background noise, making it impossible to achieve
clean separation. Similarly, Figure \cref{fig:source_separation_wind_ica}
demonstrates ICA's failure for wind burst separation, where the wind
signatures are fragmented across components rather than being cleanly
extracted. In contrast to our method's targeted separation (shown in the
main results), ICA's blind decomposition produces components that lack
physical interpretability and fail to align with the underlying seismic
source mechanisms. These results quantitatively demonstrate the
superiority of our domain-aware, multi-scale approach over traditional
blind source separation methods.

\section{Unsupervised separation of Marsquakes}
\label{app:marsquake_separation}

\begin{figure*}[!t]
  \centering
  \begin{subfigure}[t]{0.325\textwidth}
      \includegraphics[width=\textwidth]{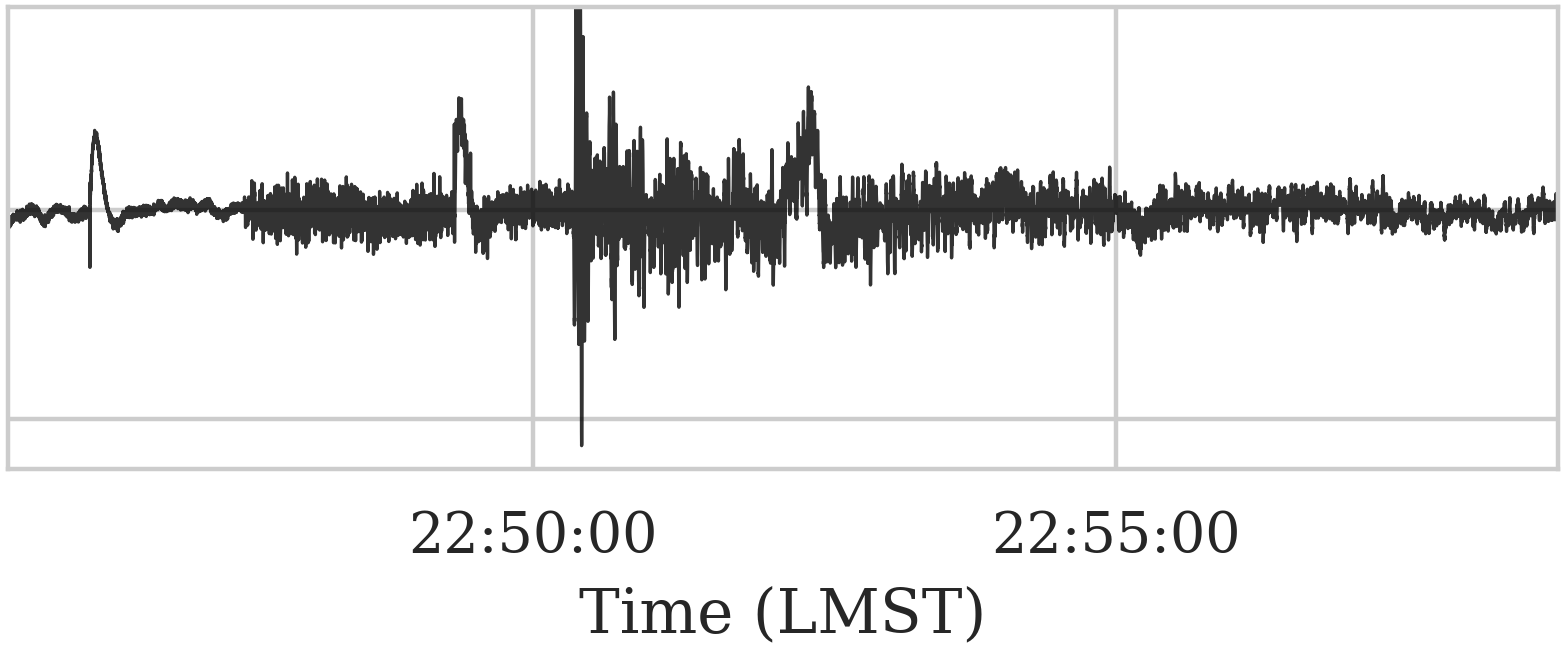}
      % \caption{}
      \caption{Raw waveform}
      \label{fig:source_separation_marsquake_real}
  \end{subfigure}\hspace{0em}
  \begin{subfigure}[t]{0.325\textwidth}
      \includegraphics[width=\textwidth]{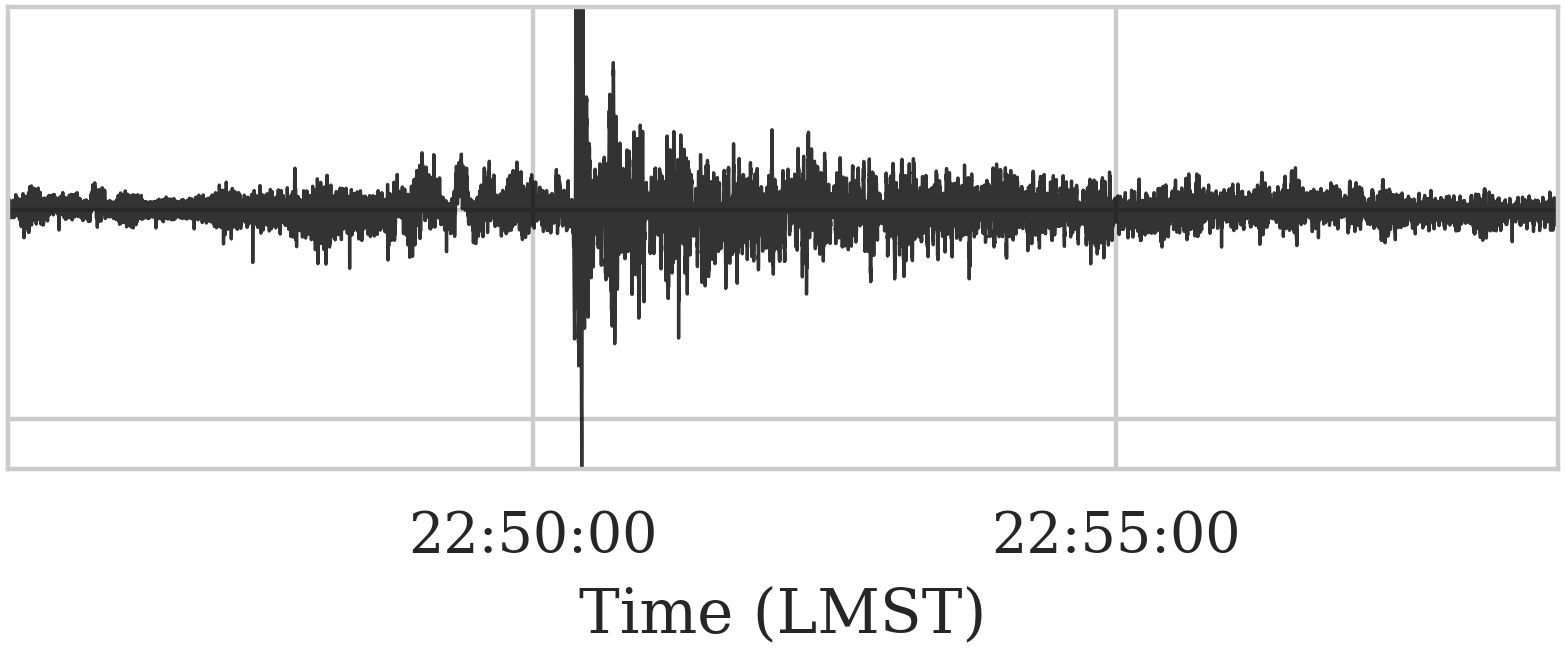}
      % \caption{}
      \caption{Separated marsquake}
      \label{fig:source_separation_marsquake_background}
  \end{subfigure}\hspace{0em}
  \begin{subfigure}[t]{0.325\textwidth}
      \includegraphics[width=\textwidth]{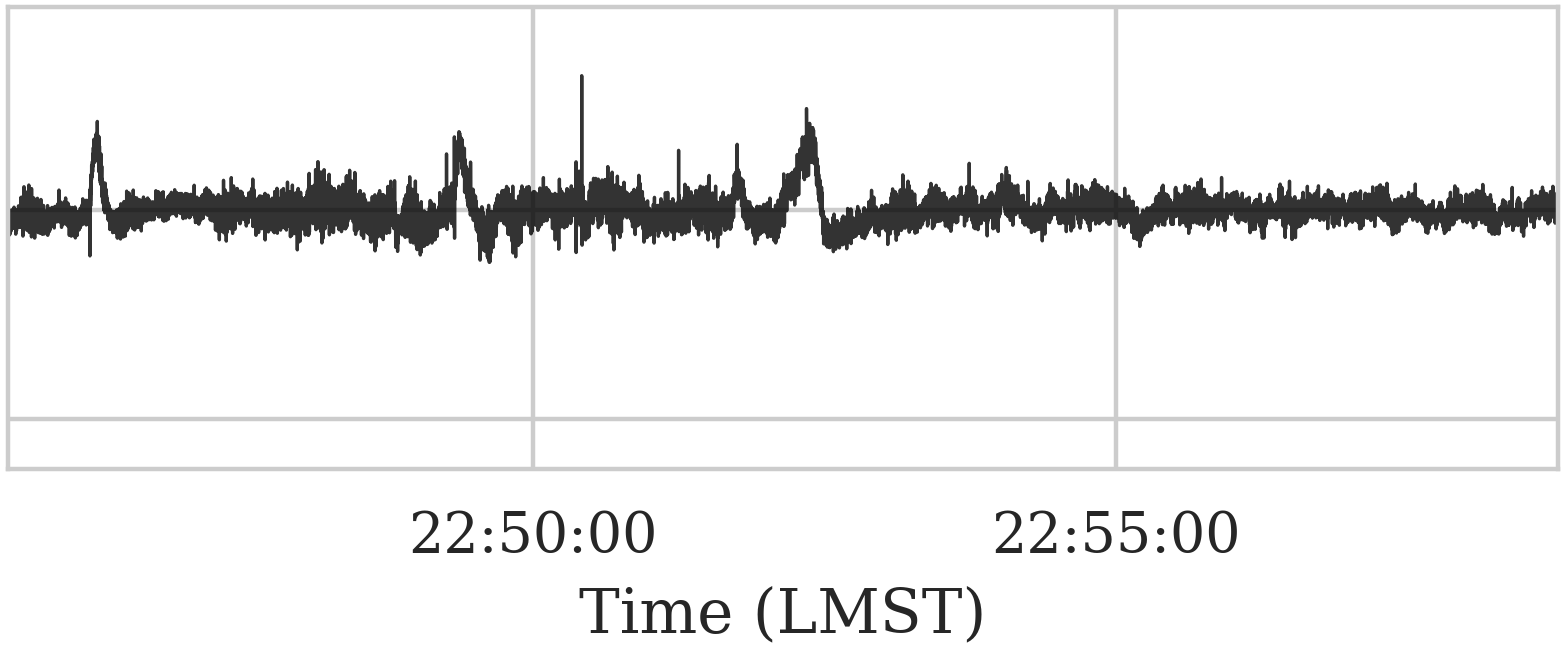}
      % \caption{}
      \caption{Separated signal}
      \label{fig:source_separation_marsquake_source}
  \end{subfigure}

  % \vspace{0.5em}
  \caption{Unsupervised separation of background noise, including transient atmospheric signals (glitches), from a marsquake (\cref{fig:source_separation_marsquake_real}) recorded by the InSight lander's seismometer on February 3, 2022 \cite{Service_2023}. Approximately eight hours of raw data from the U component were utilized for background noise separation without any explicit prior knowledge of marsquakes or glitches. The separated marsquake is shown \cref{fig:source_separation_marsquake_background}, and the separated signal in shown in \cref{fig:source_separation_marsquake_source}. The horizontal axis of the histograms represents the local mean solar time (LMST).}
  \label{fig:source_separation_marsquake}
\end{figure*}

\noindent In this section, we showcase the ability of our method to
separate signals that, due to their scarcity, do not have a dedicated
cluster within the fVAE latent space, e.g., marsquakes (cf.
\cref{fig:scale-4_broadband}). Marsquakes are of significant importance
as they provide useful information regarding the Mars subsurface,
enabling the study of Mars' interior \cite{Knapmeyer_endrunEtAl_2021,
StahlerEtAl_2021}. Here we apply our proposed unsupervised source
separation approach to isolate background noise and transient
atmospheric signals from a marsquake recorded on February 3, 2022
\cite{Service_2023}, which would allow for more accurate downstream
study of marsquakes. To achieve this, we select $80$ time windows from
each cluster in the finest timescale except clusters one and eight,
which are concentrated during the day (cf.
\cref{fig:app_cluster-1_scale-1,fig:app_cluster-8_scale-1}). The reason
for this particular choice is threefold: (i) the marsquake (cf.
\cref{fig:source_separation_marsquake_real}) is recorded during night
Martian time, so having data from the night allows for a better capture
of the background signal statistics; (ii) as marsquakes are scarce and
do not have a dedicated cluster, the chance of obtaining a time window
that is a likely sample from a cluster being a marsquake is slim,
reducing the chances of accidentally separating marsquake signals as
background noise; and (iii) we found performing source separation at the
finest scale allows for better separation of glitches, as they are also
short and transient signals.

Next, we use these time windows as representative samples from the
background noise ($\B{s}_i$ in the context of \cref{eq:ss_optimization})
with the goal of separating the clean marsquake signal. We solve the
optimization problem in \cref{eq:ss_optimization} with 1000 L-BFGS
iterations and a relative termination tolerance of 0.1. The optimization
took 40 minutes to complete. The results are depicted in
\cref{fig:source_separation_marsquake}. There are clearly three glitches
that we have successfully separated, along with the background noise.
Through this example, we demonstrated the effectiveness of our
multi-scale source separation method---based on prior information
obtained from the fVAE---when the source (the marsquake in this case) is
not a prominent source in the entire dataset, i.e., it is an outlier.

\section{Stylized example}
\label{app:stylized}

\noindent We present a multi-scale synthetic dataset used to validate
our model using ground truth sources. We first describe how the dataset
is constructed and then we present the results of our clustering and
source separation algorithm when applied to this dataset.

\subsection{Synthetic multi-scale dataset}\label{app:synthetic_dataset}

\begin{figure*}
\centering
\includegraphics[width=0.8\linewidth]{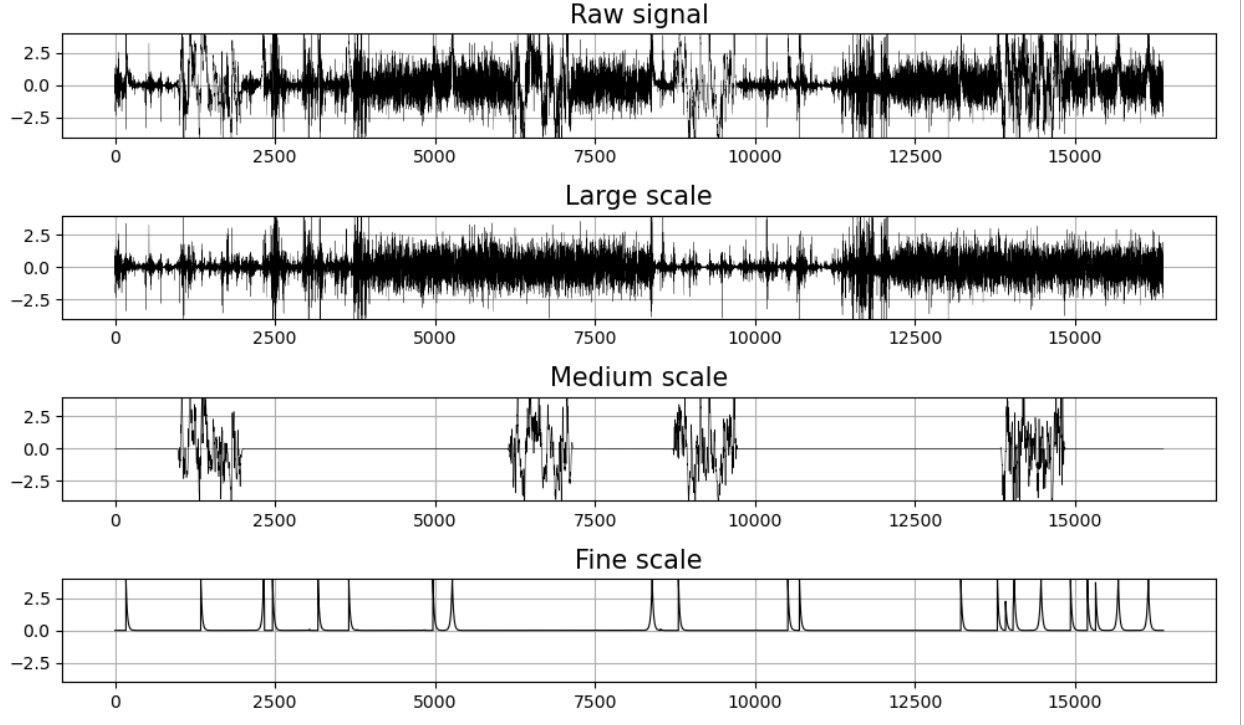}
\caption{Synthetic multi-scale dataset.}
\label{fig:synthetic-dataset}
\end{figure*}

\noindent This dataset is composed of events with three distinct timescales $w$ that are outlined below:

\begin{itemize}
\item \textbf{Large scale ($w=2^{12}$).} In this timescale, the signal alternates between two different noises to mimic the change in background noise due to day-night periodicity.
The first noise is a white noise $\B{x}_1(t)$ and the second one is a multifractal random noise $\B{x}_2(t)$ which is a non-Gaussian noise with intermittency~\cite{BacryEtAl_2001}.
The large-scale signal is then
$\B{x}_{\text{large}}(t) = (1-\theta(t))\,\B{x}_1(t) + \theta(t)\,\B{x}_2(t)$ where $\theta(t)$ is equal to $0$ on $[0,w]$ and $1$ on $[w+\eta, 2w-\eta]$ for small $\eta$, is $2w$-periodic and has a smooth junction at $t=w$ and at $t=2w$.
\item \textbf{Medium scale ($w=2^{10}$).} This timescale exhibits a single type of events which is a turbulent jet recorded in experimental conditions~\cite{ChanalEtAl_2000}.
Each event has a duration $w=2^{10}$ and is placed randomly without overlapping, with as many events as the number of ``days'' defined at large scale.
\item \textbf{Fine scales ($w=2^8$).} The signal in this timescale is a mixture of two types of transient events with exponential decay. First one $\B{x}_3(t) = e^{-|t|}$ is symmetrical in $t=0$, second one $\B{x}_4(t) = e^{-t}\mathbf{1}_{t\geq0}$ is asymmetrical.
Each of these signals is randomly positioned in the time-series with four times more occurrence than the number of ``days'' in our dataset.
\end{itemize}
The resulting dataset is a long time-series $\B{x}=\B{x}_\text{large} +
\B{x}_\text{medium} + \B{x}_\text{fine}$. \Cref{fig:synthetic-dataset}
shows one realization of this dataset along with the sources in three
scales that compose it.

\subsection{Experimental details}
\label{app:training_details_appendix}

While we are aware of the underlying sources that create this synthetic
dataset, we choose to make an uninformed decision regarding the number
of clusters per scale to mimic a more realistic setting. To this end, we
use the same architecture and hyperparameters as the fVAE trained in the
context of clustering and source separation for the Mars InSight seismic
data. Specifically, the model architecture, training procedure, and
optimization settings are directly borrowed from the Mars experiment to
ensure comparability between real and synthetic datasets (see
\cref{sec:architecture,sec:training_details} in the main text for
details).

The only distinction is that, for the synthetic data, the timescales
used for scattering covariance generation were selected to match the
durations of the three source types in our multi-scale dataset.
Specifically, we use the same window sizes described in the preceding
\cref{app:synthetic_dataset} for the pyramidal scattering spectra
computation. This ensures that the scattering features are well-adapted
to the characteristic timescales of the synthetic sources.

\subsection{Synthetic data clustering}

\begin{figure*}
    \renewcommand\thesubfigure{\roman{subfigure}}
    \captionsetup[subfigure]{skip=-10pt}
    \renewcommand{\thesubfigure}{\alph{subfigure}.\arabic{row}}
    \centering
    \setcounter{row}{1}%
    \begin{subfigure}[b]{0.245\textwidth}
        \includegraphics[width=\textwidth]{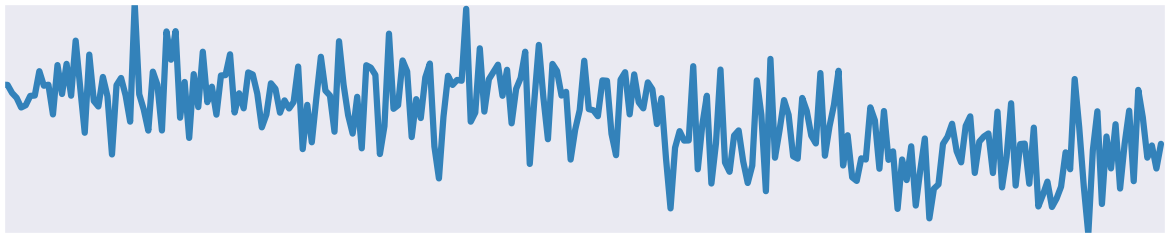}
    \vspace{0ex}\caption{}
    \label{fig:synthetic_scale_256_0_w0}
    \end{subfigure}\hspace{0em}
    \begin{subfigure}[b]{0.245\textwidth}
        \includegraphics[width=\textwidth]{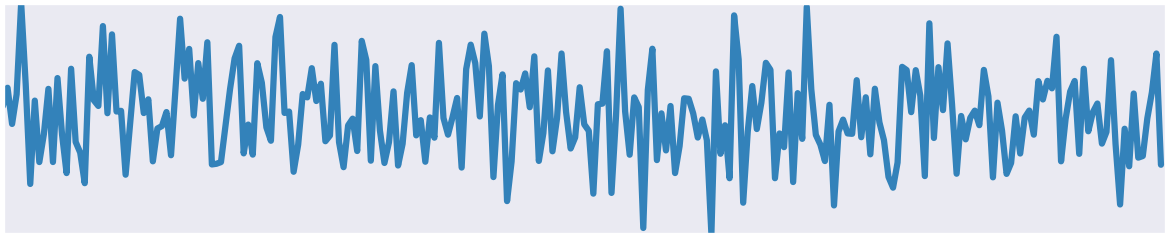}
    \vspace{0ex}\caption{}
    \label{fig:synthetic_scale_256_0_w1}
    \end{subfigure}\hspace{0em}
    \begin{subfigure}[b]{0.245\textwidth}
        \includegraphics[width=\textwidth]{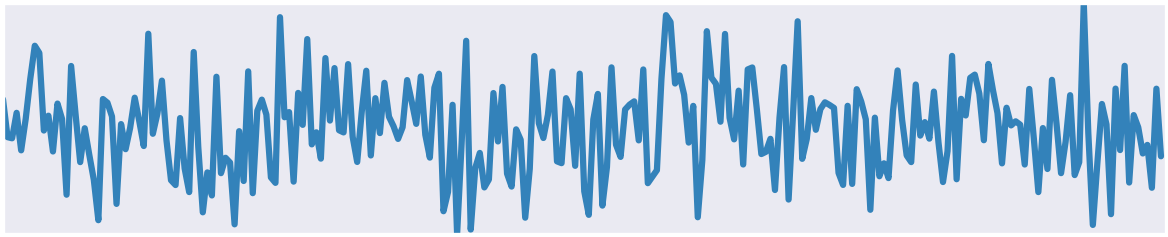}
    \vspace{0ex}\caption{}
    \label{fig:synthetic_scale_256_0_w2}
    \end{subfigure}\hspace{0em}
    \begin{subfigure}[b]{0.245\textwidth}
        \includegraphics[width=\textwidth]{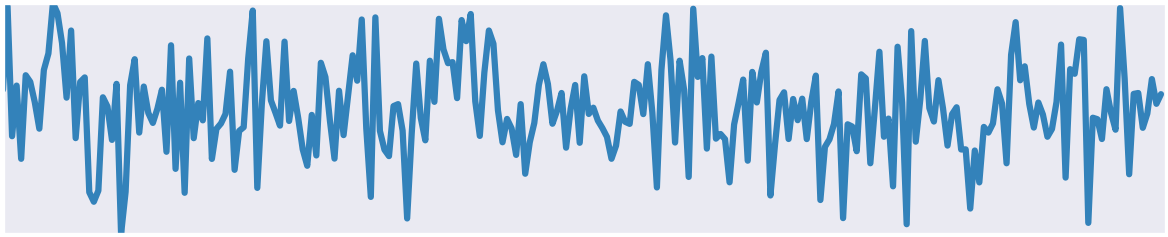}
    \vspace{0ex}\caption{}
    \label{fig:synthetic_scale_256_0_w3}
    \end{subfigure}\hspace{0em}

    \stepcounter{row}%
    \begin{subfigure}[b]{0.245\textwidth}
        \includegraphics[width=\textwidth]{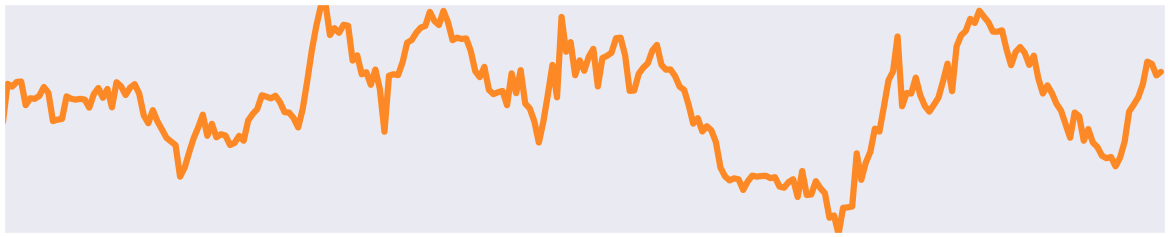}
    \vspace{0ex}\caption{}
    \label{fig:synthetic_scale_256_1_w0}
    \end{subfigure}\hspace{0em}
    \begin{subfigure}[b]{0.245\textwidth}
        \includegraphics[width=\textwidth]{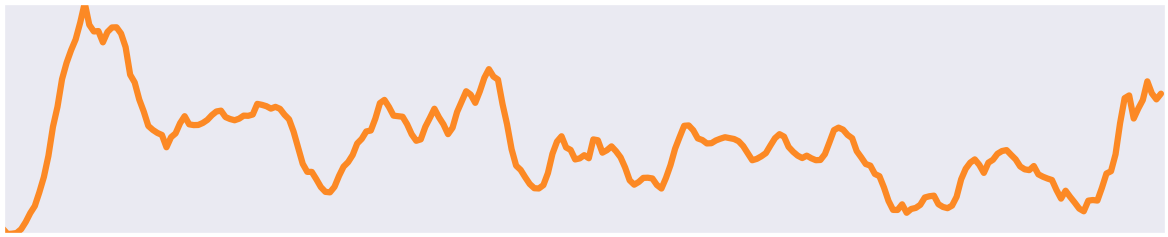}
    \vspace{0ex}\caption{}
    \label{fig:synthetic_scale_256_1_w1}
    \end{subfigure}\hspace{0em}
    \begin{subfigure}[b]{0.245\textwidth}
        \includegraphics[width=\textwidth]{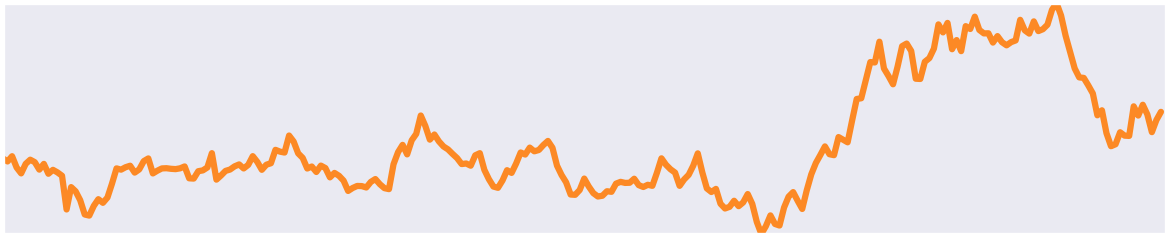}
    \vspace{0ex}\caption{}
    \label{fig:synthetic_scale_256_1_w2}
    \end{subfigure}\hspace{0em}
    \begin{subfigure}[b]{0.245\textwidth}
        \includegraphics[width=\textwidth]{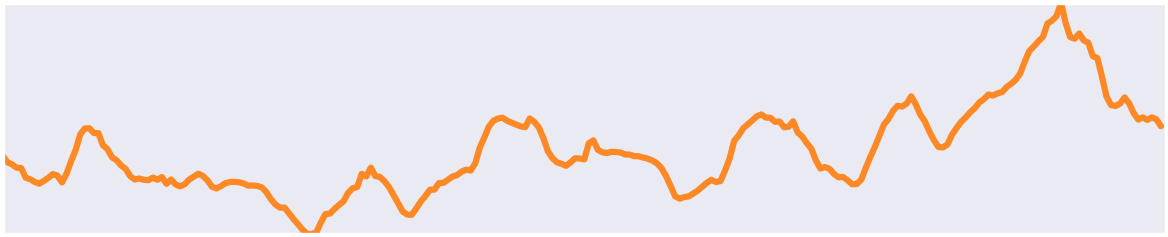}
    \vspace{0ex}\caption{}
    \label{fig:synthetic_scale_256_1_w3}
    \end{subfigure}\hspace{0em}

    \stepcounter{row}%
    \begin{subfigure}[b]{0.245\textwidth}
        \includegraphics[width=\textwidth]{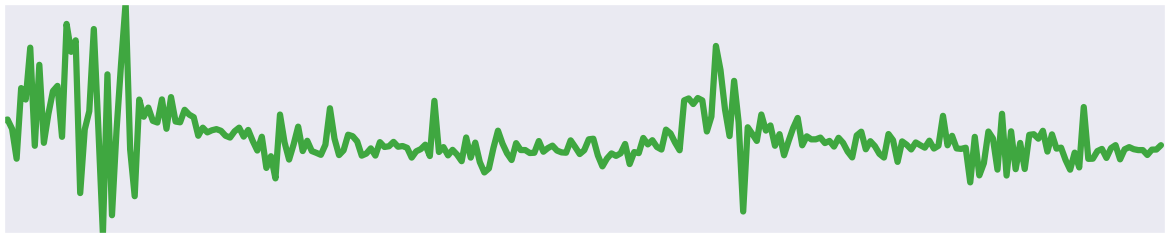}
    \vspace{0ex}\caption{}
    \label{fig:synthetic_scale_256_2_w0}
    \end{subfigure}\hspace{0em}
    \begin{subfigure}[b]{0.245\textwidth}
        \includegraphics[width=\textwidth]{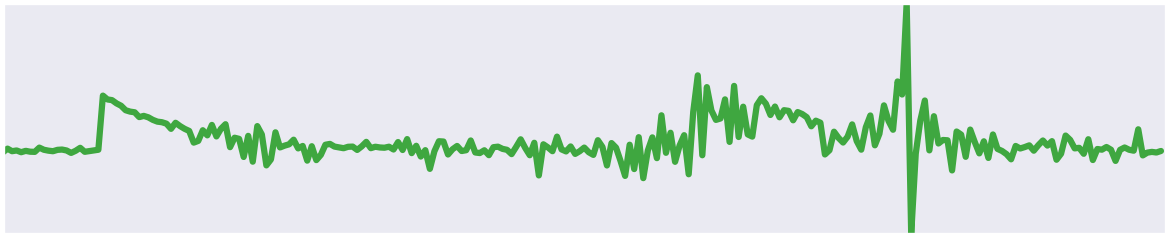}
    \vspace{0ex}\caption{}
    \label{fig:synthetic_scale_256_2_w1}
    \end{subfigure}\hspace{0em}
    \begin{subfigure}[b]{0.245\textwidth}
        \includegraphics[width=\textwidth]{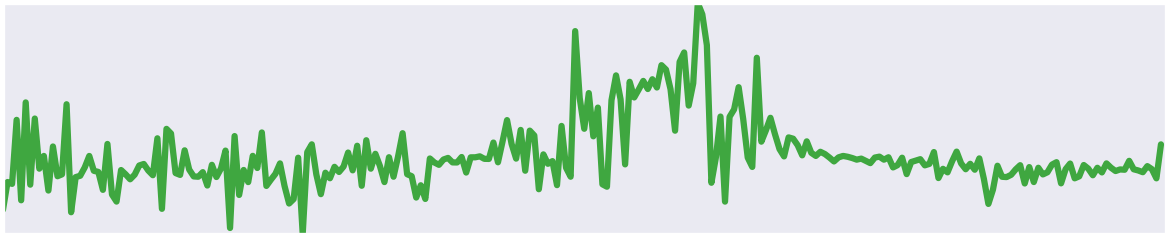}
    \vspace{0ex}\caption{}
    \label{fig:synthetic_scale_256_2_w2}
    \end{subfigure}\hspace{0em}
    \begin{subfigure}[b]{0.245\textwidth}
        \includegraphics[width=\textwidth]{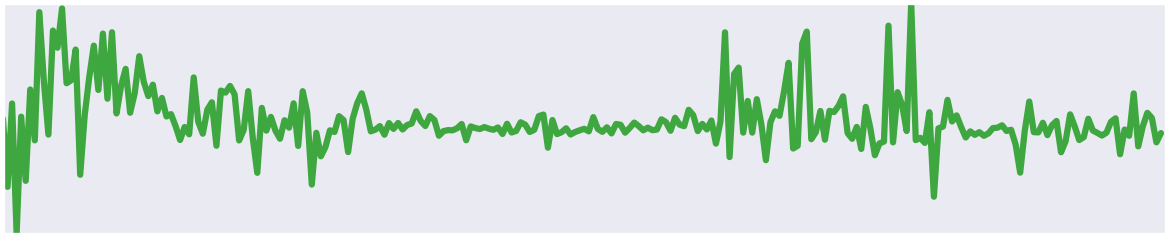}
    \vspace{0ex}\caption{}
    \label{fig:synthetic_scale_256_2_w3}
    \end{subfigure}\hspace{0em}

    \stepcounter{row}%
    \begin{subfigure}[b]{0.245\textwidth}
        \includegraphics[width=\textwidth]{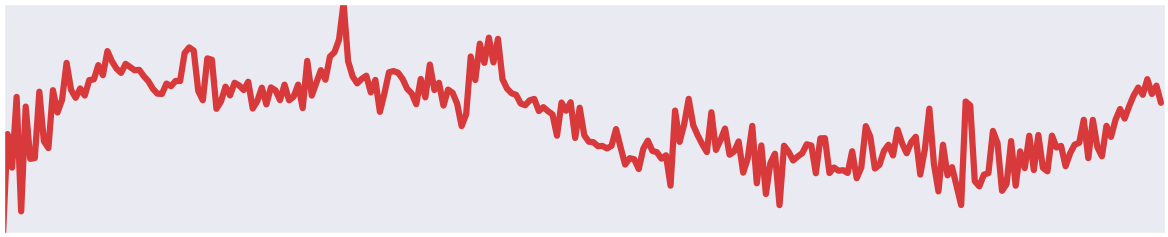}
    \vspace{0ex}\caption{}
    \label{fig:synthetic_scale_256_3_w0}
    \end{subfigure}\hspace{0em}
    \begin{subfigure}[b]{0.245\textwidth}
        \includegraphics[width=\textwidth]{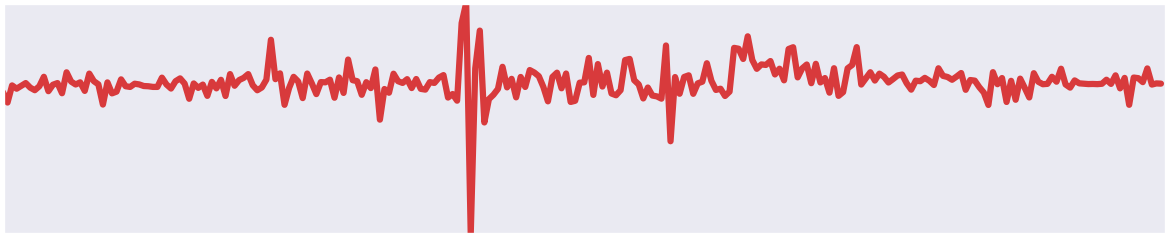}
    \vspace{0ex}\caption{}
    \label{fig:synthetic_scale_256_3_w1}
    \end{subfigure}\hspace{0em}
    \begin{subfigure}[b]{0.245\textwidth}
        \includegraphics[width=\textwidth]{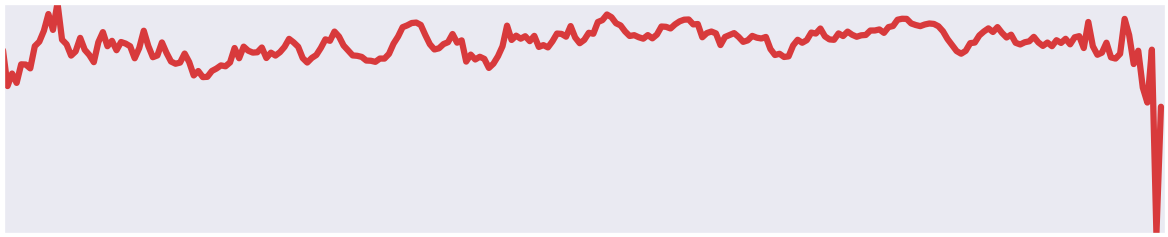}
    \vspace{0ex}\caption{}
    \label{fig:synthetic_scale_256_3_w2}
    \end{subfigure}\hspace{0em}
    \begin{subfigure}[b]{0.245\textwidth}
        \includegraphics[width=\textwidth]{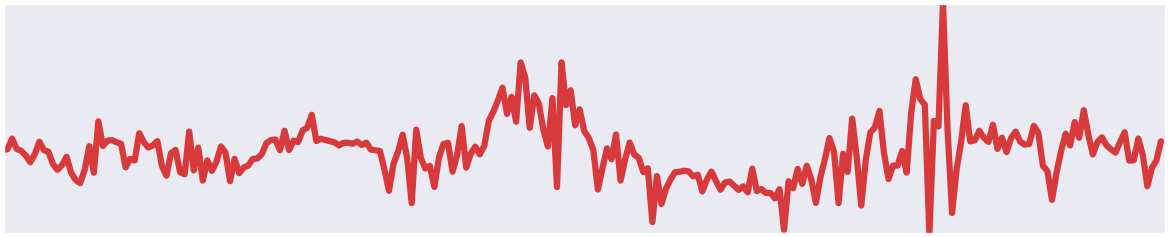}
    \vspace{0ex}\caption{}
    \label{fig:synthetic_scale_256_3_w3}
    \end{subfigure}\hspace{0em}

    \stepcounter{row}%
    \begin{subfigure}[b]{0.245\textwidth}
        \includegraphics[width=\textwidth]{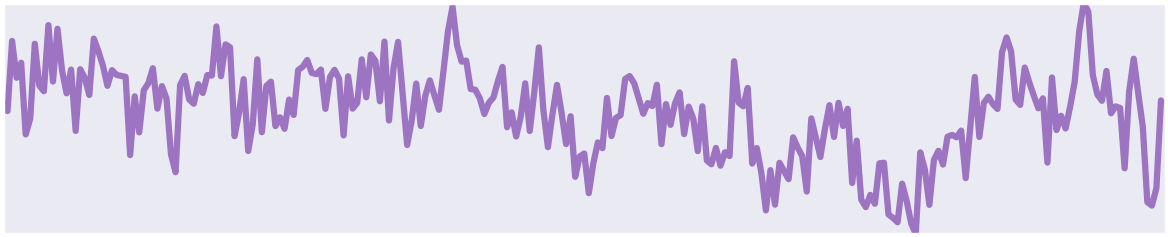}
    \vspace{0ex}\caption{}
    \label{fig:synthetic_scale_256_4_w0}
    \end{subfigure}\hspace{0em}
    \begin{subfigure}[b]{0.245\textwidth}
        \includegraphics[width=\textwidth]{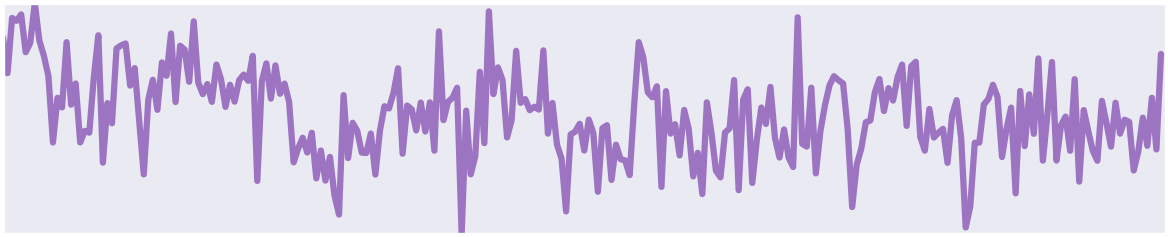}
    \vspace{0ex}\caption{}
    \label{fig:synthetic_scale_256_4_w1}
    \end{subfigure}\hspace{0em}
    \begin{subfigure}[b]{0.245\textwidth}
        \includegraphics[width=\textwidth]{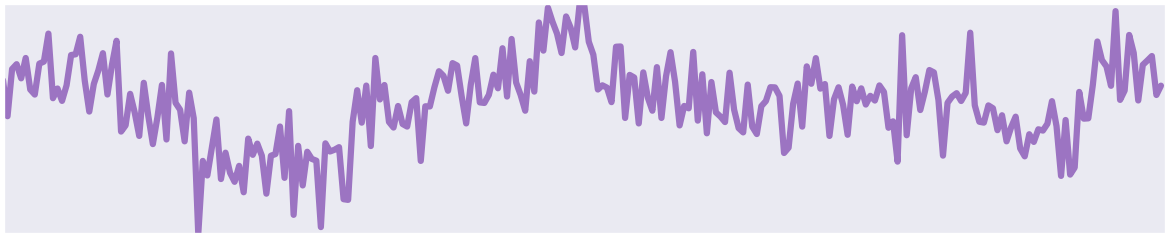}
    \vspace{0ex}\caption{}
    \label{fig:synthetic_scale_256_4_w2}
    \end{subfigure}\hspace{0em}
    \begin{subfigure}[b]{0.245\textwidth}
        \includegraphics[width=\textwidth]{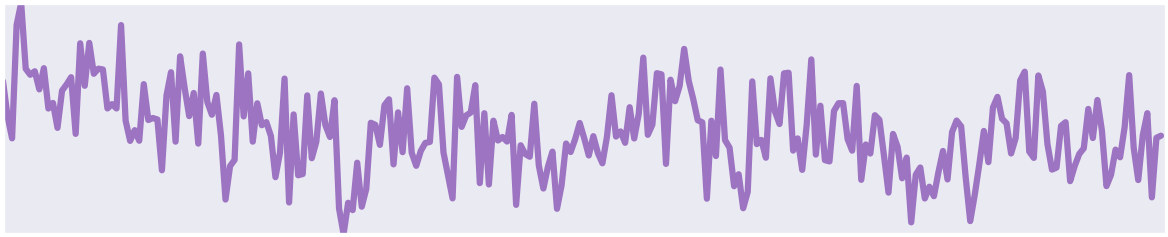}
    \vspace{0ex}\caption{}
    \label{fig:synthetic_scale_256_4_w3}
    \end{subfigure}\hspace{0em}

    \stepcounter{row}%
    \begin{subfigure}[b]{0.245\textwidth}
        \includegraphics[width=\textwidth]{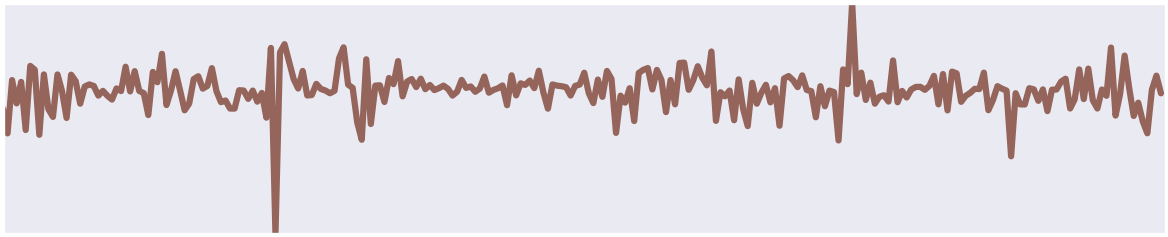}
    \vspace{0ex}\caption{}
    \label{fig:synthetic_scale_256_5_w0}
    \end{subfigure}\hspace{0em}
    \begin{subfigure}[b]{0.245\textwidth}
        \includegraphics[width=\textwidth]{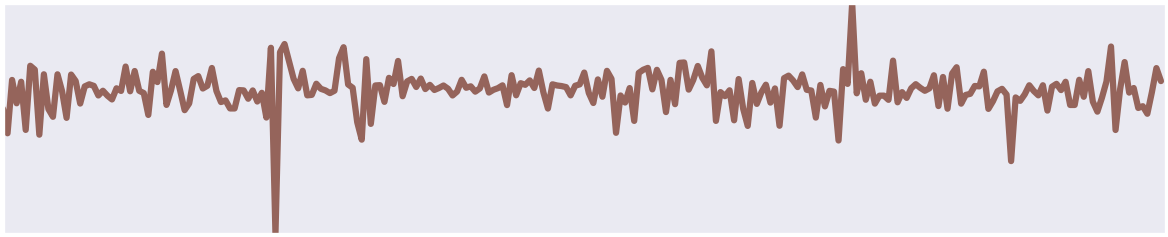}
    \vspace{0ex}\caption{}
    \label{fig:synthetic_scale_256_5_w1}
    \end{subfigure}\hspace{0em}
    \begin{subfigure}[b]{0.245\textwidth}
        \includegraphics[width=\textwidth]{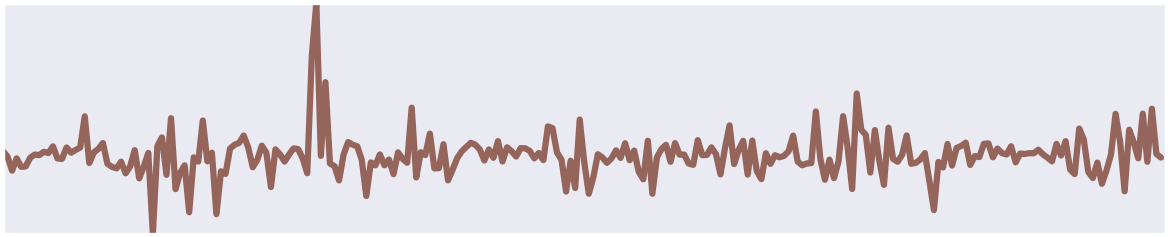}
    \vspace{0ex}\caption{}
    \label{fig:synthetic_scale_256_5_w2}
    \end{subfigure}\hspace{0em}
    \begin{subfigure}[b]{0.245\textwidth}
        \includegraphics[width=\textwidth]{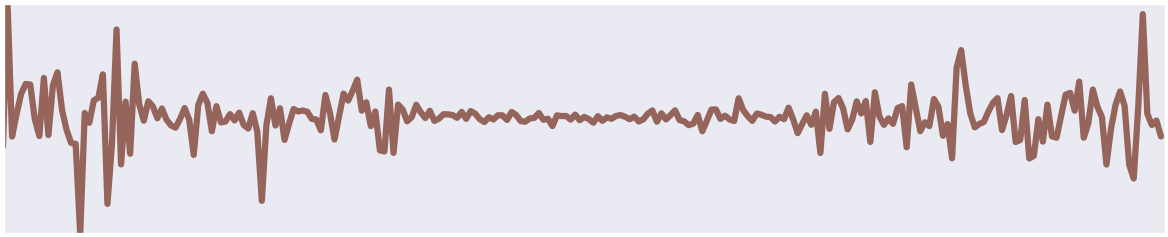}
    \vspace{0ex}\caption{}
    \label{fig:synthetic_scale_256_5_w3}
    \end{subfigure}\hspace{0em}

    \stepcounter{row}%
    \begin{subfigure}[b]{0.245\textwidth}
        \includegraphics[width=\textwidth]{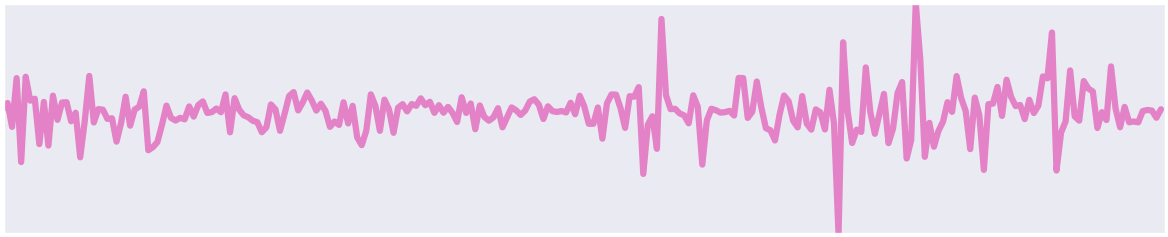}
    \vspace{0ex}\caption{}
    \label{fig:synthetic_scale_256_6_w0}
    \end{subfigure}\hspace{0em}
    \begin{subfigure}[b]{0.245\textwidth}
        \includegraphics[width=\textwidth]{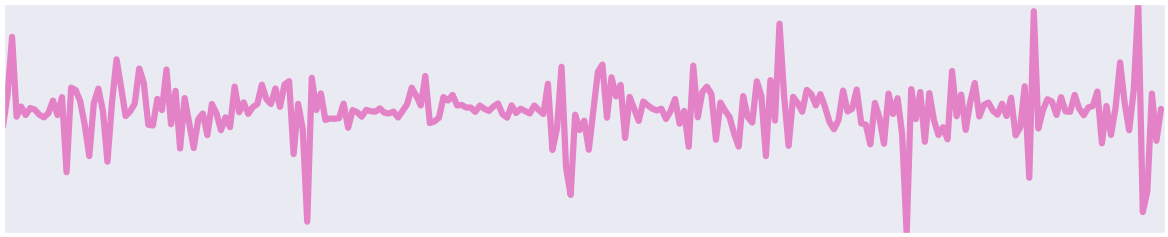}
    \vspace{0ex}\caption{}
    \label{fig:synthetic_scale_256_6_w1}
    \end{subfigure}\hspace{0em}
    \begin{subfigure}[b]{0.245\textwidth}
        \includegraphics[width=\textwidth]{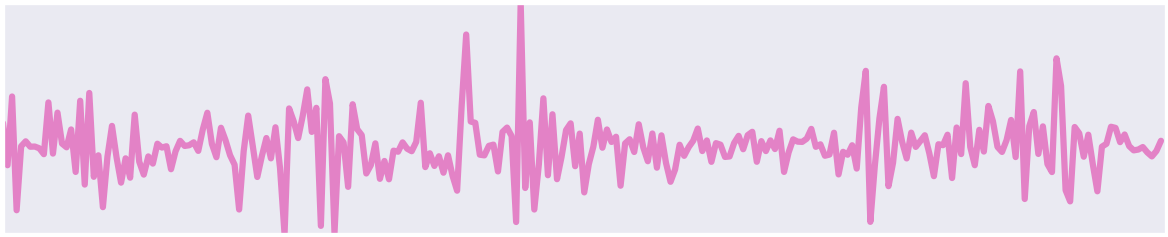}
    \vspace{0ex}\caption{}
    \label{fig:synthetic_scale_256_6_w2}
    \end{subfigure}\hspace{0em}
    \begin{subfigure}[b]{0.245\textwidth}
        \includegraphics[width=\textwidth]{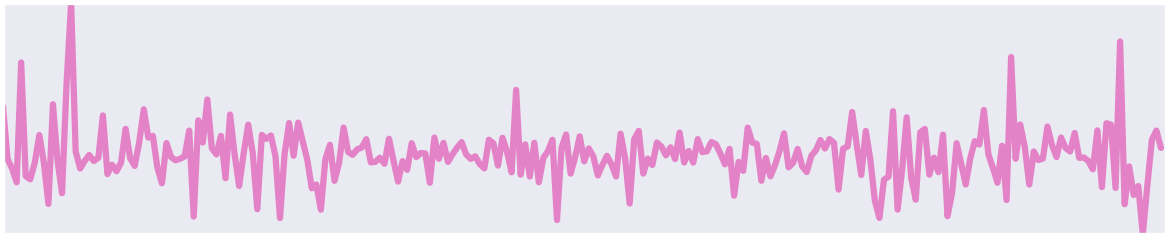}
    \vspace{0ex}\caption{}
    \label{fig:synthetic_scale_256_6_w3}
    \end{subfigure}\hspace{0em}

    \stepcounter{row}%
    \begin{subfigure}[b]{0.245\textwidth}
        \includegraphics[width=\textwidth]{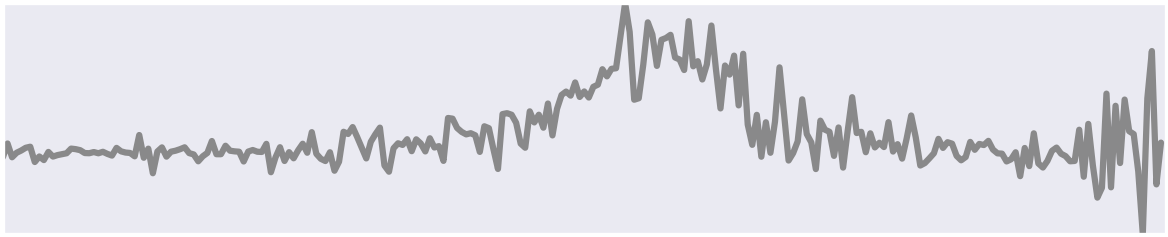}
    \vspace{0ex}\caption{}
    \label{fig:synthetic_scale_256_7_w0}
    \end{subfigure}\hspace{0em}
    \begin{subfigure}[b]{0.245\textwidth}
        \includegraphics[width=\textwidth]{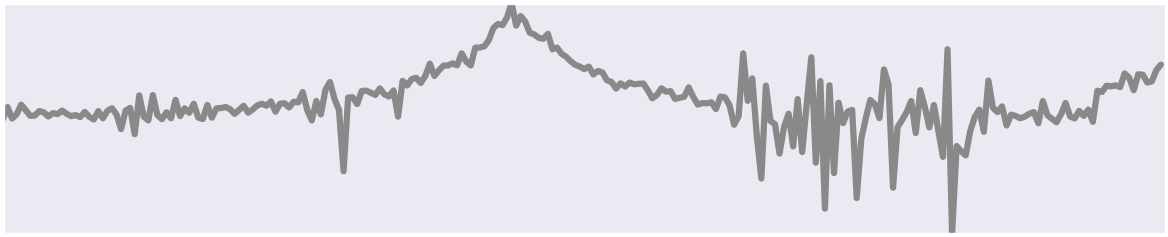}
    \vspace{0ex}\caption{}
    \label{fig:synthetic_scale_256_7_w1}
    \end{subfigure}\hspace{0em}
    \begin{subfigure}[b]{0.245\textwidth}
        \includegraphics[width=\textwidth]{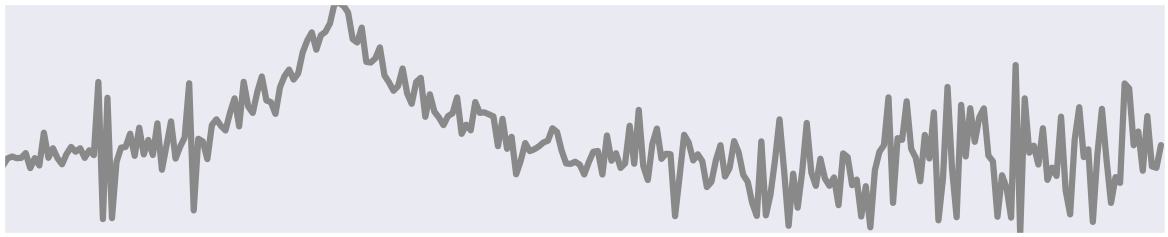}
    \vspace{0ex}\caption{}
    \label{fig:synthetic_scale_256_7_w2}
    \end{subfigure}\hspace{0em}
    \begin{subfigure}[b]{0.245\textwidth}
        \includegraphics[width=\textwidth]{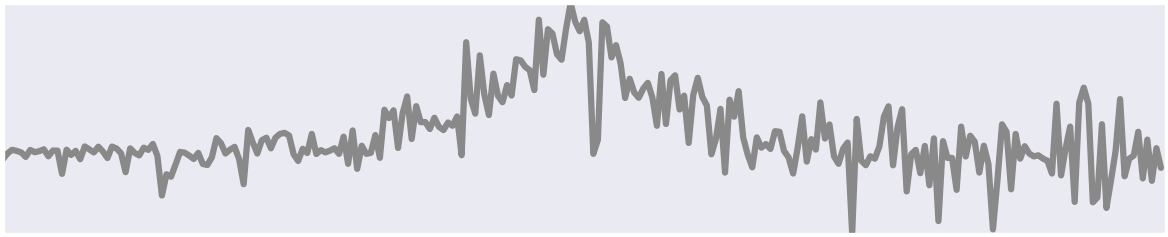}
    \vspace{0ex}\caption{}
    \label{fig:synthetic_scale_256_7_w3}
    \end{subfigure}\hspace{0em}

    \stepcounter{row}%
    \begin{subfigure}[b]{0.245\textwidth}
        \includegraphics[width=\textwidth]{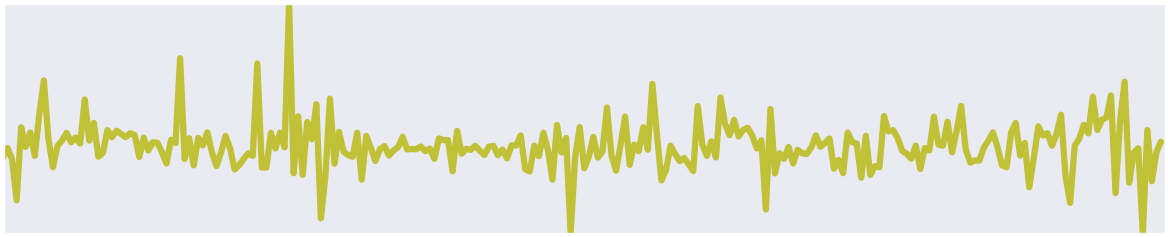}
    \vspace{0ex}\caption{}
    \label{fig:synthetic_scale_256_8_w0}
    \end{subfigure}\hspace{0em}
    \begin{subfigure}[b]{0.245\textwidth}
        \includegraphics[width=\textwidth]{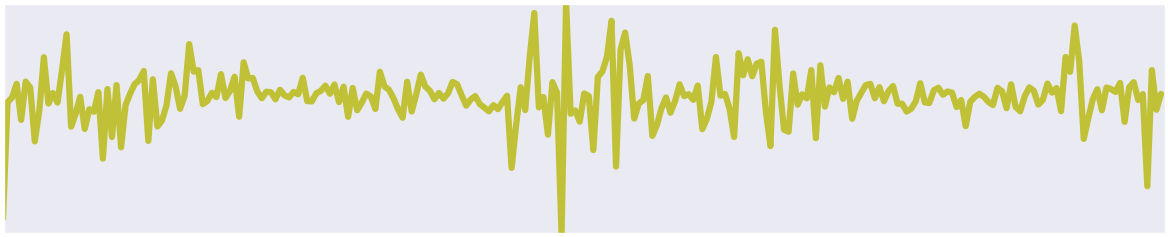}
    \vspace{0ex}\caption{}
    \label{fig:synthetic_scale_256_8_w1}
    \end{subfigure}\hspace{0em}
    \begin{subfigure}[b]{0.245\textwidth}
        \includegraphics[width=\textwidth]{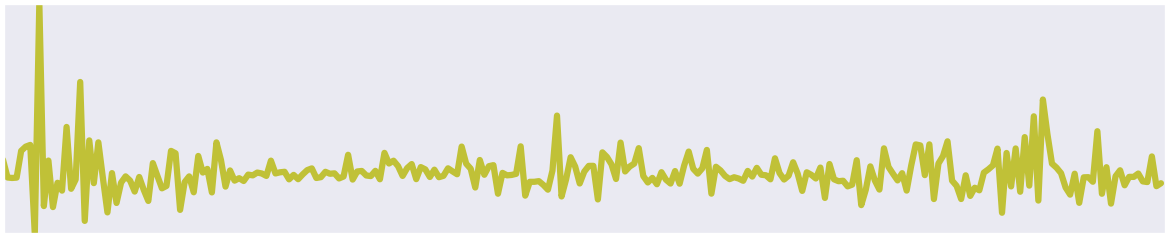}
    \vspace{0ex}\caption{}
    \label{fig:synthetic_scale_256_8_w2}
    \end{subfigure}\hspace{0em}
    \begin{subfigure}[b]{0.245\textwidth}
        \includegraphics[width=\textwidth]{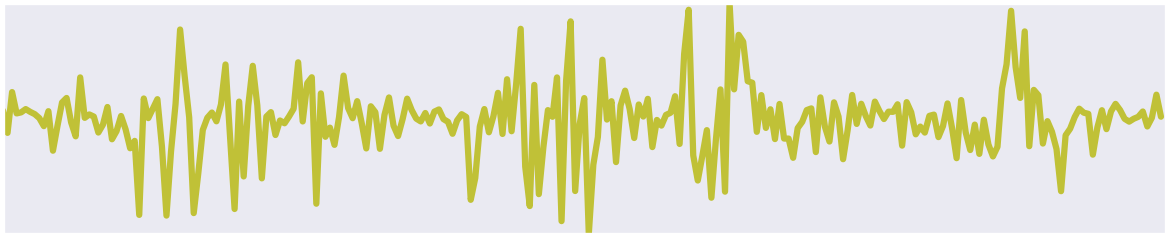}
    \vspace{0ex}\caption{}
    \label{fig:synthetic_scale_256_8_w3}
    \end{subfigure}\hspace{0em}

    \caption{Waveforms in timescale of $256$ sample, with each row from top to bottom illustrating four waveforms corresponding to clusters one to nine in this timescale, respectively.}
    \label{fig:synthetic_scale_256_waves}
\end{figure*}

\Cref{fig:synthetic_scale_256_waves} summarizes the identified clusters after training the fVAE. We can indeed identify different patterns at each timescale, some of which clearly indicate the multi-scale sources that were used to generate the data. A portion of the clusters are also comprised of at least two of this sources in combination.

\subsection{Synthetic data source separation}

\begin{figure*}
    \renewcommand\thesubfigure{\roman{subfigure}}
    \captionsetup[subfigure]{skip=-10pt}
    \renewcommand{\thesubfigure}{\alph{subfigure}.\arabic{row}}
    \centering
    \setcounter{row}{1}%
    \begin{subfigure}[b]{0.245\textwidth}
        \includegraphics[width=\textwidth]{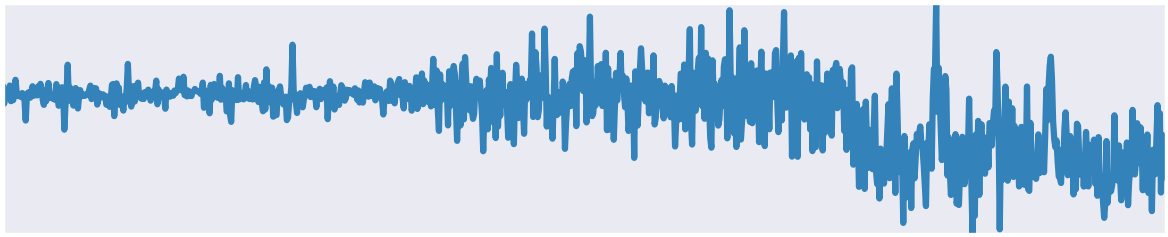}
    \vspace{0ex}\caption{}
    \label{fig:synthetic_scale_1024_0_w0}
    \end{subfigure}\hspace{0em}
    \begin{subfigure}[b]{0.245\textwidth}
        \includegraphics[width=\textwidth]{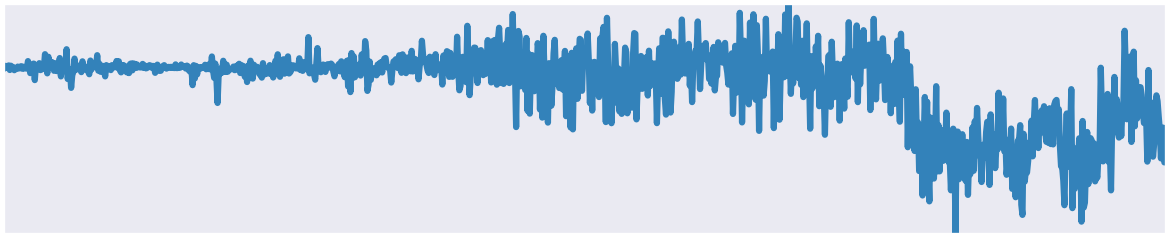}
    \vspace{0ex}\caption{}
    \label{fig:synthetic_scale_1024_0_w1}
    \end{subfigure}\hspace{0em}
    \begin{subfigure}[b]{0.245\textwidth}
        \includegraphics[width=\textwidth]{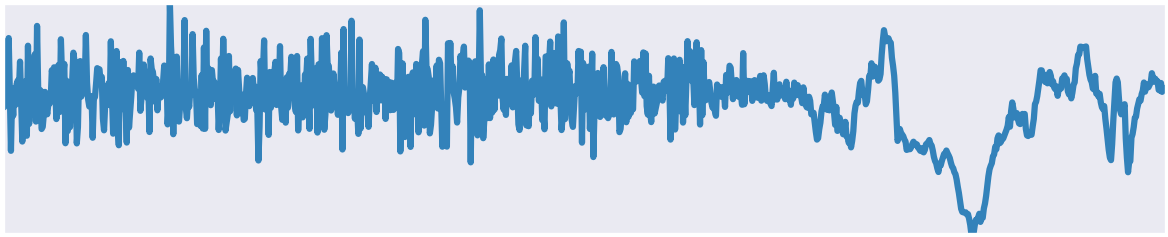}
    \vspace{0ex}\caption{}
    \label{fig:synthetic_scale_1024_0_w2}
    \end{subfigure}\hspace{0em}
    \begin{subfigure}[b]{0.245\textwidth}
        \includegraphics[width=\textwidth]{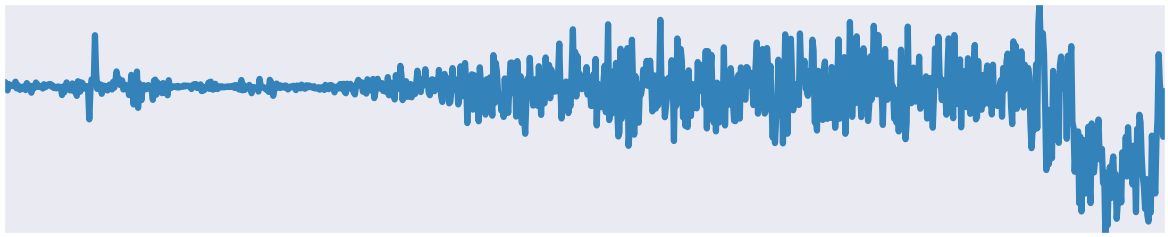}
    \vspace{0ex}\caption{}
    \label{fig:synthetic_scale_1024_0_w3}
    \end{subfigure}\hspace{0em}

    \stepcounter{row}%
    \begin{subfigure}[b]{0.245\textwidth}
        \includegraphics[width=\textwidth]{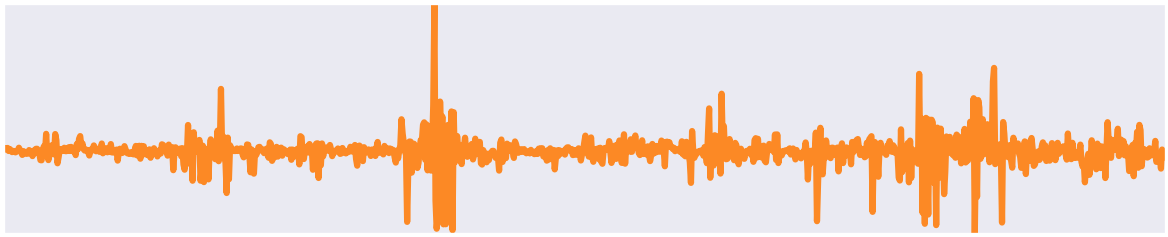}
    \vspace{0ex}\caption{}
    \label{fig:synthetic_scale_1024_1_w0}
    \end{subfigure}\hspace{0em}
    \begin{subfigure}[b]{0.245\textwidth}
        \includegraphics[width=\textwidth]{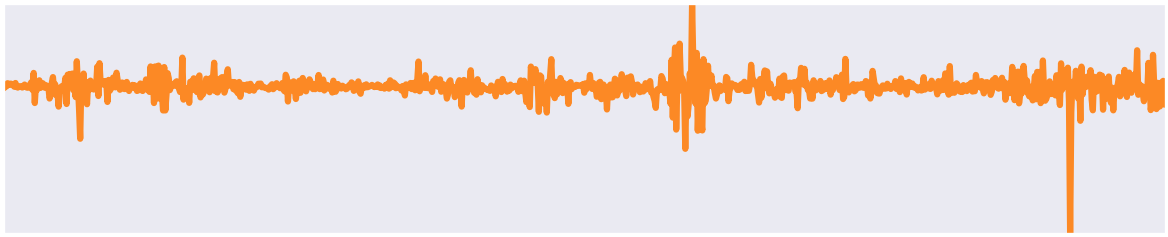}
    \vspace{0ex}\caption{}
    \label{fig:synthetic_scale_1024_1_w1}
    \end{subfigure}\hspace{0em}
    \begin{subfigure}[b]{0.245\textwidth}
        \includegraphics[width=\textwidth]{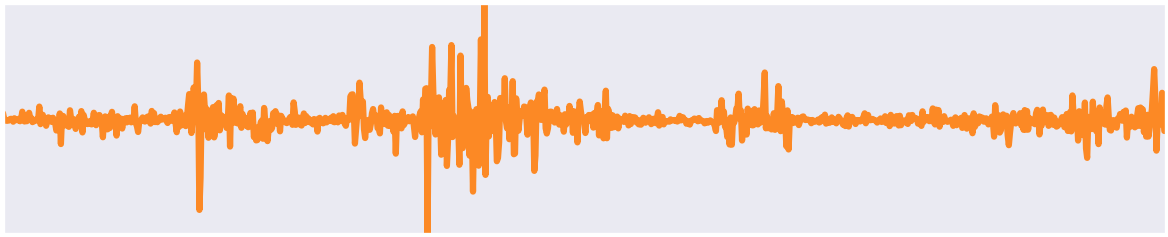}
    \vspace{0ex}\caption{}
    \label{fig:synthetic_scale_1024_1_w2}
    \end{subfigure}\hspace{0em}
    \begin{subfigure}[b]{0.245\textwidth}
        \includegraphics[width=\textwidth]{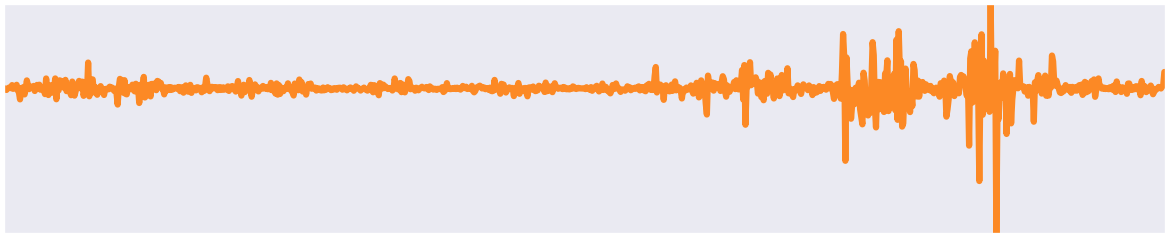}
    \vspace{0ex}\caption{}
    \label{fig:synthetic_scale_1024_1_w3}
    \end{subfigure}\hspace{0em}

    \stepcounter{row}%
    \begin{subfigure}[b]{0.245\textwidth}
        \includegraphics[width=\textwidth]{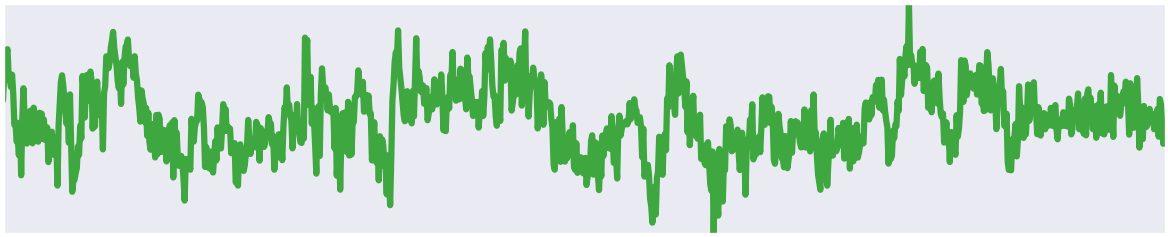}
    \vspace{0ex}\caption{}
    \label{fig:synthetic_scale_1024_2_w0}
    \end{subfigure}\hspace{0em}
    \begin{subfigure}[b]{0.245\textwidth}
        \includegraphics[width=\textwidth]{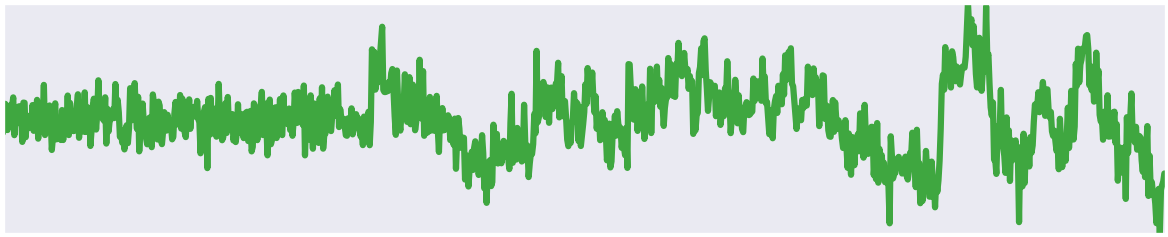}
    \vspace{0ex}\caption{}
    \label{fig:synthetic_scale_1024_2_w1}
    \end{subfigure}\hspace{0em}
    \begin{subfigure}[b]{0.245\textwidth}
        \includegraphics[width=\textwidth]{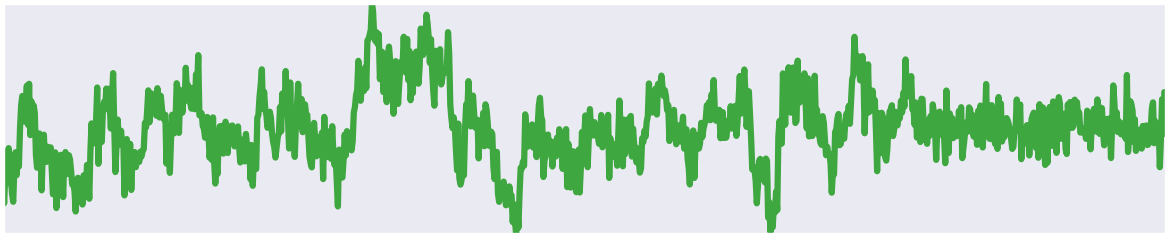}
    \vspace{0ex}\caption{}
    \label{fig:synthetic_scale_1024_2_w2}
    \end{subfigure}\hspace{0em}
    \begin{subfigure}[b]{0.245\textwidth}
        \includegraphics[width=\textwidth]{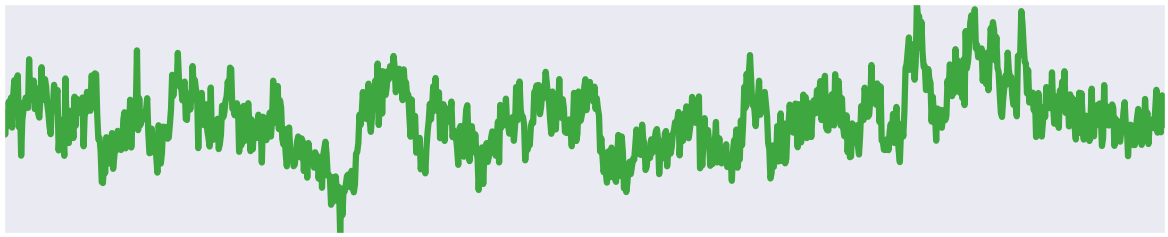}
    \vspace{0ex}\caption{}
    \label{fig:synthetic_scale_1024_2_w3}
    \end{subfigure}\hspace{0em}

    \stepcounter{row}%
    \begin{subfigure}[b]{0.245\textwidth}
        \includegraphics[width=\textwidth]{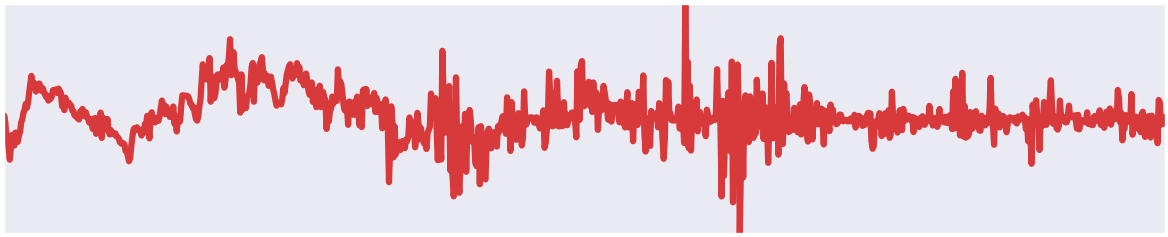}
    \vspace{0ex}\caption{}
    \label{fig:synthetic_scale_1024_3_w0}
    \end{subfigure}\hspace{0em}
    \begin{subfigure}[b]{0.245\textwidth}
        \includegraphics[width=\textwidth]{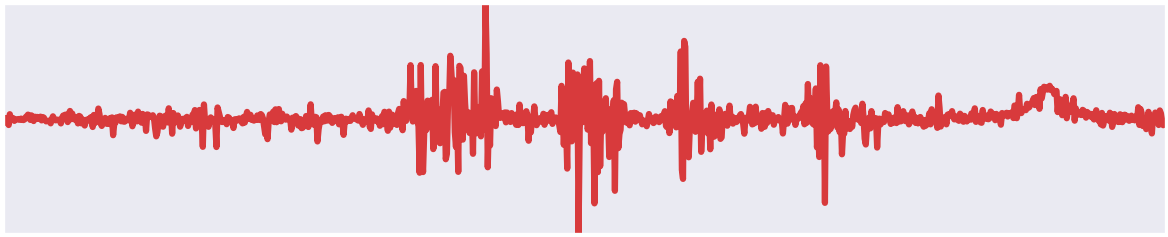}
    \vspace{0ex}\caption{}
    \label{fig:synthetic_scale_1024_3_w1}
    \end{subfigure}\hspace{0em}
    \begin{subfigure}[b]{0.245\textwidth}
        \includegraphics[width=\textwidth]{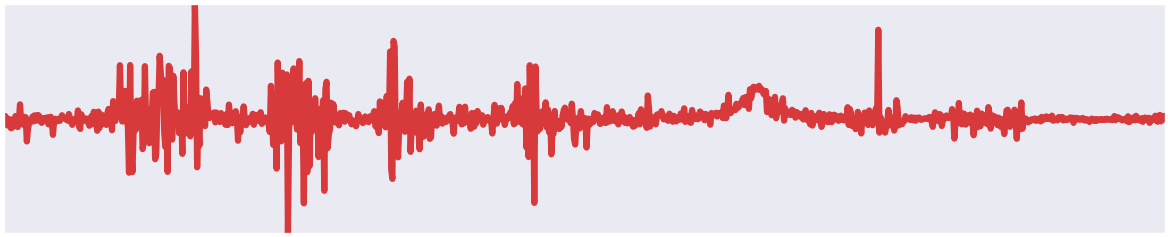}
    \vspace{0ex}\caption{}
    \label{fig:synthetic_scale_1024_3_w2}
    \end{subfigure}\hspace{0em}
    \begin{subfigure}[b]{0.245\textwidth}
        \includegraphics[width=\textwidth]{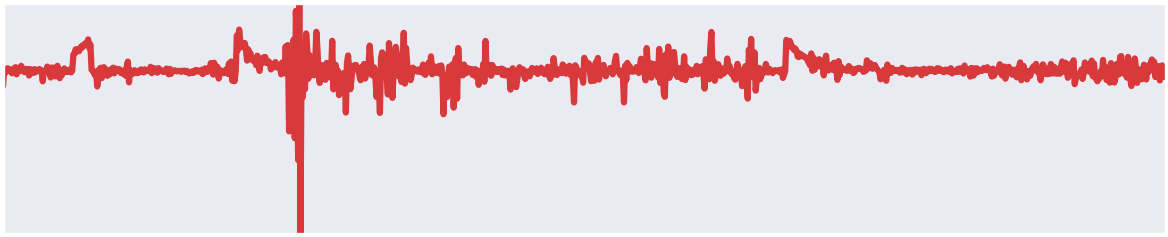}
    \vspace{0ex}\caption{}
    \label{fig:synthetic_scale_1024_3_w3}
    \end{subfigure}\hspace{0em}

    \stepcounter{row}%
    \begin{subfigure}[b]{0.245\textwidth}
        \includegraphics[width=\textwidth]{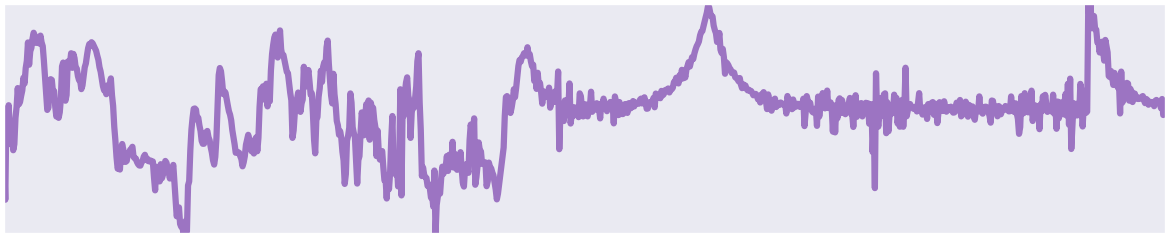}
    \vspace{0ex}\caption{}
    \label{fig:synthetic_scale_1024_4_w0}
    \end{subfigure}\hspace{0em}
    \begin{subfigure}[b]{0.245\textwidth}
        \includegraphics[width=\textwidth]{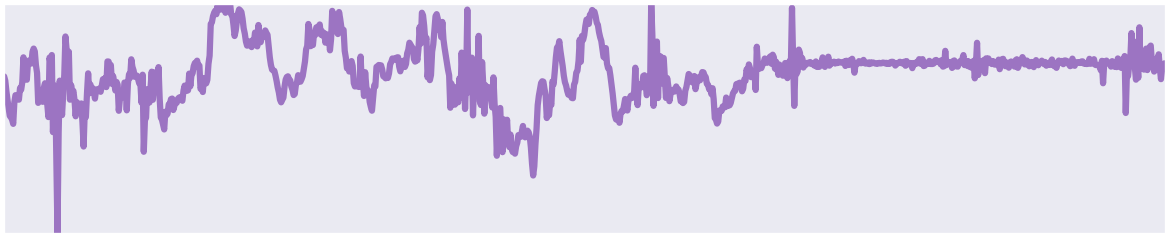}
    \vspace{0ex}\caption{}
    \label{fig:synthetic_scale_1024_4_w1}
    \end{subfigure}\hspace{0em}
    \begin{subfigure}[b]{0.245\textwidth}
        \includegraphics[width=\textwidth]{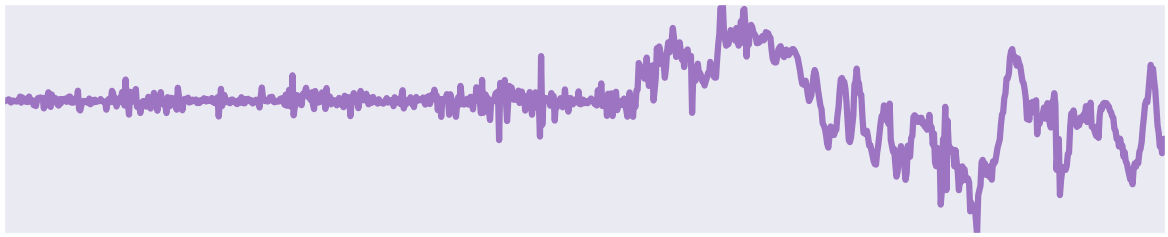}
    \vspace{0ex}\caption{}
    \label{fig:synthetic_scale_1024_4_w2}
    \end{subfigure}\hspace{0em}
    \begin{subfigure}[b]{0.245\textwidth}
        \includegraphics[width=\textwidth]{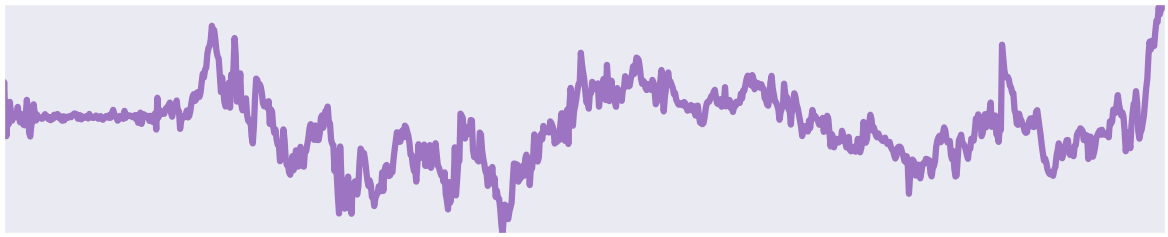}
    \vspace{0ex}\caption{}
    \label{fig:synthetic_scale_1024_4_w3}
    \end{subfigure}\hspace{0em}

    \stepcounter{row}%
    \begin{subfigure}[b]{0.245\textwidth}
        \includegraphics[width=\textwidth]{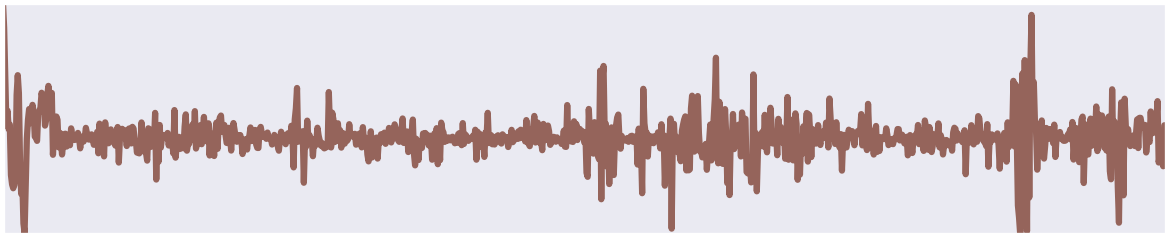}
    \vspace{0ex}\caption{}
    \label{fig:synthetic_scale_1024_5_w0}
    \end{subfigure}\hspace{0em}
    \begin{subfigure}[b]{0.245\textwidth}
        \includegraphics[width=\textwidth]{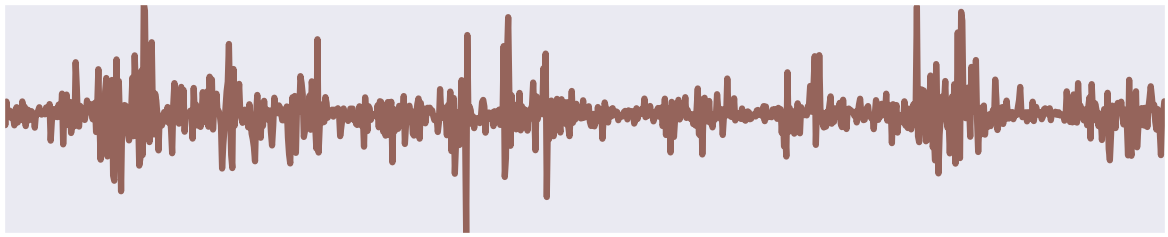}
    \vspace{0ex}\caption{}
    \label{fig:synthetic_scale_1024_5_w1}
    \end{subfigure}\hspace{0em}
    \begin{subfigure}[b]{0.245\textwidth}
        \includegraphics[width=\textwidth]{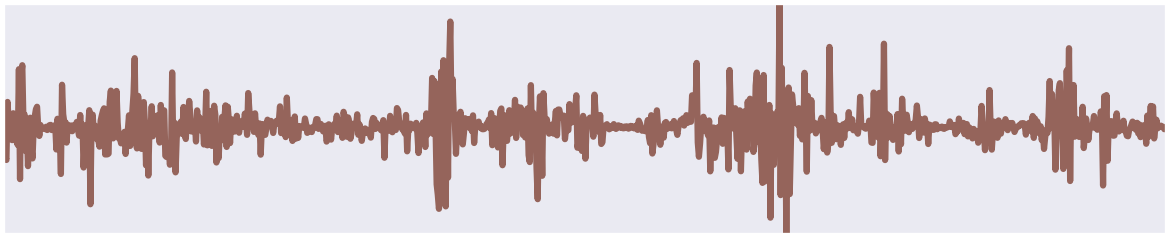}
    \vspace{0ex}\caption{}
    \label{fig:synthetic_scale_1024_5_w2}
    \end{subfigure}\hspace{0em}
    \begin{subfigure}[b]{0.245\textwidth}
        \includegraphics[width=\textwidth]{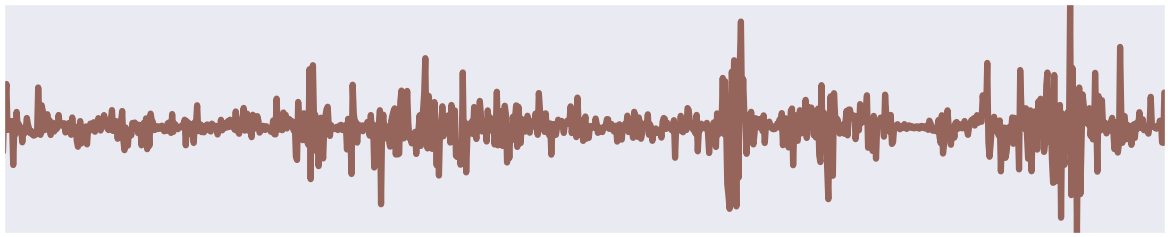}
    \vspace{0ex}\caption{}
    \label{fig:synthetic_scale_1024_5_w3}
    \end{subfigure}\hspace{0em}

    \stepcounter{row}%
    \begin{subfigure}[b]{0.245\textwidth}
        \includegraphics[width=\textwidth]{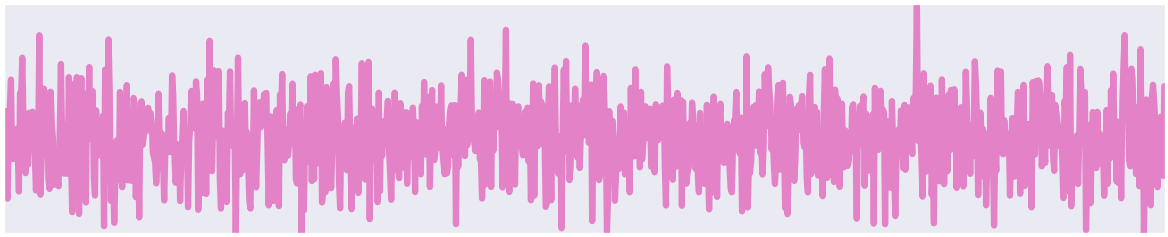}
    \vspace{0ex}\caption{}
    \label{fig:synthetic_scale_1024_6_w0}
    \end{subfigure}\hspace{0em}
    \begin{subfigure}[b]{0.245\textwidth}
        \includegraphics[width=\textwidth]{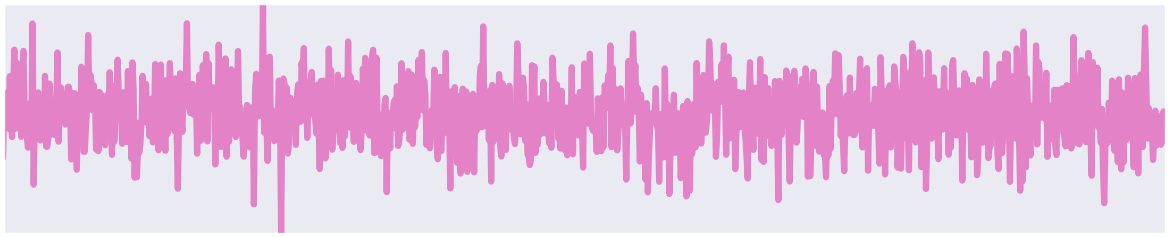}
    \vspace{0ex}\caption{}
    \label{fig:synthetic_scale_1024_6_w1}
    \end{subfigure}\hspace{0em}
    \begin{subfigure}[b]{0.245\textwidth}
        \includegraphics[width=\textwidth]{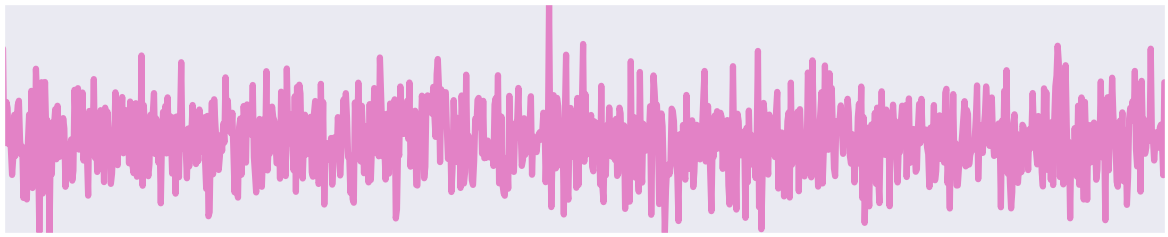}
    \vspace{0ex}\caption{}
    \label{fig:synthetic_scale_1024_6_w2}
    \end{subfigure}\hspace{0em}
    \begin{subfigure}[b]{0.245\textwidth}
        \includegraphics[width=\textwidth]{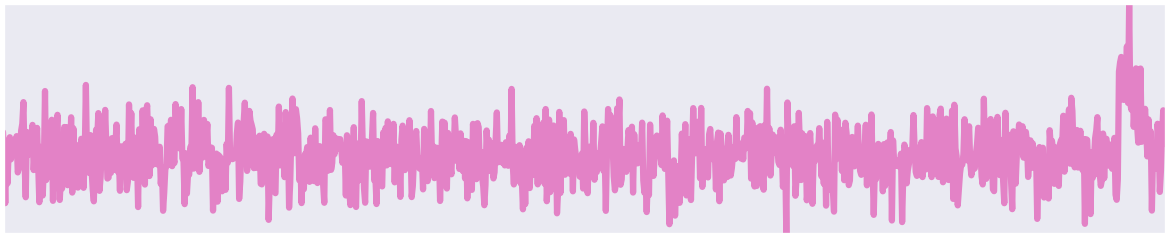}
    \vspace{0ex}\caption{}
    \label{fig:synthetic_scale_1024_6_w3}
    \end{subfigure}\hspace{0em}

    \stepcounter{row}%
    \begin{subfigure}[b]{0.245\textwidth}
        \includegraphics[width=\textwidth]{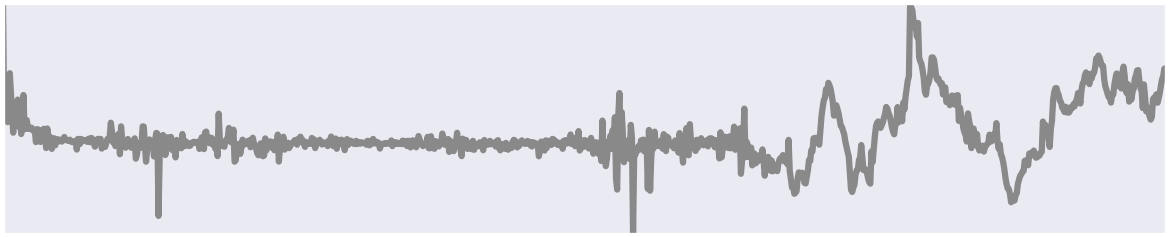}
    \vspace{0ex}\caption{}
    \label{fig:synthetic_scale_1024_7_w0}
    \end{subfigure}\hspace{0em}
    \begin{subfigure}[b]{0.245\textwidth}
        \includegraphics[width=\textwidth]{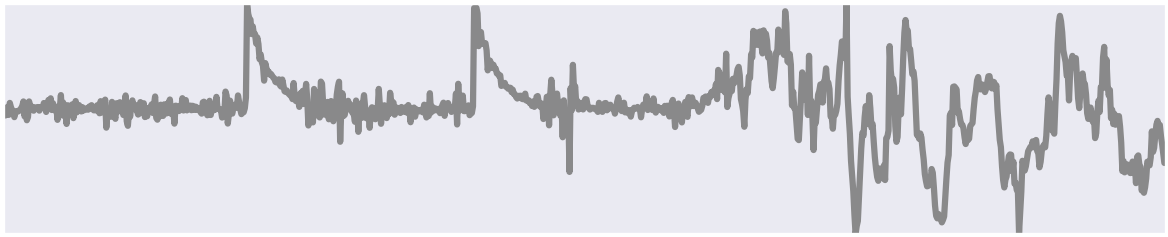}
    \vspace{0ex}\caption{}
    \label{fig:synthetic_scale_1024_7_w1}
    \end{subfigure}\hspace{0em}
    \begin{subfigure}[b]{0.245\textwidth}
        \includegraphics[width=\textwidth]{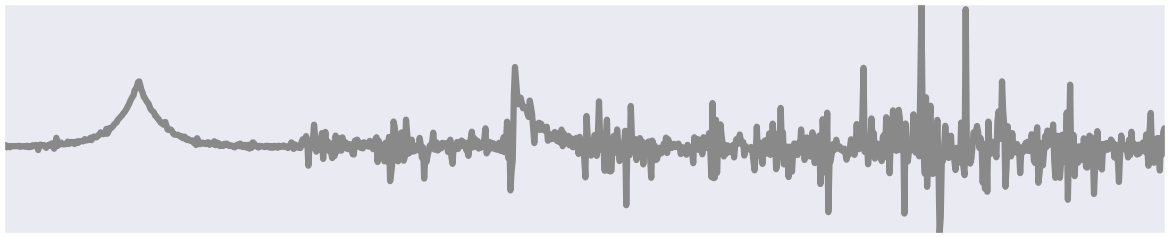}
    \vspace{0ex}\caption{}
    \label{fig:synthetic_scale_1024_7_w2}
    \end{subfigure}\hspace{0em}
    \begin{subfigure}[b]{0.245\textwidth}
        \includegraphics[width=\textwidth]{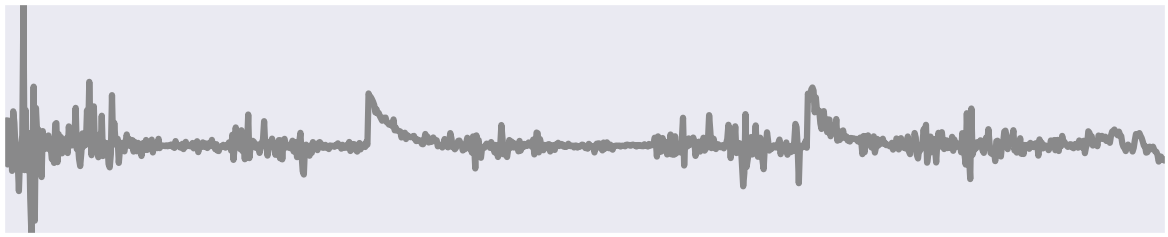}
    \vspace{0ex}\caption{}
    \label{fig:synthetic_scale_1024_7_w3}
    \end{subfigure}\hspace{0em}

    \stepcounter{row}%
    \begin{subfigure}[b]{0.245\textwidth}
        \includegraphics[width=\textwidth]{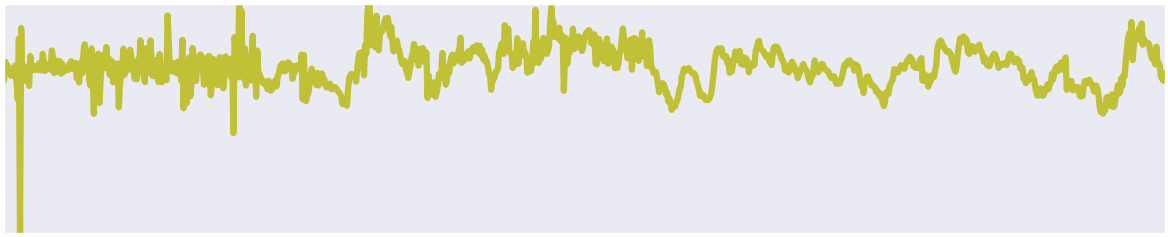}
    \vspace{0ex}\caption{}
    \label{fig:synthetic_scale_1024_8_w0}
    \end{subfigure}\hspace{0em}
    \begin{subfigure}[b]{0.245\textwidth}
        \includegraphics[width=\textwidth]{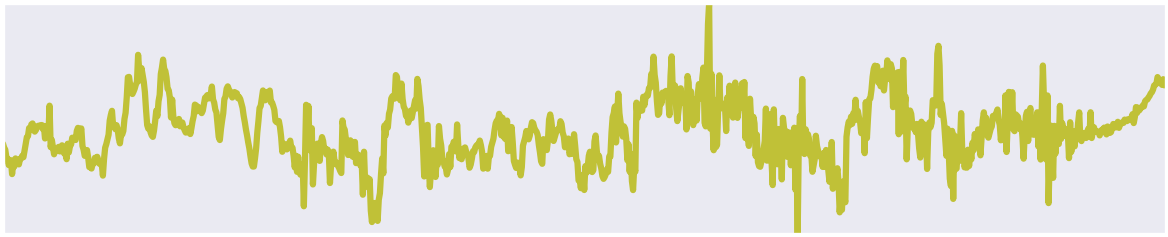}
    \vspace{0ex}\caption{}
    \label{fig:synthetic_scale_1024_8_w1}
    \end{subfigure}\hspace{0em}
    \begin{subfigure}[b]{0.245\textwidth}
        \includegraphics[width=\textwidth]{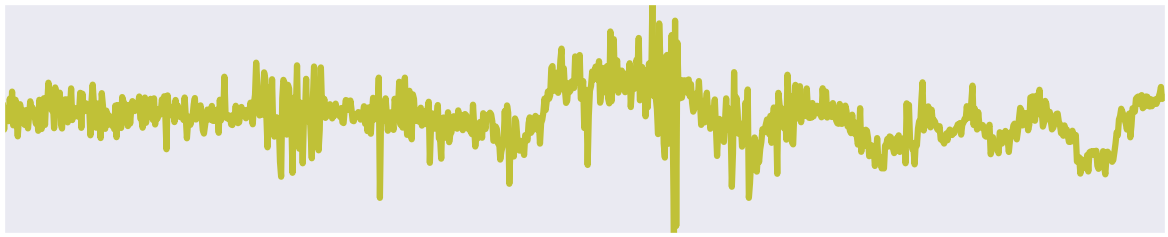}
    \vspace{0ex}\caption{}
    \label{fig:synthetic_scale_1024_8_w2}
    \end{subfigure}\hspace{0em}
    \begin{subfigure}[b]{0.245\textwidth}
        \includegraphics[width=\textwidth]{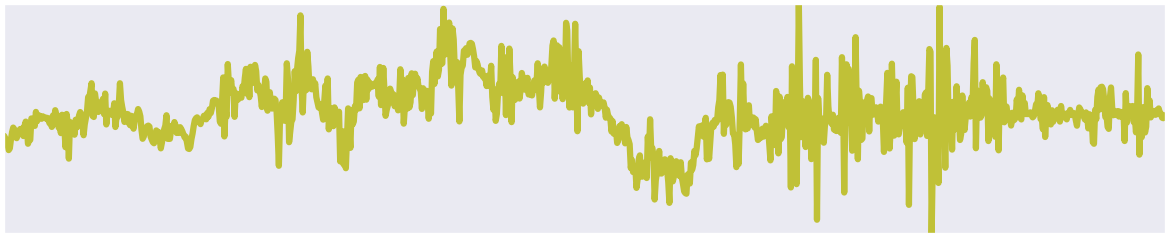}
    \vspace{0ex}\caption{}
    \label{fig:synthetic_scale_1024_8_w3}
    \end{subfigure}\hspace{0em}

    \caption{Waveforms in timescale of $1024$ sample, with each row from top to bottom illustrating four waveforms corresponding to clusters one to nine in this timescale, respectively.}
    \label{fig:synthetic_scale_1024_waves}
\end{figure*}

\begin{figure*}
    \renewcommand\thesubfigure{\roman{subfigure}}
    \captionsetup[subfigure]{skip=-10pt}
    \renewcommand{\thesubfigure}{\alph{subfigure}.\arabic{row}}
    \centering
    \setcounter{row}{1}%
    \begin{subfigure}[b]{0.245\textwidth}
        \includegraphics[width=\textwidth]{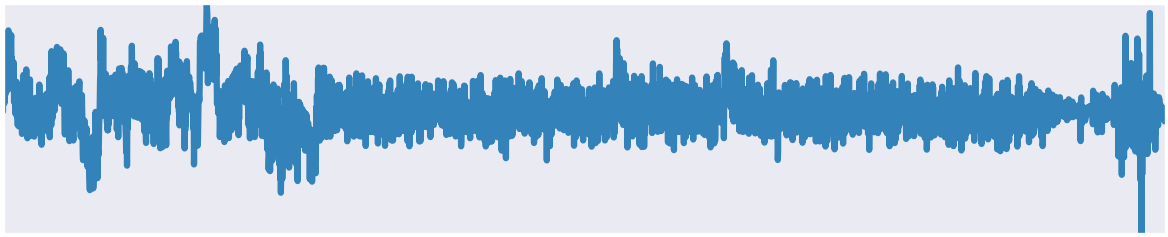}
    \vspace{0ex}\caption{}
    \label{fig:synthetic_scale_4096_0_w0}
    \end{subfigure}\hspace{0em}
    \begin{subfigure}[b]{0.245\textwidth}
        \includegraphics[width=\textwidth]{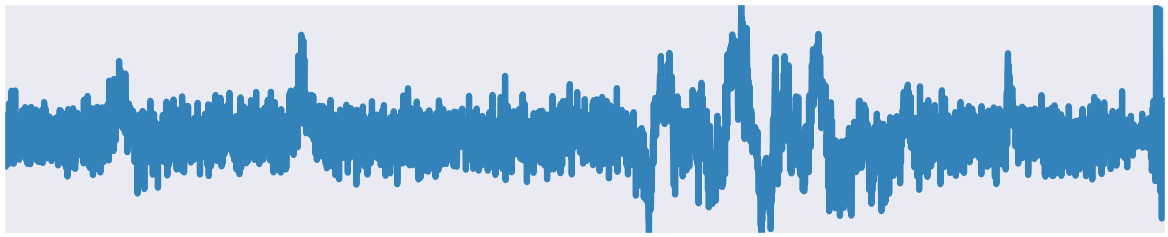}
    \vspace{0ex}\caption{}
    \label{fig:synthetic_scale_4096_0_w1}
    \end{subfigure}\hspace{0em}
    \begin{subfigure}[b]{0.245\textwidth}
        \includegraphics[width=\textwidth]{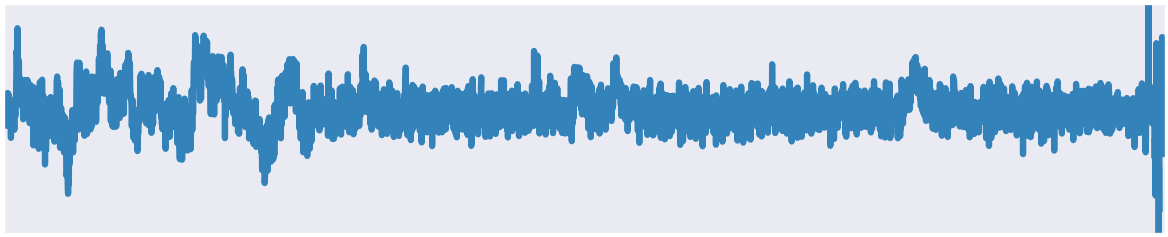}
    \vspace{0ex}\caption{}
    \label{fig:synthetic_scale_4096_0_w2}
    \end{subfigure}\hspace{0em}
    \begin{subfigure}[b]{0.245\textwidth}
        \includegraphics[width=\textwidth]{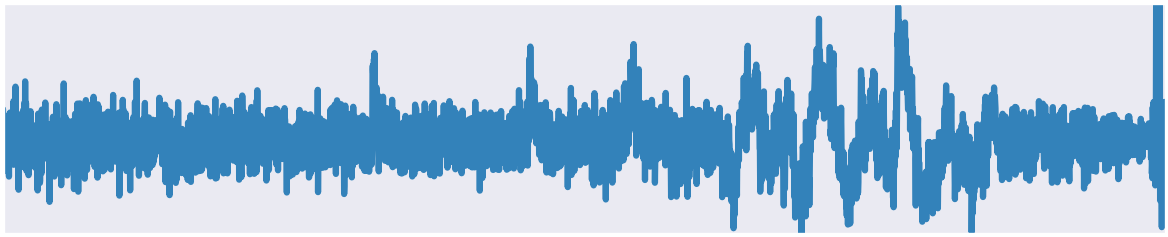}
    \vspace{0ex}\caption{}
    \label{fig:synthetic_scale_4096_0_w3}
    \end{subfigure}\hspace{0em}

    \stepcounter{row}%
    \begin{subfigure}[b]{0.245\textwidth}
        \includegraphics[width=\textwidth]{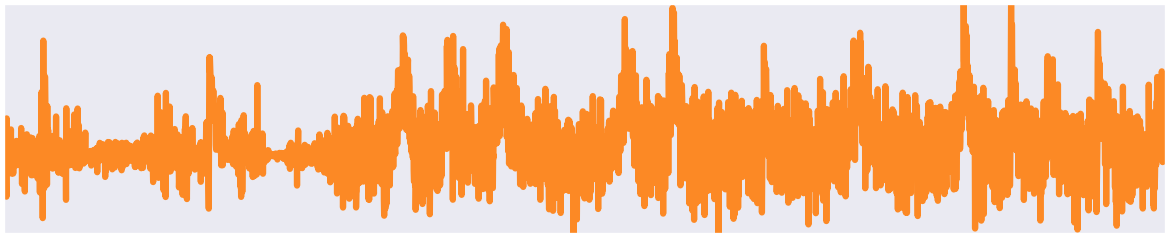}
    \vspace{0ex}\caption{}
    \label{fig:synthetic_scale_4096_1_w0}
    \end{subfigure}\hspace{0em}
    \begin{subfigure}[b]{0.245\textwidth}
        \includegraphics[width=\textwidth]{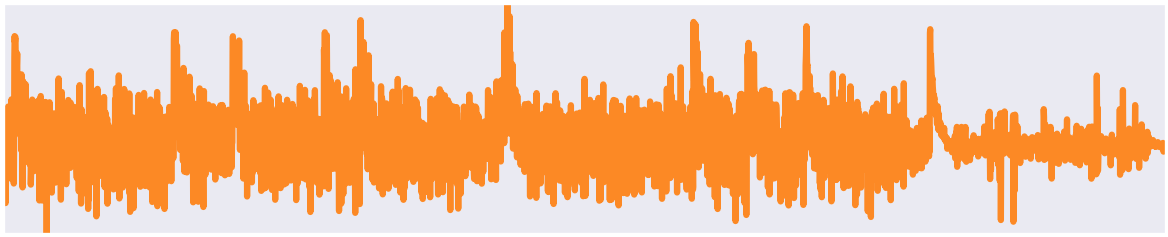}
    \vspace{0ex}\caption{}
    \label{fig:synthetic_scale_4096_1_w1}
    \end{subfigure}\hspace{0em}
    \begin{subfigure}[b]{0.245\textwidth}
        \includegraphics[width=\textwidth]{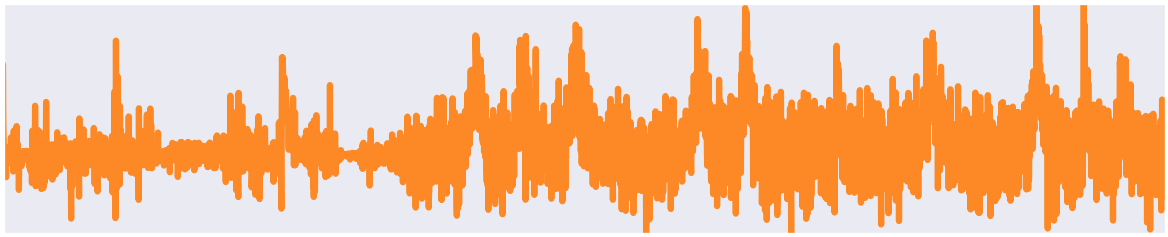}
    \vspace{0ex}\caption{}
    \label{fig:synthetic_scale_4096_1_w2}
    \end{subfigure}\hspace{0em}
    \begin{subfigure}[b]{0.245\textwidth}
        \includegraphics[width=\textwidth]{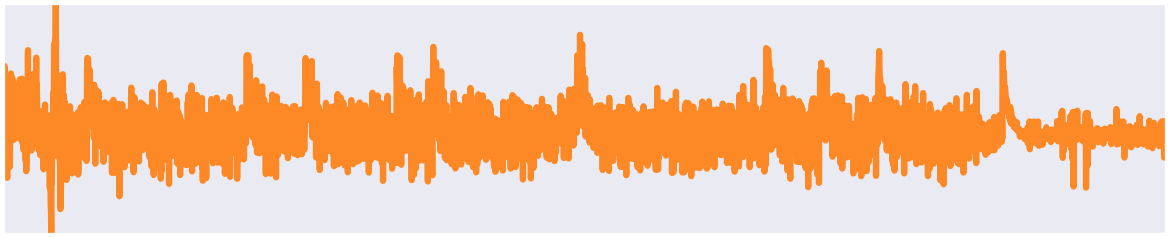}
    \vspace{0ex}\caption{}
    \label{fig:synthetic_scale_4096_1_w3}
    \end{subfigure}\hspace{0em}

    \stepcounter{row}%
    \begin{subfigure}[b]{0.245\textwidth}
        \includegraphics[width=\textwidth]{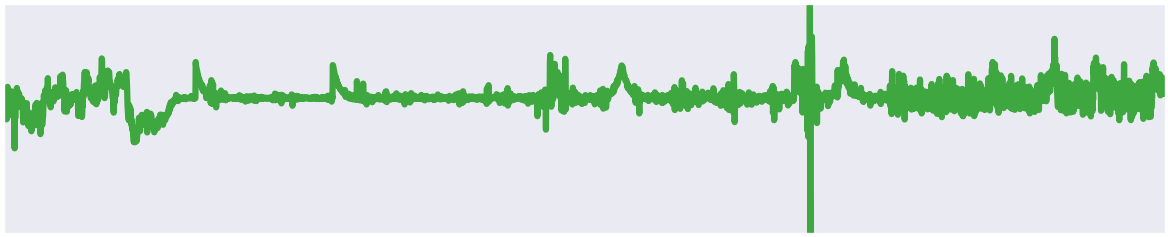}
    \vspace{0ex}\caption{}
    \label{fig:synthetic_scale_4096_2_w0}
    \end{subfigure}\hspace{0em}
    \begin{subfigure}[b]{0.245\textwidth}
        \includegraphics[width=\textwidth]{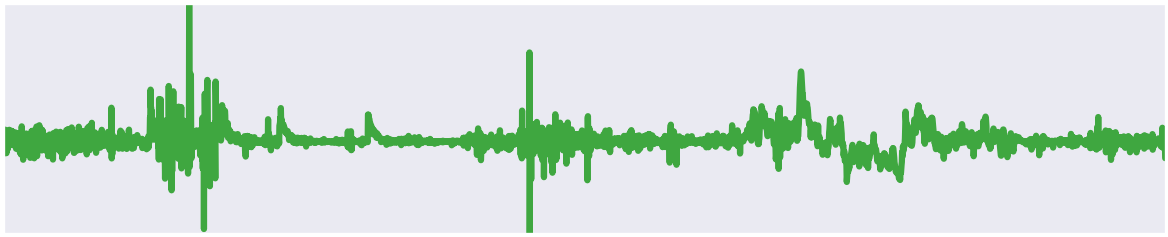}
    \vspace{0ex}\caption{}
    \label{fig:synthetic_scale_4096_2_w1}
    \end{subfigure}\hspace{0em}
    \begin{subfigure}[b]{0.245\textwidth}
        \includegraphics[width=\textwidth]{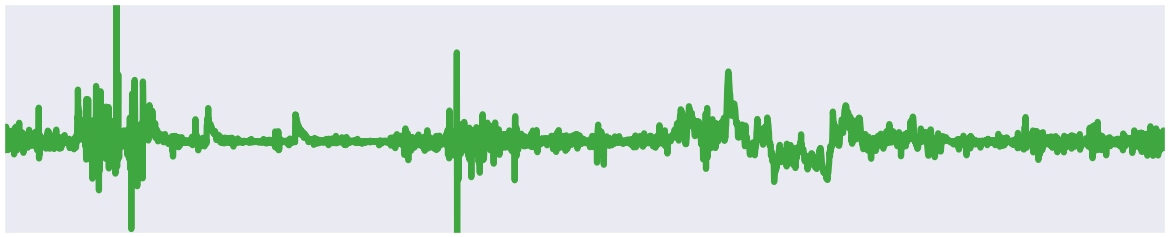}
    \vspace{0ex}\caption{}
    \label{fig:synthetic_scale_4096_2_w2}
    \end{subfigure}\hspace{0em}
    \begin{subfigure}[b]{0.245\textwidth}
        \includegraphics[width=\textwidth]{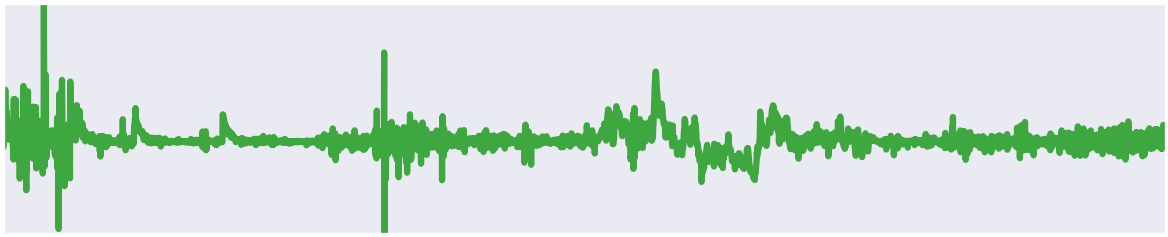}
    \vspace{0ex}\caption{}
    \label{fig:synthetic_scale_4096_2_w3}
    \end{subfigure}\hspace{0em}

    \stepcounter{row}%
    \begin{subfigure}[b]{0.245\textwidth}
        \includegraphics[width=\textwidth]{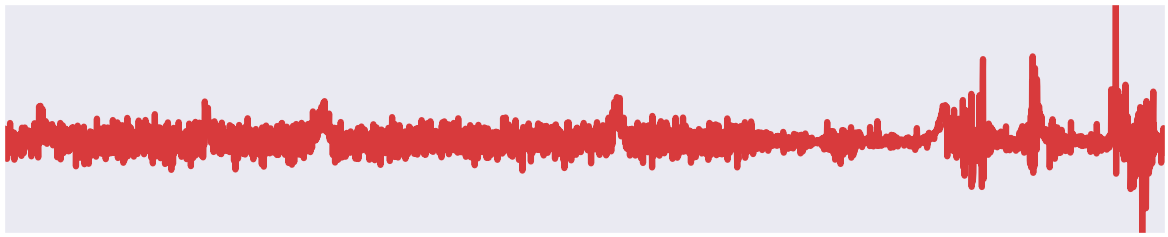}
    \vspace{0ex}\caption{}
    \label{fig:synthetic_scale_4096_3_w0}
    \end{subfigure}\hspace{0em}
    \begin{subfigure}[b]{0.245\textwidth}
        \includegraphics[width=\textwidth]{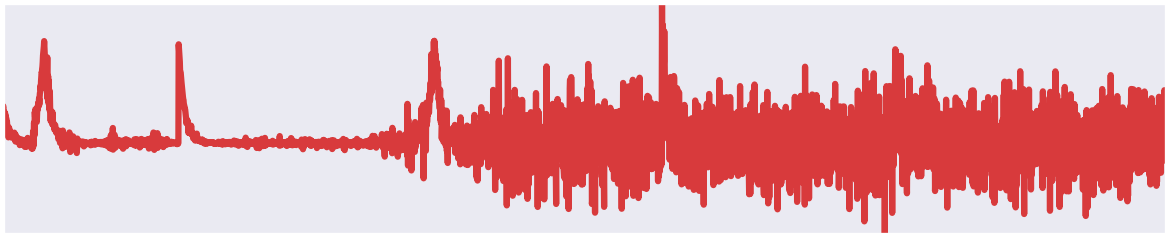}
    \vspace{0ex}\caption{}
    \label{fig:synthetic_scale_4096_3_w1}
    \end{subfigure}\hspace{0em}
    \begin{subfigure}[b]{0.245\textwidth}
        \includegraphics[width=\textwidth]{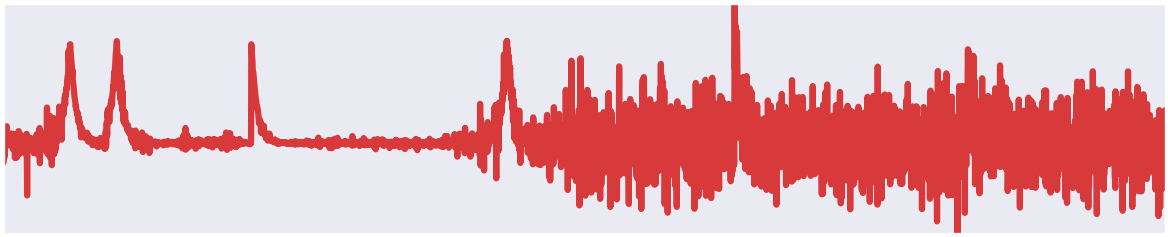}
    \vspace{0ex}\caption{}
    \label{fig:synthetic_scale_4096_3_w2}
    \end{subfigure}\hspace{0em}
    \begin{subfigure}[b]{0.245\textwidth}
        \includegraphics[width=\textwidth]{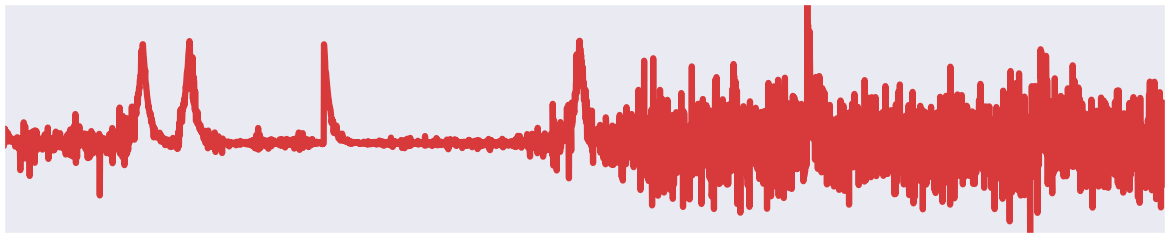}
    \vspace{0ex}\caption{}
    \label{fig:synthetic_scale_4096_3_w3}
    \end{subfigure}\hspace{0em}

    \stepcounter{row}%
    \begin{subfigure}[b]{0.245\textwidth}
        \includegraphics[width=\textwidth]{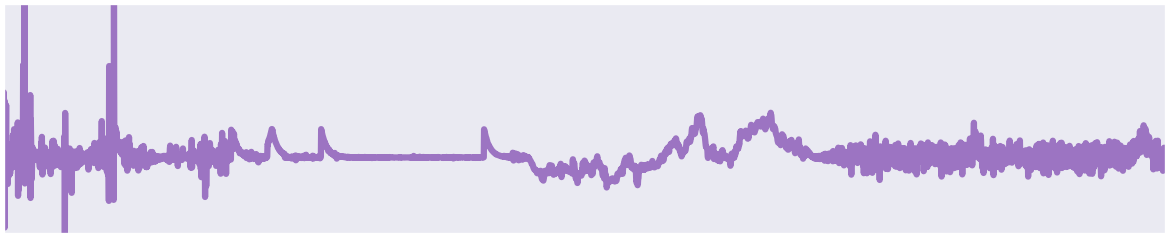}
    \vspace{0ex}\caption{}
    \label{fig:synthetic_scale_4096_4_w0}
    \end{subfigure}\hspace{0em}
    \begin{subfigure}[b]{0.245\textwidth}
        \includegraphics[width=\textwidth]{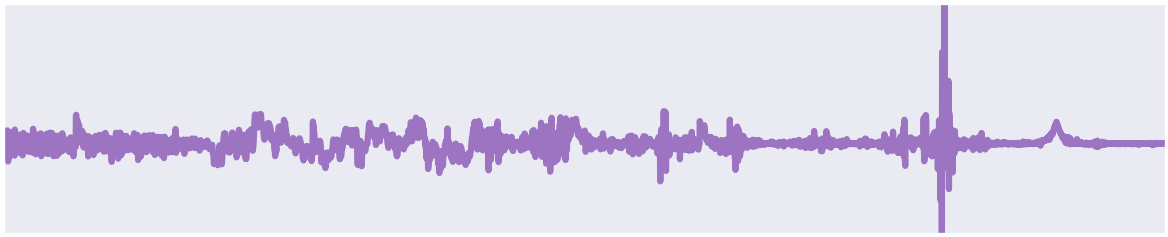}
    \vspace{0ex}\caption{}
    \label{fig:synthetic_scale_4096_4_w1}
    \end{subfigure}\hspace{0em}
    \begin{subfigure}[b]{0.245\textwidth}
        \includegraphics[width=\textwidth]{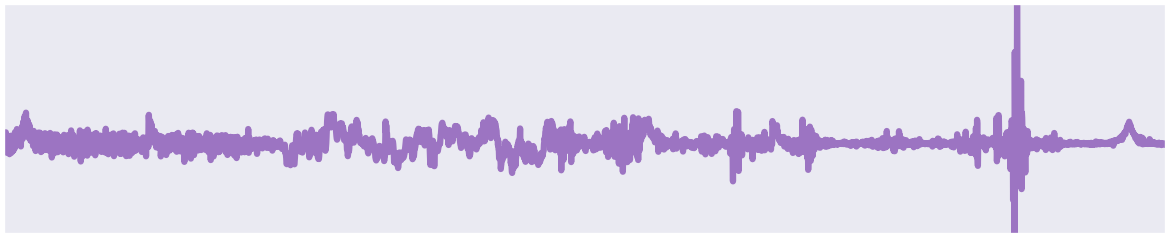}
    \vspace{0ex}\caption{}
    \label{fig:synthetic_scale_4096_4_w2}
    \end{subfigure}\hspace{0em}
    \begin{subfigure}[b]{0.245\textwidth}
        \includegraphics[width=\textwidth]{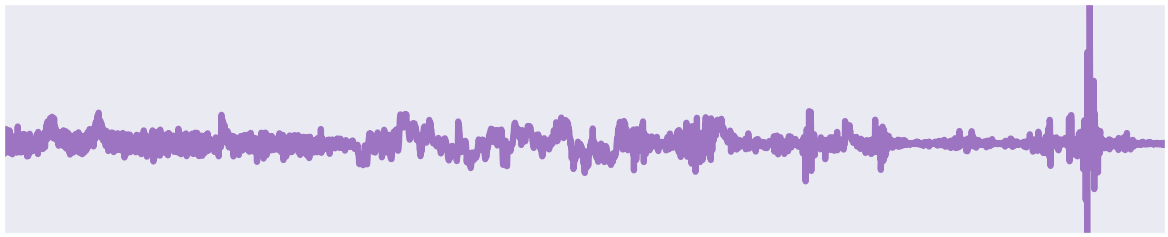}
    \vspace{0ex}\caption{}
    \label{fig:synthetic_scale_4096_4_w3}
    \end{subfigure}\hspace{0em}

    \stepcounter{row}%
    \begin{subfigure}[b]{0.245\textwidth}
        \includegraphics[width=\textwidth]{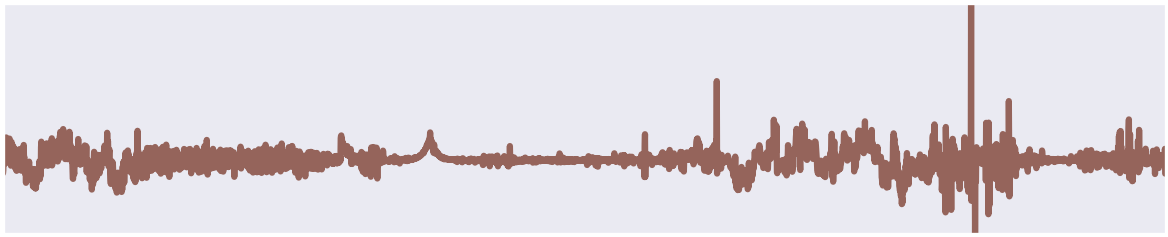}
    \vspace{0ex}\caption{}
    \label{fig:synthetic_scale_4096_5_w0}
    \end{subfigure}\hspace{0em}
    \begin{subfigure}[b]{0.245\textwidth}
        \includegraphics[width=\textwidth]{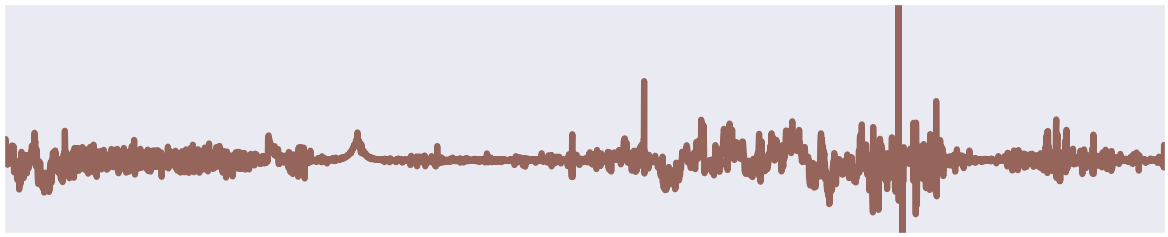}
    \vspace{0ex}\caption{}
    \label{fig:synthetic_scale_4096_5_w1}
    \end{subfigure}\hspace{0em}
    \begin{subfigure}[b]{0.245\textwidth}
        \includegraphics[width=\textwidth]{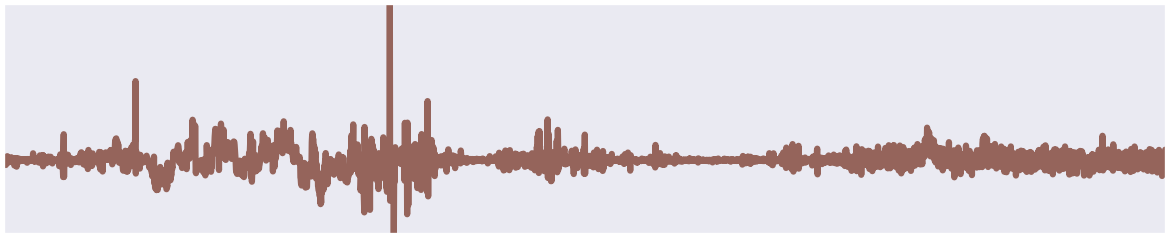}
    \vspace{0ex}\caption{}
    \label{fig:synthetic_scale_4096_5_w2}
    \end{subfigure}\hspace{0em}
    \begin{subfigure}[b]{0.245\textwidth}
        \includegraphics[width=\textwidth]{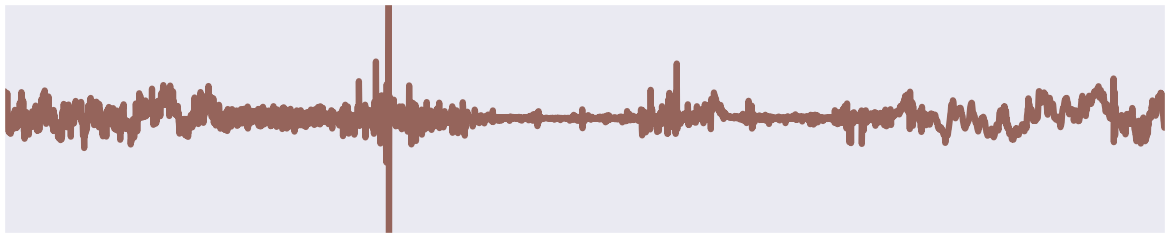}
    \vspace{0ex}\caption{}
    \label{fig:synthetic_scale_4096_5_w3}
    \end{subfigure}\hspace{0em}

    \stepcounter{row}%
    \begin{subfigure}[b]{0.245\textwidth}
        \includegraphics[width=\textwidth]{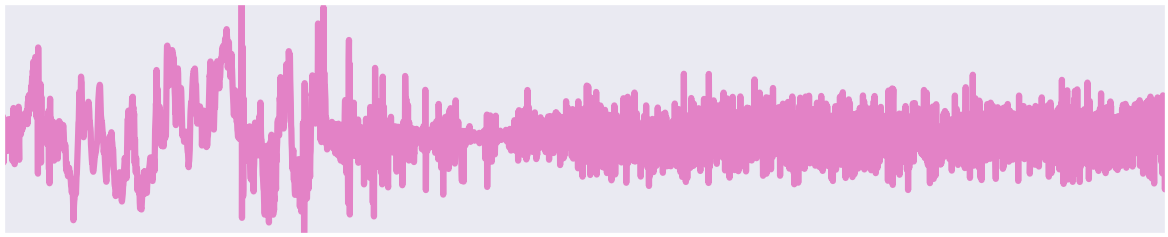}
    \vspace{0ex}\caption{}
    \label{fig:synthetic_scale_4096_6_w0}
    \end{subfigure}\hspace{0em}
    \begin{subfigure}[b]{0.245\textwidth}
        \includegraphics[width=\textwidth]{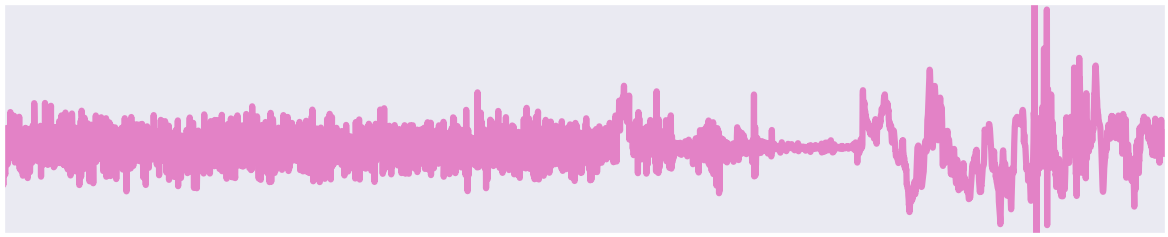}
    \vspace{0ex}\caption{}
    \label{fig:synthetic_scale_4096_6_w1}
    \end{subfigure}\hspace{0em}
    \begin{subfigure}[b]{0.245\textwidth}
        \includegraphics[width=\textwidth]{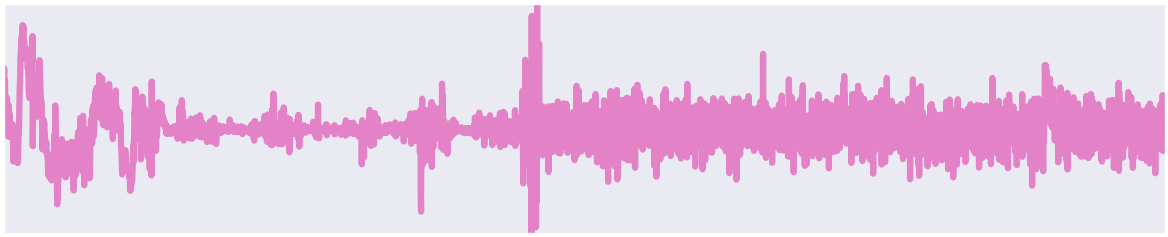}
    \vspace{0ex}\caption{}
    \label{fig:synthetic_scale_4096_6_w2}
    \end{subfigure}\hspace{0em}
    \begin{subfigure}[b]{0.245\textwidth}
        \includegraphics[width=\textwidth]{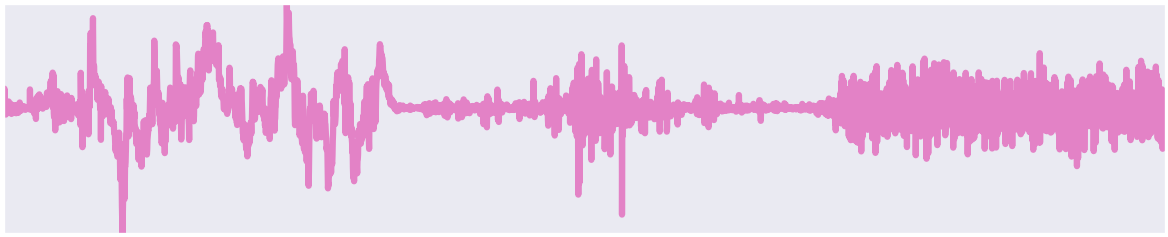}
    \vspace{0ex}\caption{}
    \label{fig:synthetic_scale_4096_6_w3}
    \end{subfigure}\hspace{0em}

    \stepcounter{row}%
    \begin{subfigure}[b]{0.245\textwidth}
        \includegraphics[width=\textwidth]{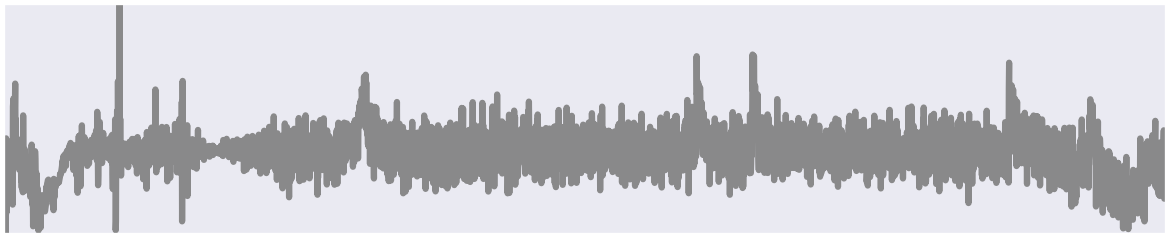}
    \vspace{0ex}\caption{}
    \label{fig:synthetic_scale_4096_7_w0}
    \end{subfigure}\hspace{0em}
    \begin{subfigure}[b]{0.245\textwidth}
        \includegraphics[width=\textwidth]{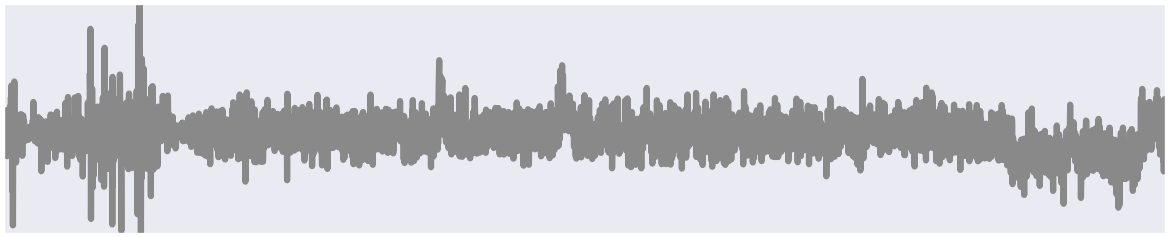}
    \vspace{0ex}\caption{}
    \label{fig:synthetic_scale_4096_7_w1}
    \end{subfigure}\hspace{0em}
    \begin{subfigure}[b]{0.245\textwidth}
        \includegraphics[width=\textwidth]{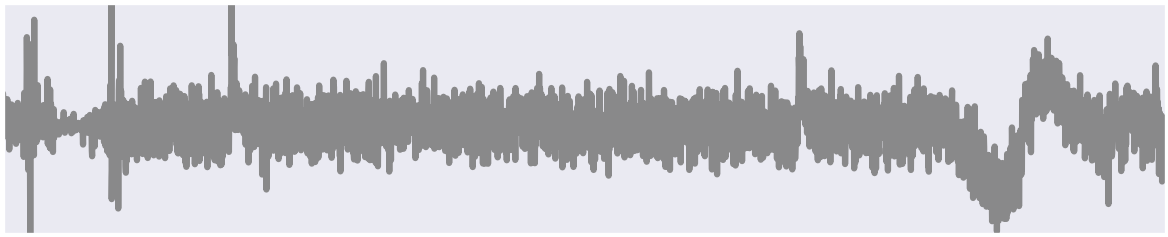}
    \vspace{0ex}\caption{}
    \label{fig:synthetic_scale_4096_7_w2}
    \end{subfigure}\hspace{0em}
    \begin{subfigure}[b]{0.245\textwidth}
        \includegraphics[width=\textwidth]{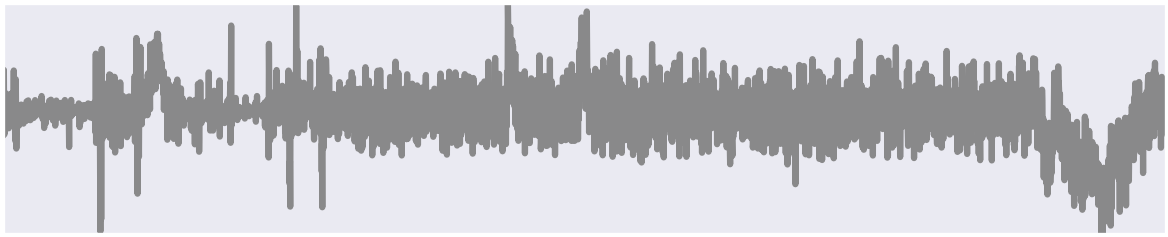}
    \vspace{0ex}\caption{}
    \label{fig:synthetic_scale_4096_7_w3}
    \end{subfigure}\hspace{0em}

    \stepcounter{row}%
    \begin{subfigure}[b]{0.245\textwidth}
        \includegraphics[width=\textwidth]{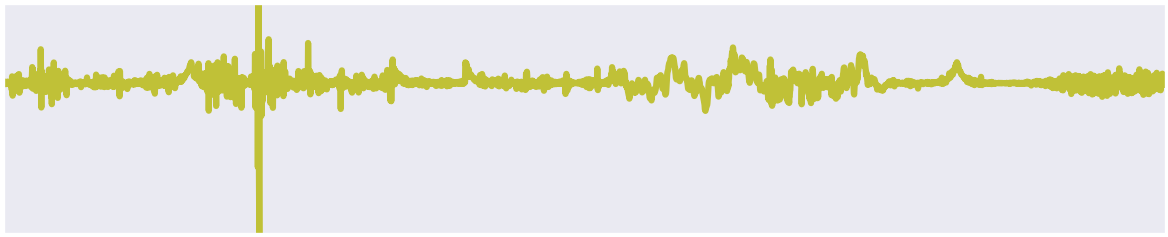}
    \vspace{0ex}\caption{}
    \label{fig:synthetic_scale_4096_8_w0}
    \end{subfigure}\hspace{0em}
    \begin{subfigure}[b]{0.245\textwidth}
        \includegraphics[width=\textwidth]{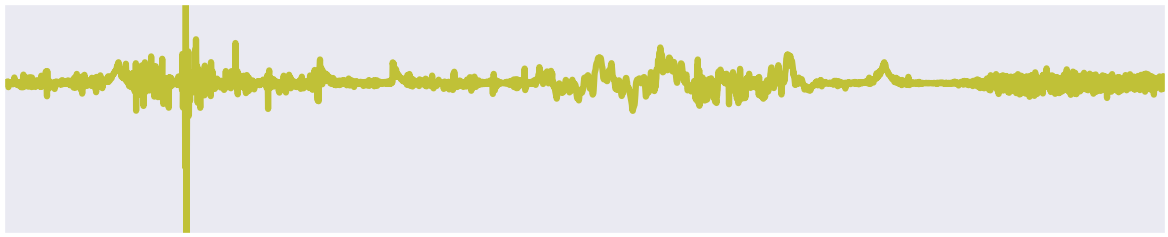}
    \vspace{0ex}\caption{}
    \label{fig:synthetic_scale_4096_8_w1}
    \end{subfigure}\hspace{0em}
    \begin{subfigure}[b]{0.245\textwidth}
        \includegraphics[width=\textwidth]{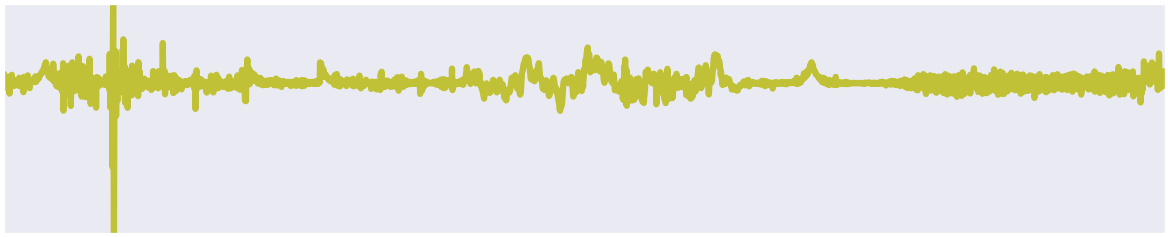}
    \vspace{0ex}\caption{}
    \label{fig:synthetic_scale_4096_8_w2}
    \end{subfigure}\hspace{0em}
    \begin{subfigure}[b]{0.245\textwidth}
        \includegraphics[width=\textwidth]{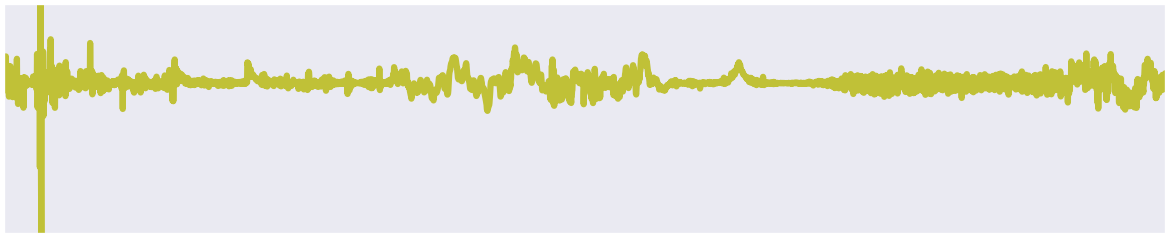}
    \vspace{0ex}\caption{}
    \label{fig:synthetic_scale_4096_8_w3}
    \end{subfigure}\hspace{0em}

    \caption{Waveforms in timescale of $4096$ sample, with each row from top to bottom illustrating four waveforms corresponding to clusters one to nine in this timescale, respectively.}
    \label{fig:synthetic_scale_4096_waves}
\end{figure*}

\begin{figure*}[t]
    \centering

    \begin{subfigure}[b]{0.32\textwidth}
        \includegraphics[width=\textwidth]{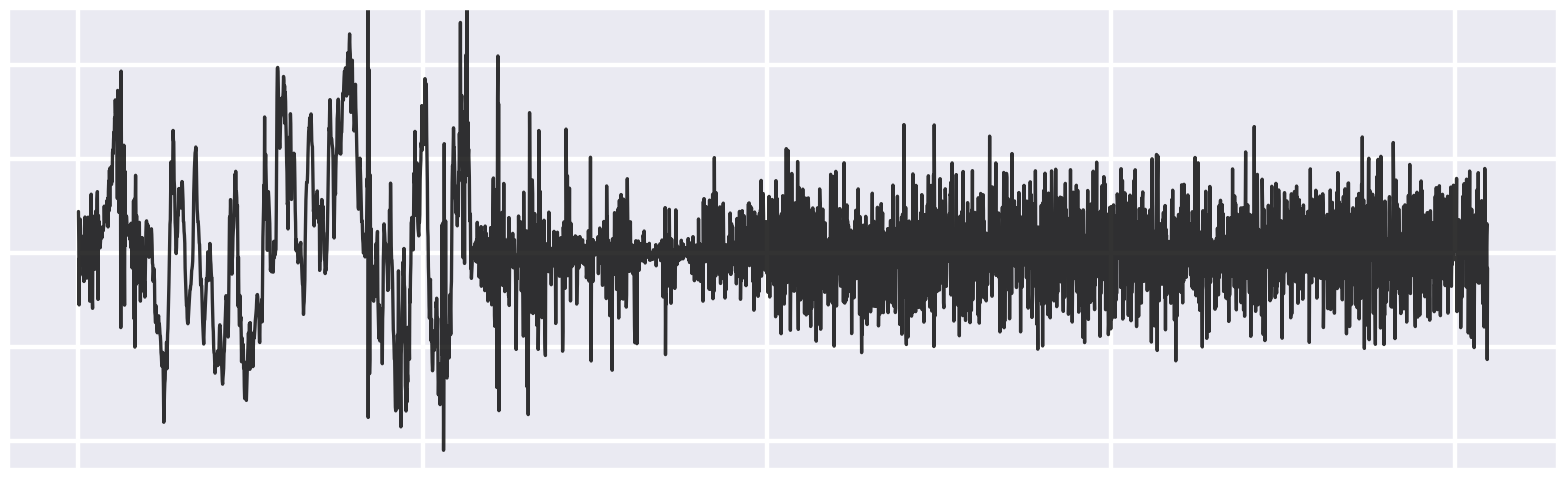}
    \end{subfigure}\hspace{0em}
    \begin{subfigure}[b]{0.32\textwidth}
        \includegraphics[width=\textwidth]{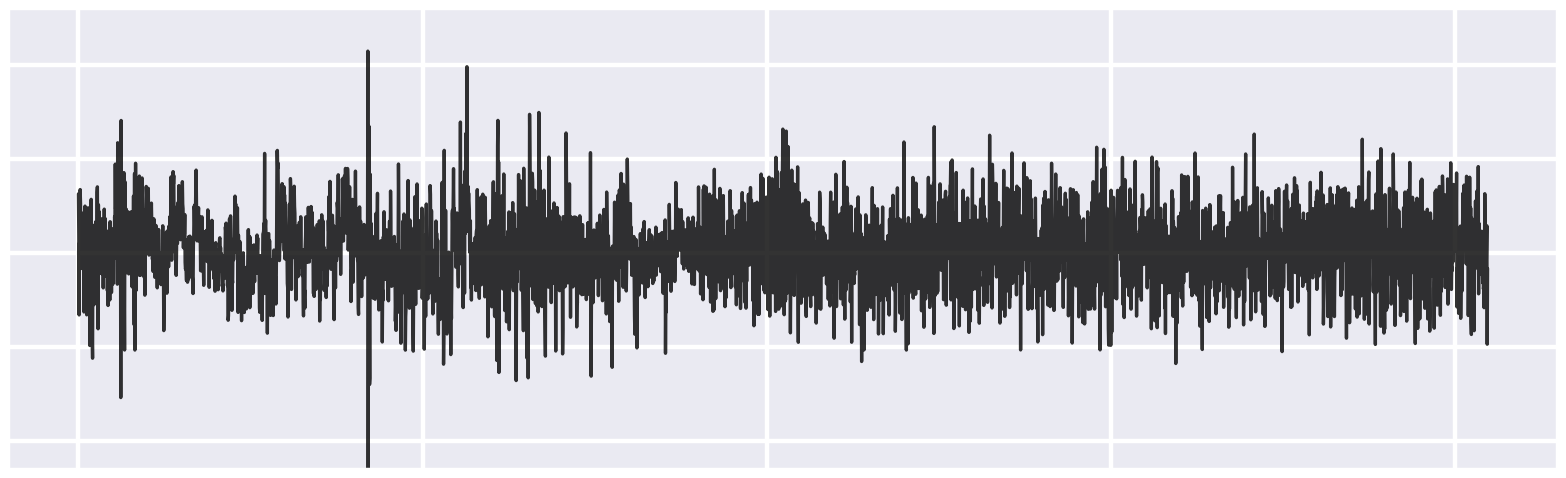}
    \end{subfigure}\hspace{0em}
    \begin{subfigure}[b]{0.32\textwidth}
        \includegraphics[width=\textwidth]{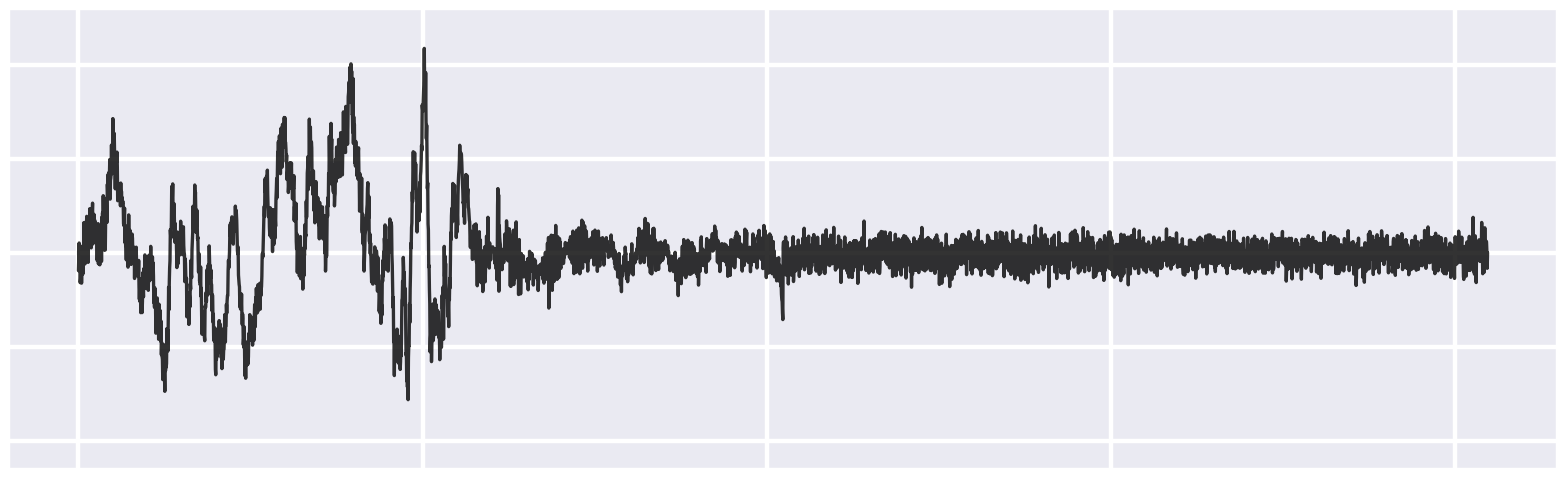}
    \end{subfigure}

   \begin{subfigure}[b]{0.32\textwidth}
        \includegraphics[width=\textwidth]{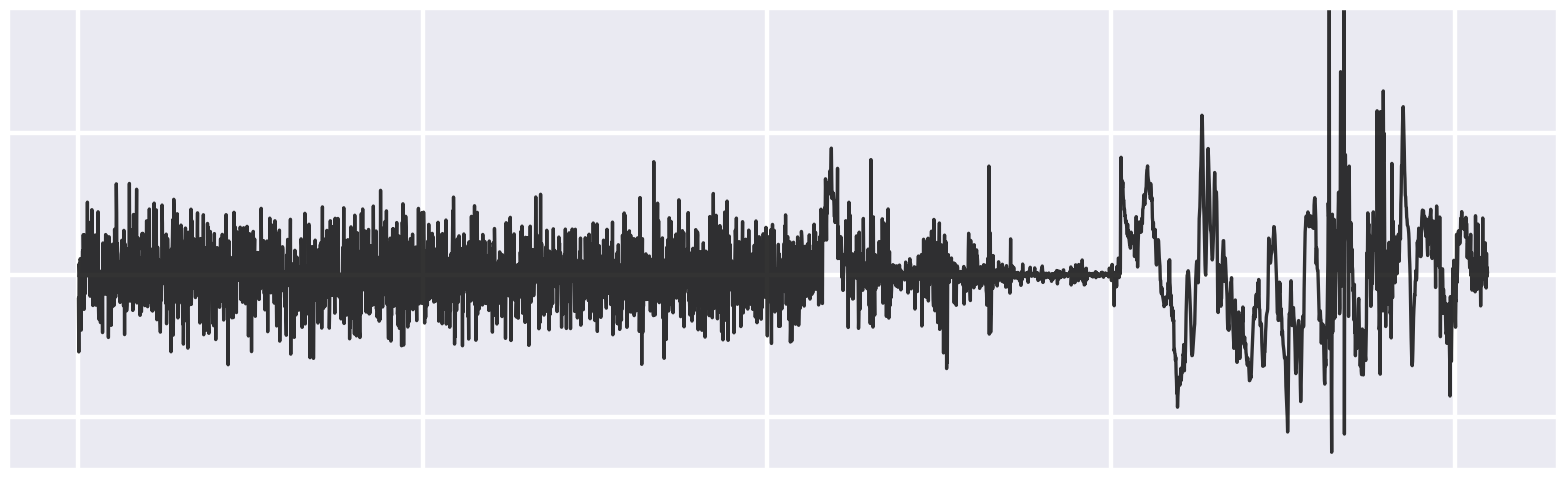}
    \end{subfigure}\hspace{0em}
    \begin{subfigure}[b]{0.32\textwidth}
        \includegraphics[width=\textwidth]{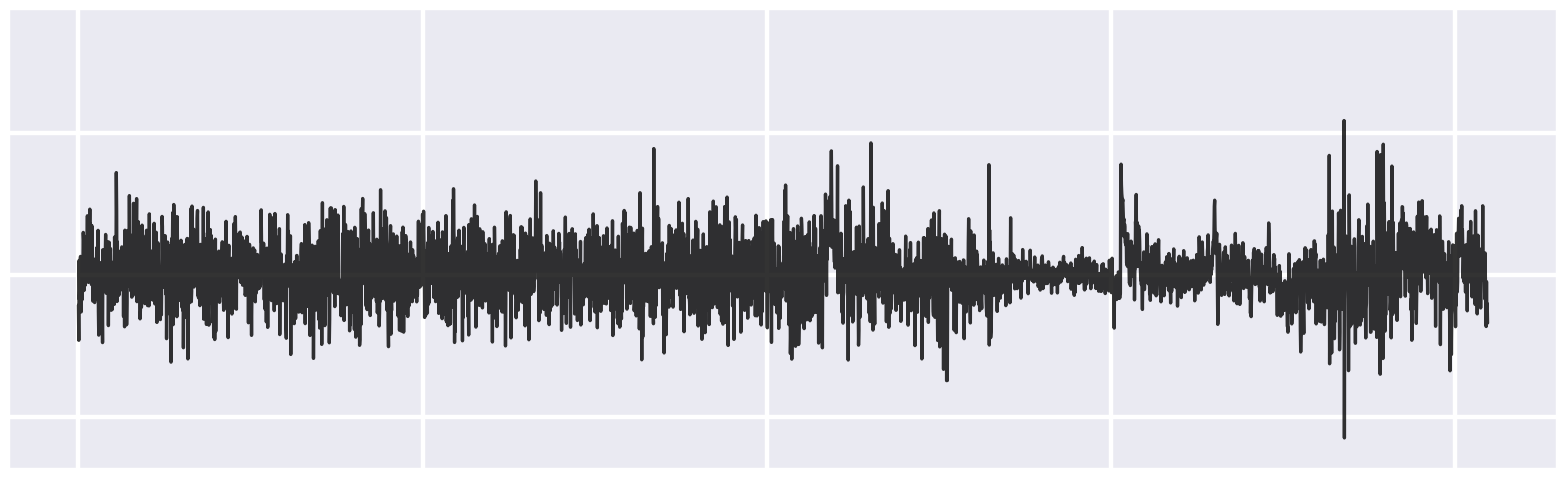}
    \end{subfigure}\hspace{0em}
    \begin{subfigure}[b]{0.32\textwidth}
        \includegraphics[width=\textwidth]{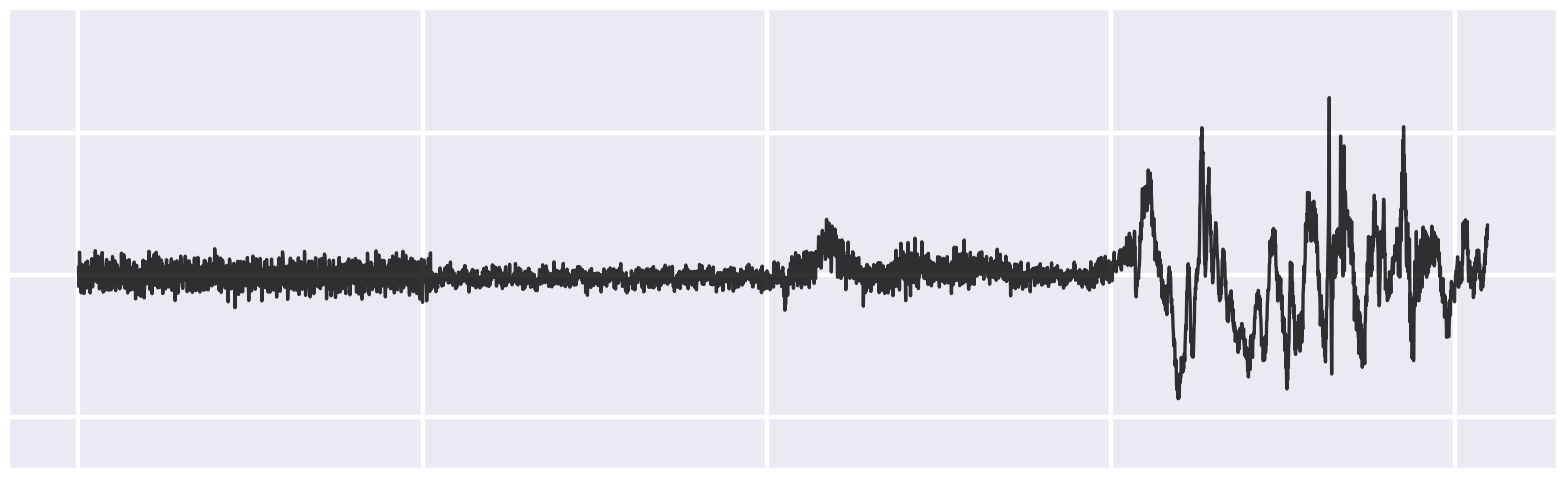}
    \end{subfigure}

   \begin{subfigure}[b]{0.32\textwidth}
        \includegraphics[width=\textwidth]{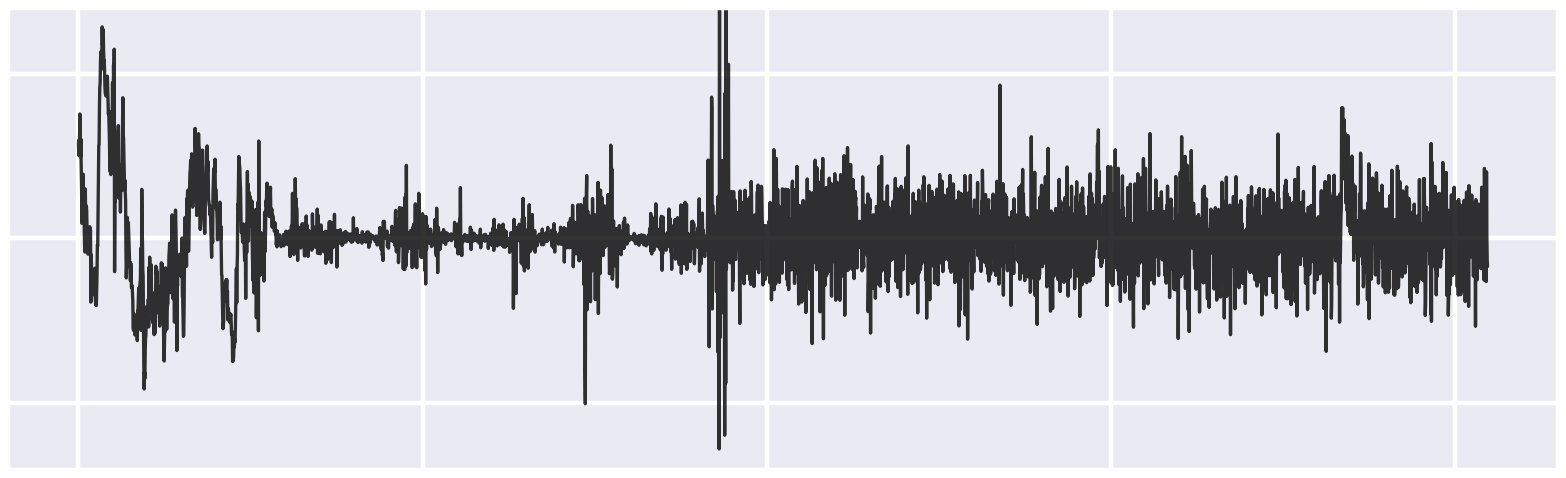}
        \caption{Mixed waveform}
    \end{subfigure}\hspace{0em}
    \begin{subfigure}[b]{0.32\textwidth}
        \includegraphics[width=\textwidth]{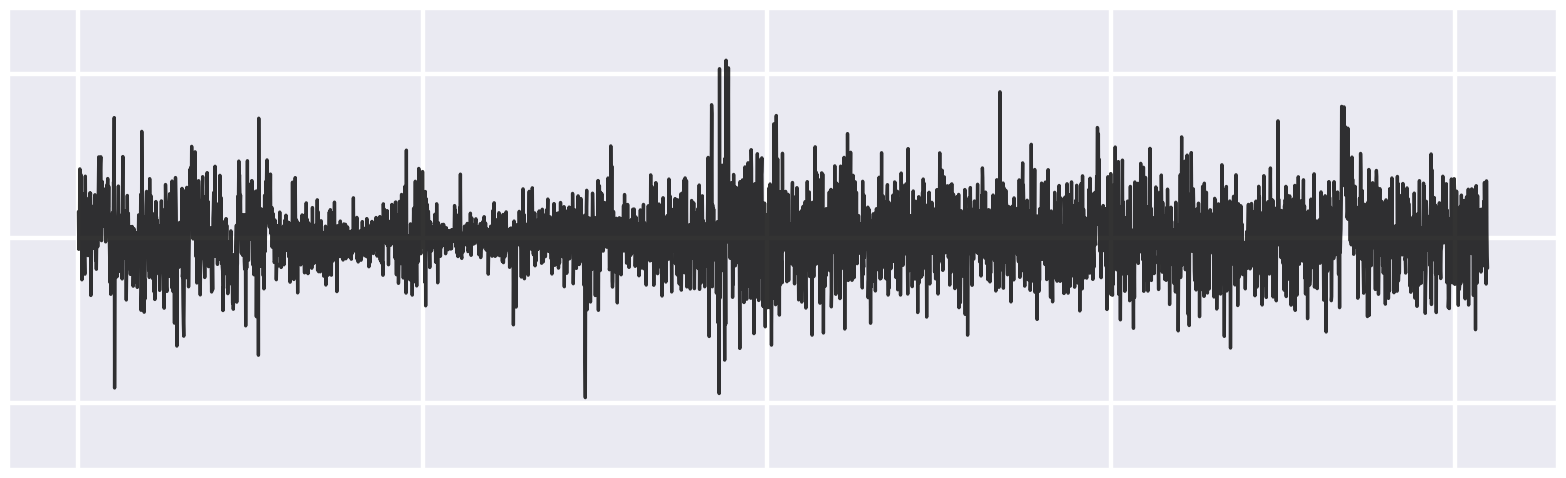}
        \caption{After source separation}
    \end{subfigure}\hspace{0em}
    \begin{subfigure}[b]{0.32\textwidth}
        \includegraphics[width=\textwidth]{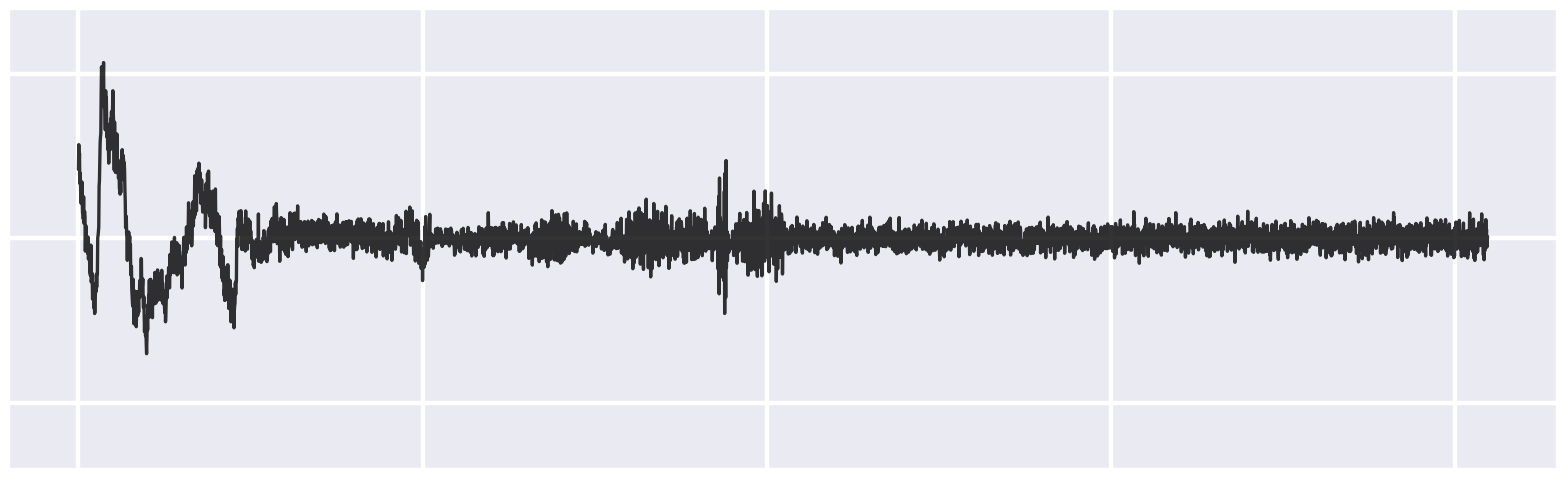}
        \caption{Medium-scale event separated}
    \end{subfigure}
    \caption{Removing the medium-scale source.}
    \label{fig:synthetic_srcsep_g_4096_6_n_1024_6}
\end{figure*}

Similar to the case with data from Mars, we can identify certain patterns in the obtained clusters and use another cluster as prior information in order to perform per-cluster source separation. Here we exemplify this by separating the medium-scale source that we added from cluster seven in in the largest scale (pink waveforms in \cref{fig:synthetic_scale_4096_waves}). To achieve this, we select the pink cluster in \cref{fig:synthetic_scale_1024_waves} as prior cluster as it contains similar background noise to the target cluster but without the medium-scale source. The results of this source separation is shown in \cref{fig:synthetic_srcsep_g_4096_6_n_1024_6} for three waveforms. According to the results, our approach has been able to successfully identify and separated the medium-scale source from a set of target waveforms with minimal distortion to the other sources.

  \end{appendices}

  \end{document}